%% file: main.tex
\documentclass{article}
\usepackage{makecell}
\usepackage{array} 
\usepackage{tabularx}
\usepackage{subcaption}

\PassOptionsToPackage{numbers, compress, sort}{natbib}

\usepackage{xargs}
\usepackage[pdftex,dvipsnames]{xcolor}
\usepackage{xspace}

\newcommandx{\unsure}[2][1=]{\todo[linecolor=red,backgroundcolor=red!25,bordercolor=red,#1]{#2}}
\newcommandx{\change}[2][1=]{\todo[linecolor=blue,backgroundcolor=blue!25,bordercolor=blue,#1]{#2}}
\newcommandx{\info}[2][1=]{\todo[linecolor=OliveGreen,backgroundcolor=OliveGreen!25,bordercolor=OliveGreen,#1]{#2}}
\newcommandx{\improvement}[2][1=]{\todo[linecolor=Plum,backgroundcolor=Plum!25,bordercolor=Plum,#1]{#2}}
\newcommandx{\thiswillnotshow}[2][1=]{\todo[disable,#1]{#2}}

\usepackage[preprint]{neurips_2025}

\usepackage[utf8]{inputenc} 
\usepackage[T1]{fontenc}    
\usepackage{graphicx} 
\usepackage{hyperref}       
\usepackage{url}            
\usepackage{booktabs}       
\usepackage{amsfonts}       
\usepackage{nicefrac}       
\usepackage{microtype}      
\usepackage{xcolor}         
\usepackage{paralist}

\input{macros}

\title{A clinically meaningful foundation model evaluation for structured electronic health records}

\author{%
     Vincent Jeanselme$^\dagger$\thanks{Corresponding authors: Vincent Jeanselme (vj2292@cumc.columbia.edu), Zilin Jing (zj2398@cumc.columbia.edu), Shalmali Joshi (sj3261@cumc.columbia.edu)} 
     \And
     Zilin Jing$^{\dagger*}$ 
     \And
     Aparajita Kashyap$^\dagger$ 
     \And
     Chao Pang$^\dagger$
     \And
     Florent Pollet\thanks{Equal contribution}
     \And
     Young Sang Choi$^\ddag$
     \And 
     Xinzhuo Jiang$^\ddag$
     \And
     Yuta Kobayashi$^\ddag$
     \And
     Yanwei Li$^\ddag$
     \And
     Sara Matijevic\thanks{Equal contribution}
     \And
     Karthik Natarajan
     \And
     Shalmali Joshi$^*$
     \And
     \vspace{-20pt}\\
     Department of Biomedical Informatics\\
     Columbia University \\
     New York, USA
}

\begin{document}

\maketitle

\begin{abstract}
Foundation models (FMs) promise to address core limitations of traditional supervised machine learning: (i) reliance on large amounts of labeled data, (ii) task specificity, and (iii) poor transportability. Despite methodological advances in structured electronic health record (EHR) foundation models, no systematic benchmark has validated whether these models meaningfully deliver on these promises. We introduce a benchmark of 14 clinically meaningful prediction tasks spanning patient prognosis and early diagnosis of acute and chronic conditions. We benchmark 6 state-of-the-art EHR FMs beyond population-level discrimination, emphasizing the need for evaluating their calibration and fairness, with rigorous controls for data contamination and reproducibility across more than 6 million patients from Columbia University Irving Medical Center and MIMIC-IV. Our benchmark identifies that FMs deliver on some of their promises. In particular, top-performing FMs outperform traditional baselines on discriminative performance, especially under limited labeled data, and exhibit more equitable performance across socio-medical groups. However, these models may underperform in low-prevalence settings, as pretraining losses may discard discriminative information about such conditions, and present lower calibration under limited labeled data. Further, cross-institutional transportability remains a challenge for structured EHR FMs. Together, these findings advance our understanding of EHR FMs’ potential for clinical utility, highlight critical gaps that remain to be addressed, and provide a reproducible framework to track progress.
\end{abstract}

\input{content/1.introduction}
\input{content/2.results}

\input{content/3.discussion}
\input{content/4.methodology}

\section*{Author contributions}
SJ, VJ, and YSC conceived and designed the study.
AK, CP, FP, XJ, YL, and ZJ developed the benchmark framework.
VJ developed the evaluation framework.
AK, FP, XJ, YK, YL, and ZJ performed data curation and preprocessing, including conversion of CUMC and MIMIC-IV datasets to MEDS format.
AK, CP, FP, SM, YK, and ZJ conducted the experiments.
VJ analyzed the results.
AK, CP, SJ, VJ, and YSC drafted the manuscript.
All authors reviewed, edited, and approved the final manuscript.

\section*{Acknowledgments}
VJ, YSC, ZJ, and SJ would like to acknowledge partial support from NIH R01MH137679. AK would like to acknowledge the support of NIH T15LM007079. YK and SJ acknowledge partial support from the SPIRIT Precision Pilot Award at Columbia. ZJ and SJ would like to acknowledge partial support from NIH R01LM014380. SJ would like to acknowledge partial support from the SNF Center for Precision Psychiatry and Mental Health, Columbia. SM would like to acknowledge partial support from the UKRI Turing AI Fellowship (Grant Ref: EP/V023233/1) and the UK Engineering and Physical Science Research Council Doctoral Studentship (Grant Ref: EP/S02428X/1, Project Ref: 2722161).
Any opinions, findings, conclusions, or recommendations in this manuscript are those of the authors and do not reflect the views, policies, endorsements, expressed or implied, of any aforementioned funding agencies/institutions.

\section*{Competing interests}
All authors declare no financial or non-financial competing interests. 

\section*{Data availability}
The clinical data used in this benchmark cannot be shared due to patient privacy restrictions. To support reproducibility, we provide an experimental setup on the publicly available MIMIC-IV dataset.

\section*{Code availability}
The code for the proposed benchmark and task definitions is publicly available at \url{https://github.com/reAIM-Lab/ehr_foundation_model_benchmark/}.

\bibliographystyle{naturemag}
\bibliography{references}

\newpage
\appendix
\input{content/appendix}

\end{document}

%% file: macros.tex
\makeatletter
\long\def\@makecaption#1#2{
  \vskip 0.8ex
  \setbox\@tempboxa\hbox{\small {\bf #1:} #2}
  \parindent 1.5em  
  \dimen0=\hsize
  \advance\dimen0 by -3em
  \ifdim \wd\@tempboxa >\dimen0
  \hbox to \hsize{
    \parindent 0em
    \hfil
    \parbox{\dimen0}{\def\baselinestretch{0.96}\small
      {\bf #1.} #2
    }
    \hfil}
  \else \hbox to \hsize{\hfil \box\@tempboxa \hfil}
  \fi
}
\makeatother

\newenvironment{proof-of-claim}{\noindent{\bf Proof of Claim}
  \hspace*{1em}}{\qed\bigskip\\}

\newenvironment{proof-sketch}{\noindent{\bf Sketch of Proof}
  \hspace*{1em}}{\qed\bigskip\\}
\newenvironment{proof-idea}{\noindent{\bf Proof Idea}
  \hspace*{1em}}{\qed\bigskip\\}
\newenvironment{proof-of-lemma}[1][{}]{\noindent{\bf Proof of Lemma {#1}}
  \hspace*{1em}}{\qed\bigskip\\}
  \newenvironment{proof-of-corollary}[1][{}]{\noindent{\bf Proof of Corollary {#1}}
  \hspace*{1em}}{\qed\bigskip\\}
\newenvironment{proof-of-proposition}[1][{}]{\noindent{\bf
    Proof of Proposition {#1}}
  \hspace*{1em}}{\qed\bigskip\\}
\newenvironment{proof-of-theorem}[1][{}]{\noindent{\bf Proof of Theorem {#1}}
  \hspace*{1em}}{\qed\bigskip\\}
\newenvironment{inner-proof}{\noindent{\bf Proof}\hspace{1em}}{
  $\bigtriangledown$\medskip\\}
\newenvironment{proof-attempt}{\noindent{\bf Proof Attempt}
  \hspace*{1em}}{\qed\bigskip\\}

\newcounter{example}

\newenvironment{example*}[1][]{
  \ifthenelse{\isempty{#1}}{%
    \noindent \textbf{Example:}\hspace*{.05em}
  }{%
    \noindent \textbf{Example} ({#1})\textbf{:}\hspace*{.05em}
  }
}{%
  $\diamond$ \bigskip
}

\newcounter{remark}

\newenvironment{remark*}[1][]{
  \ifthenelse{\isempty{#1}}{%
    \noindent \textbf{Remark:}\hspace*{.05em}
  }{%
    \noindent \textbf{Remark} ({#1})\textbf{:}\hspace*{.05em}
  }
}{%
  $\diamondsuit$ \bigskip
}

\newcommand{\statrv}[1][]{
  \ifthenelse{\isempty{#1}}{%
    X
  }{%
    X_{#1}
  }
}

\makeatletter
\long\def\@makecaption#1#2{
  \vskip 0.8ex
  \setbox\@tempboxa\hbox{\small {\bf #1:} #2}
  \parindent 1.5em  
  \dimen0=\hsize
  \advance\dimen0 by -3em
  \ifdim \wd\@tempboxa >\dimen0
  \hbox to \hsize{
    \parindent 0em
    \hfil 
    \parbox{\dimen0}{\def\baselinestretch{0.96}\small
      {\bf #1.} #2
    } 
    \hfil}
  \else \hbox to \hsize{\hfil \box\@tempboxa \hfil}
  \fi
}
\makeatother

\newcommand{\cumc}{CUMC\xspace}
\newcommand{\mimic}{MIMIC-IV\xspace}

\newcommand{\gpt}{\textsc{GPT}\xspace}

\newcommand{\bert}{BERT\xspace}

\newcommand{\behrt}{BEHRT\xspace}

\newcommand{\los}{\texttt{Long LOS}\xspace}

\newcommand{\motor}{MOTOR\xspace}

\newcommand{\cehrgpt}{CEHR-GPT\xspace}

\newcommand{\cehrbert}{CEHR-BERT\xspace}

\newcommand{\mamba}{MAMBA\xspace}

\newcommand{\corebehrt}{CORE-BEHRT\xspace}

\newcommand{\femr}{FEMR\xspace}

\newcommand{\medstab}{\textsc{MEDS-Tab}\xspace}

\newcommand{\llama}{\textsc{LLaMA}\xspace}

\newcommand{\stanfordehr}{{Stanford-EHR}\xspace}

\newcommand{\mambatransport}{\mamba-\textsc{Transport}\xspace}

%% file: content/1.introduction.tex
\section{Introduction}
Machine learning (ML) has long promised to improve health outcomes by enabling early prediction of disease risk~\citep{Ahsan2022-mlearlydx}, opportunistic screening~\citep{al2023machine, pickhardt2023opportunistic}, and outcome-based risk stratification~\cite{zeiberg2019machine}.
However, while ML models and artificial intelligence (AI)-augmented clinical workflows achieve expert-level performance on many such tasks~\cite{rodriguez2019stand,eisemann2025nationwide}, they are rarely implemented in real-world clinical settings due to operational factors~\cite{beede2020human}, high deployment and maintenance costs~\cite{Rajagopal2024-mlopscost}, and limited generalizability~\cite{wynants2020prediction}.
This inability to generalize across different tasks and patient populations primarily stems from the learning paradigm of traditional supervised learning methods, which require training separate models for each task of interest on curated, labeled data. 
This approach requires extensive human annotation, limiting both the scale and diversity of the training data and, consequently, the range of medical conditions studied.

Foundation models (FMs) represent a new paradigm in ML~\citep{bommasani2021opportunities}. 
Pre-trained on large amounts of unlabeled data through self-supervised learning~\citep{devlin2019bert, henderson2023foundation, radford2018improving}, these models learn representations independently of any specific task. 
These models can then be ``fine-tuned'' to adapt to specific tasks, or, alternatively, used to extract representations then leveraged for various downstream applications~\citep{radford2018improving, team2023gemini, guo2025foundation,moor2023foundation}.
FMs have demonstrated state-of-the-art performance across natural language~\citep{touvron2023llama}, multimodal vision-language~\citep{dai2023instructblip}, multivariate time-series~\citep{das2024decoder}, and biological data~\citep{fu2025foundation}.
In particular, these models aim to address the limitations of traditional ML by
(i) reducing the need for human annotations: FMs require no label for pretraining, and task-specific post-training promises to be more data efficient than traditional task-specific models, 
(ii) improving generalizability across tasks: the same FM can be used for post-training for various downstream tasks~\citep{renc2025foundation, henderson2023foundation}, and 
(iii) promising improved transportability across datasets as the model size and data volume scale~\citep{Bian2025-imagingrobustness, Feng2025-genomicrobustness}. 

The increasing availability of structured Electronic Health Records (EHRs) presents an opportunity to leverage FMs in healthcare~\citep{wornow2023shaky}.
Reflecting patients' interactions with the healthcare system, these data capture patients' health and clinical decision-making thereof, offering a unique opportunity for real-time monitoring and diagnosis at the point of care.
Multiple EHR FMs have been proposed in recent years to tackle the unique challenges of EHRs, such as data standardization and interoperability~\citep{Timilsina2025-harmonizingfms}, irregular temporality~\citep{pang2021cehr, steinberg2024motor}, and long context windows to better capture physiological progression~\citep{wornowcontext}, and have demonstrated competitive performance across medical tasks~\citep{bommasani2021opportunities, he2024foundation}. 

Despite the proliferation of EHR FMs, there is a critical lack of consensus on evaluation to quantify their potential for clinical utility~\citep{wornow2023shaky}.  
While recent efforts to harmonize EHR data and standardize AI pipelines have improved evaluation~\citep{arnrich2024medical, hripcsak2015observational, doi:10.1056/AIra2501253, stang2010advancing, xu2025aces}, existing evaluations have often been limited to a single institution (e.g., MIMIC-IV), a limited set of downstream tasks (e.g., mortality prediction with inconsistent definitions), a single measure of performance (e.g., AUROC)~\citep{wornow2024ehrshot, chen2023multimodal}, and often a limited set of FMs.
Such limited evaluations 
(i) prevent robust conclusions of which tasks, clinical settings, and populations benefit from structured EHR FMs, 
(ii) make replicability and reproducibility challenging and resource-intensive, and 
(iii) preclude end-users, such as researchers at healthcare institutions, from a clear understanding of robust design choices and adaptivity to their own settings, a barrier further exacerbated by limited computational resources to recreate such comparisons.
Consequently, this lack of benchmarks leaves the community without a standardized measure of progress and an inability to make informed decisions on infrastructure investment and deployment~\citep{wang2025crisis, Unknown2025-ul}.

Our work addresses this gap through a robust and reproducible evaluation of EHR FMs. The benchmark is designed around (i) diverse clinically-relevant tasks, meaning task and cohort definitions that align with clinical use cases, (ii) a comprehensive assessment, beyond discrimination, particularly calibrated risk and subpopulation fairness analysis, which are critical for fair deployment, and (iii) reproducible frameworks to measure progress and enable systematic evaluation on new datasets and FMs. 
Specifically, our benchmark's primary contributions are:
We introduce $14$ curated, clinically meaningful downstream prediction tasks comprising $11$ phenotypes and $3$ patient prognoses, spanning chronic conditions, acute events, and healthcare operations outcomes such as readmission. 
For each condition, the studied cohort is restricted to patients for whom the prediction is clinically relevant, based on prior diagnoses that represent risk factors for the phenotype of interest, i.e., within an `at-risk' cohort. This task definition improves clinical actionability, increases task difficulty, and is curated to prevent data leakage (see Appendix~\ref{apd:phenotypes} for detailed cohort descriptions for all tasks). 

We evaluate $6$ FMs across key design choices such as tokenization and pre-training strategies developed at major healthcare institutions, and assess the transportability of EHR-based FMs.
The proposed evaluation enables a robust and fair comparison across these design choices, with manual and automated curation to prevent data contamination between pre-training, linear probing, fine-tuning, and downstream evaluation.
Beyond discriminative performance, which is often used as the only evaluation metric, we present a broad set of metrics to capture calibration and fairness with respect to healthcare-specific subgroup identifiers, such as healthcare utilization.
Given that compute itself is a critical bottleneck in healthcare, we evaluate the computational cost of pretraining such FMs on site-specific data, and the direct utility of transporting EHR FMs without local fine-tuning.

The benchmark operates on the increasingly robust infrastructure for standardizing downstream tasks and FM evaluations~\citep{arnrich2024medical, doi:10.1056/AIra2501253, xu2025aces}\footnote{\url{https://www.ohdsi.org/data-standardization/}} with standardized MEDS-formatted data~\citep{doi:10.1056/AIra2501253}. The full benchmark is publicly available on GitHub\footnote{\url{https://github.com/reAIM-Lab/ehr_foundation_model_benchmark/}} and is intended as a prescriptive framework for future assessments of structured and unstructured EHR FMs. 

Applying this benchmark on \cumc and \mimic yields four key insights into how FMs generalize across tasks and patient populations. 
First, top-performing FMs outperform traditional baselines in discrimination but underperform on calibration under limited labeled data. This weakens their applicability in clinical settings where reliable risk estimates are critical for decision-making. 
Second, by pretraining on large amounts of data, FMs extract robust representations across diverse groups, yielding fairer predictions. 
Third, FMs provide the most benefit under limited labeled data, presenting an opportunity for diseases where large-scale annotated data are hard to obtain. However, these models underperform for low-prevalence conditions, as pretraining losses may not capture these conditions.
Finally, cross-institutional transportability remains unmet: variations in clinical practice and patient characteristics result in insufficient distributional overlap, limiting the performance of externally trained FMs. 
Together, these findings highlight that realizing the clinical potential of EHR FMs requires not only technical innovation but also rigorous evaluation frameworks and standardized data pipelines.

%% file: content/2.results.tex
\section{Results}
\label{sec:results}

\subsection{Experimental setting}
We pre-train all $6$ state-of-the-art structured EHR FMs independently on two structured EHR datasets represented in the MEDS format: \cumc and \mimic. Using these models, we then extract embeddings on a test set. In a process known as linear probing, we train logistic regression heads from the extracted representations to create task-specific models. Independently, we train task-specific supervised models to predict the label of interest from the raw data. First, we compare all models along population-level discrimination, calibration, and fairness metrics across race, gender, and healthcare utilization. For tasks where relative improvement over task-specific baselines is low, we evaluate whether full fine-tuning of top-performing FMs can close the performance gap. Finally, we assess transportability using a model trained on \stanfordehr~and tested on \cumc. Table~\ref{tab:demographics} summarizes the population characteristics of all considered datasets: \cumc, \mimic, and \stanfordehr (for transportability only). Our methodology is detailed in Section~\ref{sec:benchmarking}.

\subsection{Are structured EHR FM embeddings predictive of clinically meaningful outcomes?} 
Our first evaluation assesses how embeddings extracted from FMs perform across diverse downstream tasks using linear probing. Tables~\ref{tab:leaderboard_cumc} (a), (b) summarize average model rankings for the two datasets (\cumc and \mimic). In these settings, models are pretrained on each dataset, then linearly probed for the task of interest by training a logistic regression with $100,000$ labeled datapoints. 

In both datasets, \motor and \cehrgpt achieve better discrimination than the best supervised baseline. However, while \motor achieves the best discriminative performance, the model ranks among the lowest in calibration among FMs. This shows that while FMs can outperform on discrimination, they may underperform on calibration. Notably, supervised baselines achieve strong calibration performance in \cumc but rank last in \mimic. Note that the calibration difference is small in this large-data regime, and the best baseline provides a more consistent calibration benefit in the small-data regime (see Appendix Tables~\ref{tab:leaderboard_perf_cumc} and~\ref{tab:leaderboard_perf_mimic}).


Beyond discrimination and calibration, Table~\ref{tab:leaderboard_cumc} provides evidence that pre-training on large, diverse populations improves fairness relative to task-specific models. While FMs still exhibit differential group performance, their average fairness ranking improves over traditional supervised baselines on both datasets.

Computation time and inference FLOPs are reported in Appendix~\ref{apd:resources}.

\begin{table}[!ht]
    \caption{Average performance rank across all tasks stratified by metrics of interest on \cumc and \mimic. Models are sorted by average ranking across the three metrics.}
    \label{tab:leaderboard_cumc}
    \centering
        \footnotesize
        \setlength{\tabcolsep}{2pt}
        \setlength\extrarowheight{-3pt}
        \begin{subtable}[b]{0.45\textwidth}
            \centering
            \caption{\cumc}
            \begin{tabular}{lccc}
                FMs & Calibration & Discrimination & Fairness\\
                \midrule
                \motor & 4.71 & \textbf{1.79} & \textbf{3.20} \\
                \cehrgpt & \textbf{2.86} & 3.43 & 3.71 \\
                \mamba & 3.36 & 3.86 & 3.32 \\
                \llama & 3.00 & 3.79 & 3.80 \\
                Best Baseline & 3.00 & 3.57 & 6.12 \\
                \cehrbert & 4.64 & 4.71 & 3.90 \\
                \corebehrt & 6.43 & 6.86 & 3.95 \\
            \end{tabular}
        \end{subtable}
        \hfill 
        \begin{subtable}[b]{0.45\textwidth}
            \centering
            \caption{\mimic}
            \begin{tabular}{lccc}
                FMs & Calibration & Discrimination & Fairness\\
                \midrule
                \motor & 4.67 & \textbf{1.33} & \textbf{3.08} \\
                \cehrgpt & 4.17 & 2.00 & 3.58 \\
                \llama & \textbf{2.83} & 4.00 & 3.67 \\
                \mamba & 3.00 & 4.50 & 3.75 \\
                \cehrbert & 3.83 & 5.00 & 3.67 \\
                Best Baseline & 4.67 & 4.33 & 6.33 \\
                \corebehrt & 4.83 & 6.83 & 3.92 \\
            \end{tabular}
        \end{subtable}
\end{table}

\subsection{How do structured EHR FMs perform with limited labeled data?}
One promise of FMs is their reduced requirement for labeled data to achieve superior performance. After pre-training on large amounts of data, the extracted representations are expected to generalize to downstream tasks, even with limited labels. To test this assumption, we measure performance by increasing the number of labeled data from $100$ to $100,000$ under linear probing. 

\begin{figure}[!ht]
    \centering
    \begin{subfigure}{0.49\textwidth}
        \includegraphics[width=200pt]{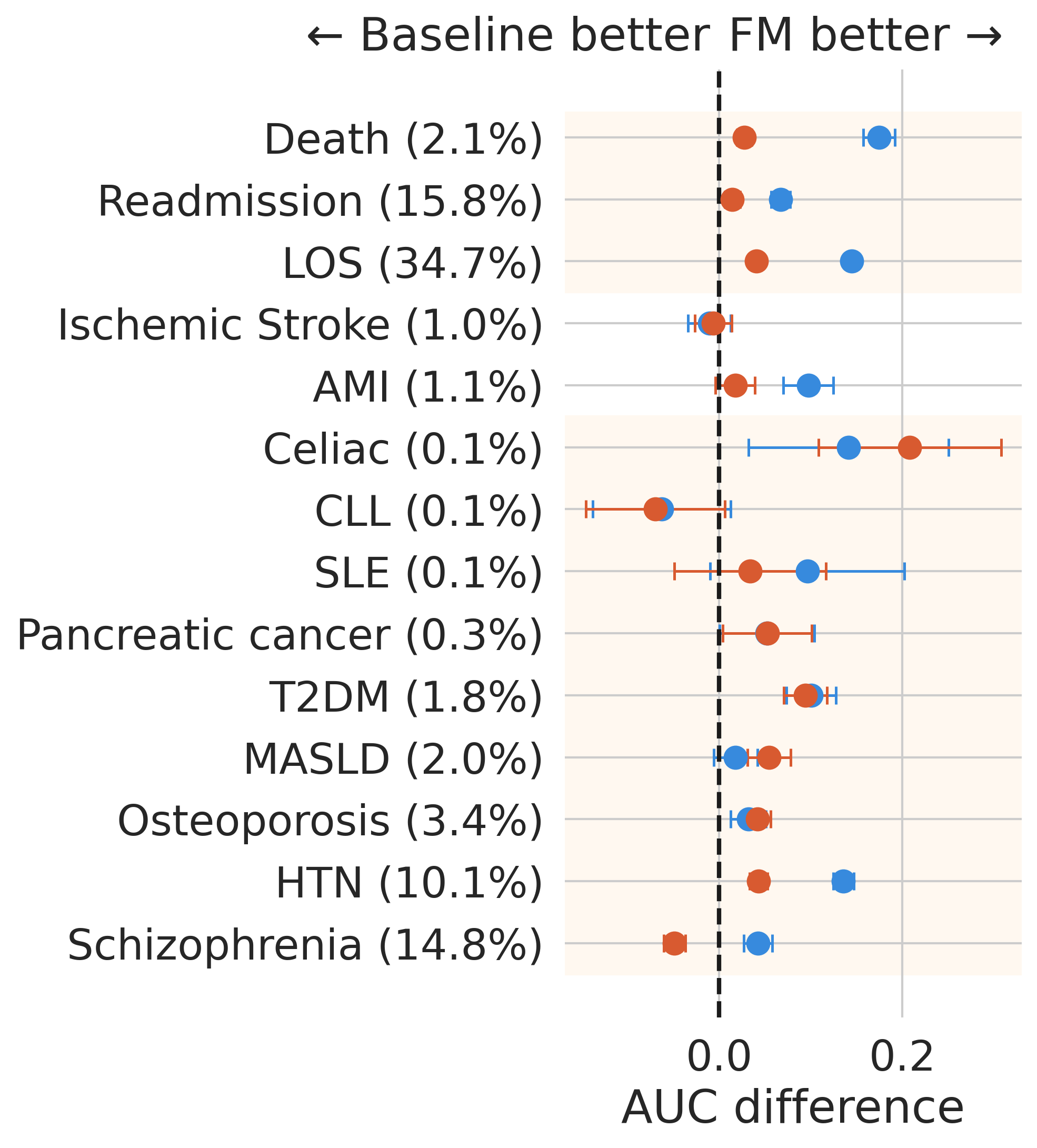}
        \caption{Performance gap on \cumc}
    \end{subfigure}
    \begin{subfigure}{0.49\textwidth}
        \includegraphics[width=200pt]{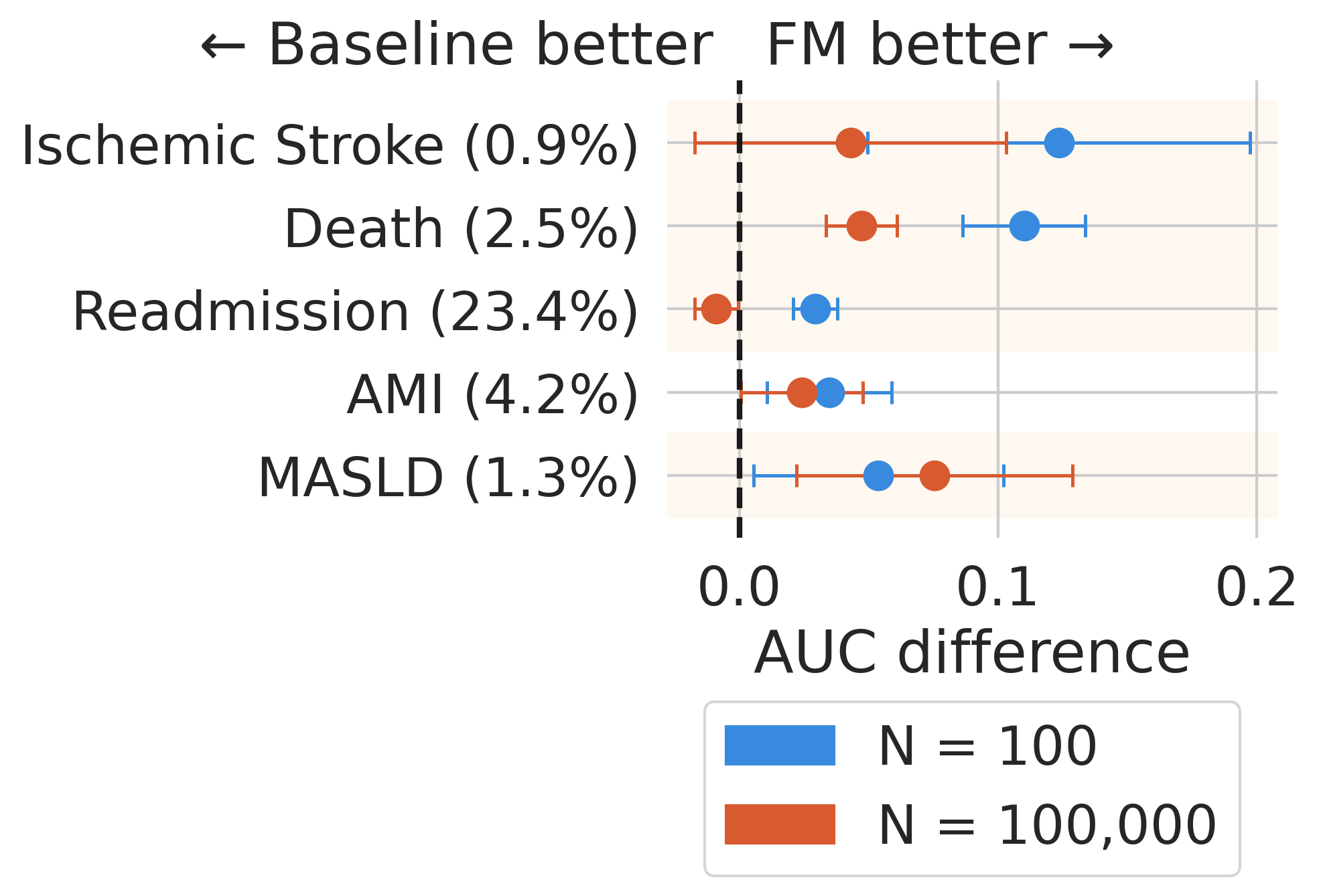}
        \caption{Performance gap on \mimic}
    \end{subfigure}
    \caption{Discriminative performance gap between the best-performing foundation model (FM) and the best baseline on \cumc and \mimic. Numbers in parentheses indicate the prevalence of positive cases for each task.}
    \label{fig:gap}
\end{figure}

Figures~\ref{fig:gap} (a), (b) show the gap in discriminative performance between the best-performing FM and the best baseline when using linear probing on $N$ data points. Appendices~\ref{appendix:additional_results:cumc} and~\ref{appendix:additional_results:mimic} detail the performance for a range of $N$ for each task, model, and metric on both datasets. 
A positive gap indicates that the best FM outperforms the best task-specific baseline. 

When trained with a small amount of labeled data, FMs systematically outperform the best baseline, except for low-prevalence conditions. This finding is particularly relevant in clinical settings where labeled data is scarce, and FMs may offer the most value precisely where traditional supervised approaches are most limited. 

However, FMs present limited gains in predicting Ischemic Stroke and Chronic lymphocytic leukemia (CLL), for which the prevalence is below 1\%. This low prevalence may prevent the model from capturing these events during pretraining, resulting in representations that fail to distinguish patients affected by these conditions.  
Increasing the number of labels from $100$ to $100,000$ often reduces the gap between the best-performing FM and task-specific baselines, diminishing the value of using FMs over more traditional baselines.

\subsection{Does FM fine-tuning further improve performance over task-specific baselines?} 
\begin{figure}[!ht]
    \centering
    \begin{subfigure}{\textwidth}
        \includegraphics[height=100pt]{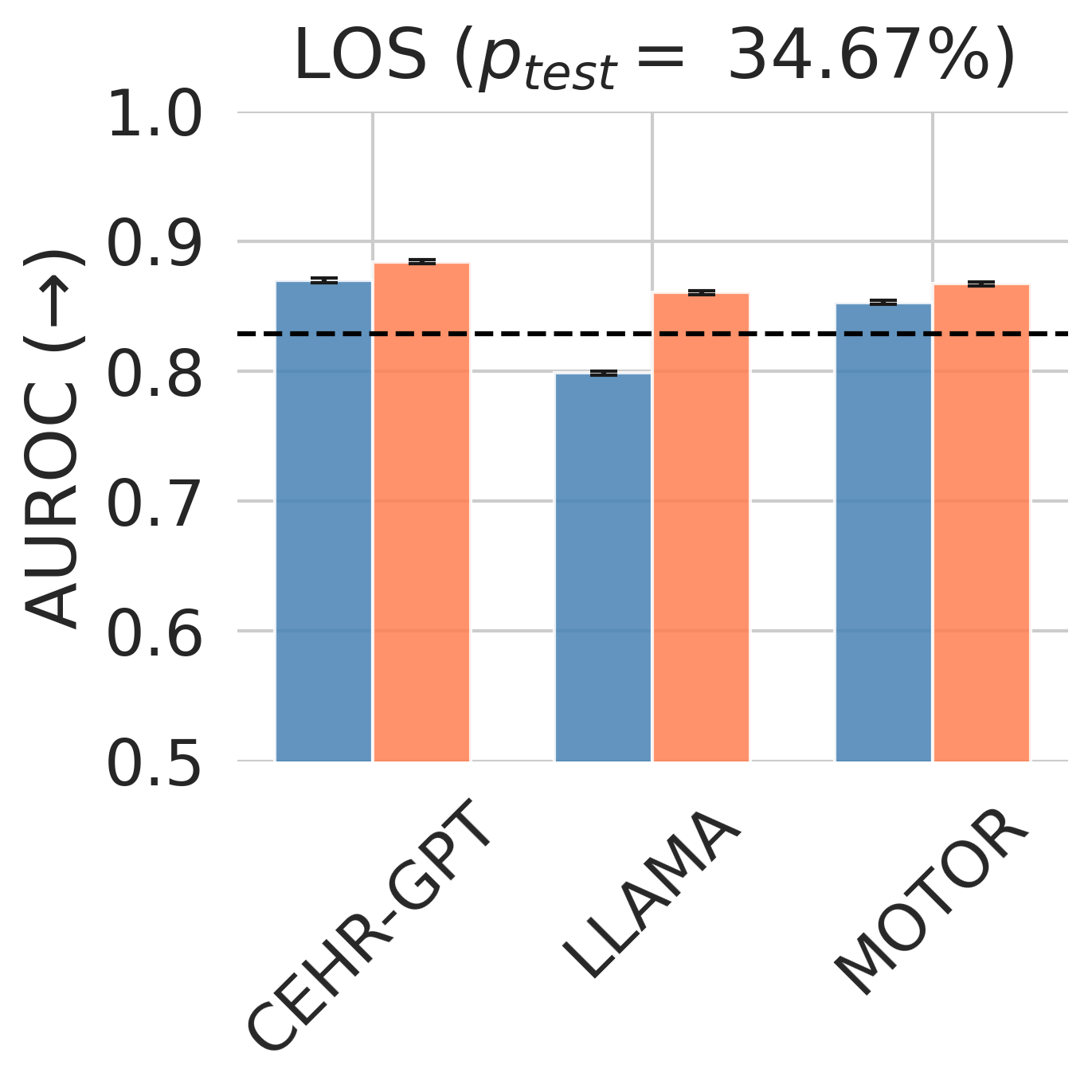}
        \includegraphics[height=100pt]{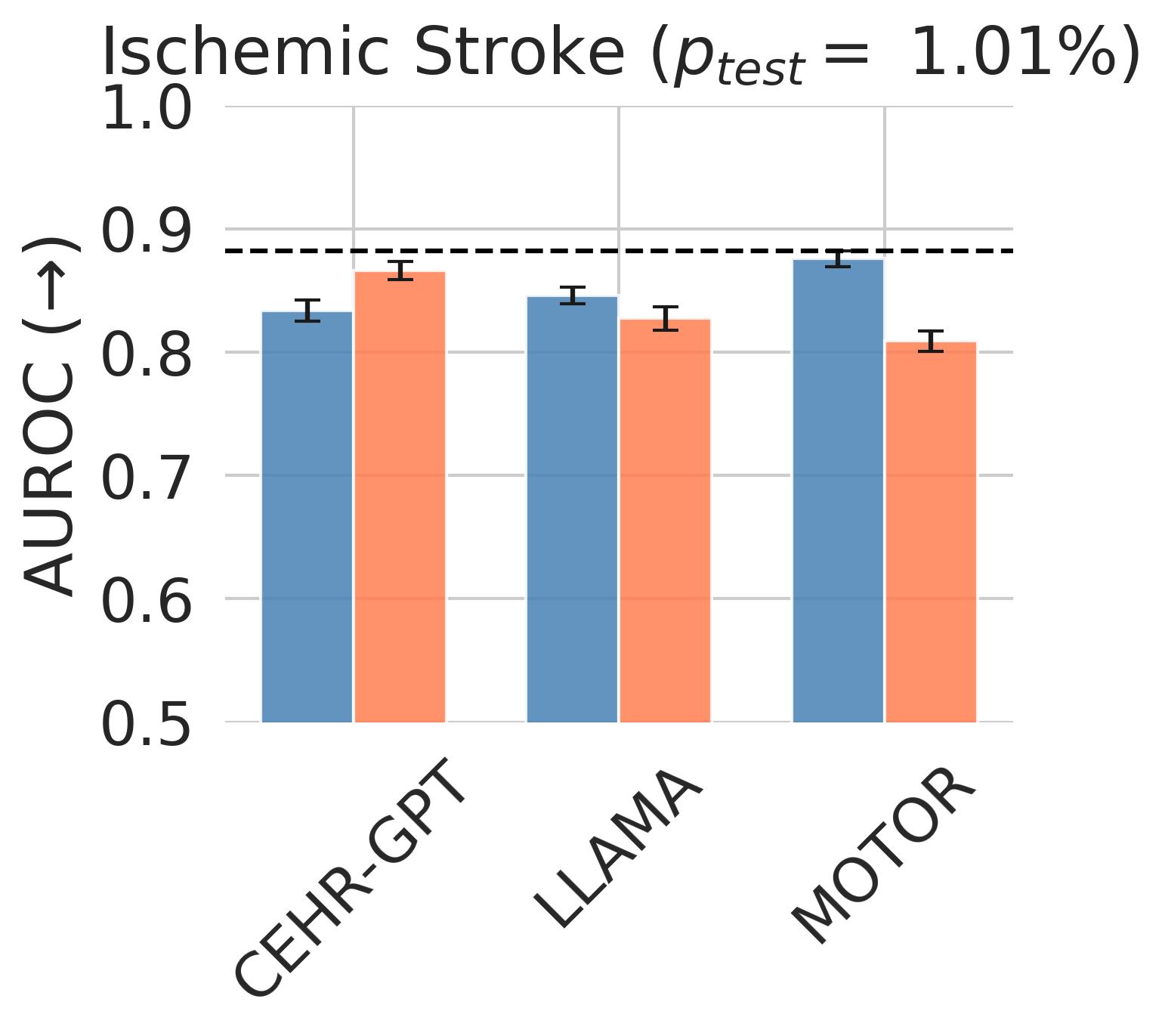}
        \includegraphics[height=100pt]{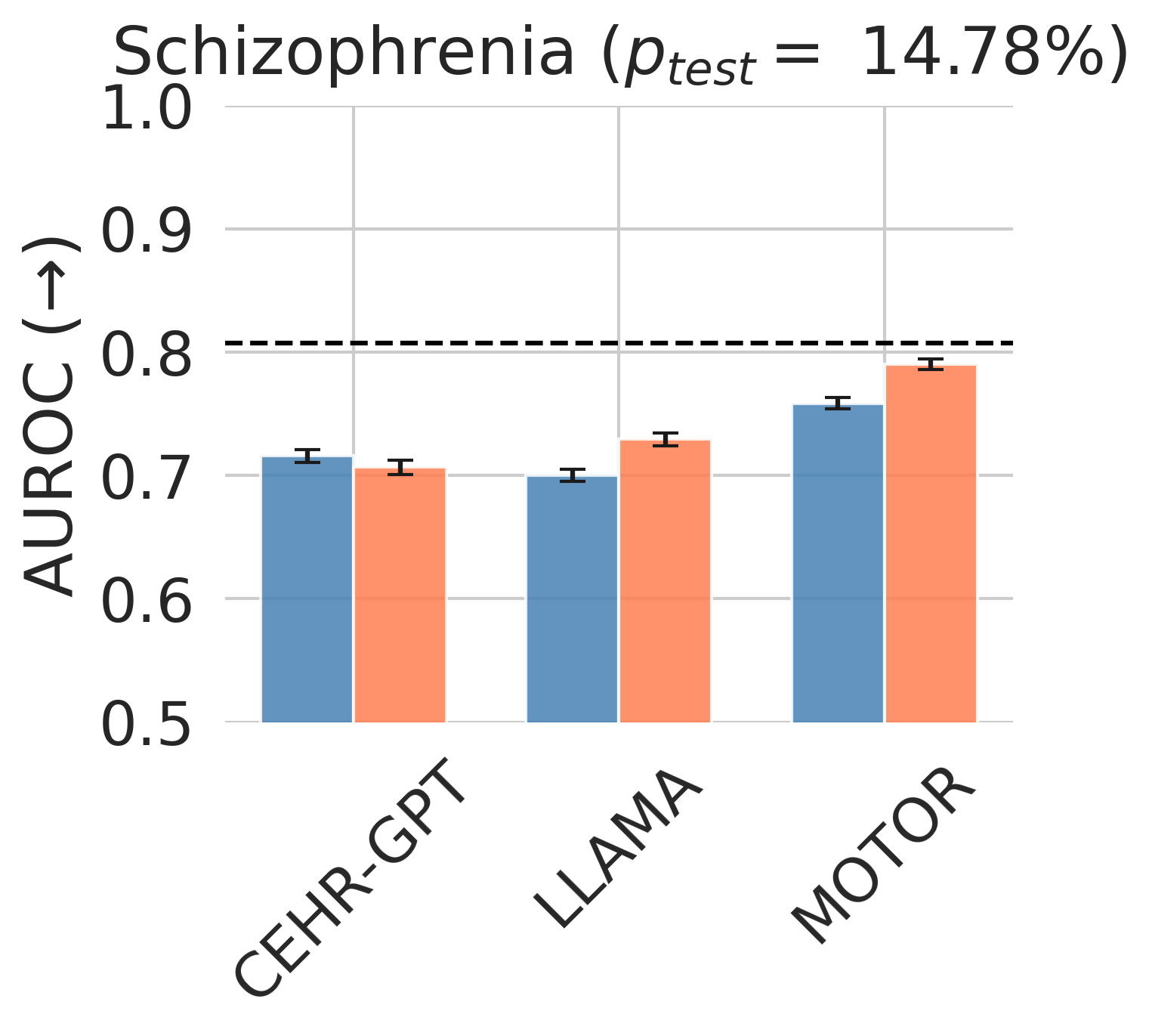}
        \includegraphics[height=100pt]{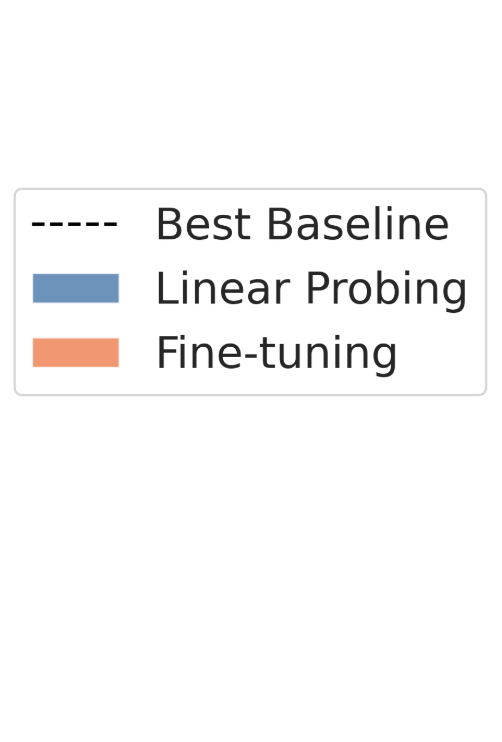}
        \caption{Fine-tuning performance on \cumc}
    \end{subfigure}
    \begin{subfigure}{\textwidth}
        \includegraphics[height=100pt]{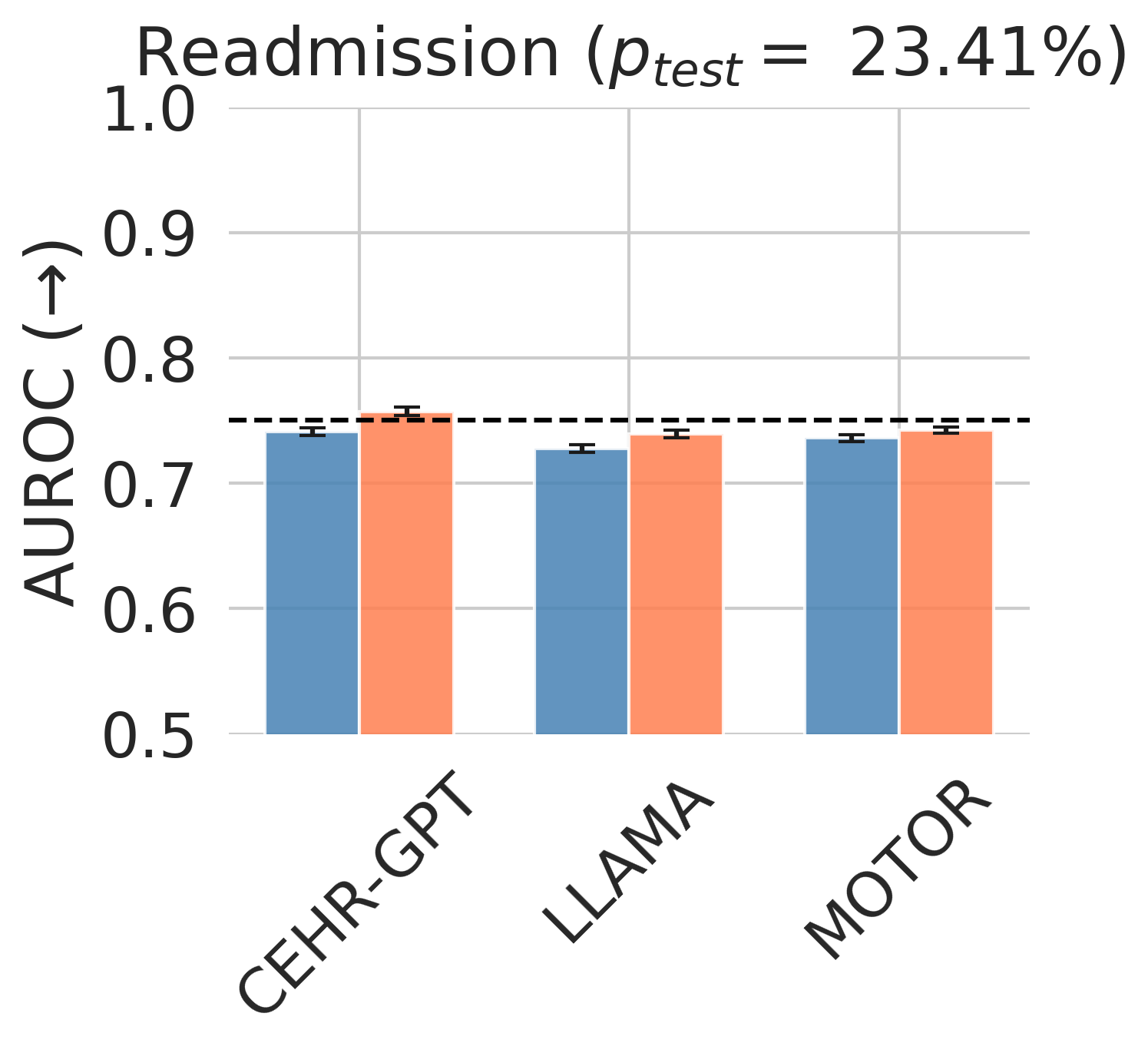}
        \includegraphics[height=100pt]{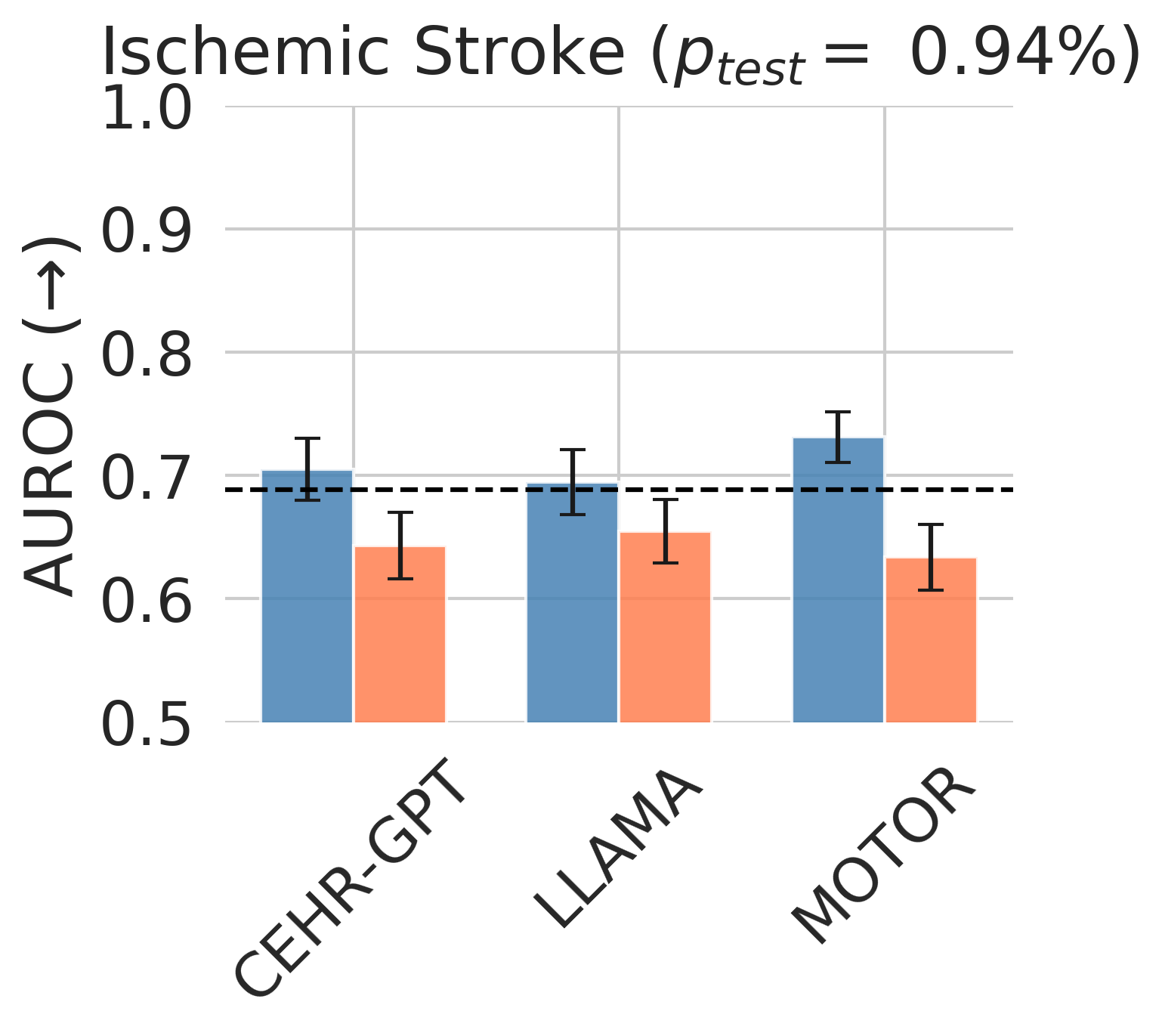}
        \includegraphics[height=100pt]{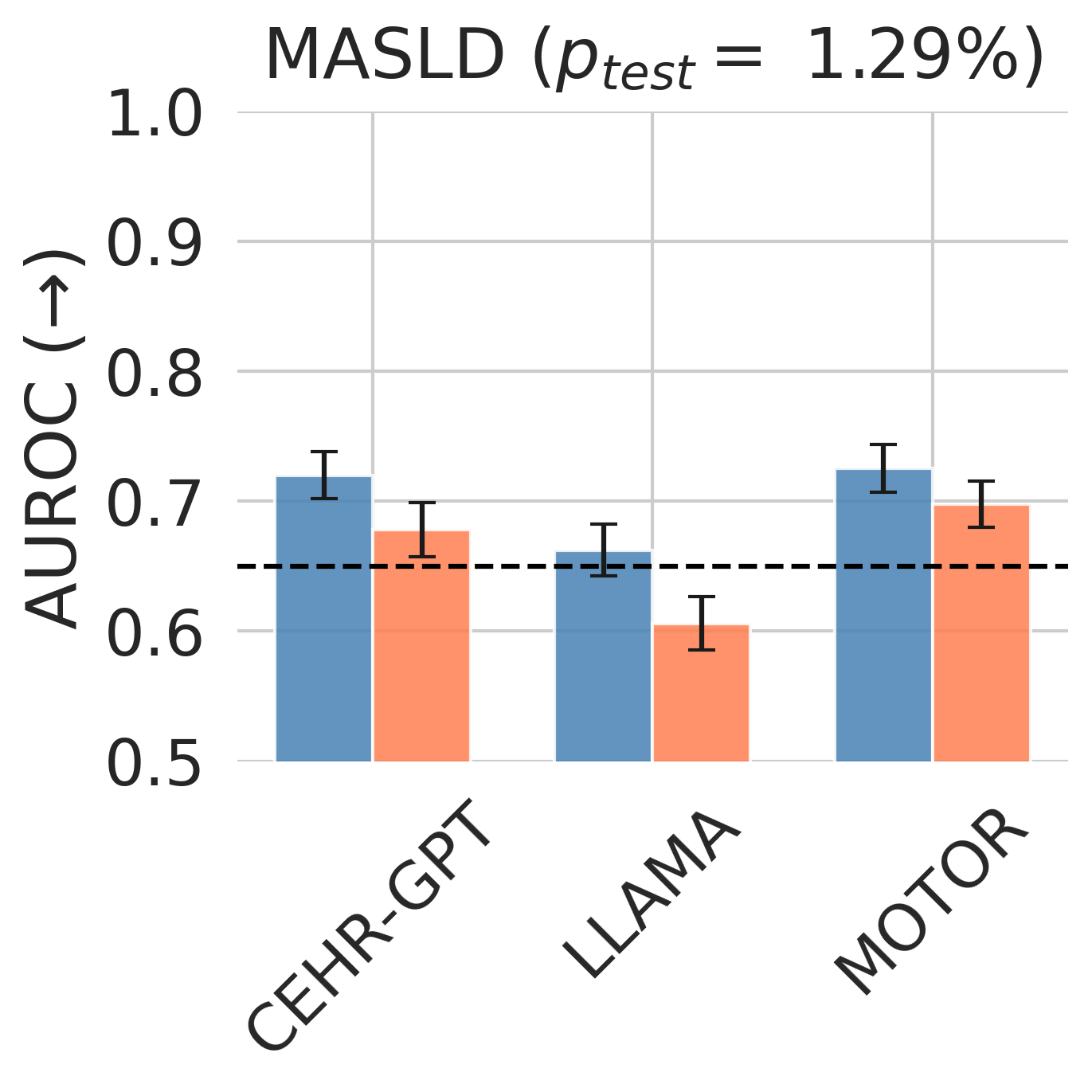}
        \includegraphics[height=100pt]{figures/legend_finetune.png}
        \caption{Fine-tuning performance on \mimic}
    \end{subfigure}
    \caption{Discriminative performance comparison between fine-tuned and linear probing of FMs.}
    \label{fig:fine_tuning}
\end{figure}

While the previous analyses focused on linear probing, fine-tuning FMs may offer further performance improvements. 
However, this comes at the cost of increased computational requirements and a greater need for labeled data. 
Figure~\ref{fig:fine_tuning} compares the performance of linear probing and fine-tuning for three top-performing FMs on three tasks selected because at least one FM underperformed the supervised baseline when using linear probing. 

In the subset of tasks evaluated, we find that fine-tuning does not consistently close the performance gap relative to traditional supervised approaches, even when applied to top-performing FMs with $100,000$ training samples. This is particularly pronounced under low-prevalence conditions, as shown for Ischemic Stroke, where fine-tuning failed to improve performance for \llama~and \motor. Similar results are observed for the low-prevalence conditions Ischemic Stroke and MASLD in \mimic, where fine-tuning worsens performance. This observation may stem either from pretraining failing to capture these conditions or from fine-tuning on a limited number of labels resulting in overfitting.

\subsection{Are FMs transportable across healthcare institutions?}
A key promise of FMs is their ability to generalize across institutions. 
A FM pretrained at one hospital could be used to generate patient representations at another site and be directly applied to downstream tasks, eliminating the need for local training and reducing the computational cost of deploying clinical models.

\begin{figure}[!ht]
    \centering
    \begin{subfigure}{0.25\textwidth}
        \includegraphics[height=80pt]{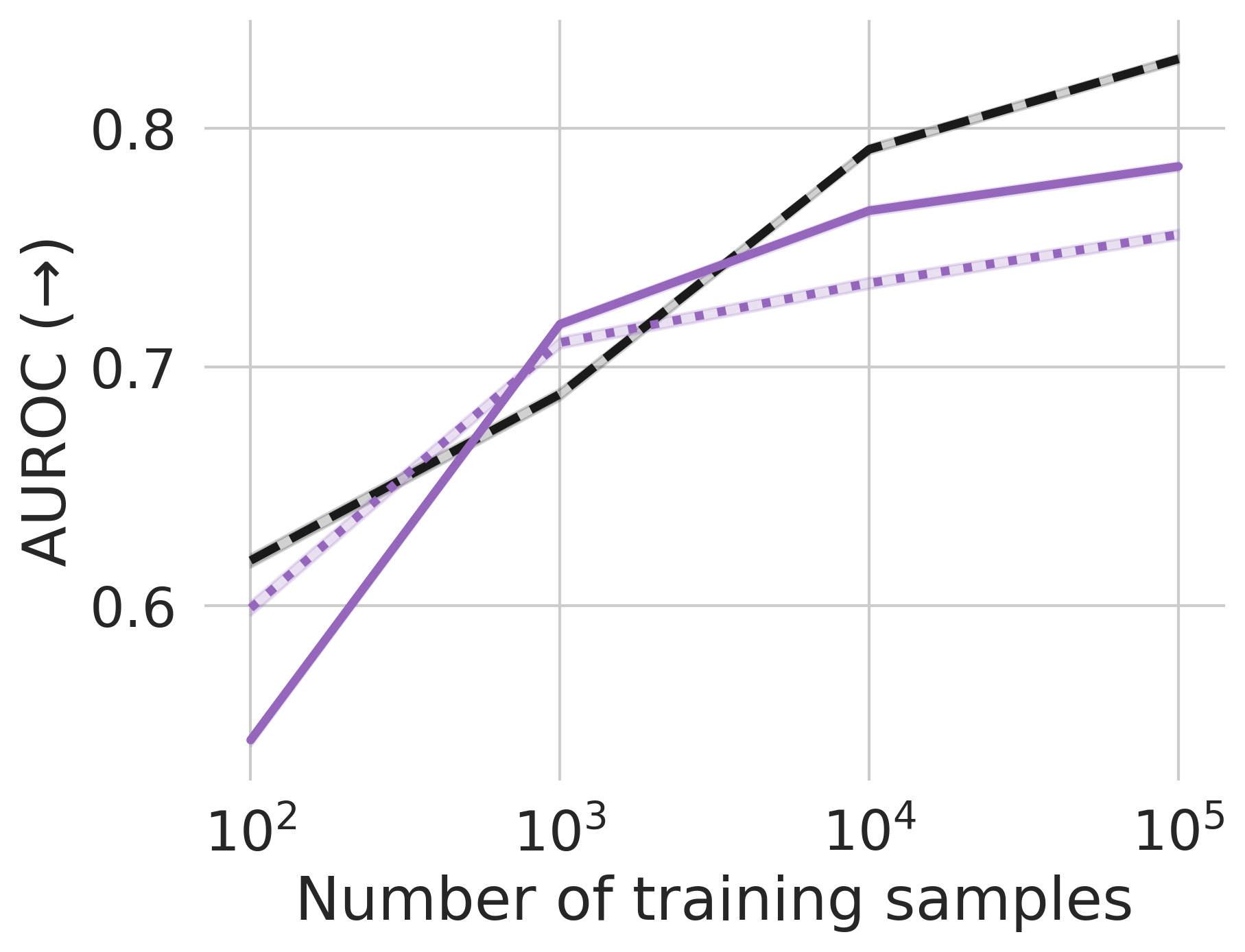}
        \caption{LOS}
    \end{subfigure}
    \begin{subfigure}{0.25\textwidth}
         \includegraphics[height=80pt]{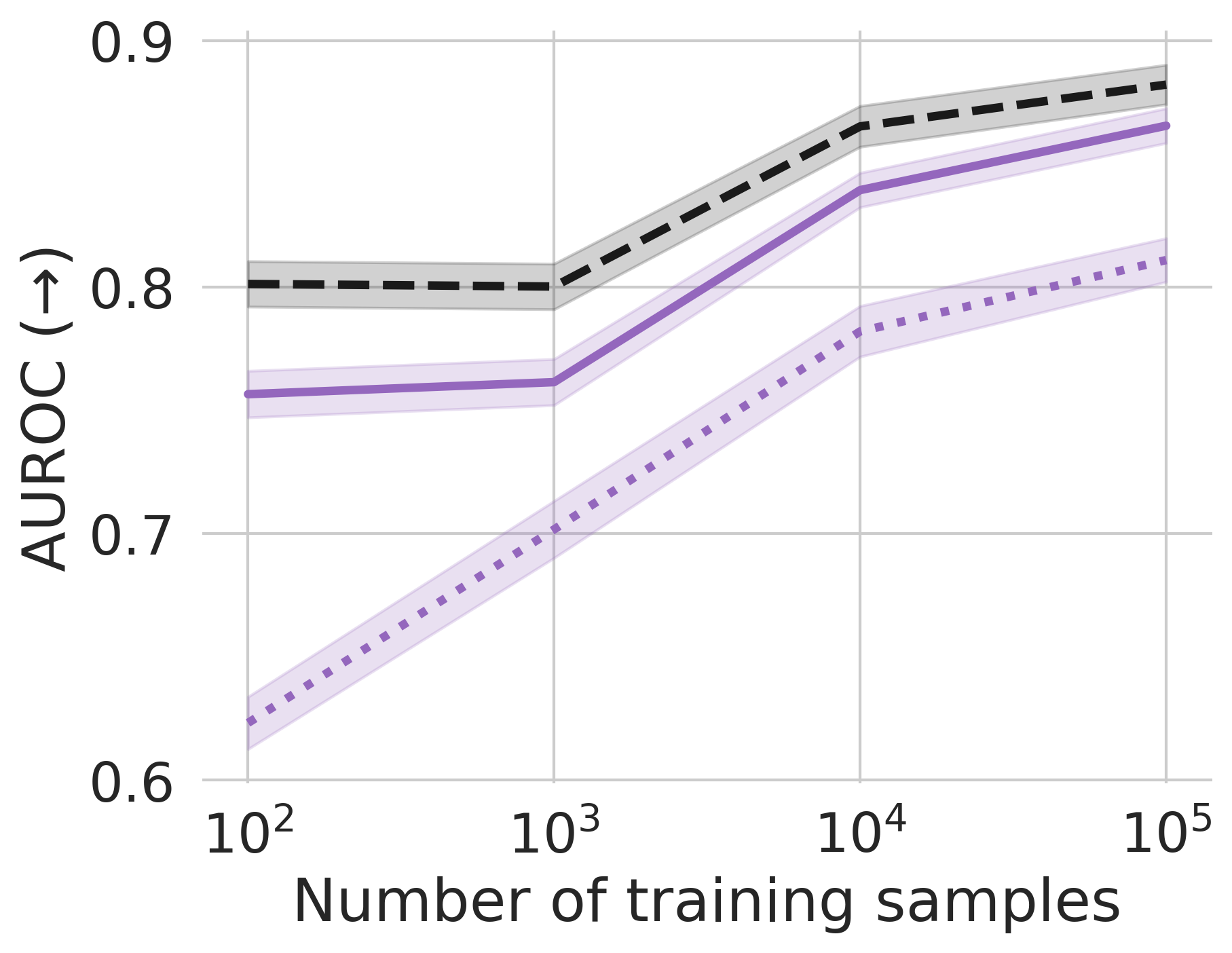}
        \caption{Ischemic Stroke}
    \end{subfigure}
    \begin{subfigure}{0.25\textwidth}
        \includegraphics[height=80pt]{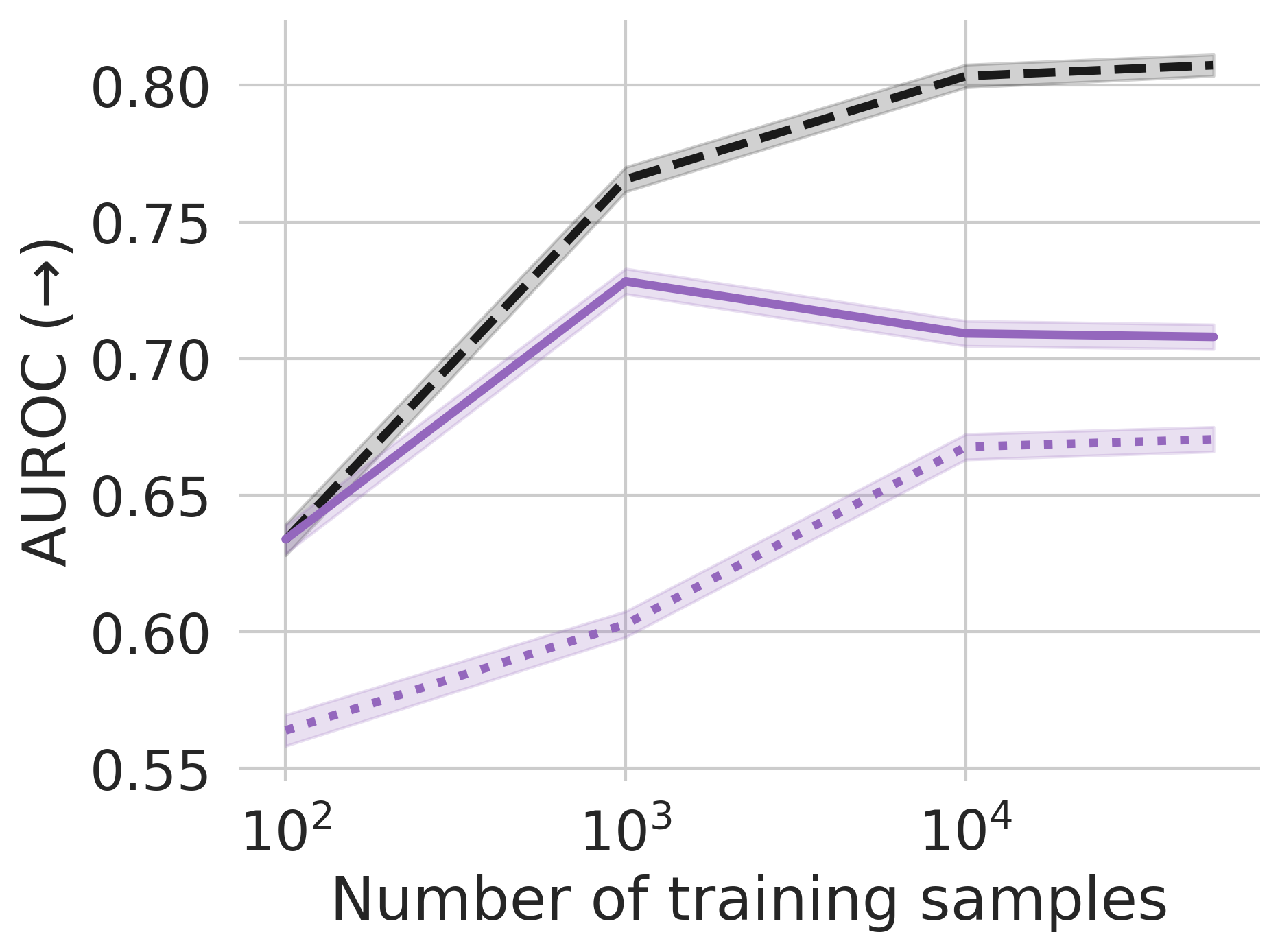}
        \caption{Schizophrenia}
    \end{subfigure}
    \includegraphics[height=100pt]{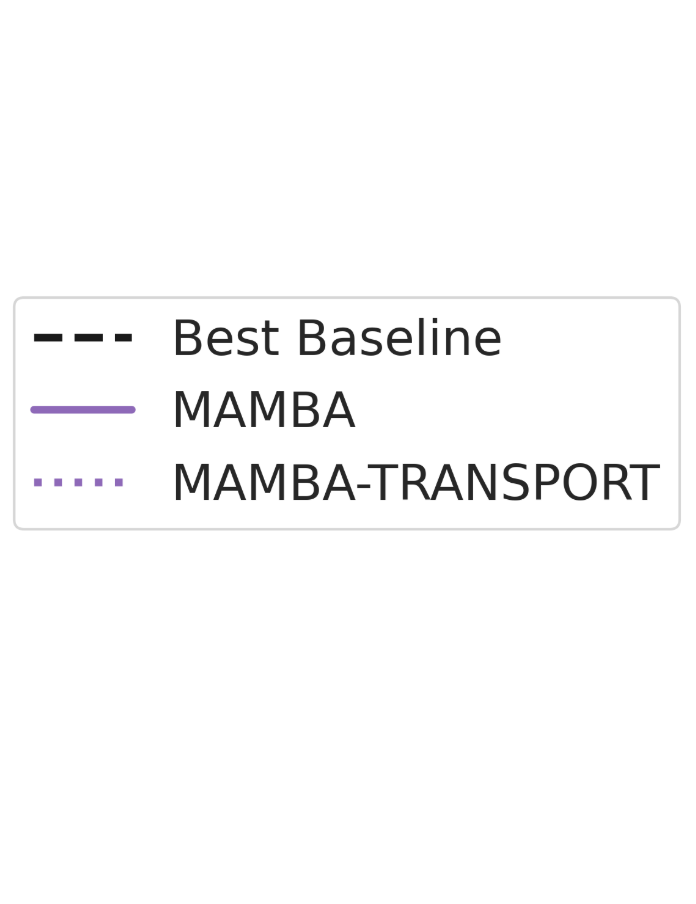}
    \caption{Transport performance under increasing numbers of labeled data.}
    \label{fig:transport}
\end{figure}

We evaluate this property using a \mamba model pretrained on \stanfordehr (\mamba-\textsc{Transport}) and run inference on \cumc with local linear probing. We compare this transported model to the same model pre-trained and tested on \cumc in Figure~\ref{fig:transport} (Performance on all other tasks is detailed in Appendix~\ref{app:transport}). 

Across tasks, the transported model consistently underperforms its locally trained counterpart and the task-specific baseline. This result calls into question the robustness of FMs to new medical environments, where differences in EHR recording practices, coding, and clinical workflows may fundamentally limit their transportability. For example, healthcare utilization varies significantly across institutions, with \cumc averaging 118.4 distinct events per year and \stanfordehr averaging only 28.6 (Appendix \ref{app:datasets}). These differences likely reflect both population shifts and differing recording practices.

%% file: content/3.discussion.tex
\section{Discussion}

The proposed benchmark for systematic evaluation of structured EHR FMs across diverse tasks and datasets surfaces critical shortcomings of state-of-the-art structured EHR FMs. 
While FMs present important benefits in low-label regimes and improve predictive fairness, they fail to consistently outperform task-specific baselines and do not reliably transport across institutions.

Across both linear probing and fine-tuning, FMs fail to consistently outperform task-specific baselines. This finding challenges the implicit assumption that data-scale and the pretraining paradigm directly translate into improved performance. 
While FMs improve discriminative performance when post-trained with limited data, they offer limited benefit when more data is available. Importantly, when linear probing falls short, fine-tuning does not bridge this gap, despite substantially higher computational costs. 
These results underline that 
representations learned using existing pretraining strategies for structured EHRs may not always learn embeddings with discriminative signals necessary for meaningful downstream tasks. In particular, low-prevalence conditions may not be captured by such a representation and consequently, the embeddings result in underperformance in this setting.

The top-performing FM presents lower calibration than task-specific baselines under limited labeled data. 
While discrimination is important for prioritization in clinical practice, practitioners often require risk quantification to inform medical recommendations. The lower calibration of FMs compared with traditional baselines suggests that this paradigm may yield biased risk estimates.
This phenomenon may result from non-calibrated pretraining losses. 
Optimizing for representativeness may fail to capture the true prevalence of clinical conditions at the cost of calibration. 

A critical dimension of our benchmark is to extend traditional evaluation to capture potential group disparities. 
FMs improve fairness relative to task-specific baselines, particularly in healthcare utilization. 
Pretraining on large, diverse patient populations results in representations that generalize across groups. 
However, this finding should be interpreted cautiously. 
Our analysis focuses on a limited set of groups with high missingness rates in recorded identifiers. 
Further, fairness for one group does not guarantee similar properties across other clinically relevant axes. 
The observed fairness gains are therefore promising but not sufficient to ensure equitable deployment across diverse patient populations.

FMs do not transport reliably across institutions. The application of \mamba model, pretrained in \stanfordehr, and tested in \cumc consistently underperformed the locally pretrained model. The promises of the FM paradigm may fall short when applied to hospitals with different populations and clinical and recording practices. 
Beyond differences in healthcare utilization, event frequency, and data density (see Appendix~\ref{app:datasets}), data harmonization and preprocessing practices across institutions remain a fundamental barrier to ensuring that the benefits of these models reach under-resourced institutions and the populations they serve.

Across datasets, \motor, which explicitly accounts for the temporal structure of EHR data, shows a consistent performance advantage. This points to temporal modeling as a promising architectural direction, though the current proliferation of FMs makes it difficult to isolate which design choices drive gains. Systematic benchmarking of each design choice can fill this gap. The proposed evaluation would therefore be a step towards the principled design of future FMs.

When interpreting these results, note that the benchmark has the following limitations.
First, while our benchmark considers a broad range of tasks and metrics that capture critical evaluation desiderata in healthcare, this work does not consider time-to-event or regression tasks, and is intended for retrospective evaluation, not quantifying the impact on downstream clinical decisions~\citep{wornow2023shaky}. Collaboration with domain experts and clinical trials remains crucial to ensure additional validation and safe deployment~\citep{shah2023creation, joshi2025ai}.
Second, our analysis focuses on models available at the time of the study, with a maximum parameter count of $\sim120$ million and a fixed embedding dimensionality due to compute constraints. As new models may present different behavior, we invite the community to contribute to this benchmark. Additionally, model size is often linked to performance and generalization capacity~\citep{kaplan2020scaling}; future work should evaluate the impact of these choices on model performance. 
Third, our evaluation of transportability is limited to a single model and external center, as publicly available FMs are scarce due to privacy concerns.

Through a novel benchmark for the systematic evaluation of structured EHR FMs, our work highlights key directions for future work.
First, the development of future FMs needs to better align with the true underlying occurrence of events. The current lack of calibration suggests that the obtained representations may not capture rare conditions in medical data.  
Enabling full transportability may require further process-level, and practice-level harmonization beyond data representation, so that residual lack of generalizability is attributable to population-level differences. 
The proposed benchmark is made publicly available and is designed to accommodate new modalities, datasets, tasks, and models. We invite the community to build on it as a shared reference point for measuring progress in EHR FM research, allowing researchers to make informed design choices and practitioners to deploy the most relevant tools for their clinical problem.

%% file: content/4.methodology.tex
\section{Methodology}
\label{sec:benchmarking}
We benchmark $6$ state-of-the-art structured EHR FMs on \cumc and \mimic MEDS-transformed data. 
This study was approved under IRB-AAAV2068 at Columbia University. 

\subsection{Preprocessing}
\subsubsection{Columbia University Irving Medical Center (CUMC)}
{\bf{Data.}} \cumc is a large urban academic medical center in New York City. \cumc acts as a quaternary care center capable of providing highly specialized care for individuals in the greater New York metropolitan area. The data, therefore, encode a mixture of specialist care (inpatient and outpatient), primary care, and emergency services over a diverse population.
While race is sparsely recorded (resp. $61.7\%$ and $75.1\%$ missing), the collected data reflect a diverse population with $7.3\%$ of all patients identifying as non-white. Further, the database presents $55.7\%$ women and $44.0\%$ men. Additional demographic information can be found in Table \ref{tab:demographics}.

For all patients in \cumc, the database encompasses patient demographics, visit details for inpatient and outpatient care, conditions (billing diagnoses and problem lists), medications (outpatient prescriptions,  inpatient medication orders, and administrations), medical devices, clinical measurements (such as laboratory tests and vital signs), and other clinical observations like symptoms. Our analysis uses longitudinal health records for $\sim6.7$ million patients spanning $1986$ to $2023$, focusing on prescriptions, procedure events, lab events, patient visits, and setting information. All data have been de-identified and standardized according to the Observational Medical Outcomes Partnership Common Data Model (OMOP CDM).

{\bf{Standardization of measurement units.}} Similar measurements can be recorded differently in EHRs. 
We standardize laboratory measurement values and units. We identify the most common unit for each lab and convert, where possible, the unit and associated lab value to this unit.

{\bf{Standardization of diagnosis codes.}} We convert ICD-9 codes to ICD-10 using the General Equivalence Mappings (GEM) provided by the Centers for Medicare and Medicaid Services (CMS)\footnote{\url{https://www.cms.gov/medicare/coding/icd10/downloads/icd-10_gem_fact_sheet.pdf}}. When multiple ICD-10 codes correspond to a given ICD-9 code, we map to the most frequent ICD-10 code. 
If the mapped ICD-10 code is not present in our dataset, we map to the parent ICD code capturing a more general condition. 
Any ICD-9 codes absent from the GEM database are dropped.

{\bf{Handling missingness.}} For laboratory tests with a concept code but no associated numerical lab value ($14.3\%$ of labs) or no unit ($5.3\%$ of labs), we keep the laboratory event but indicate that the lab value is missing. Numerical values without associated ICD codes ($6.23\%$) are dropped.

{\bf{MEDS Conversion.}}
We use MEDS-ETL\footnote{Version 0.3.9 - \url{https://github.com/Medical-Event-Data-Standard/meds_etl/tree/0.3.9}} to convert data from OMOP to MEDS, which represents data as a patient id, time stamp, concept code, and associated value (see Appendix~\ref{apd:preprocessing}).

\subsubsection{MIMIC-IV}
We also benchmark all models on MIMIC-IV, a public Intensive Care Unit (ICU) dataset containing approximately 364k hospital admissions from Beth Israel Deaconess Medical Center between 2008 and 2022 \citep{johnson2023mimic}. We preserve the original mixed coding system (eg: ICD-9, ICD-10, etc) and follow the previous MEDS-ETL pipeline to convert raw data into MEDS formats.

\subsection{Downstream task definitions}
Downstream evaluations of FMs should reflect the diversity of downstream applications. In healthcare, these span operational tasks associated with patient outcomes, such as mortality as prevalent in prior benchmarking efforts~\citep{wornow2024ehrshot,wornowcontext, chen2023multimodal,mcdermott2024event, doi:10.1056/AIra2501253}. Despite their prevalence, definitions of these outcomes and cohorts have varied widely across publications, prohibiting potential comparison and limiting reproducibility and transportability~\citep{doi:10.1056/AIra2501253, mcdermott2021reproducibility}. Our work contributes reproducible task definitions, leveraging the OHDSI cohort builder. 
In addition to these tasks, EHR models are frequently used to phenotype both acute and chronic conditions that require timely interventions. 
We propose $11$ phenotypes across diverse clinical settings and reflective of chronic and acute disease courses~\citep{linden2022prevalence,blecker2021hospitalizations}. Across all tasks, crucially, we maintain any class imbalance present in the true data --- rather than downsampling negative instances --- to ensure downstream clinical utility.

{\textbf{In-hospitalization mortality (\texttt{Death}).}} The cohort includes patients with at least one hospitalization lasting longer than $48$ hours, using the admission time as the reference. The prediction is made at 48 hours after admission, referred to as \texttt{prediction-time}. Patients who die during the same hospitalization are labeled positive, while those who are discharged alive are labeled as negative. 
 As in other tasks, we require that all patients have at least $2$ years of observation prior to the prediction time (see Figure~\ref{fig:phenotype} in Appendix~\ref{appendix:phenotype_definitions}).

{\bf{$\mathbf{30}$-day readmission (\texttt{Readmission}).}} 
We predict a $30$-day all-cause readmission following hospital discharge. The \texttt{prediction-time} is defined as the discharge time, and any readmissions occurring on the same day are excluded. All patients in the cohort must have at least $2$ years of observation prior to the prediction time and must not be censored within $30$ days following discharge.

{\bf{Prolonged length-of-stay (\los).}}
We predict whether a hospitalization will last longer than seven days, with the prediction made $48$ hours after admission. Patients are required to have at least two years of observation prior to the \texttt{prediction-time}.

{\bf{Phenotyping.}} We defined $11$ phenotypes to assess model performance. Rather than relying solely on ICD codes to define tasks, we established at-risk and case cohorts for each phenotype using clinically validated phenotyping definitions. We introduce phenotype-specific rules about temporality to distinguish between the chronic prediction case (e.g., predicting disease onset) and the acute prediction case (e.g., predicting a future adverse event) to ensure clinical utility. Our phenotypes span major health events such as cancer, autoimmune conditions, and chronic conditions such as type-2 diabetes. A detailed definition of these phenotypes is provided in Appendix \ref{apd:phenotypes}. Our use of an \emph{at-risk} cohort, rather than a general population cohort, for disease prediction increases task difficulty and clinical utility. All patients with no data during the observation period were excluded to ensure that all models were trained and evaluated on the same set of patients. For \mimic, as all patients are admitted through the emergency department or ICU, we focus on predicting acute conditions. We select two acute conditions (AMI and Stroke) and one chronic condition (MASLD) for \mimic.

\subsection{Structured EHR FMs}
We select FMs with promising reported performance that cover a diverse range of pre-training objectives, tasks, and tokenization strategies used in the literature, as summarized in Table~\ref{tab:considered_models}. 
The following describes the specific design choices made by these models and the adjustments enforced to ensure that all models have the same embedding dimensionality ($768$) and a similar parameter count ($120 \pm 5\%$ million). Detailed information on any model-specific preprocessing is provided in Appendix~\ref{appendix:model_details} and on how temporal information is integrated in Appendix~\ref{apd:temporal_encoding}. Context lengths were chosen as the smallest power of $2$ that covers at least $99\%$ of the visits observed in the dataset, as illustrated in Appendix~\ref{apd:context_lengths_choice}.

\begin{table}[ht]
    \footnotesize
    \centering
    \setlength{\tabcolsep}{4pt}
    \caption{EHR foundation models evaluated in this work.}
    \label{tab:considered_models}
    \begin{tabular}{ccccccc}
    \toprule
     \textbf{Design Choices} & \textbf{\makecell{\textsc{Cehr}\\\bert}} & \textbf{\makecell{\textsc{Cehr}\\\gpt}} & \textbf{\makecell{\textsc{Core}\\\behrt}} & \textbf{\textsc{Llama}} & \textbf{\textsc{Mamba}} & \textbf{\motor} \\ \cmidrule{1-7}
     Architecture & \bert & \gpt & \bert & \textsc{Llama} & \textsc{Mamba} & \makecell{Custom} \\
     \makecell{Context \\ length} & 2048 & 2048 & 2048 & 8192 & 8192 & 8192 \\
     \makecell{Learning \\ objective} & MLM & NTP & MLM & NTP & NTP & TTE \\
     \makecell{Parameter count \\ (in Millions)} & 114 & 119 & 119 & 124 & 120 & 117 \\
     Tokenization & Concept & Concept & Concept & \makecell{Concept \\ \& Value} & \makecell{Concept \\ \& Value} & \makecell{Concept \\ \& Ancestor} \\
     
     \makecell{Temporal \\ encoding} & \makecell{Age, \\ Timestamp, \\ Inter-event, \\ Position} & \makecell{Age, \\ Inter-event} & \makecell{Age, \\ Timestamp, \\ Position}  & Position & None & \makecell{Age, \\ Timestamp} \\
    \makecell{Data \\ Source} & \makecell{Demo., visits,\\ diagnoses,\\ procedures,\\meds} &\makecell{Demo., visits,\\ diagnoses,\\ procedures\\ meds} & \makecell{Demo., visits,\\ diagnoses \\ meds}  & All & All & All \\
     \bottomrule
    \end{tabular}
    \vspace{2pt}
    \newline
    MLM: Masked language modeling. NTP: Next token prediction. TTE: Time to event.\\ Inter-event: time difference between consecutive events that occur on different days.
\end{table}

{\bf{\cehrbert~\citep{pang2021cehr}.}}
\cehrbert (Chronological Electronic Health Record \bert) aims to leverage EHR temporal information. To this end, \cehrbert introduced Artificial Time Tokens (ATTs), inserted between neighboring events to represent the time intervals to capture fine-grained temporal differences more effectively. 
\cehrbert combines concept embeddings with both absolute and relative time embeddings, which reflect the patient's age at the time of the event. \cehrbert focuses on structured EHR inputs, specifically using demographic, visit, diagnosis, procedure, and medication data to model temporal clinical trajectories. 

{\bf{\cehrgpt}~\citep{pang2024cehr, pang2025cehrxgptscalablemultitaskfoundation}.} 
\cehrgpt is a \gpt-2 model~\citep{radford2019language} trained on the patient representation introduced by \cehrbert, to enable generating synthetic structured EHR data.  
\cehrgpt differs from the previous model through: (i) demographic tokens placed at the start of each patient sequence, (ii) day tokens inserted between neighboring visits to represent time intervals in days, and (iii) day tokens placed within inpatient visits to preserve the duration of hospital stays, to capture patient timelines. \cehrgpt only operates on a subset of OMOP domains, including demographics, visits, diagnoses, procedures, and medications.

{\bf{\corebehrt~\citep{odgaard2024core}.}}
\corebehrt is a \bert-based model for early-diagnosis prediction, which stores patient trajectories as \emph{sentences} with separator tokens to denote the time elapsed between visits. 
\corebehrt incorporates medication, in addition to diagnosis information used in its precursor model \behrt~\citep{li2020behrt}. However, it does not use information about procedures or laboratory tests. 
Unlike \cehrbert, which uses Time2Vec\citep{kazemi_time2vec_2019} to encode temporality, \corebehrt incorporates temporality via patient age.

{\bf{\llama~\citep{grattafiori2024llama}.}}
\llama is a general-purpose FM using a transformer-based architecture analogous to \gpt~\citep{radford2018improving} but differs in the choice of normalization layers. Additionally, \llama does not incorporate time but uses rotary positional embeddings (RoPE) in the attention mechanism to encode the order of medical events. It represents a state-of-the-art transformer architecture. 

{\bf{\mamba~\citep{gu2023mamba}.}}
\mamba~is a selective state-space model that replaces attention blocks with state-space layers. It can be interpreted as a generalized version of gated Recurrent Neural Networks that capture continuous time-varying inputs. This architecture reduces computational time complexity to linear time compared to transformers and has been shown to better handle long sequences with temporal dependencies, a common occurrence in EHRs~\cite{wornowcontext}. 

{\bf{\motor~\citep{steinberg2024motor}.}} 
 \motor~is a Transformer model pre-trained using a piecewise exponential time-to-event (TTE) pretraining objective. Unlike other architectures that rely on traditional pre-training tasks such as masked language modeling (MLM) or next-token prediction (NTP), \motor~employs TTE predictions to predict around $8,000$ medical codes. This training strategy enables \motor~to leverage a patient’s historical context to predict future events across varying time intervals. 

\subsection{Baselines}
We compare the previous methodologies with two featurization strategies, followed by traditional supervised training. These baselines comprise tabular EHR benchmarks.

{\bf{\femr \cite{steinberg2020languagemodelseffectivepatient}}} includes normalized age and event counts between the patient’s first recorded event and the prediction time. These features are then used to train Logistic Regression and LightGBM \cite{10.5555/3294996.3295074}. 

{\bf{\medstab \citep{oufattole2024medstabautomatedtabularizationbaseline}}} extends the previous set of features using customizable time windows with multiple aggregation functions, such as count, value, min, and max. The extracted features are then used to train a Logistic Regression and an XGBoost model~\cite{Chen_2016}.

For each task, we report the best-performing model among these four baselines as the ‘Best Baseline’. Detailed hyperparameter grids are deferred to Appendix~\ref{appendix:baselines}.

\subsection{Training}
\label{sec:empirical_setting}
We adopt a local pre-training approach for all FMs, followed by post-training. To avoid data leakage, our evaluation relies on a fixed patient split, in which 60\% of patients are used for training in \cumc (80\% for \mimic), 10\% for hyperparameter tuning during pre-training and post-training, and the rest for the proposed evaluation. As our evaluation tasks range from $5,000$ to over $2$ million records, we cap the training set used for linear probing at $100,000$ records and the evaluation set at $50,000$ records per task. 

\subsubsection{Pre-training}

{\bf{Local.}}
We pretrain all models \textit{from scratch} independently on \cumc and \mimic data to generate patient embeddings. To this end, we use the same pre-processing, tokenization, hyperparameters, and training objectives as in the original works. To ensure the same convergence levels for all models despite model-specific losses and learning rates, we use a 0.1-1\% change in relative loss as a stopping criterion. 

{\bf{External.}} We use open weights from a \textsc{Mamba} model pretrained on \stanfordehr, which we refer to as {\mambatransport}, on downstream tasks on our EHR data without local fine-tuning. 
 Although \motor also provides open weights, we could not evaluate its transportability due to insufficient information on lab measurement units in the original pre-training datasets.

\subsubsection{Linear probing} 
Linear probing consists of fitting a logistic regression model using representations extracted from frozen FMs~\citep{marks2025closer}. This common approach is \begin{inparaenum} \item[(i)] less computationally expensive and data-hungry than further fine-tuning \item[(ii)] enables evaluating pretraining design choices on downstream performance.\end{inparaenum}~To ensure optimal linear probing performance, we employ a 5-fold cross-validation to select the $\ell_2$-penalty \(\lambda\) on a logarithmic scale between \(10^{-4}\) and \(10^{4}\). 

\subsubsection{Fine-tuning} 
Fine-tuning consists of further training a FM using labeled data~\citep{bommasani2021opportunities}. As these models are often trained for next-token prediction, one must first add a classification head with a final Softmax based on the intermediary representation. Following the linear probing approach, a single classification layer is appended before training the model. However, unlike linear probing, the classification loss is backpropagated through both the classification network and the base FM. As fine-tuning is sensitive to hyperparameters, we perform a grid search over the fine-tuning learning rate (set to \{0.05, 0.1, 0.2, 0.3, 0.5\} of the pretraining rate) and the ratio of the classification head learning rate to the fine-tuning rate, drawn from \{1, 2, 5\}, reflecting the goal of slowing the base model's updates relative to the classification head. We fix the scheduler to linear decay with a one-epoch warmup, selected via a preliminary search on a single model, and train for up to 10 epochs with early stopping (patience 3 on validation log-loss), choosing the final configuration by validation log-loss.

\subsection{Evaluation}
For all metrics, confidence intervals were obtained via test-set bootstrapping with $100$ iterations. Bootstrap is performed at the observation level rather than the patient level, as the model is intended to support repeated predictions. Estimated confidence intervals therefore reflect the uncertainty associated with repeated-measurement settings. Note that this approach may underestimate the uncertainty associated with performance on entirely new patients, as it does not account for between-patient variability.

{\bf{Performance metrics.}} As both discrimination and calibration are critical to medical applications, our benchmark quantifies discrimination using Area Under the Receiver Operator Curve (AUROC) and calibration using Expected Calibration Error (ECE) and Brier Score (Brier). 

{\bf{Fairness metrics.}} As medical datasets can reflect historical socio-medical biases~\citep{chen2020ethical}, our work studies the risk of reinforcing such inequities by quantifying group fairness~\citep{barocas2023fairness}. Specifically, we measure the previous metrics stratified by reported sex, race (Black, White, Other), and healthcare utilization\footnote{We represent healthcare utilization as empirical tertiles of the number of medical encounters a patient has in a year, defined as any day with a recorded visit, condition, procedure, observation, or laboratory test.}, and report the associated maximal absolute difference across groups, denoted as $\Delta := \max_g |d_g - d_{\neg g}|$, where $d$ is a performance metric and $g$ is a given group.

{\bf{Ranking.}} Performance was computed for each task. Rankings were then computed for each task, averaged across socio-medical groups for fairness gaps, and then averaged across tasks.

{\bf{Inference computational cost.}} Due to resource constraints and the response time requirement associated with clinical deployment, inference time and costs are critical for usability. We report inference FLOPs as a proxy for cost (see Appendix~\ref{apd:resources} for compute resources used in this evaluation).

%% file: content/appendix.tex
\section{Existing EHR foundation models}\label{apd:lit_review}

Table~\ref{tab:review:ehr_foundations} describes existing EHR FMs that have been proposed in the literature, their associated characteristics, and evaluation metrics.
\begin{table}[htbp]
    \footnotesize
    \setlength{\tabcolsep}{1pt} 
    \setlength\extrarowheight{3pt}
    \caption{Structured EHR foundation models, pre-training/training datasets, tokenization strategy, and evaluation measures (grouped by training data, chronologically ordered).}
    \label{tab:review:ehr_foundations}
    \centering
    \begin{tabular}{@{} c c c c c c c c c @{}}
        \toprule
        \textbf{Foundation} & \textbf{Dataset} & \multicolumn{3}{c}{\textbf{Tokenization}} & \textbf{Pretrain Loss} & \multicolumn{3}{c}{\textbf{Evaluation}} \\
        \cmidrule(lr){3-5} 
        \cmidrule(lr){7-9}
        \textbf{Model} & & \textbf{Time} & \textbf{Concept} & \textbf{Value} & & \textbf{Disc.} & \textbf{Cal.} & \textbf{Fair.} \\
        \midrule
        
        \textsc{Behrt}~\citep{li2020behrt} & \makecell{Clinical Practice\\Research Datalink} & \makecell{PE \& Age-\\Embedding} & Code-based & None & MLM & \checkmark & & \\
        
        Med-\textsc{BERT}~\citep{rasmy2021med} & \makecell{Cerner Health\\Facts} & \makecell{PE \& Age-\\Embedding} & Code-based & None & MLM &\checkmark & & \\

        \textsc{Clmbr} ~\citep{steinberg2021language}  & \makecell{\textsc{Ehrshot}} & \makecell{RoPE} & Code-based & None & NTP & \checkmark & & \\
        Med\textsc{GPT}~\citep{kraljevic2021medGPT}& \makecell{London Hospital\\+ MIMIC-III} & RoPE & \makecell{Text \& Code \\based} & None & NTP & \checkmark & & \\
        
        \cehrbert~\citep{pang2021cehr} & CUMC-NYP & Time2vec+ATT
        & Code-based & None & MLM & \checkmark & \checkmark & \\
        
        GatorTron~\citep{yang2022large} & \makecell{UF Health,\\Pubmed,\\Wikipedia\\+ MIMIC III}& None & Text-based & None & MLM \& SOP & \checkmark & & \\
        
        GenHPF~\citep{hur2023genhpf} & \makecell{MIMIC-III, IV\\+ eICU} & ALiBi & Text-based & 
        \makecell{Digit Place\\Embedding} & SSL & \checkmark  &  & \\

        Foresight~\citep{kraljevic2024foresight} & \makecell{London Hospital\\+ MIMIC-III} & Sinusoidal & \makecell{Text \& Code \\based} & None & NTP &\checkmark & & \\

        \motor~\citep{steinberg2024motor} & \makecell{\textsc{Stanford-Ehr}\\+ MERATIVE} & RoPE & Code-based & None & TTE & \checkmark & \checkmark & \checkmark  \\
        
        TransformEHR~\citep{yang2023transformehr} & VA Bedford & Sinusoidal & Code-based & None & Visit MLM& \checkmark & & \checkmark  \\
        
        \textsc{Ebcl}~\citep{jeong2023event} & MIMIC-III  & Custom & Code-based & CLIP & EBCL &\checkmark & & \\

        \cehrgpt~\citep{pang2024cehr} & CUMC-NYP & ATT
        & Code-based & None & NTP & &  & \\

        \corebehrt~\citep{odgaard2024core} & \makecell{Capital Region\\of Denmark} & RoPE & Code-based & None & MLM & \checkmark & & \\
        \textsc{EhrMamba}~\citep{fallahpour2024ehrmamba} & MIMIC-IV& Time2vec & Code-based & None & NTP & \checkmark & & \\
        
        \textsc{Ethos}~\citep{renc2024zero} & MIMIC-IV& \makecell{Age quantiles\\ + ATT} & Code-based & Quantiles & NTP & \checkmark & & \\

        \makecell{Long-context \\ \llama,\mamba ~\citep{wornowcontext}}  & \makecell{\textsc{Transport}} & \makecell{RoPE} & Code-based & None & NTP & \checkmark & & \\
        \bottomrule
    \end{tabular}
    \vspace{4pt}
    \parbox{\textwidth}{\footnotesize \textit{Notes:} Disc.: Discrimination; Cal.: Calibration. Fair.:Fairness. Checkmarks ($\checkmark$) indicate the aspect was evaluated in the paper. PE: Positional embedding. ATT: Artificial Time Tokens. MLM: Masked language modeling. NTP: Next token prediction. SOP: Sentence order prediction. SSL: Self-supervised learning. TTE: Time to event. EBCL: Event-based contrastive learning}
\end{table}

\newpage
\section{Phenotype Definitions}\label{appendix:phenotype_definitions}
For most phenotypes, we used the Observational Health Data Sciences and Informatics (OHDSI) Phenotype Library\footnote{\url{https://ohdsi.github.io/PhenotypeLibrary/}} to generate the at-risk and case cohorts. The phenotype library is a publicly accessible and version-controlled catalog of phenotypes generated by members of the OHDSI community. We evaluated our phenotype algorithms using OHDSI PheValuator~\citep{swerdel2019phevaluator}. All phenotypes are available in the GitHub repository as JSON files and are fully reproducible on any dataset that employs the OMOP-CDM.

The at-risk cohorts were defined primarily through three methods: (i) identifying risk factors for the condition through UpToDate\footnote{\url{https://www.uptodate.com/contents/search}}, an evidence-based clinical resource that provides point-of-care medical information summaries about a wide variety of medical topics; (ii) identifying non-descendant diagnostic codes linguistically related to the condition of interest through GloVe embeddings ~\citep{pennington2014glove,lee2021glovephenotyping}; and (iii) leveraging the ICD hierarchy to identify subchapters of codes related to the condition of interest. 

\begin{figure}[!ht]
    \centering
    \includegraphics[width=0.6\textwidth]{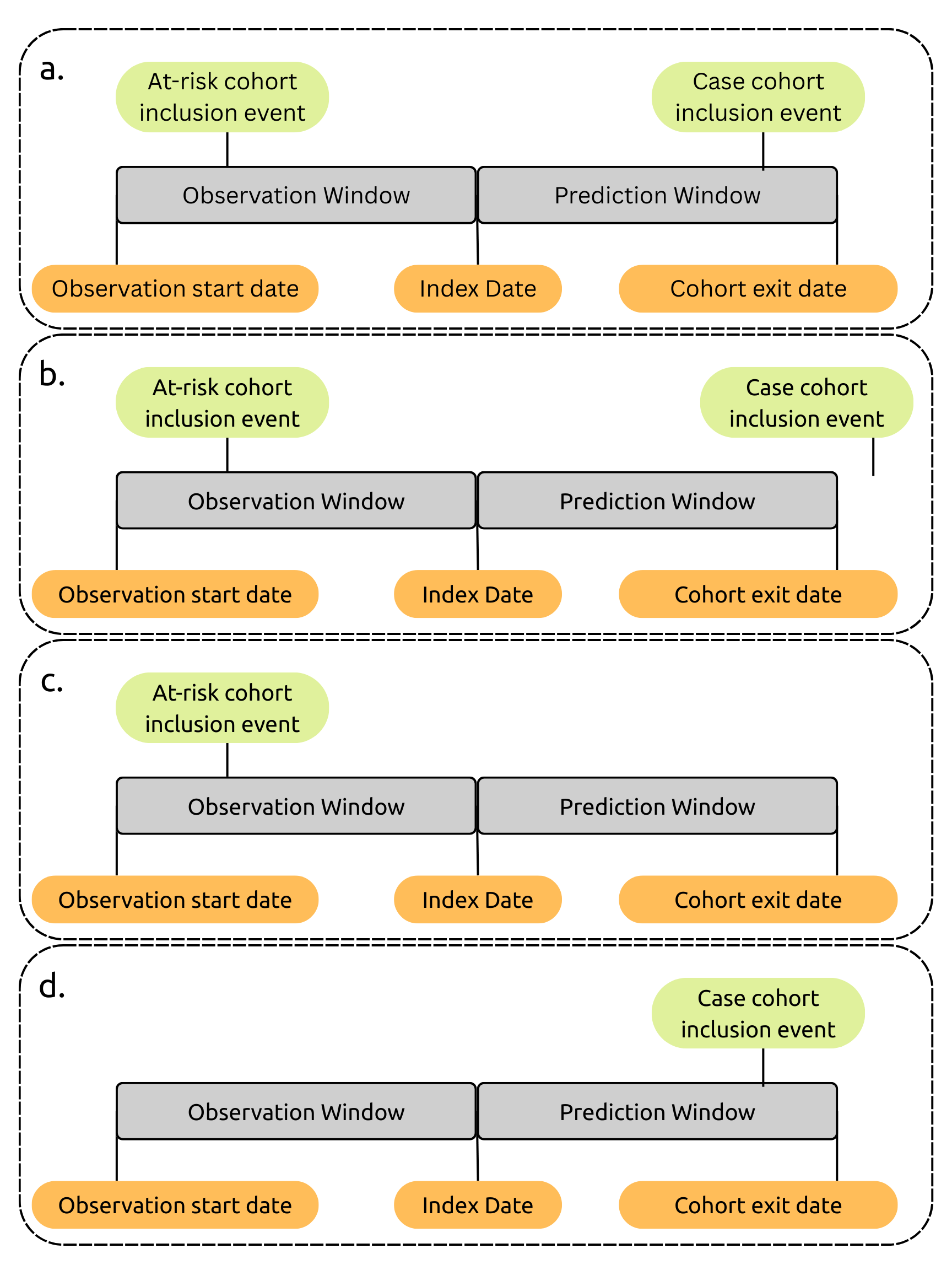}
    \caption{The proposed phenotypes are structured using consistent temporal components: (observation entry, index, and cohort exit) and clear inclusion criteria for the at-risk and case cohorts.
    (a) Example of an individual included in the case cohort, as their case-cohort inclusion event occurs during the prediction window. 
    (b-c) Examples of individuals included in the at-risk, but not the case cohort, since they have an at-risk inclusion event but do not have a case cohort inclusion event during the prediction window. 
    (d) Example of an individual not included in the at-risk or case-cohorts, as they are missing an at-risk inclusion event.}
    \label{fig:phenotype}
\end{figure}

Each phenotype has three major temporal components: cohort entry date, index date, and cohort exit date (Figure~\ref{fig:phenotype}). The observation period is defined as the time between the cohort entry date and the index date. The model uses data in the observation period to predict the outcome. The time from the end of the index date to the cohort exit date comprises the prediction period, where the goal of the model is to predict whether or not the target diagnosis occurs in this period. 

For all disease phenotypes, the observation start date corresponds with the patient's first entry in the database. The patient enters the at-risk cohort when the inclusion criteria for being at-risk are met, and the entrance event to the at-risk cohort is named "at-risk cohort inclusion event". Index event, i.e., the patient encounter when the model predicts whether the patient belongs in the case or control group for the phenotype at that time, will always be defined after the patient enters the at-risk cohort, i.e., after the at-risk cohort inclusion event. The prediction window is 1 year for all phenotypes. The exact specifications of all phenotypes, including specific concept codes, are stored in JSON files in the GitHub repository. 

\subsection{Celiac Disease}\label{apd:phenotypes}
Celiac is a chronic disease. The inclusion event for the at-risk cohort can be a family history of celiac \citep{UptoDateCeliac}, or a related code as identified by GloVE~\cite{pennington2014glove} embeddings. Inclusion in the case cohort is based on a celiac disease phenotype created as part of the \href{https://data.ohdsi.org/HERACharacterization/}{HERA characterization study} and requires a diagnosis code of celiac disease (SNOMED code 396331005, ICD-10 K90.0, or ICD-9 579.0).

Patients who satisfy the inclusion criteria for the case cohort prior to the at-risk cohort are not included in the phenotyping task.

\subsection{Acute Myocardial Infarction (AMI)}
AMI, also known as a heart attack, is an acute condition. The inclusion event for the at-risk cohort is one of the AMI risk factors from \citep{UptoDateAMI}, a related code identified by GloVe embeddings, or a diagnosis from the ``Ischemic heart diseases'' chapter (ICD10: I20-I25). The inclusion event for the AMI case cohort is an appropriate diagnostic code (SNOMED: 22298006) during an emergency room or inpatient visit. We specifically omit the SNOMED code 1755008 (prior myocardial infarction) from the inclusion event. Additionally, we require a 180-day washout period between AMI diagnoses to ensure that each diagnosis refers to a separate event. This phenotype was also used in~\citep{Suchard2019}. 

Patients who satisfy the inclusion criteria for the case cohort less than 7 days prior to entry of the at-risk cohort are not included in the phenotyping task.

\subsection{Systemic Lupus Erythematosus (SLE)}
SLE, or lupus, is a chronic disease. We derive the phenotype for SLE from the OHDSI-generated phenotype, which has been previously validated via chart review \citep{swerdel_ramcharran_hardin_2023}. The entry event for the at-risk cohort includes: differential diagnoses for SLE \citep{UptoDateSLE}, related codes generated by GloVe embeddings, and any code within the "Systemic connective tissue disorders" subchapter of ICD10 (M30-M36). Additionally, the validated phenotype for SLE includes a set of concepts titled "Signs and symptoms suggestive of SLE"; any of these condition occurrences can also constitute an entry event for the at-risk cohort. The entry event for the case cohort starts with at least one condition occurrence from the predefined concept set "Signs and symptoms suggestive of SLE" or a drug exposure from the concept set “SLE treatments”. In both cases, this initial event must be followed by a diagnosis from the SLE concept set within 90 days. Due to its chronic nature, only the first diagnosis of SLE is considered during phenotyping. 

Patients who satisfy the inclusion criteria for the case cohort prior to the at-risk cohort are not included in the phenotyping task.

\subsection{Pancreatic Cancer}
Pancreatic cancer is also a chronic disease. The at-risk cohort entry event is defined as a common differential diagnosis and symptoms for pancreatic cancer \citep{UptoDatePC}, a related diagnosis code derived from GloVe embeddings, or a diagnosis code from the "Malignant neoplasms of digestive organs" subchapter (ICD10: C15-C26). The inclusion event for the case cohort is a diagnosis of "Neoplasm of pancreas" (SNOMED: 126859007) with no diagnoses of "Benign neoplasm of pancreas" (SNOMED: 92264007) and "Benign tumor of exocrine pancreas" (SNOMED: 271956003). We defined our pancreatic cancer cohort following the "Phenotyping and Validation of Cancer Diagnoses" \citep{PancreaticCancerPresentation}.

Patients who satisfy the inclusion criteria for the case cohort prior to the at-risk cohort will not be included in the phenotyping task.

\subsection{Hypertension (HTN)}

Hypertension is also a chronic disease. The entry event for the at-risk cohort includes risk factors identified in \citep{UptoDateHTNEpi, UptoDateHTNAdult} and condition codes from the "Hypertensive diseases" ICD-10 subchapter (I10-I1A). No GloVe embeddings were found for hypertensive disorder. The inclusion event for the case cohort is the first diagnosis of hypertensive disorder (SNOMED: 38341003) or its descendants.

Patients who satisfy the inclusion criteria for the case cohort prior to the at-risk cohort will not be included in the phenotyping task.

\subsection{Metabolic Dysfunction-Associated Steatotic Liver Disease (MASLD)}
MASLD is a chronic disease. The at-risk event is an occurrence of one of the MASLD risk factors: insulin resistance or Type 2 diabetes, metabolic syndrome,
obesity, dyslipidemia, sarcopenia, hypertension, hypertriglyceridemia, polycystic ovarian syndrome, chronic kidney disease, hypobetalipoproteinemia, lysosomal acid lipase deficiency, defects in mitochondrial fatty acid oxidation \cite{Huh2022, Chung2025}. Inclusion into the case cohort is based on the PheKB phenotype algorithm\footnote{\url{https://phekb.org/phenotype/non-alcoholic-fatty-liver-disease-nalfd-alcoholic-fatty-liver-disease-ald}}, which consists of ICD9 (571.5, 571.8, 571.9) and ICD10 codes (K75.81, K76.0, K76.9). The diagnosis of hepatic steatosis from imaging, medications, and clinical notes is not considered for simplicity.

\subsection{Ischemic Stroke}
Ischemic stroke is an acute condition. The inclusion event for the at-risk cohort includes the following risk factors: previous stroke, transient ischemic attack, hypertension, high cholesterol, heart disease, diabetes, obesity, and sickle cell disease \cite{CDCStrokeRisk2024}. The inclusion event for the case cohort is an ischemic stroke code (ICD10: I$63^*$) that occurs during an inpatient or emergency room visit \cite{Suchard2019}.

\subsection{Osteoporosis}
Osteoporosis is a chronic disease. The at-risk cohort entry event includes any of the following risk criteria: women aged 65 and older, men aged 70 and older, anyone 50 or older who has had a fracture or with the following risk factors, such as White or Asian women, smokers and heavy drinkers, who weigh less than 125 pounds, who have undergone bariatric surgery, with kidney failure, inflammatory bowel disease, rheumatoid arthritis, liver disease or an eating disorder or who take oral corticosteroids daily or other high-risk medications (e.g., thyroid hormone replacement, immunosuppressant drugs, warfarin) \cite{UCSD_OsteoporosisRisk}. The inclusion event for the case cohort is a diagnosis code for "Osteoporosis" (SNOMED: 80502) or any of its descendants.

\subsection{Chronic Lymphoid Leukemia (CLL)}
Inclusion events for the CLL at-risk cohort consist of any of the following risk factors: age greater than or equal to 50 with a CLL symptom, or with a family history of CLL. CLL symptoms include weight loss, splenomegaly, shortness of breath, loss of appetite, localized enlarged lymph nodes, and fatigue \cite{ACS_CLLRiskFactors}. The inclusion event for the case cohort was generated using the "Chronic lymphoid leukemia" concept set, excluding "Acute lymphoblastic leukemia". The phenotype definition was created as part of the HERA characterization study\footnote{\url{https://data.ohdsi.org/HERACharacterization/}}.

\subsection{Type 2 Diabetes Mellitus (T2DM)}
T2DM is a chronic disease. The inclusion event for the at-risk cohort is one of the following: over 35 years old and overweight or obese, with prediabetes and a family history of diabetes, or with a history of gestational diabetes \cite{NIDDK_Type2DiabetesRiskFactors}. The case cohort was generated from the OHDSI Legend study~\citep{Suchard2021LEGENDT2DM}. The cohort includes patients with a T2DM diagnosis on two separate days, or with one T2DM diagnosis and one antidiabetes drug, or one T2DM diagnosis and two high HbA1C lab tests (test results between 6.5 and 30) at least 7 days before the diagnosis date.

\subsection{Schizophrenia}
Schizophrenia is a chronic disease. Inclusion into the at-risk cohort requires a non-schizophrenia spectrum psychotic disorder. The case cohort requires a subsequent diagnosis of schizophrenia or schizoaffective disorder. Individuals with schizophreniform disorder were excluded from both cohorts, as their ambiguous diagnostic status could constitute data leakage. We further require that the individual be between 10 and 35 years old at the prediction time, as this is the typical age range of schizophrenia onset. All individuals had at least 3 years of observation, with at least one diagnosis or prescription in the last 3 years of observation. The schizophrenia phenotype, including the 3-year observation lookback, was previously validated in~\citep{finnerty2024prevalence}.

As with all chronic diseases, the observation start date is the start of the patient's observed data in the database. The cohort exit date is either a schizophrenia diagnosis or the end of observed time (if the patient does not have schizophrenia. The index date is 90 days prior to the cohort exit date, leaving a 90-day censoring time to avoid temporal data leakage. We use the full patient history (limited only by the model's context length) to predict the diagnostic transition from psychosis to schizophrenia because of the relatively small cohort size: 636 patients with a diagnostic transition from psychosis to schizophrenia and 2,634 with psychosis only. 

\newpage
\section{Datasets overview}
\label{app:datasets}

Table~\ref{tab:demographics} summarizes demographic distributions of \mimic, \cumc, and \stanfordehr data. 
While all models were tested on our dataset, \mamba-\textsc{Transport} was trained on \stanfordehr~and then tested on \cumc to assess the model's transportability. Table~\ref{tab:events_comparison_datasets} summarizes the frequency of all event types in those datasets. 

\begin{table}[!ht]
    \footnotesize
    \centering
    \setlength{\tabcolsep}{4pt}
    \caption{Demographic distribution of \mimic, \cumc and \stanfordehr (EHRSHOT) data.}
    \label{tab:demographics}
    \begin{tabular}{ccccc}
    \toprule
      & & \makecell{\mimic\\~\\
      $\%$ (n)} & \makecell{\cumc\\~ \\
      $\%$ (n)} & \makecell{\stanfordehr\\(EHRSHOT) \\ $\%$ (n)}\\
      \midrule
      & \makecell{\textbf{ Female}}\ & 52.7 (191,984) & 55.7 (3,744,185) & 51.1 (3,441) \\
      \textbf{Sex} & \makecell{\textbf{ Male}}\ & 47.3 (172,643)&44.0 (2,955,940) & 48.9 (3,298) \\
      & \makecell{\textbf{Missing}}\ & 0.0 (0) & 0.3 (16,814) & 0 (0) \\
      \midrule
      & \makecell{\textbf{Hispanic or Latino}}\ & 3.6 (13,303) &8.3 (557,173) & 15.4 (1,038) \\
      \textbf{Ethnicity} & \makecell{\textbf{Not Hispanic or Latino}}\ & 54.1 (197,088) & 16.6 (1,118,232) & 77.7 (5,236) \\
      & \makecell{\textbf{Missing}}\ & 42.3 (154,236) & 75.1 (5,041,534) & 6.9 (465) \\
      \midrule

      & \textbf{Asian} & 2.7 (9,741) &1.2 (83,697) & 15.5 (1,043)\\
      & \makecell{\textbf{Black or African American}} & 7.9 (28,981) & 5.9 (396,677) & 4.4 (298)\\
      & \makecell{\textbf{Native Hawaiian or Other Pacific Islander}} & 0.1 (266) &0.1 (6,534) & 1.1 (74)\\
      \textbf{Race} & \makecell{\textbf{Native American or Alaskan Native}} & 0.1 (502) &0.1 (7,443) & 0.4 (25)\\
      & \textbf{White} & 41.2 (150,073) &16.9 (1,133,364) & 55.4 (3,736)\\
      & \textbf{Other} & 3.5 (12,643) &14.1 
 (944,342) & 0 (0) \\
      & \textbf{Missing} & 44.5 (162,421) &61.7 (4,144,882) & 23.2 (1,563)\\
    
    \bottomrule
    \end{tabular}
\end{table}

\begin{table}[!ht]
    \footnotesize
    \centering
    \setlength{\tabcolsep}{4pt}
    \caption{Means (standard deviation) of the number of events differentiated by data types available in \mimic, \cumc and \stanfordehr (EHRSHOT) data. Note that in \mimic, healthcare utilization is calculated as the number of events per year. For \cumc and \stanfordehr, healthcare utilization is calculated as the number of events that occur \textit{on distinct days} per year. Note that this difference is due to the differing contexts of hospitalization data (\mimic) and more general EHRs.}
    \label{tab:events_comparison_datasets}
    \begin{tabular}{cccc}
    \toprule
      & \mimic & \cumc & \stanfordehr (EHRSHOT) \\
      \cmidrule{1-4}
      \makecell{\textbf{Healthcare utilization}\\(distinct events per year)} &16816.6 (58562.9) & 118.4 (162.1) & 28.6 (59.1) \\
      \cmidrule{1-4}
      \makecell{\textbf{Conditions}\\(occurrences per year)} & 169.2 (414.6) & 98.9 (392.9) & 57.4 (243.8) \\
      \makecell{\textbf{Medications}\\(occurrences per year)} & 1046.2 (3077.3) & 84.3 (508.6) & 133.2 (583.5) \\
      \makecell{\textbf{Procedures}\\(occurrences per year)} & 82.1 (316.9) & 45.8 (222.7) & 85.6 (398.6) \\
      \makecell{\textbf{Lab Tests}\\(occurrences per year)} &14190.4 (52040.3) & 817.0 (4823.0) & 1,717.5 (7,142.5) \\
    \bottomrule
    \end{tabular}
\end{table}

\begin{figure}[!ht]
    \centering
    \includegraphics[width=0.5\textwidth]{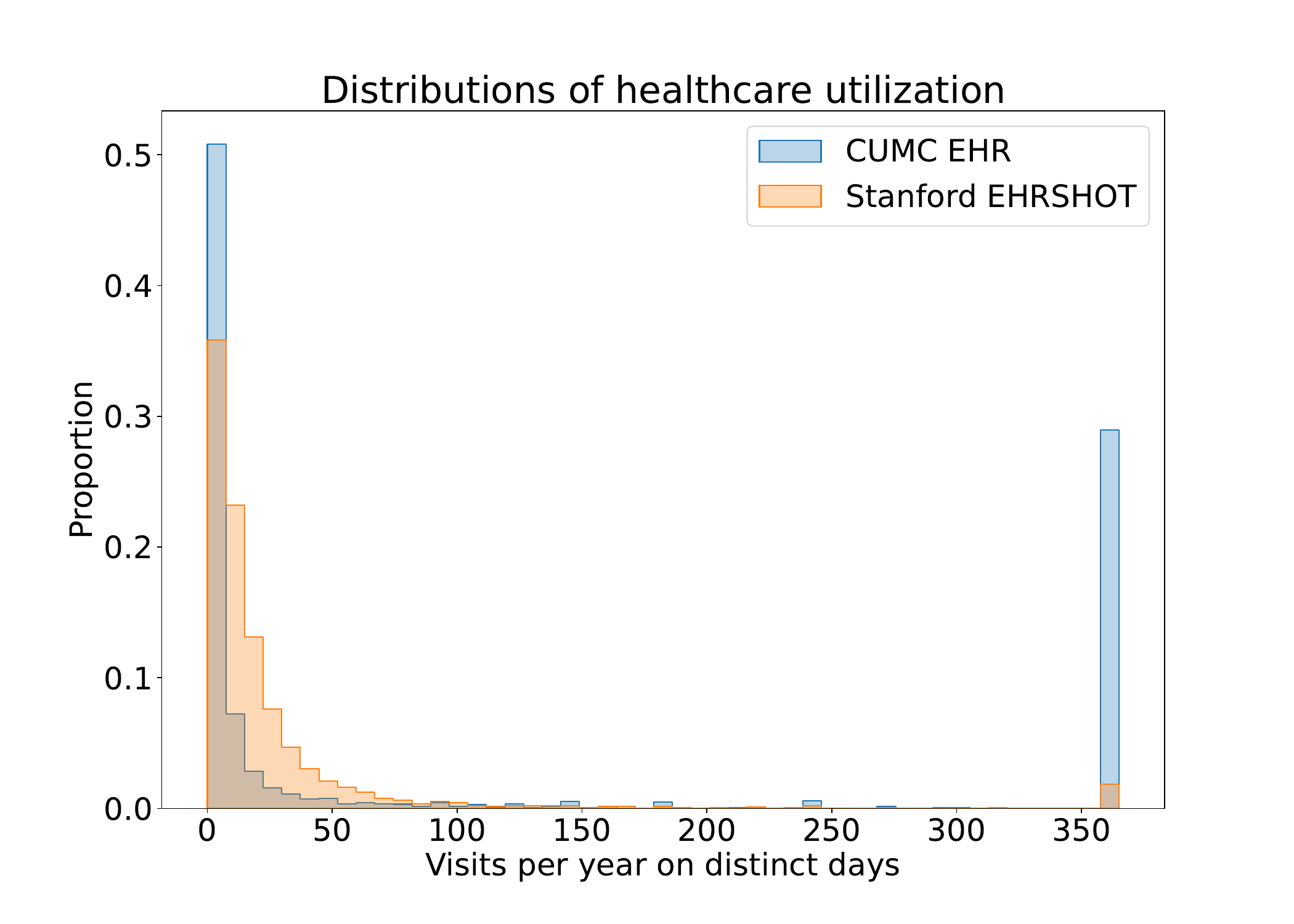}
    \caption{Healthcare utilization across datasets.}
    \label{fig:hcucomparison}
\end{figure}

Figure~\ref{fig:hcucomparison} summarizes the healthcare utilization characteristics on \cumc and \stanfordehr (EHRSHOT), a publicly available subset of the original \stanfordehr~data.
Healthcare utilization was measured as the number of distinct interactions with the healthcare system per year. We consider any event (condition, drug, procedure, measurement, observation, or lab test) as part of a ``distinct interaction'', but constrain it to a maximum of one ``interaction'' per day (e.g., if a person receives two diagnoses and a prescription on the same day, we assume this all comes from the same visit). We divide the number of distinct interactions by the number of years the person is observed within the dataset. We find that, relative to \stanfordehr (EHRSHOT), the \cumc dataset has more patients with either extremely low or extremely high (e.g., every day) healthcare utilization.

\begin{figure}[!ht]
    \centering
    \includegraphics[width=0.8\textwidth]{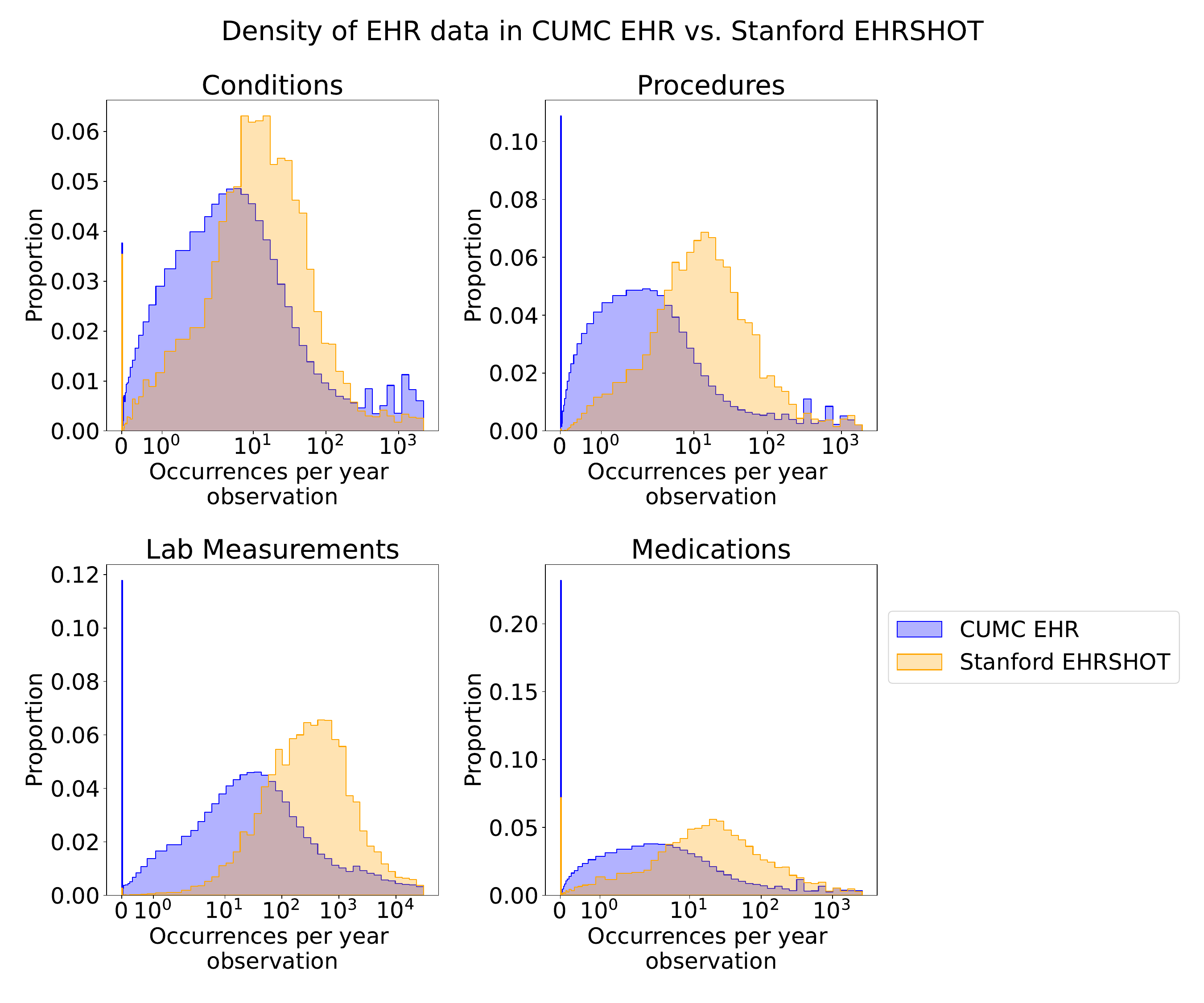}
    \caption{Record type densities across datasets.}
    \label{fig:datadensity_comparison}
\end{figure}

Figure~\ref{fig:datadensity_comparison} presents the density of the different record types.
In order to compare the density of data at each institution, we measure the number of occurrences of each data type (a. conditions, b. medications, c. procedures, d. lab tests) per year of observation for each patient. Unlike with the healthcare utilization metric, we count multiple instances of a data type separately (e.g., all diagnoses made on the same day will contribute to the "density" measurement). We find that the \stanfordehr (EHRSHOT) data is denser than the \cumc data across all measured data types.

Table~\ref{tab:dataset_split_details} summarizes the number of patients and medical events for pretraining, tuning, and test splits. 
\begin{table}[ht]
\centering
\caption{Dataset Split of MIMIC-IV and \cumc}
\label{tab:dataset_split_details}
    \begin{tabular}{ccccc}
        \toprule
        \textbf{Split Name} & \multicolumn{2}{c}{\textbf{MIMIC-IV}} & \multicolumn{2}{c}{\textbf{\cumc}} \\
        \cmidrule(lr){2-3} \cmidrule(lr){4-5}
        & {\# Patients} & {\# Events} & {\# Patients} & {\# Events} \\
        \midrule
        Training Set   & 291,702 & 579,667,440  & 3,996,578 & 1,562,316,866\\
        Tuning Set & 36463  &  71,789,378 & 705,279 & 273,247,565 \\
        Test Set       & 36462  & 71,290,747 & 2,015,082 & 781,462,164 \\
        \bottomrule
    \end{tabular}
\end{table}

\newpage
\section{Models-specific preprocessing and pre-training}~\label{appendix:model_details}
The following describes any model-specific preprocessing and parameters used for the models.

\subsection{MEDS-processing}
\label{apd:preprocessing}
For each individual, medical events are extracted from all relevant OMOP domain tables, including \texttt{person}, \texttt{visit\_occurrence}, \texttt{condition\_occurrence}, \texttt{drug\_exposure}, \texttt{procedure\_occurrence}, \texttt{measurement}, \texttt{device\_exposure}, \texttt{observation}, and \texttt{death}.  Demographic information is placed at the beginning of each patient sequence, followed by clinical events ordered chronologically. For non-numeric records, the \texttt{subject\_id}, \texttt{concept\_code}, and \texttt{timestamp} are extracted into the MEDS event format. For numeric or categorical events, such as measurement records, associated values and units are stored in the \texttt{numeric\_value}, \texttt{text\_value}, and \texttt{unit} fields, respectively.

By design, MEDS-ETL prioritizes source concept codes over standard concept codes during conversion. Standardized concept codes come from vocabularies defined as standard by the OHDSI community. As a result, OMOP uses standard concept IDs (e.g., SNOMED for conditions), whereas MEDS uses source concept IDs (e.g., ICD-9/10). We use the MEDS format for \motor, \femr, \medstab, and \texttt{Context Clues}, as these models were originally developed and extensively evaluated using the MEDS. Although \cehrbert and \cehrgpt support the MEDS data format, the reliance on source concept codes may result in sub-optimal model performance due to the many-to-many mappings between terminologies. Moreover, the original authors of these models explicitly stated that they used standard concept codes extracted from OMOP, not source codes. For fairness, we use OMOP when training and evaluating \cehrbert and \cehrgpt. Patient sequences used for pre-training and evaluation were extracted from OMOP  using the respective original code bases\footnote{\url{https://github.com/knatarajan-lab/CEHRBERT_data}}\citep{pang2021cehr, pang2024cehr}. EHR preprocessing for \corebehrt is also conducted without the MEDS conversion using the custom processing and tokenization pipeline \citep{odgaard2024core}. 

\subsection{Encoding temporal information} \label{apd:temporal_encoding}
The considered FMs integrate temporal information using different encoding mechanisms. Table~\ref{tab:temporal-encoding} details, for each temporal information, the associated encoding strategy.
\begin{table}[!ht]
\centering
\footnotesize
\caption{Temporal Information Input and Encoding Mechanisms Across Models}
\label{tab:temporal-encoding}
\begin{tabular}{ccccc}
\toprule
\textbf{Model} & \textbf{Age} & \textbf{Time Stamp} & \textbf{Inter-event} & \textbf{Position}\\
\midrule
\motor       & Sinusoidal & Time Features & None              & None \\
\cehrbert     & Time2Vec     & Time2Vec               & Time token   & Sinusoidal\\

\cehrgpt      & Age Token    & None                   & Time token   & None \\

\corebehrt    & Time2Vec     & Time2Vec               & None              & Sinusoidal\\

\mamba         & None         & None                   & None              & None \\
\llama         & None         & None                   & None              & Sinusoidal \\
\bottomrule
\end{tabular}
\vspace{2pt}
\newline
Inter-event: time difference between consecutive events that occur on different days. 
\end{table}

\motor incorporates two timestamp features --- absolute time and its square --- by replacing the last two dimensions of the hidden states. It integrates sinusoidal age embeddings at each layer of the \femr decoder using RoPE, allowing age and time information to influence the model throughout its depth.

\cehrbert integrates timestamp, age, and visit positional embeddings through a feedforward network before passing them to the encoder layers. Additionally, it inserts time tokens between consecutive visits and between inpatient events occurring on different days to capture the temporal irregularity of EHR sequences.

\cehrgpt encodes temporal information by augmenting patient sequences with explicit time-related tokens. It introduces a start-year token and an age token at the beginning of each trajectory, and inserts fine-grained day tokens between visits and between inpatient events on different days. These temporal embeddings, corresponding to year, age, and day tokens, are learned directly through the model’s next-token prediction objective.

\corebehrt adds time embeddings, age embeddings, and visit positional embeddings to the concept embeddings before passing them to the \textsc{BERT} encoder layers. Additionally, it applies RoPE positional embeddings to the hidden states at every encoder layer.

\llama does not incorporate age, absolute time, or inter-event temporal information, relying solely on the default RoPE positional embeddings. \mamba similarly does not encode any explicit temporal signals. 

\subsection{Context length selection}
\label{apd:context_lengths_choice}
\begin{figure}[!h]
    \centering
    \includegraphics[width=0.5\textwidth]{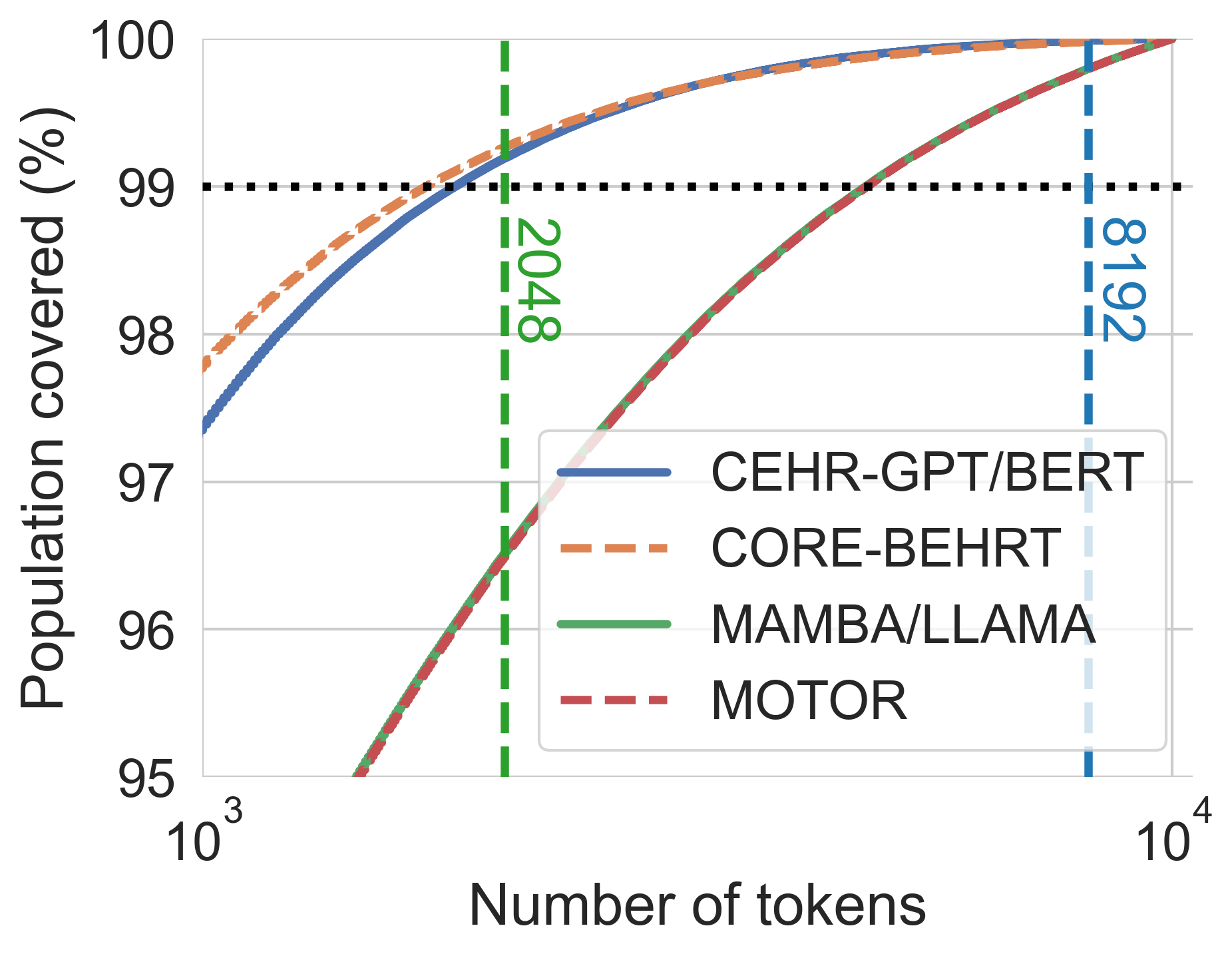}
    \caption{Percentage of the population covered given different context windows.}
    \label{fig:Distributions}
\end{figure}

As tokenization choices result in different lengths of embedded data, we selected each model's context lengths based on the percentage of visits covered by a given choice. Figure~\ref{fig:Distributions} illustrates the frequency of visits for each number of tokens. 
Using these figures, we selected the context lengths presented in Table~\ref{tab:considered_models} to cover at least 99\% of the visits.

\subsection{Pre-training hyperparameters}
All other pre-training hyperparameters are described in Table~\ref{tab:hyper}.
\begin{table}[!ht]
    \footnotesize
    \centering
    \setlength{\tabcolsep}{2pt}
    \caption{Considered foundation models for electronic health records.}
    \label{tab:hyper}
    \begin{tabular}{ccccccc}
    \toprule
     \textbf{Hyperparameter} & \textbf{\makecell{\textsc{CEHR}\\\textsc{BERT}}} & \textbf{\makecell{\textsc{CEHR}\\\textsc{GPT}}} & \textbf{\makecell{\textsc{Core}\\\textsc{Behrt}}} & \textbf{\llama} & \textbf{\mamba} & \textbf{\motor} \\ \cmidrule{1-7}
     Architecture & \textsc{BERT} & \textsc{GPT} & \textsc{BERT} & \llama & \mamba & \makecell{Custom} \\
     \makecell{Embedding \\ size} & 768 & 768 & 768 & 768 & 768 & 768\\
     \makecell{Learning \\ rate} & $0.0002$ & $0.0002$ & $0.00005$ & $0.0002$ & $0.0002$ & $0.00001$\\
     \makecell{Weight \\ decay} & $0.01$ & $0.1$ & $0.01$ & $0.1$ & $0.1$ & $0.1$\\
     Optimizer &\makecell{\texttt{Adam} \\$\lambda_1=0.9$ \\ $\lambda_2=0.999$} & \makecell{\texttt{Adam} \\$\lambda_1=0.9$ \\ $\lambda_2=0.95$} & \makecell{\texttt{AdamW} \\$\lambda_1=0.9$ \\ $\lambda_2=0.999$} & \makecell{\texttt{AdamW} \\$\lambda_1=0.9$ \\ $\lambda_2=0.95$} & \makecell{\texttt{AdamW} \\$\lambda_1=0.9$ \\ $\lambda_2=0.95$} & \makecell{\texttt{AdamW} \\$\lambda_1=0.9$ \\ $\lambda_2=0.95$} \\
     \bottomrule
    \end{tabular}
\end{table}

\newpage
\subsection{Fine-tuning hyperparameters}
The hyperparameters for fine-tuning are provided in Table~\ref{tab:hparam_grid}.

\begin{table}[!ht]
\centering
\caption{Fine-tuning hyperparameters.}
\label{tab:hparam_grid}
\begin{tabular}{p{0.4\linewidth} p{0.6\linewidth}}
\hline
\textbf{Hyperparameter} & \textbf{Values (grid)} \\
\hline
Finetuned LR/Pretrained LR  & \{$0.05, 0.1, 0.2, 0.3, 0.5\}$ of the pretraining \\
Classification Head LR/Finetuned LR  & \{$1, 2, 5\}$ \\
Warmup epoch & $1$ \\
Scheduler & Linear Decay \\
Batch size & Same as pretraining \\
Maximum training epochs & $10$ \\
Early stopping patience & $3$ \\
Early stopping metric & Validation Log-loss \\
Overall selection & Validation Log-loss\\
\hline
\end{tabular}
\end{table}

\newpage
\subsection{Baselines}
\label{appendix:baselines}

\paragraph{\femr-LightGBM} 
Hyperparameter search for LightGBM is presented in \autoref{tab:femr-lightgbm-hp} and summarized below, based on the initial code of \femr\footnote{\url{https://github.com/ChaoPang/femr/tree/omop_meds_v3_tutorial}}.

\begin{table}[ht]
    \footnotesize
    \centering
    \setlength{\tabcolsep}{3pt}
    \caption{Hyperparameter Search Space for FEMR LightGBM}
    \label{tab:femr-lightgbm-hp}
    \begin{tabular}{cccc}
        \toprule
        \textbf{Parameter} & \textbf{Type} & \textbf{Range} & \textbf{Scale} \\ \cmidrule{2-4}
        \texttt{lambda\_l1} & Float & $[10^{-8}, 10.0]$ & Log \\ 
        \texttt{lambda\_l2} & Float & $[10^{-8}, 10.0]$ & Log \\ 
        \texttt{num\_leaves} & Integer & $[2, 256]$ & Linear \\ 
        \texttt{feature\_fraction} & Float & $[0.4, 1.0]$ & Linear \\
        \texttt{bagging\_fraction} & Float & $[0.4, 1.0]$ & Linear \\ 
        \texttt{bagging\_freq} & Integer & $[1, 7]$ & Linear \\
        \texttt{min\_child\_samples} & Integer & $[5, 100]$ & Linear \\ 
        \bottomrule
    \end{tabular}
\end{table}

\paragraph{\medstab -XGBoost}
Featurization, using \medstab \ 0.1, was performed on each task. Extracted features consist of counts, sums, squared sums, minimum and maximum values over multiple time windows of 1, 7, 30, 365 days, and the full length of stay prior to prediction times. Then, an XGBoost model was trained with hyperparameters selected over 100 draws from a search over parameters described in \autoref{tab:hydra-model-hp}.

\paragraph{\femr-Logistic Regression and \medstab-Logistic Regression} \label{baseline_logistic}
A standard 10-fold cross-validation was employed to select the optimal inverse of the $l2$ penalty (\(\lambda\)), using values chosen on a logarithmic scale between \(10^{-4}\) and \(10^{4}\).

\begin{table}[!ht]
    \footnotesize
    \centering
    \setlength{\tabcolsep}{3pt}
    \caption{Hyperparameter Search Space for \medstab}
    \label{tab:hydra-model-hp}
    \begin{tabular}{cccc}
        \toprule
        \textbf{Parameter} & \textbf{Type} & \textbf{Range} & \textbf{Scale} \\ \cmidrule{2-4}
        \texttt{model.eta} & Float & $[0.001, 1]$ & Log \\
        \texttt{model.lambda} & Float & $[0.001, 1]$ & Log \\
        \texttt{model.alpha} & Float & $[0.001, 1]$ & Log \\
        \texttt{model.subsample} & Float & $[0.5, 1]$ & Linear \\
        \texttt{model.min\_child\_weight} & Float & $[10^{-2}, 100]$ & Linear \\
        \texttt{model.max\_depth} & Integer & $[2, 16]$ & Linear \\
        \texttt{training\_params.num\_boost\_round} & Integer & $[100, 1000]$ & Linear \\
        \texttt{training\_params.early\_stopping\_rounds} & Integer & $[1, 10]$ & Linear \\
        \texttt{tabularization.min\_code\_inclusion\_count} & Integer & $[10, 1000000]$ & Log \\
        \bottomrule
    \end{tabular}
\end{table}

\clearpage
\section{Compute resources}
\label{apd:resources}
All experiments were run on a server with 4 NVIDIA H100 NVL GPUs with 96 GB of VRAM, 2 Intel(R) Xeon(R) Platinum 8480+ CPUs (56 cores each) with 2 TB of memory.

\subsection{Pretraining}
Pretraining applies only to the foundation models; the baselines are not pretrained, denoted ``--'' in Table~\ref{tab:pretrain_time}. As these numbers do not include prior iterations of the models, our total estimate is about $1{,}400$ compute hours.  
\begin{table}[ht]
    \footnotesize
    \centering
    \setlength{\tabcolsep}{3pt}
    \caption{Pretraining total running time (in days).}
    \begin{tabular}{cccccccc}
    \toprule
     \textbf{Metrics} &
     \textbf{Baseline}& \textbf{\makecell{\textsc{CEHR}\\\textsc{\textsc{BERT}}}} & \textbf{\makecell{\textsc{CEHR}\\\textsc{GPT}}} &\textbf{\motor}  & \textbf{\llama} & \textbf{\mamba} & \textbf{\makecell{\textsc{Core}\\\textsc{\textsc{Behrt}}}} \\ \cmidrule{2-8}
    \cumc & -- & 1.3 & 0.7 & 1.5 & 1.3 & 1.9 & 3 \\
    \mimic & -- & 0.1 & 0.2 & 0.2 & 0.4 & 0.3 & 0.5 \\
    \bottomrule
    \end{tabular}
    \label{tab:pretrain_time}
\end{table}

\subsection{Linear Probing}
Table~\ref{tab:train_time} reports the estimated linear-probing time for the final run. For the baselines, the reported time is featurization plus model fitting.
\begin{table}[ht]
    \footnotesize
    \centering
    \setlength{\tabcolsep}{3pt}
    \caption{Linear probing total running time (in days).}
    \begin{tabular}{cccccccc}
    \toprule
     \textbf{Metrics} &
     \textbf{Baseline}& \textbf{\makecell{\textsc{CEHR}\\\textsc{\textsc{BERT}}}} & \textbf{\makecell{\textsc{CEHR}\\\textsc{GPT}}} &\textbf{\motor}  & \textbf{\llama} & \textbf{\mamba} & \textbf{\makecell{\textsc{Core}\\\textsc{\textsc{Behrt}}}} \\ \cmidrule{2-8}
    \cumc & 11 & 0.2 & 0.2 & 0.3 & 0.3 & 0.6 & 0.3 \\
    \mimic & 6 & 0.1 & 0.3 & 0.2 & 0.2 & 0.4 & 0.1 \\
    \bottomrule
    \end{tabular}
    \label{tab:train_time}
\end{table}

\subsection{Fine-tuning}
Table~\ref{tab:train_time_ft} reports the estimated fine-tuning time, computed as the time for one epoch multiplied by the total number of epochs across the full grid search.
\begin{table}[ht]
    \footnotesize
    \centering
    \setlength{\tabcolsep}{3pt}
    \caption{Fine-tuning total running time (in days) for Stroke.}
    \begin{tabular}{cccc}
    \toprule
     \textbf{Metrics} &
      \textbf{\makecell{\textsc{CEHR}\\\textsc{GPT}}} &\textbf{\motor}  & \textbf{\llama} \\ \cmidrule{2-4}
   \mimic & 0.8 & 0.9 & 1.9\\ 
   \cumc & 0.3 & 0.6 & 1.8 \\ 
    \bottomrule
    \end{tabular}
    \label{tab:train_time_ft}
\end{table}

\newpage
\subsection{Inference}
As computing resources may be a limitation for deploying FMs in healthcare, Table~\ref{tab:flops} reports the number of FLOPs for inference, corresponding to the computational requirement at deployment. Inference cost is identical whether the model is applied through linear probing or fine-tuning, since both perform the same forward pass through the model; we therefore report a single inference figure per model.
\begin{table}[ht]
    \footnotesize
    \centering
    \setlength{\tabcolsep}{3pt}
    \caption{Inference FLOPs.}
    \begin{tabular}{cccccccc}
    \toprule
     \textbf{Metrics} & \textbf{\makecell{\textsc{CEHR}\\\textsc{\textsc{BERT}}}} & \textbf{\makecell{\textsc{CEHR}\\\textsc{GPT}}} &\textbf{\motor}  & \textbf{\llama} & \textbf{\mamba} & \textbf{\makecell{\textsc{Core}\\\textsc{\textsc{Behrt}}}} \\ \cmidrule{2-7}
    Inference FLOPs & $2.02e^{16}$ & $1.78e^{16}$ & $5.1e^{16}$ &$7.53e^{16}$&$9.32e^{16}$&$1.52e^{16}$\\
    \bottomrule
    \end{tabular}
    \label{tab:flops}
\end{table}

\newpage
\section{\cumc detailed results}~\label{appendix:additional_results:cumc}

\subsection{Average performance comparison}
Section~\ref{sec:results} presents the average ranking across tasks. Similarly, Table~\ref{tab:leaderboard_perf_cumc} presents the average performance across tasks, exploring the impact of a small training set on FM performance. While performances are lower in this setting, FMs provide an edge on discrimination but underperform on calibration.
\begin{table}[!ht]
    \footnotesize
    \centering
    \setlength{\tabcolsep}{2pt}
    \caption{Average performance on the proposed dataset stratified per metrics, sorted by average ranking across all metrics.}
    \label{tab:leaderboard_perf_cumc}
    \begin{tabular}{cccc ccc}
        \toprule
        Foundation & \multicolumn{3}{c}{Large data regime} & \multicolumn{3}{c}{Small data regime} \\
        \cmidrule(lr){2-4} \cmidrule(lr){5-7}
         Model & Calibration & Discrimination & Fairness &Calibration & Discrimination & Fairness \\
        \cmidrule{2-7}
            \motor & 0.01 (0.02) & 0.79 (0.08) & 0.06 (0.06) & 0.08 (0.06) & 0.69 (0.09) & 0.06 (0.06) \\
            \cehrgpt & 0.01 (0.03) & 0.76 (0.08) & 0.07 (0.08) & 0.07 (0.04) & 0.64 (0.11) & 0.08 (0.07) \\
            \mamba & 0.01 (0.02) & 0.75 (0.07) & 0.05 (0.05) & 0.08 (0.04) & 0.63 (0.07) & 0.06 (0.07) \\
            \llama & 0.01 (0.02) & 0.75 (0.08) & 0.07 (0.06) & 0.07 (0.04) & 0.61 (0.05) & 0.07 (0.06) \\
            Best Baseline & 0.01 (0.01) & 0.76 (0.09) & 0.14 (0.11) & 0.04 (0.02) & 0.63 (0.09) & 0.13 (0.12) \\
            \cehrbert & 0.01 (0.03) & 0.73 (0.08) & 0.09 (0.09) & 0.07 (0.04) & 0.63 (0.07) & 0.08 (0.07) \\
            \corebehrt & 0.02 (0.04) & 0.68 (0.09) & 0.08 (0.08) & 0.08 (0.04) & 0.59 (0.08) & 0.07 (0.07) \\
         \bottomrule
    \end{tabular}
\end{table}

\newpage
\subsection{Task-specific performances}
\begin{figure}[ht]
    \makebox[\textwidth][l]{
    \includegraphics[height=100px]{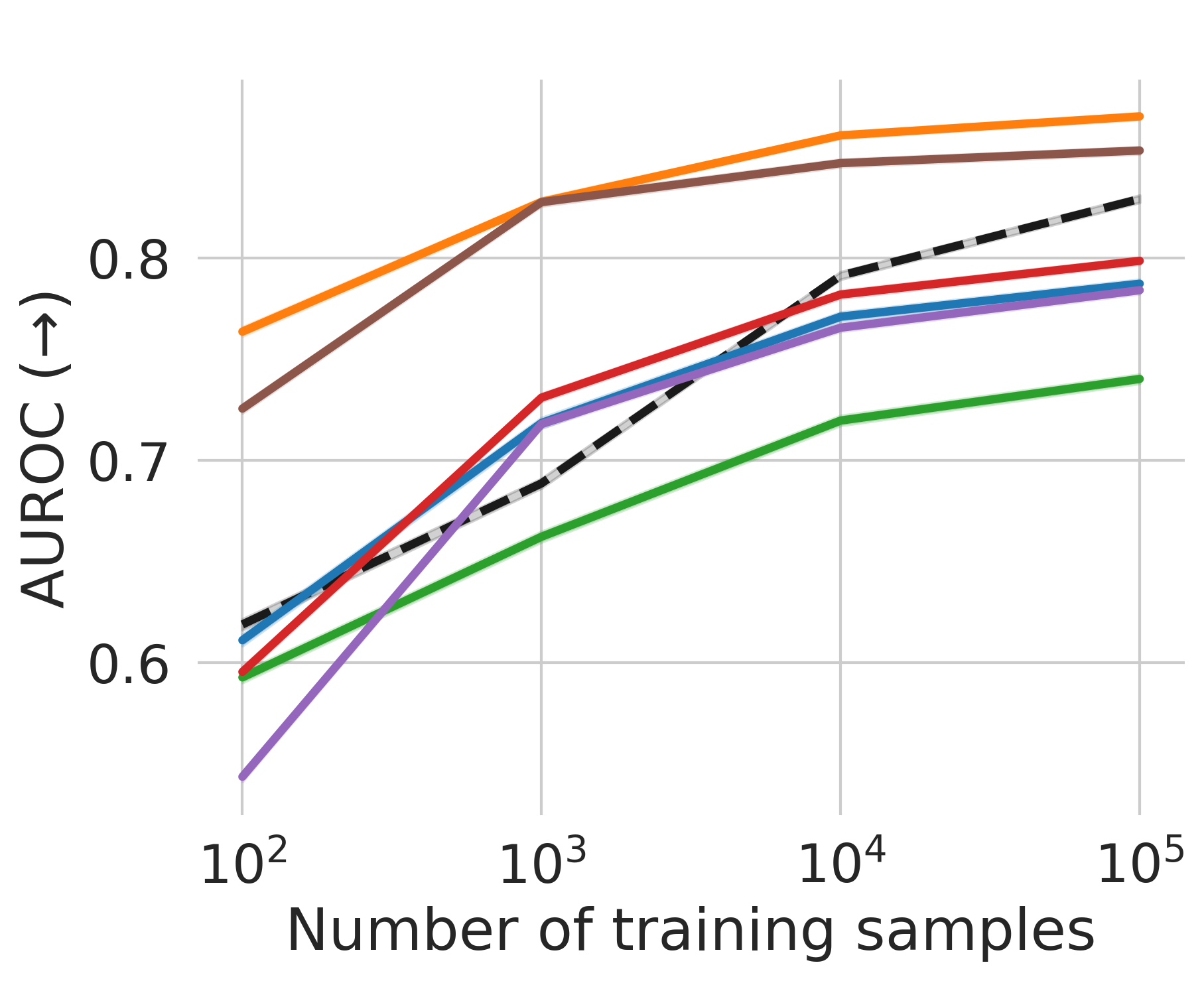}    
    \includegraphics[height=100px]{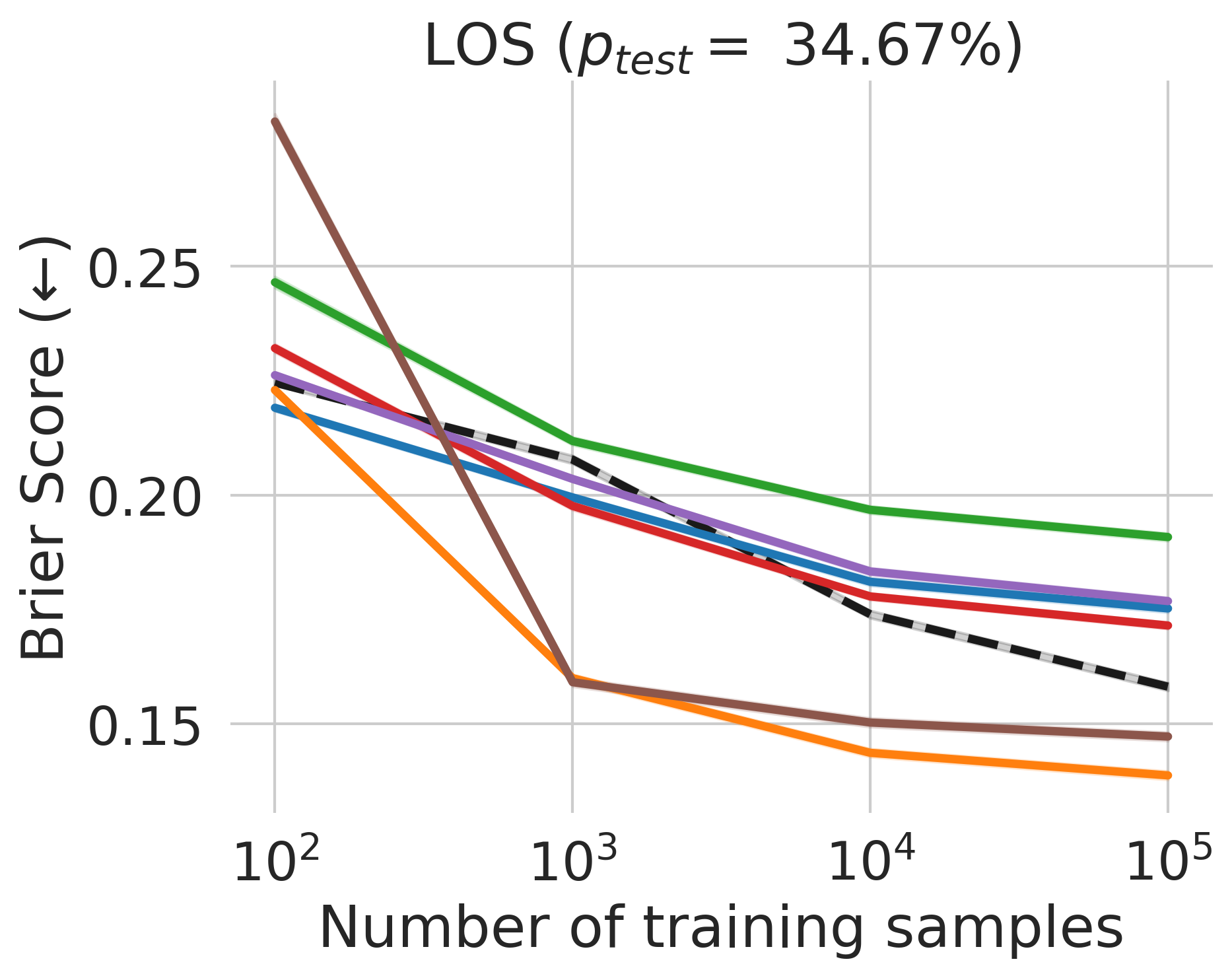}
    \includegraphics[height=100px]{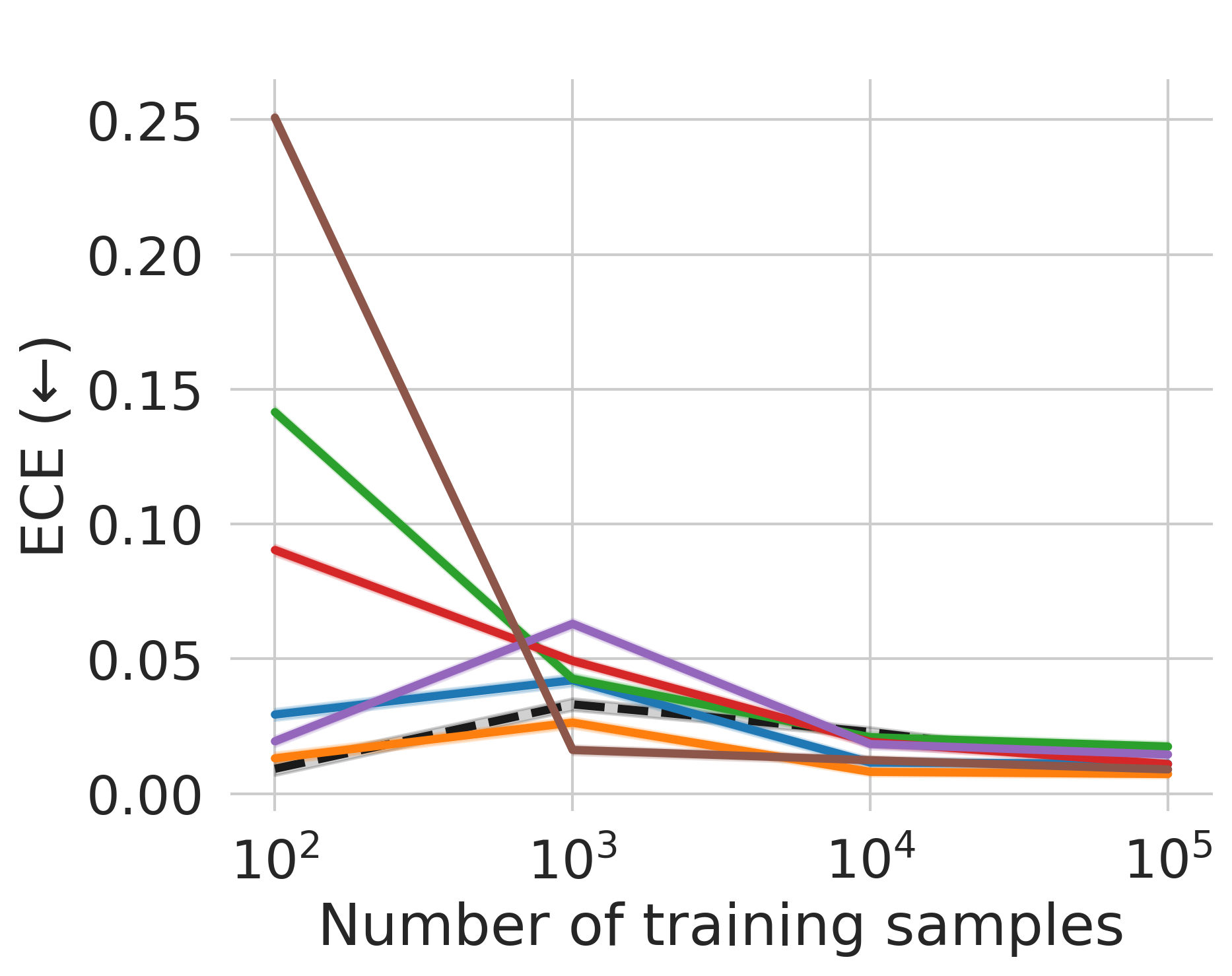}}

    \makebox[\textwidth][l]{
    \includegraphics[height=100px]{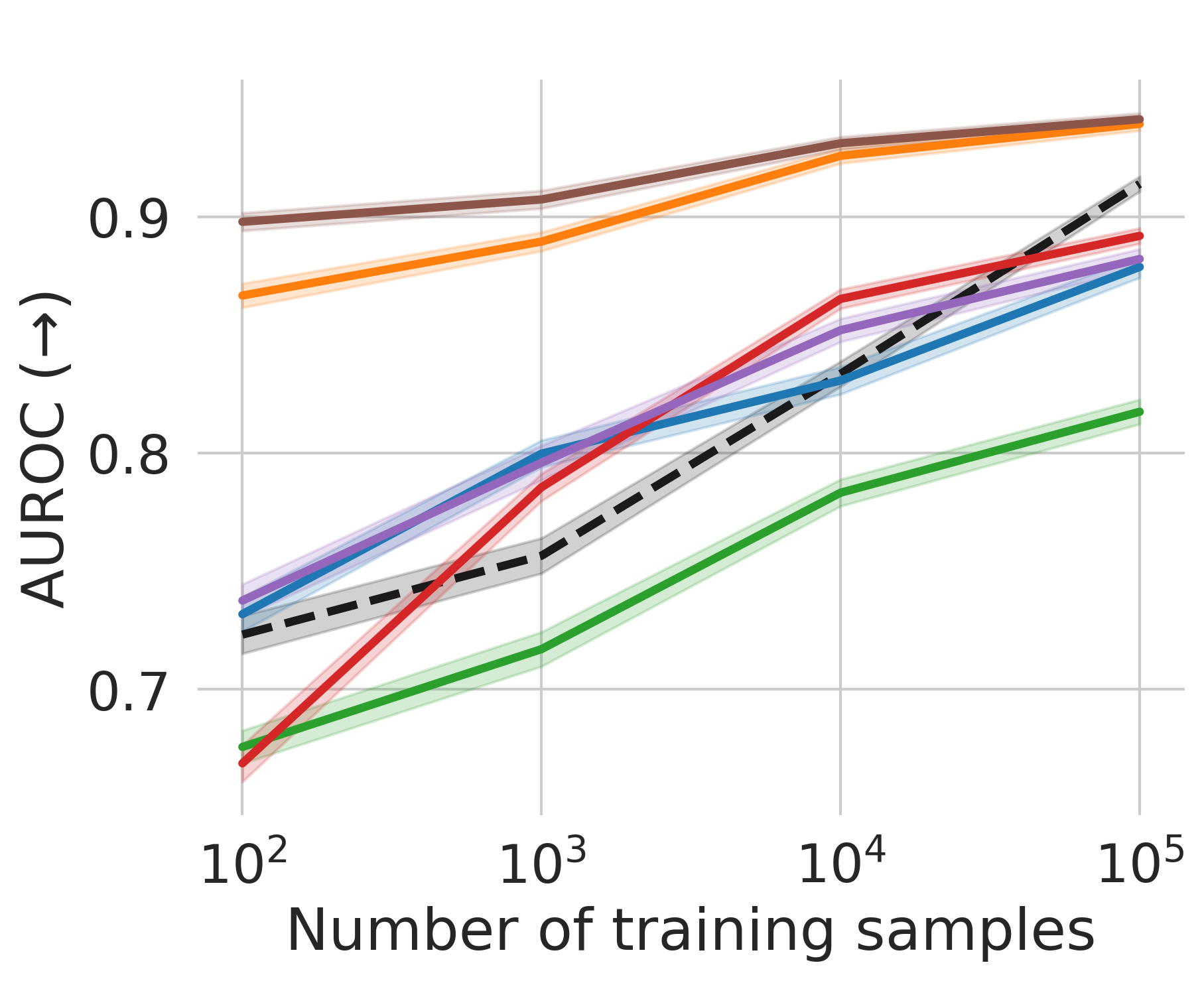}
    \includegraphics[height=100px]{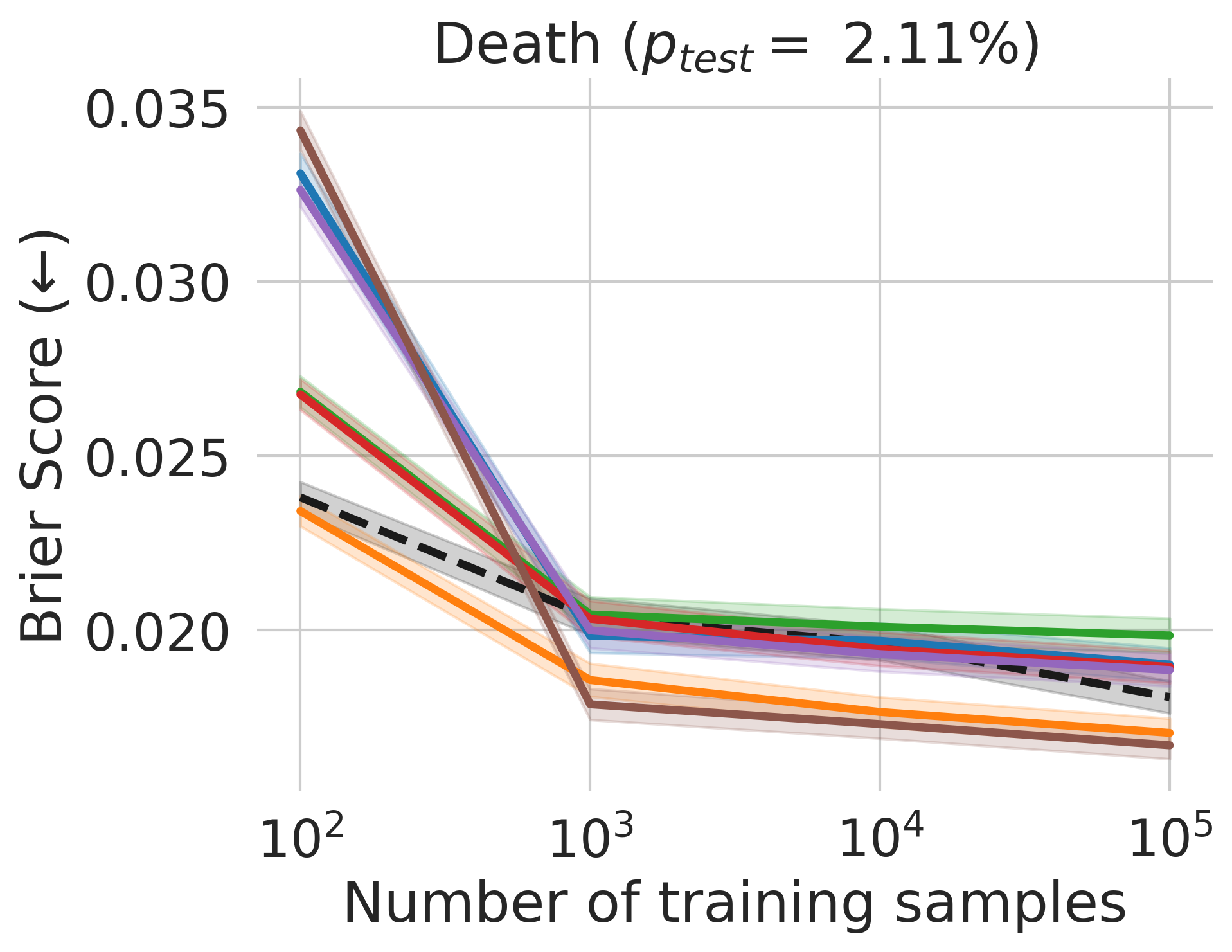}
    \includegraphics[height=100px]{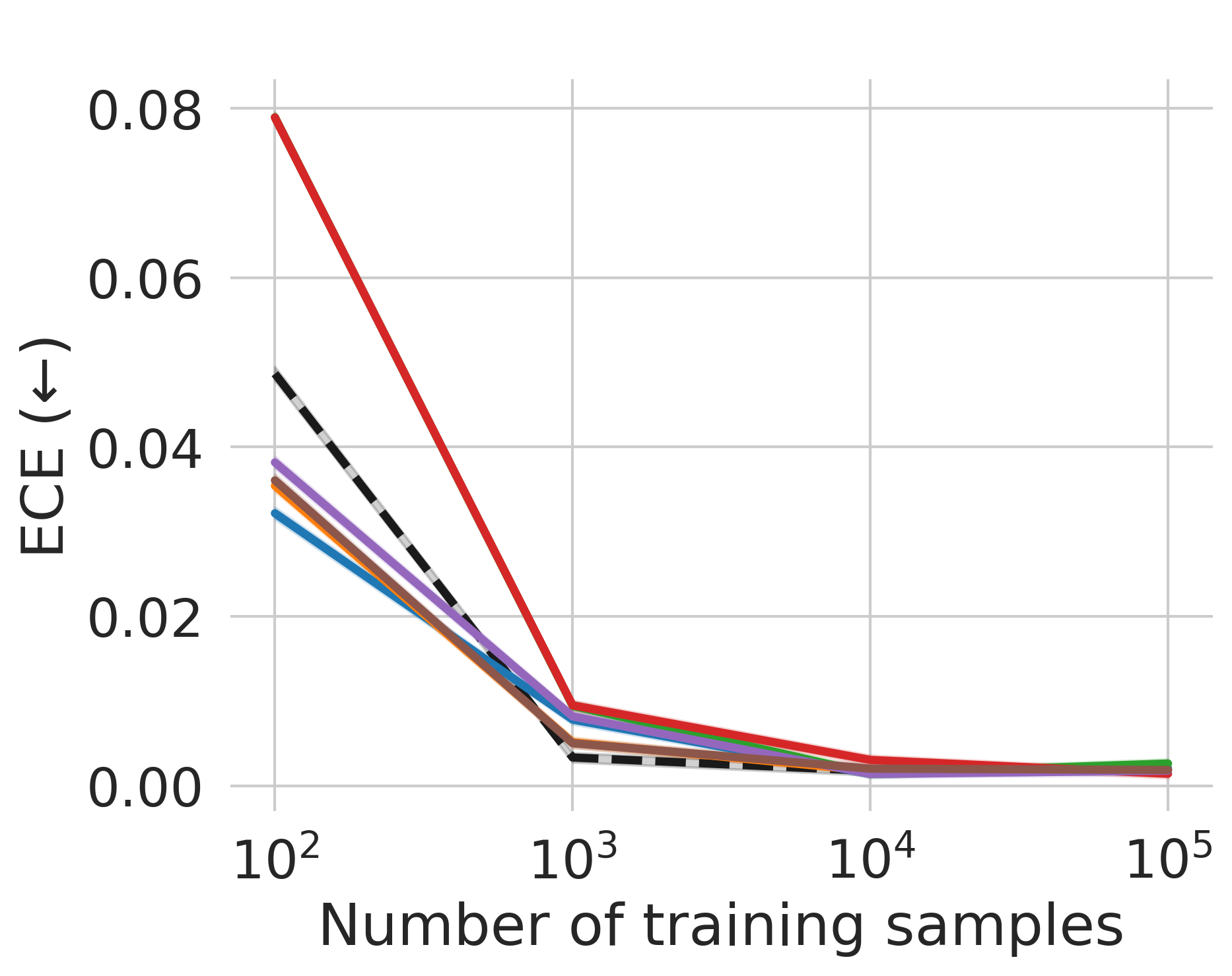}
    \includegraphics[height=100px]{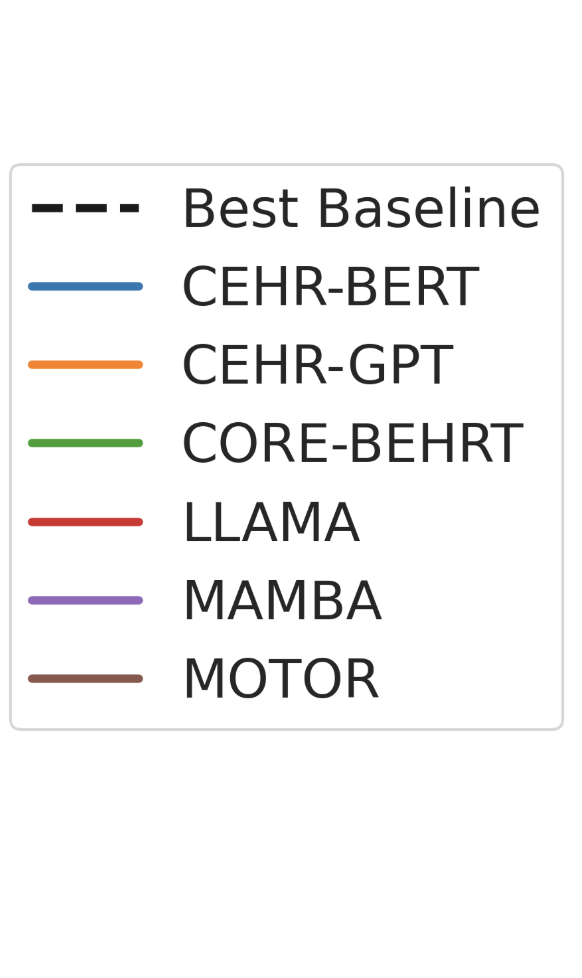}}

    \makebox[\textwidth][l]{
    \includegraphics[height=100px]{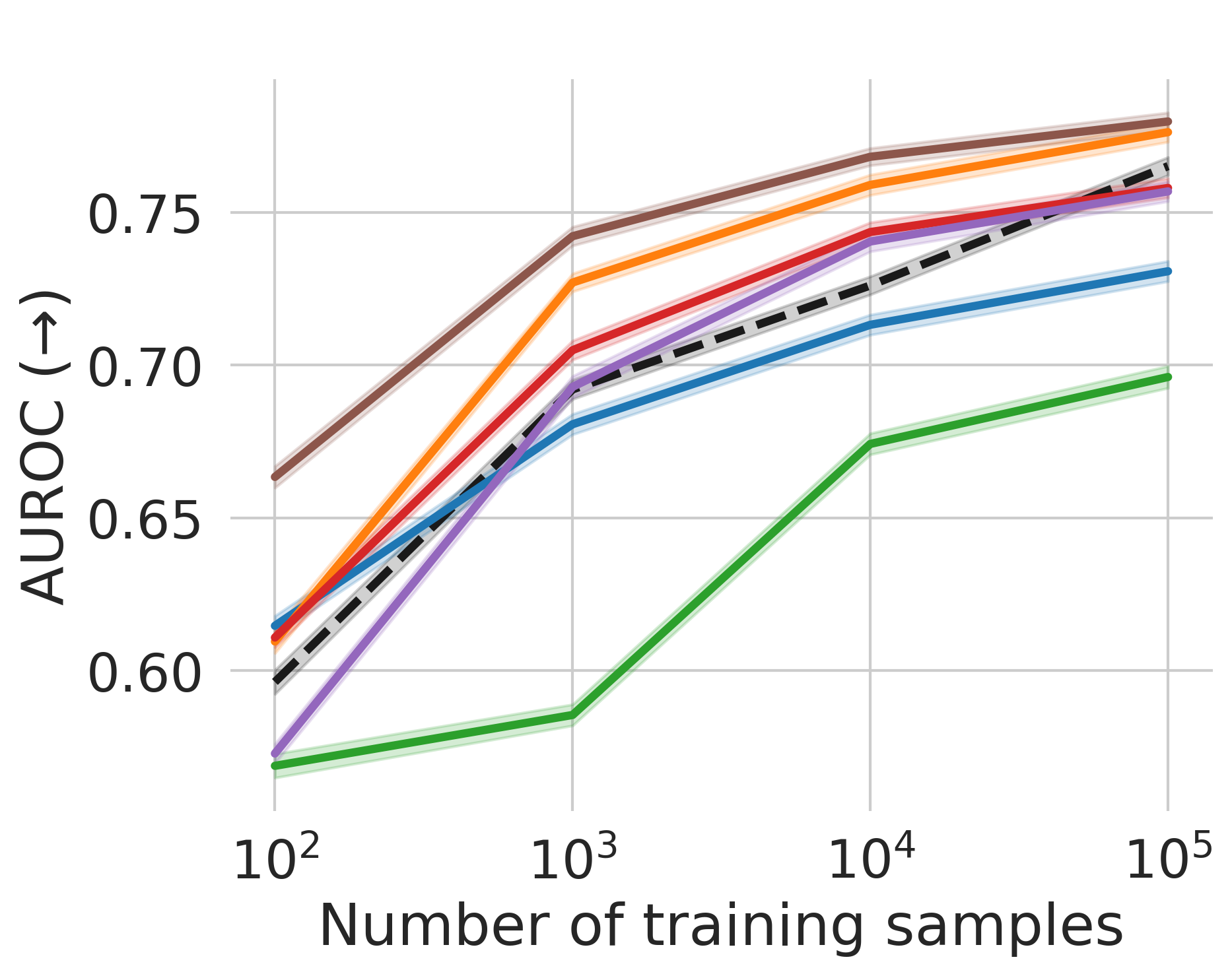}
    \includegraphics[height=100px]{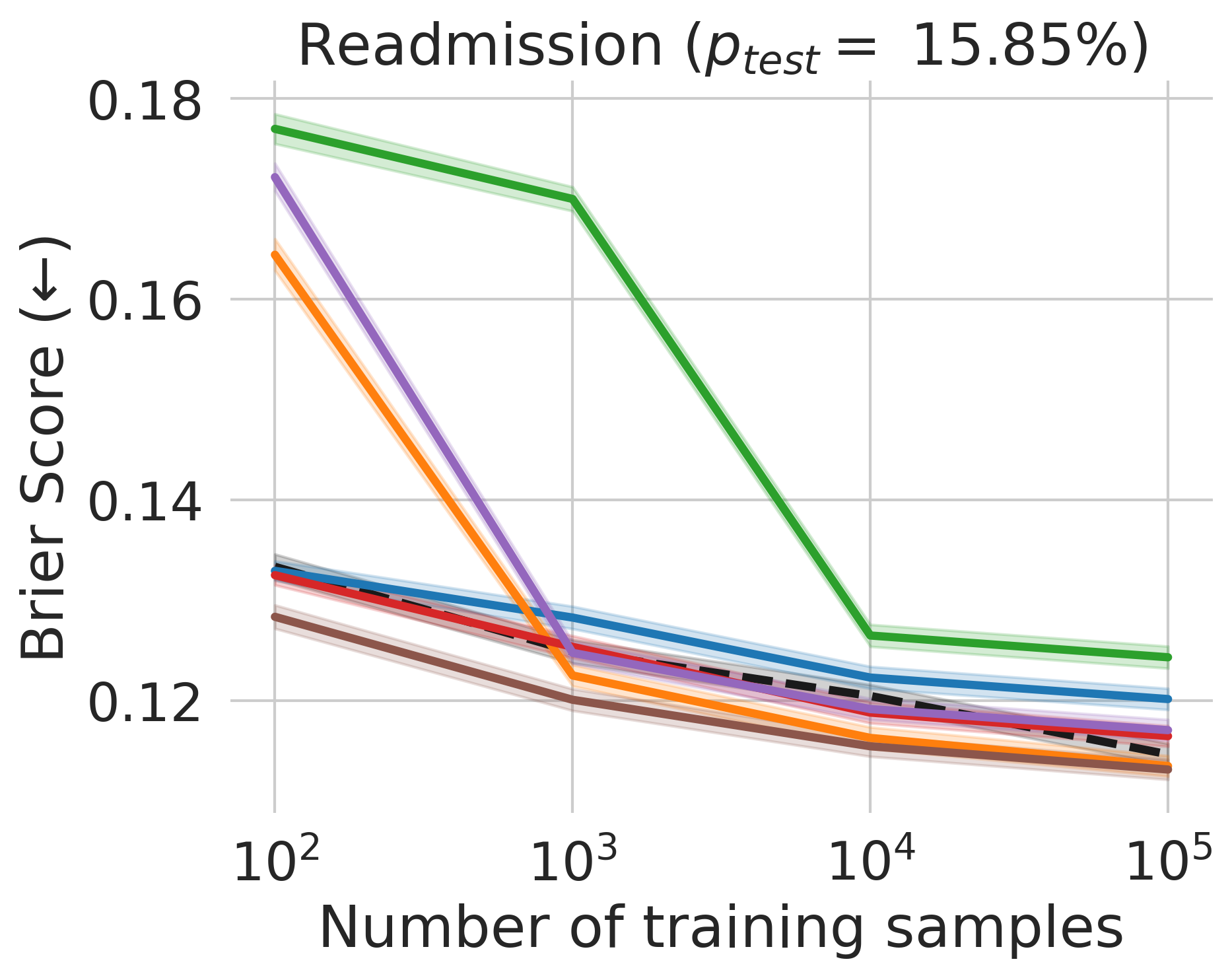}
    \includegraphics[height=100px]{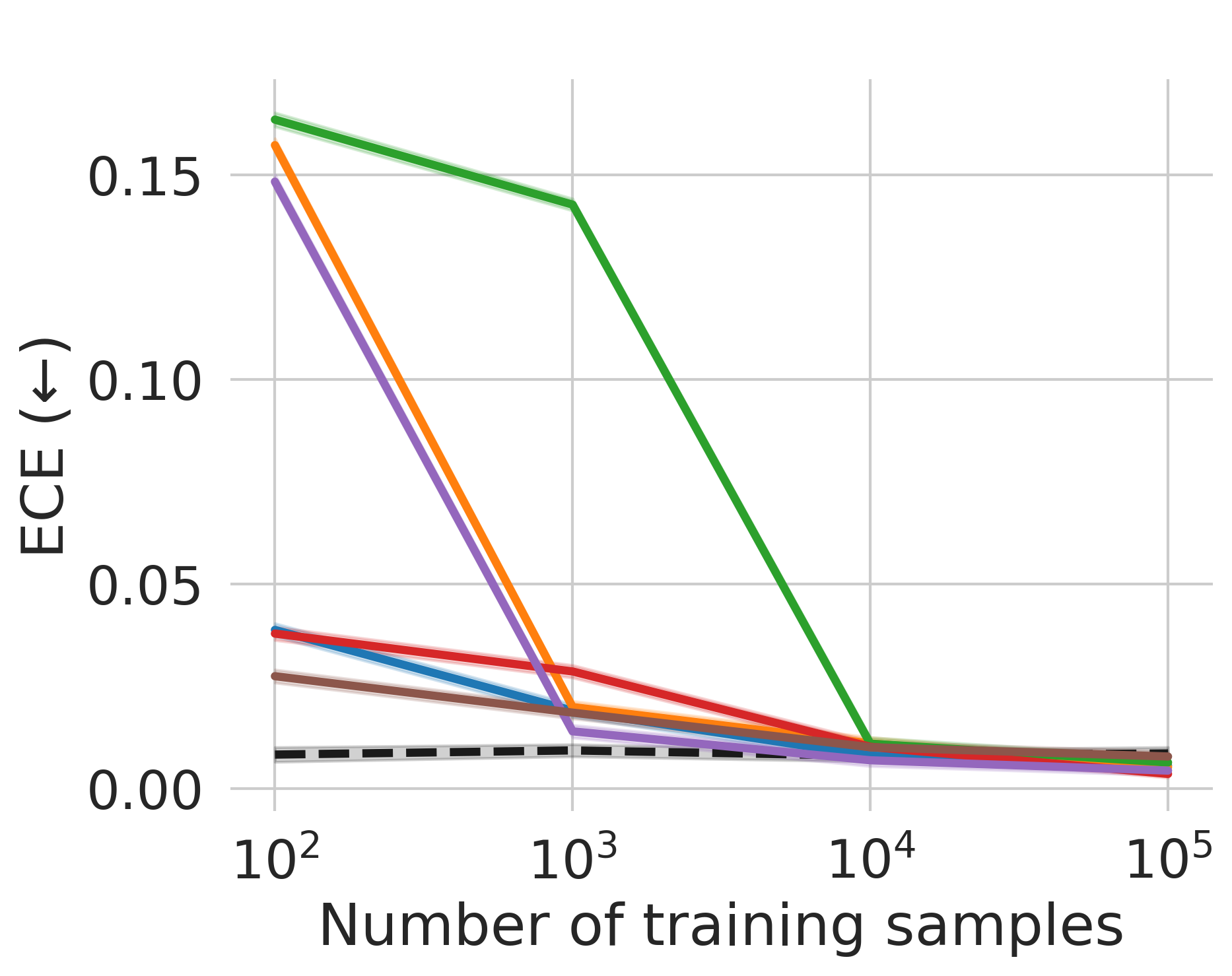}}

    \centering
    \caption{Discriminative performances and calibration results for all outcomes for an increasing number of training points (Shaded area represents bootstrapped standard deviation).}
    \label{fig:all_outcomes:cumc}
\end{figure}

\begin{figure}[ht]
    \makebox[\textwidth][l]{
    \includegraphics[height=100px]{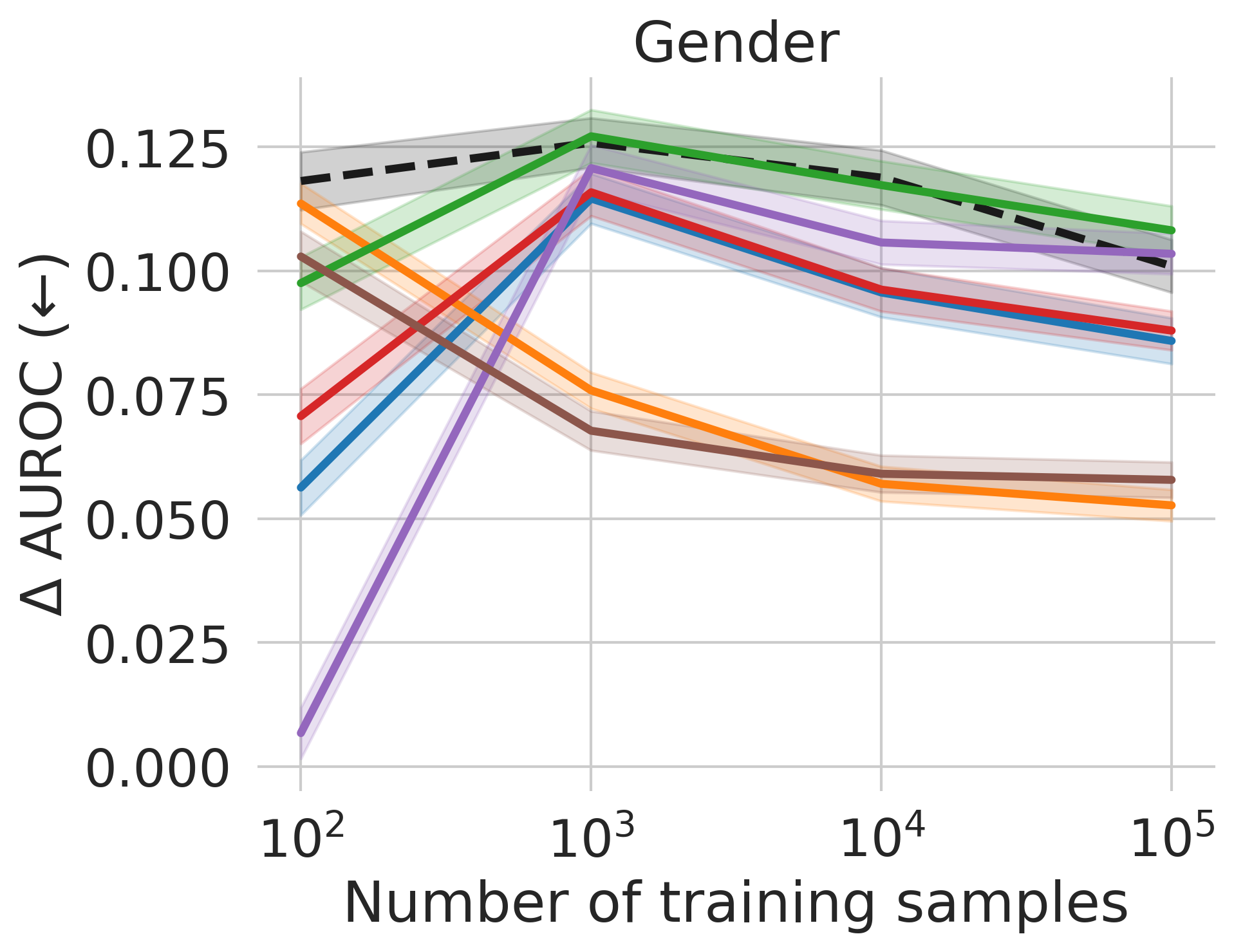}
    \includegraphics[height=100px]{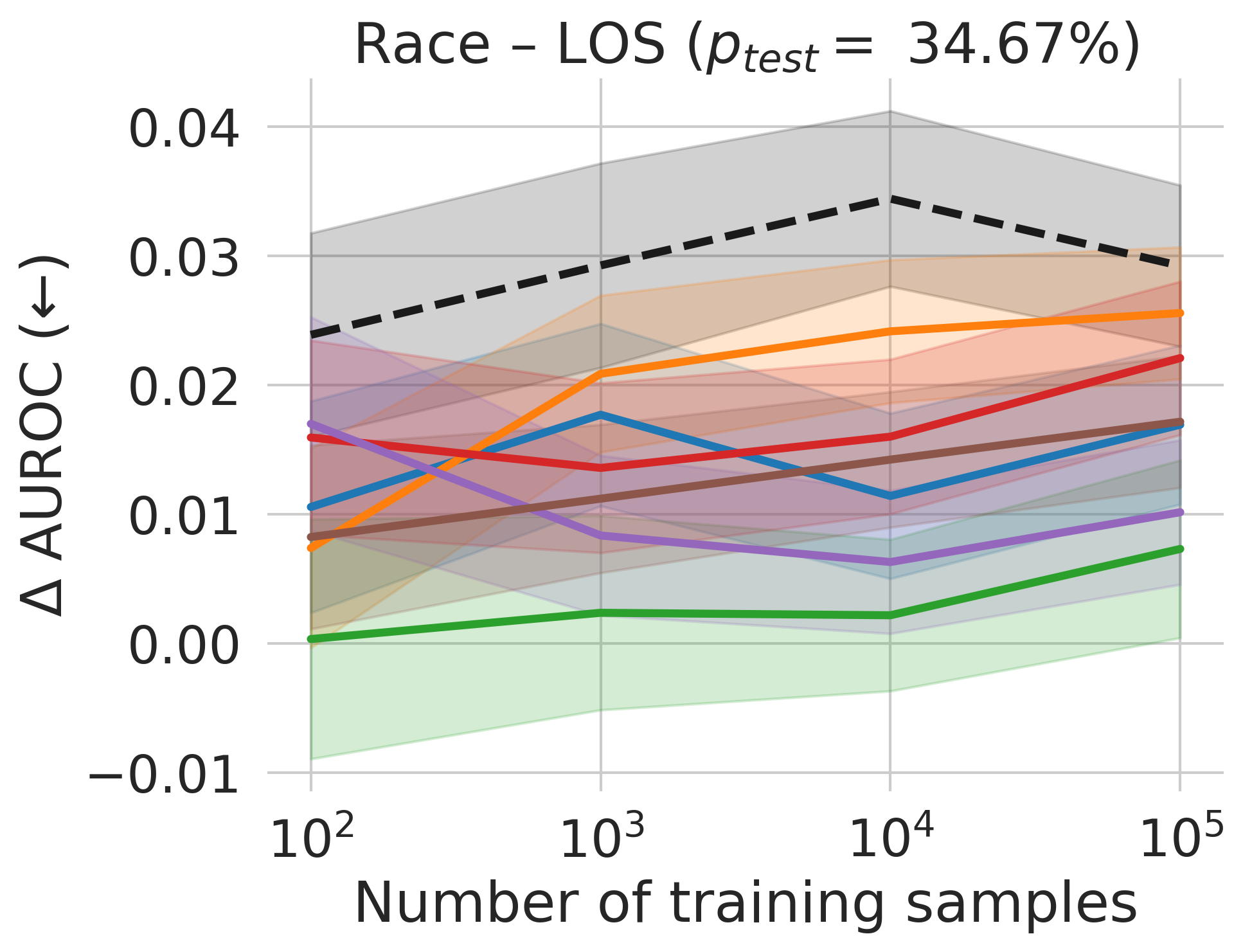}
    \includegraphics[height=100px]{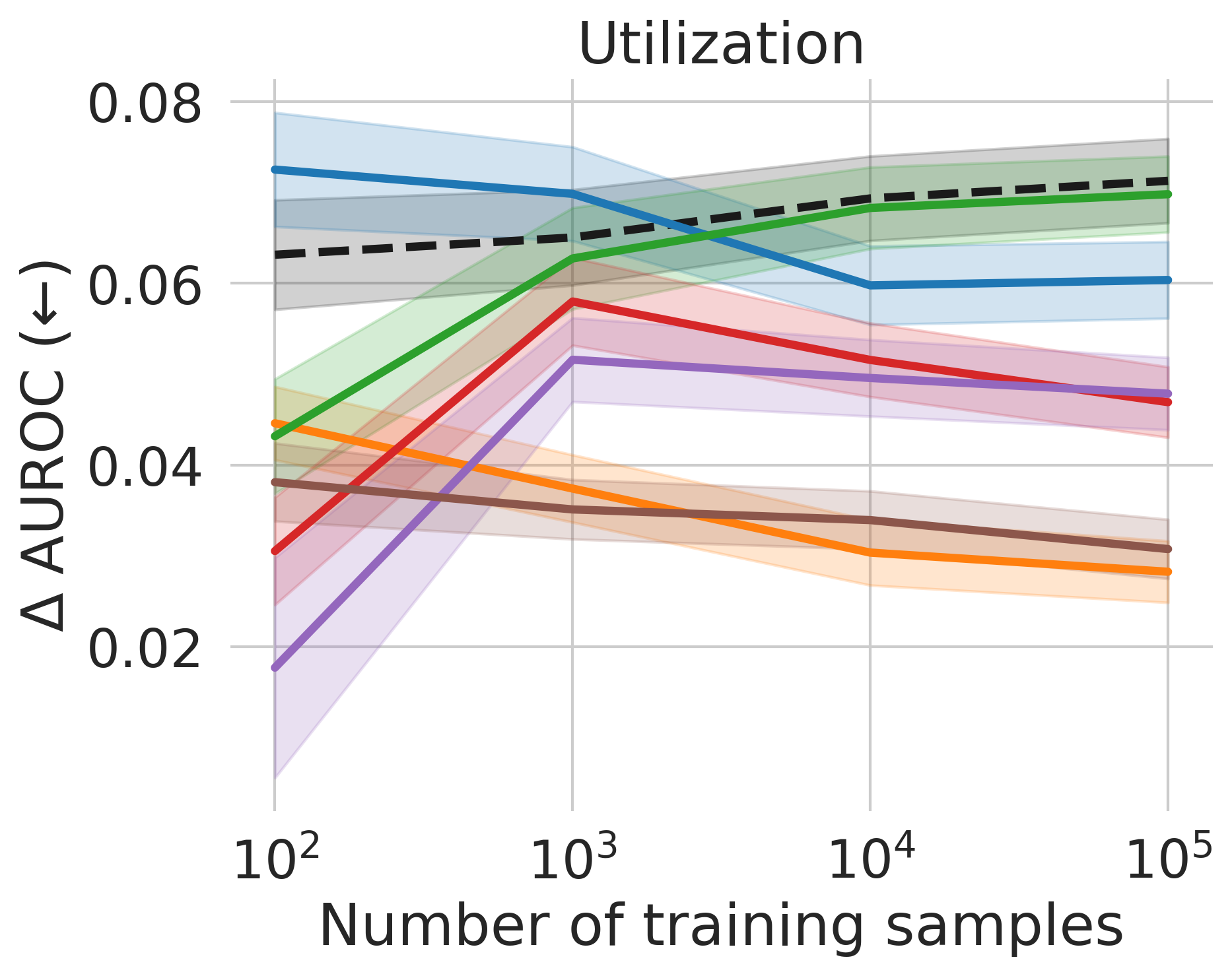}}

    \makebox[\textwidth][l]{
    \includegraphics[height=100px]{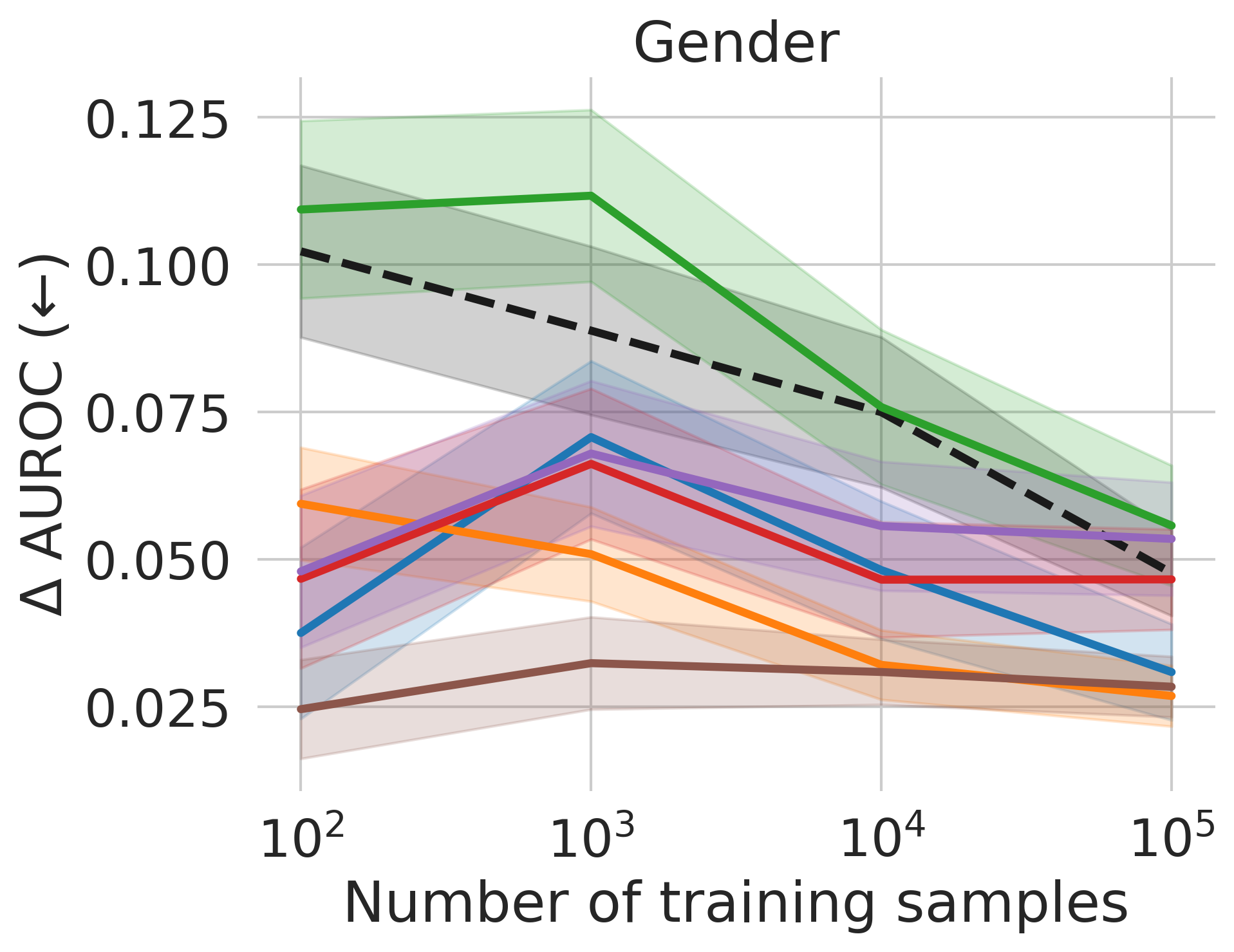}
    \includegraphics[height=100px]{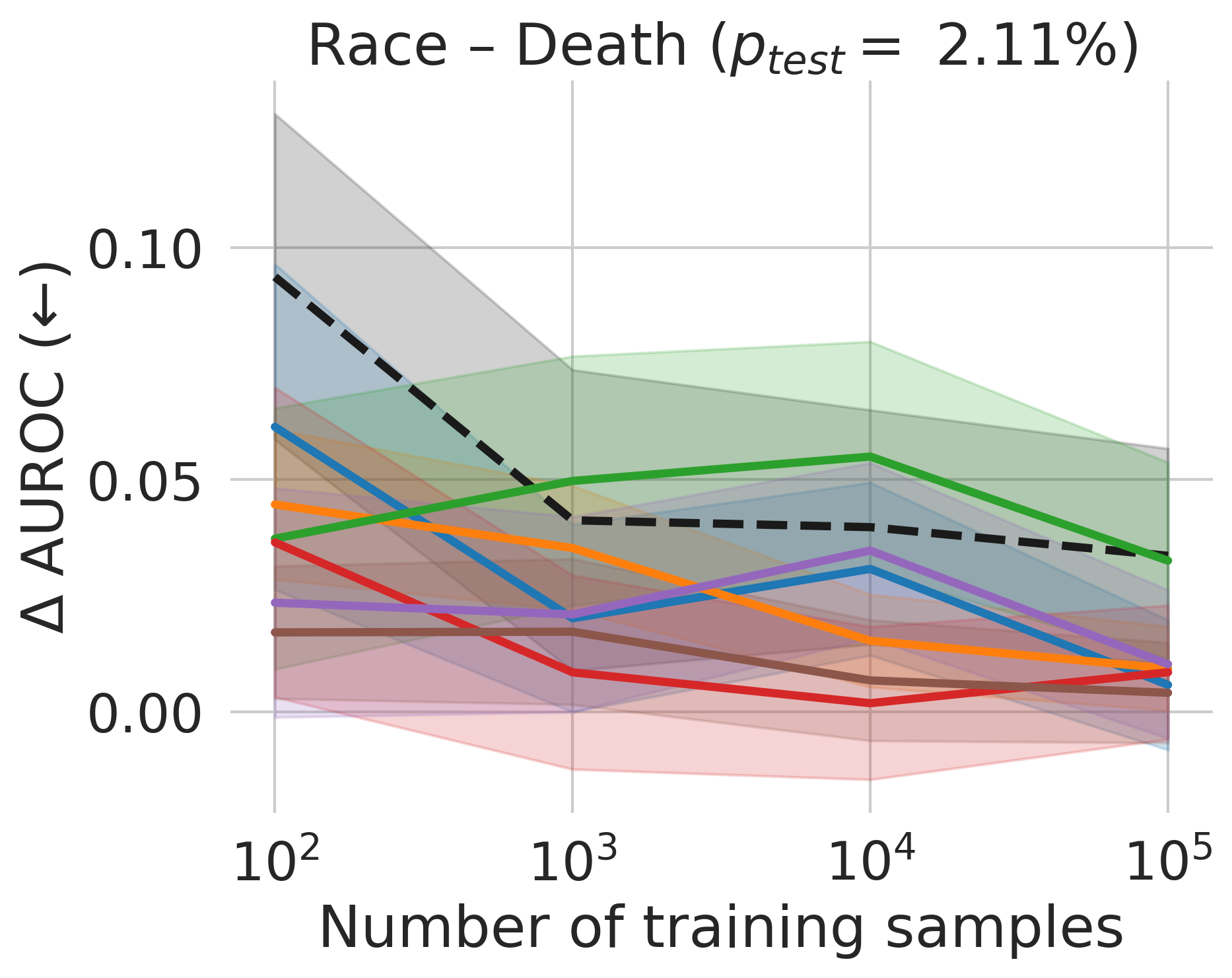}
    \includegraphics[height=100px]{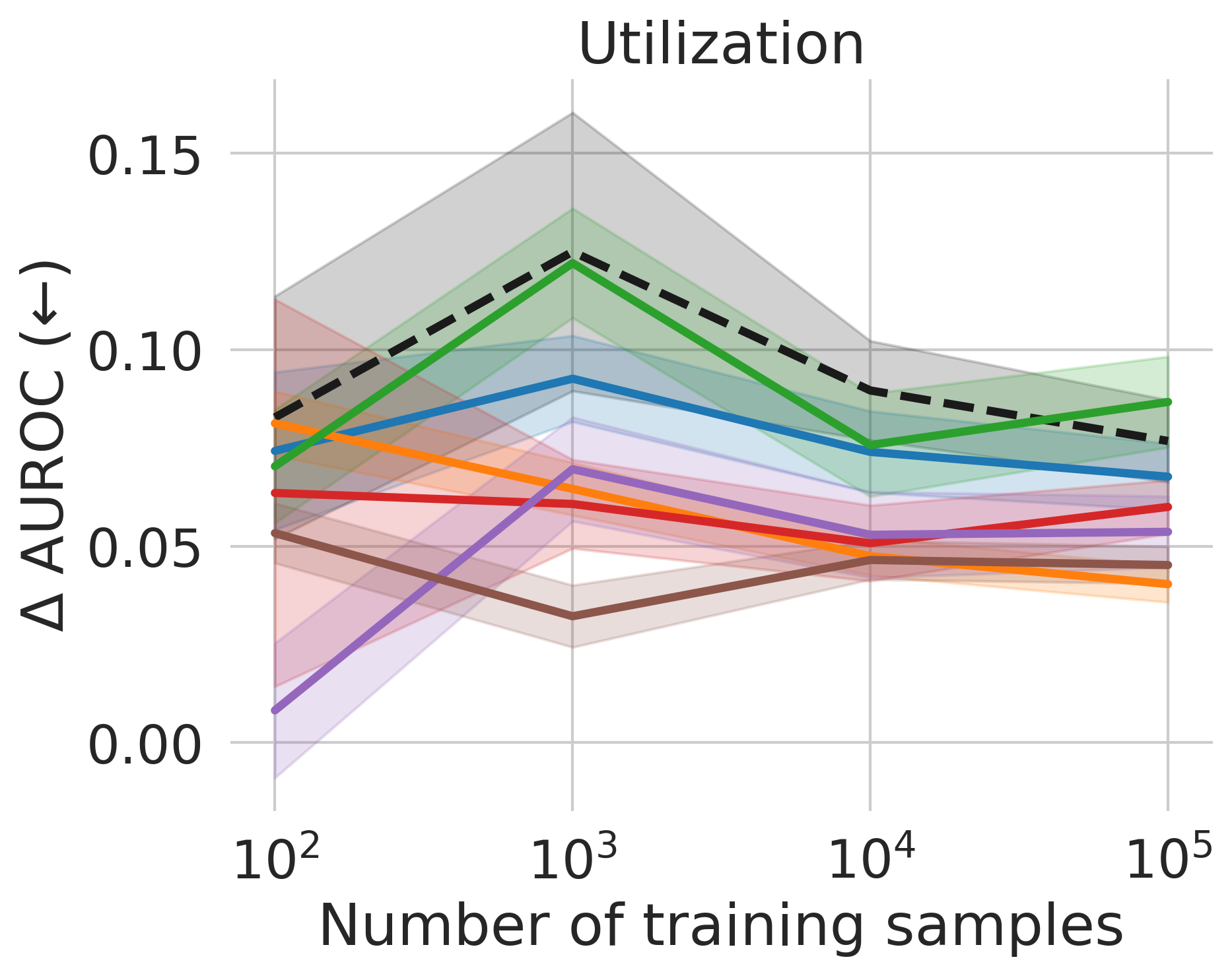}
    \includegraphics[height=100px]{figures/legend.png}}

    \makebox[\textwidth][l]{
    \includegraphics[height=100px]{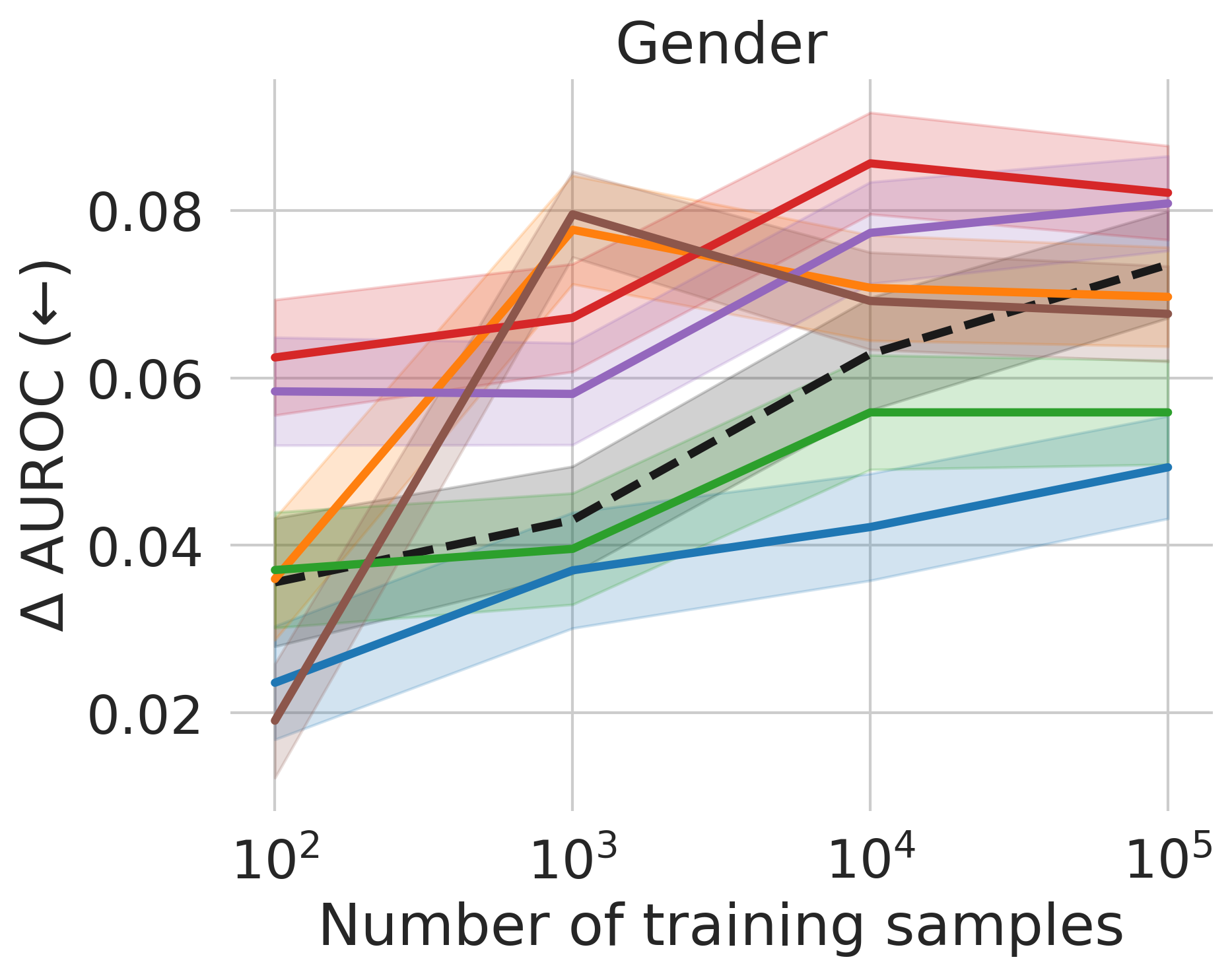}
    \includegraphics[height=100px]{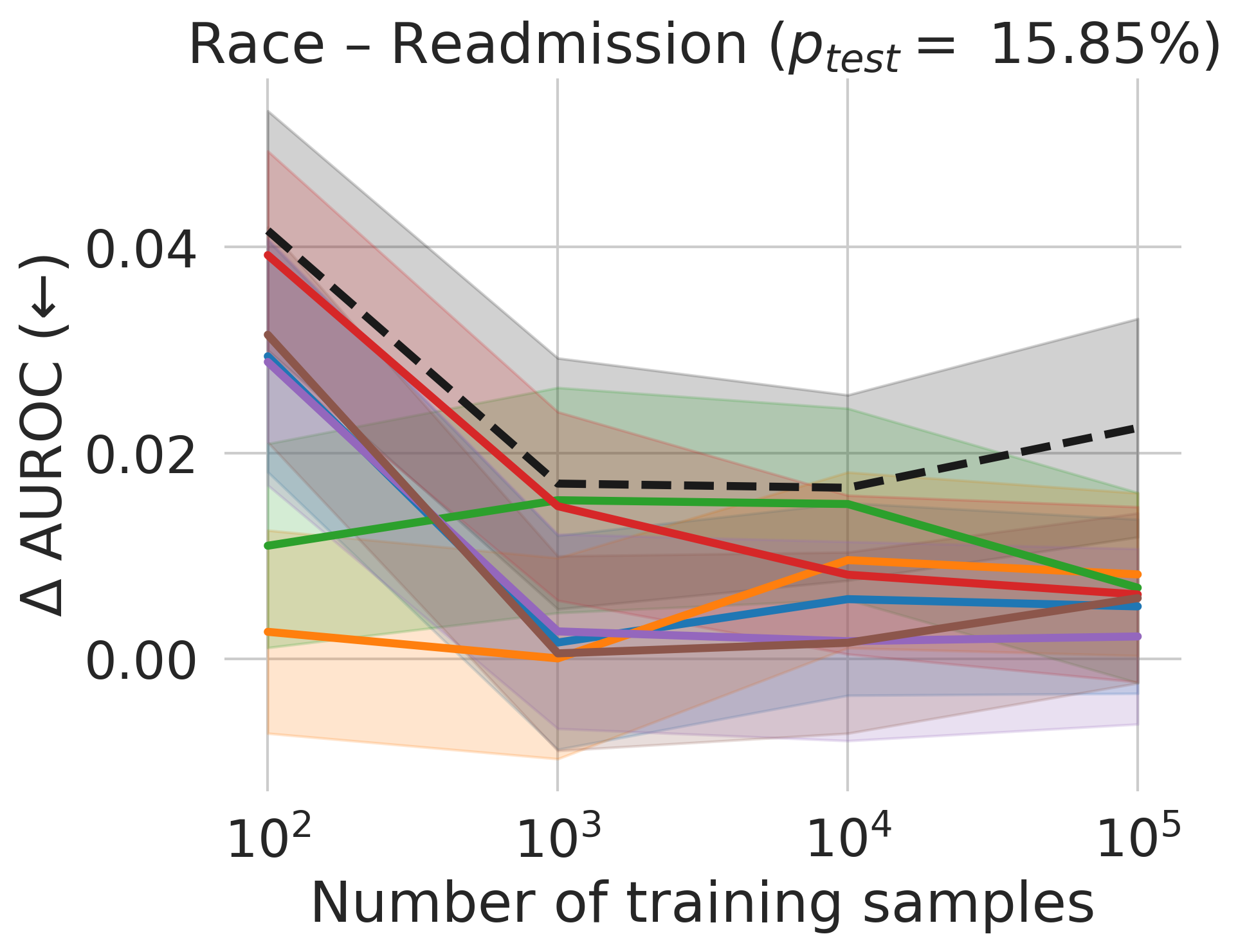}
    \includegraphics[height=100px]{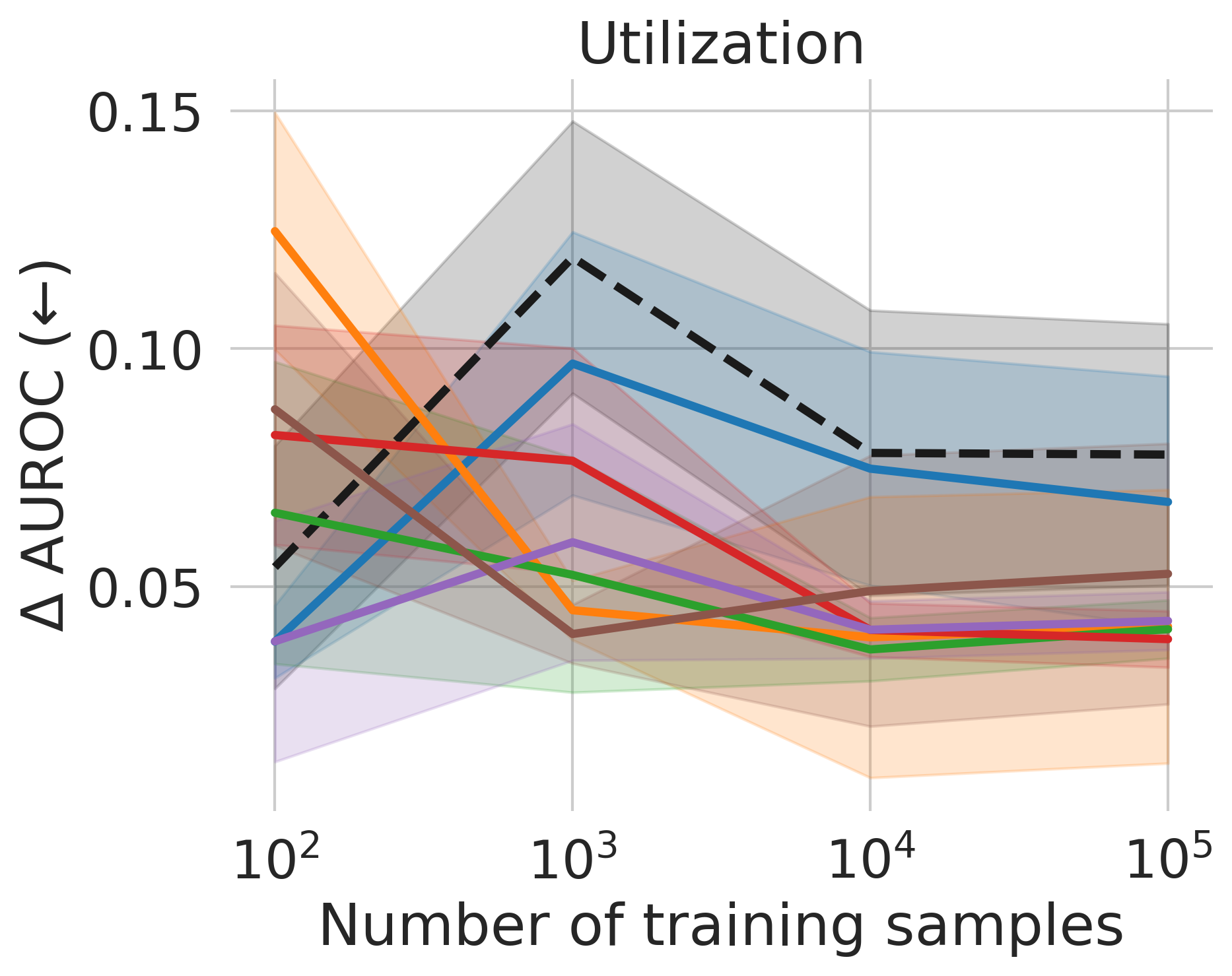}
    }

    \centering
    \caption{Discriminative performances and calibration results for all outcomes for an increasing number of training points (Shaded area represents bootstrapped standard deviation).}
    \label{fig:all_outcomes:cumc:fairness}
\end{figure}

\begin{figure}[ht]
    \makebox[\textwidth][l]{
    \includegraphics[height=100px]{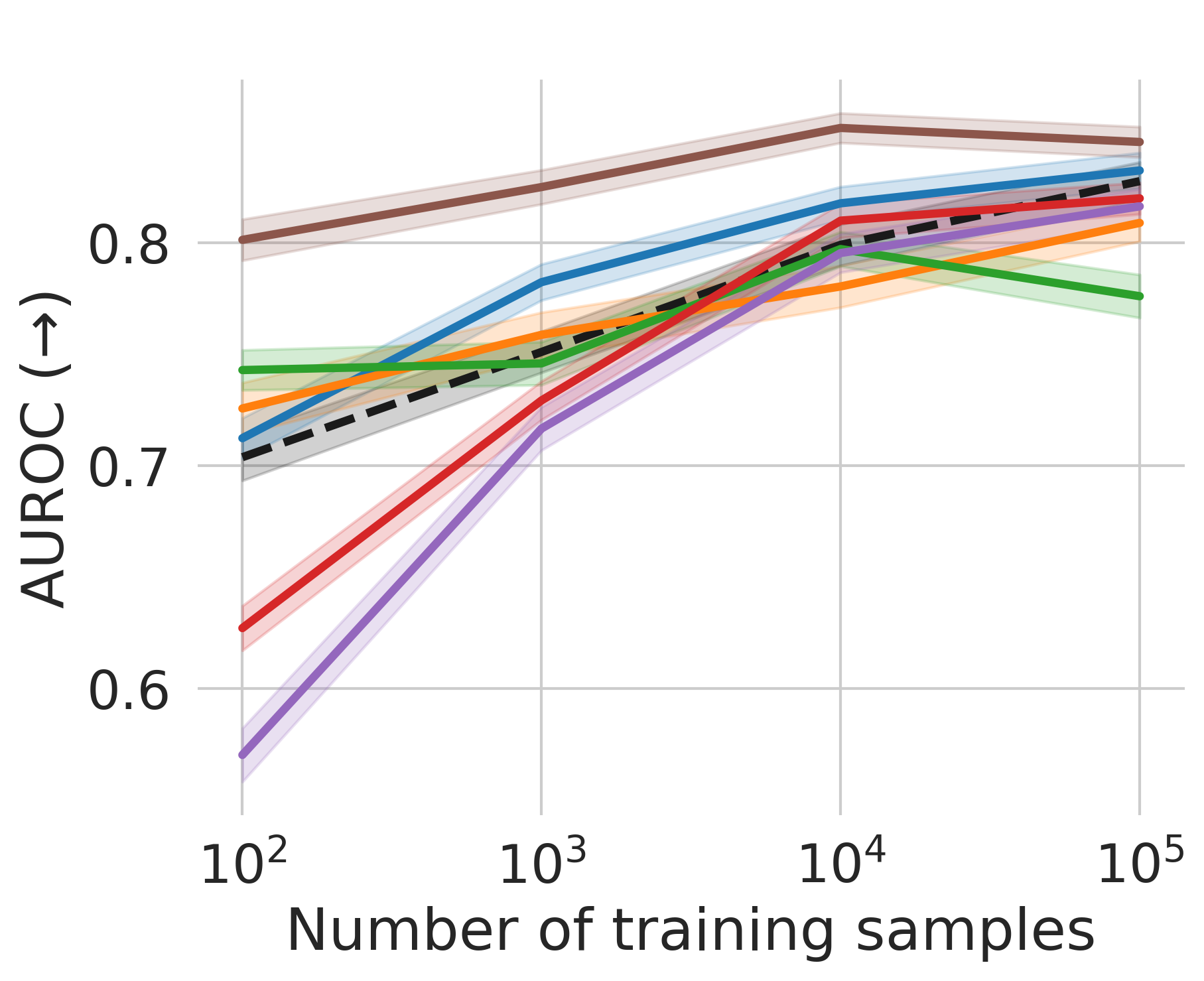}    
    \includegraphics[height=100px]{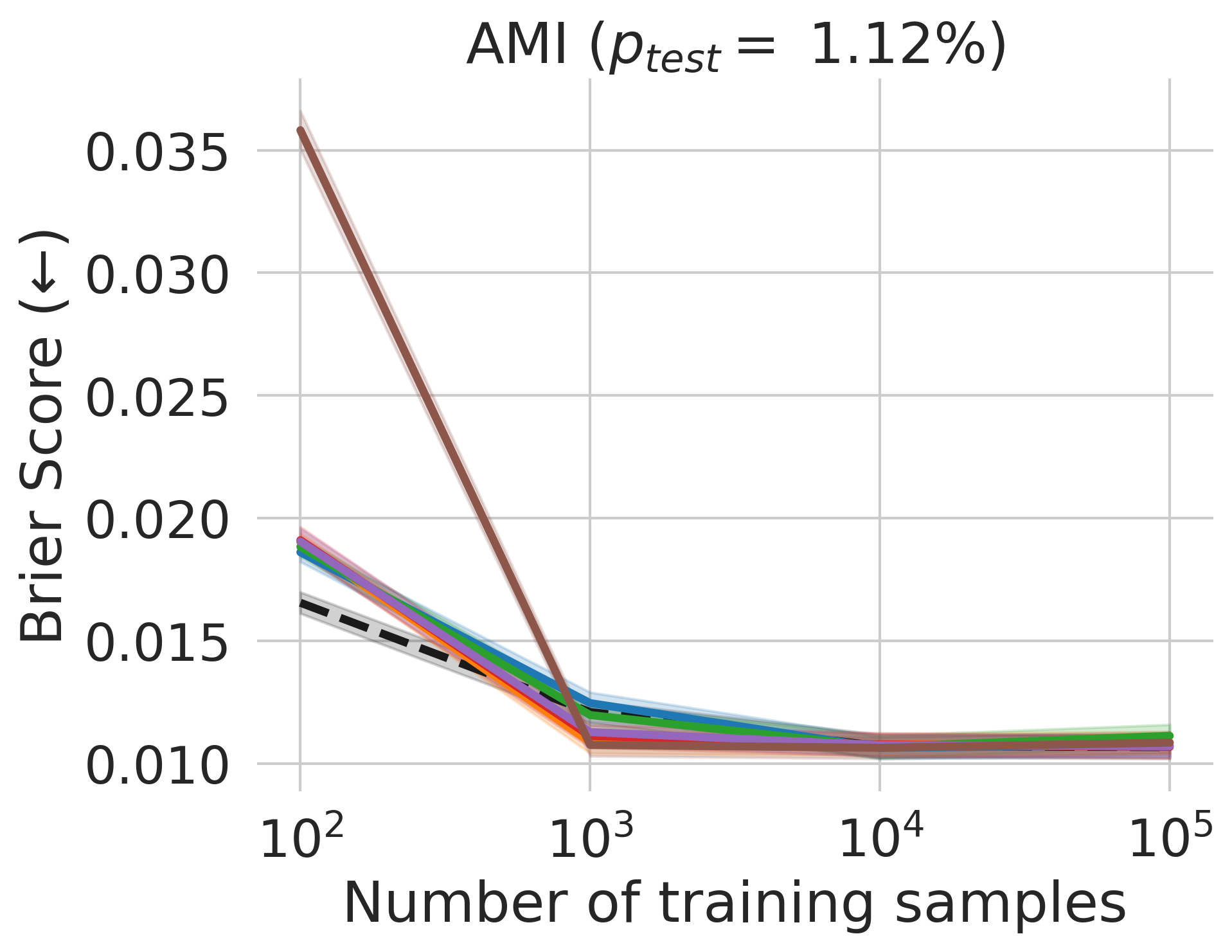}
    \includegraphics[height=100px]{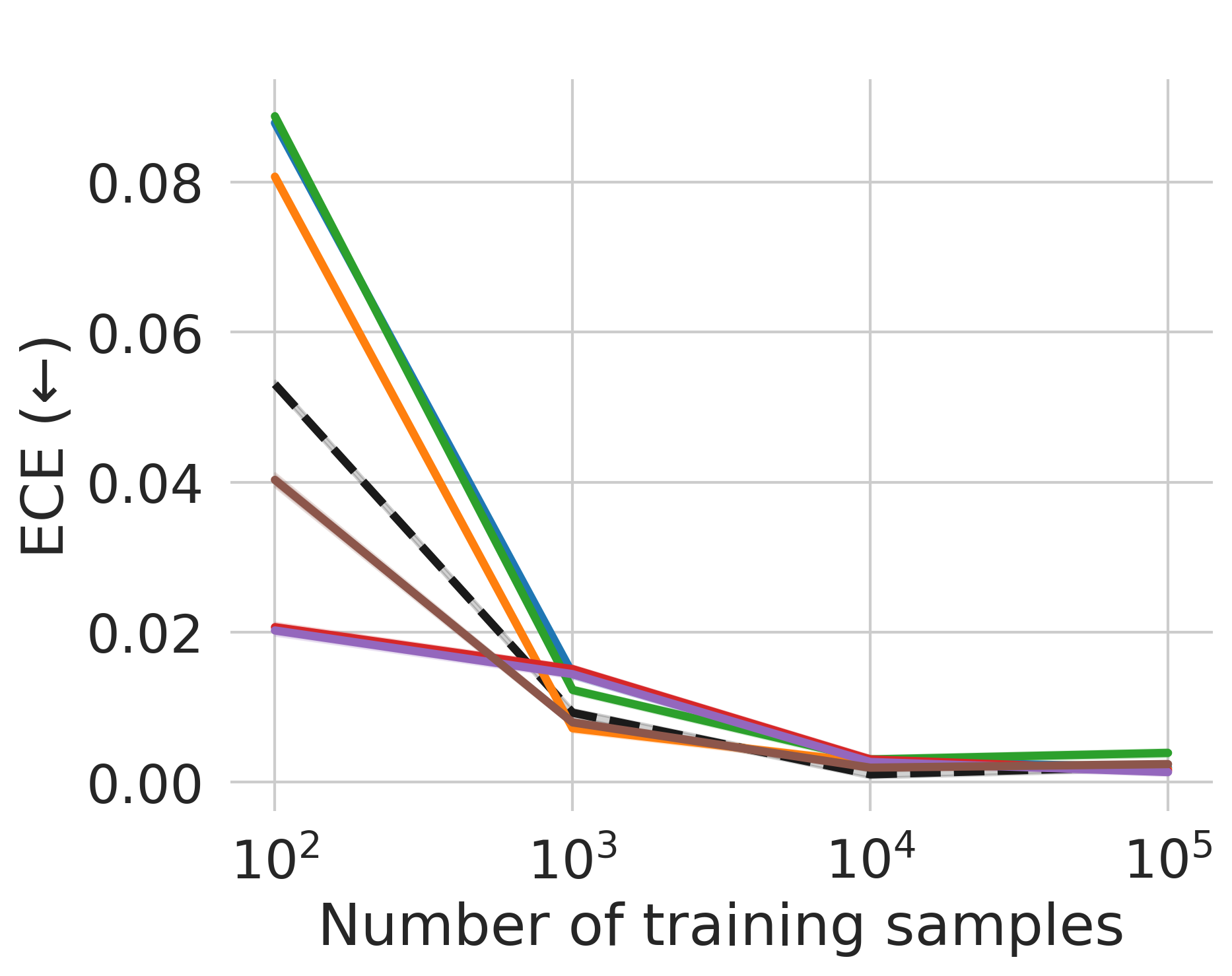}}

    \makebox[\textwidth][l]{
    \includegraphics[height=100px]{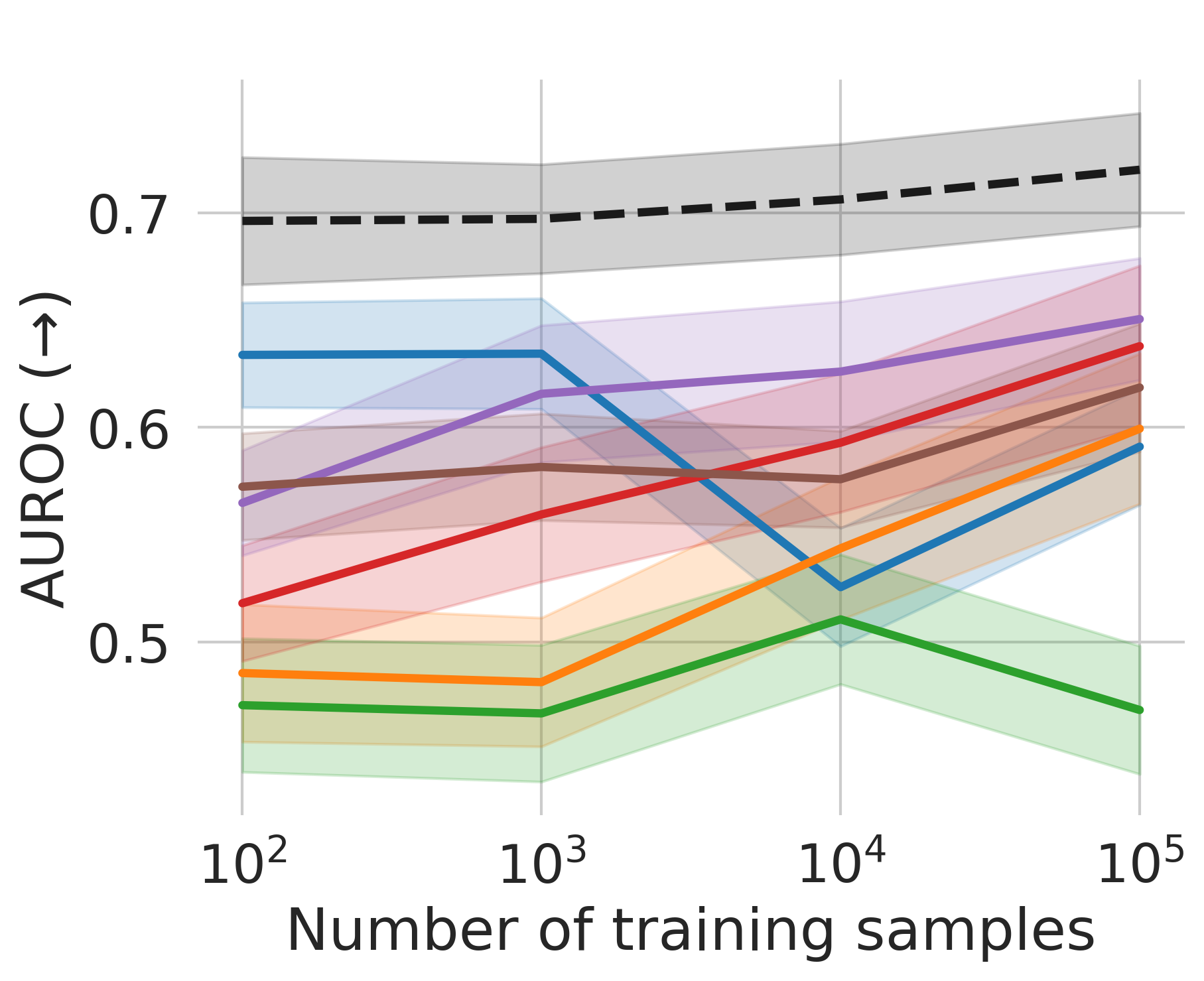}
    \includegraphics[height=100px]{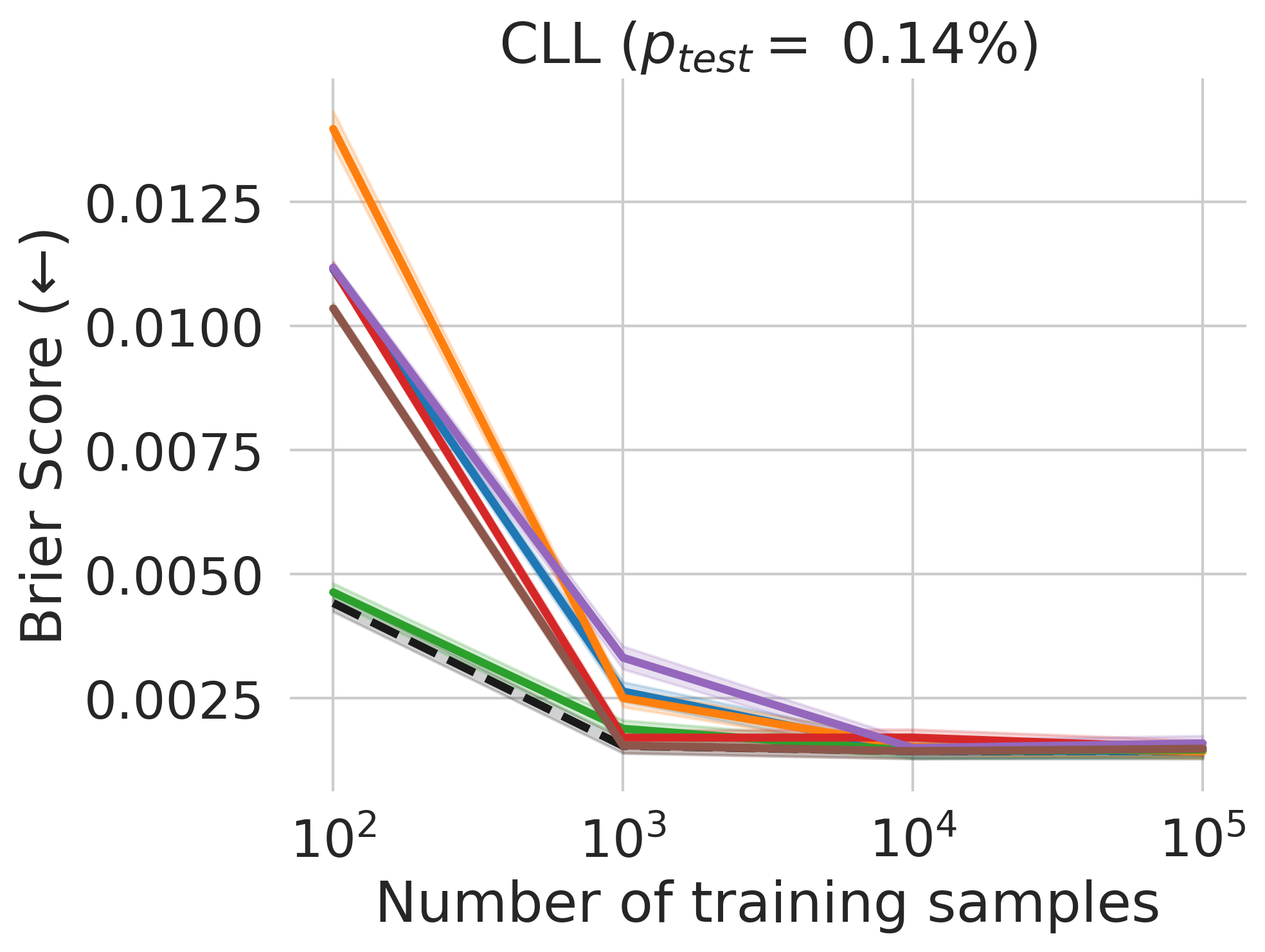}
    \includegraphics[height=100px]{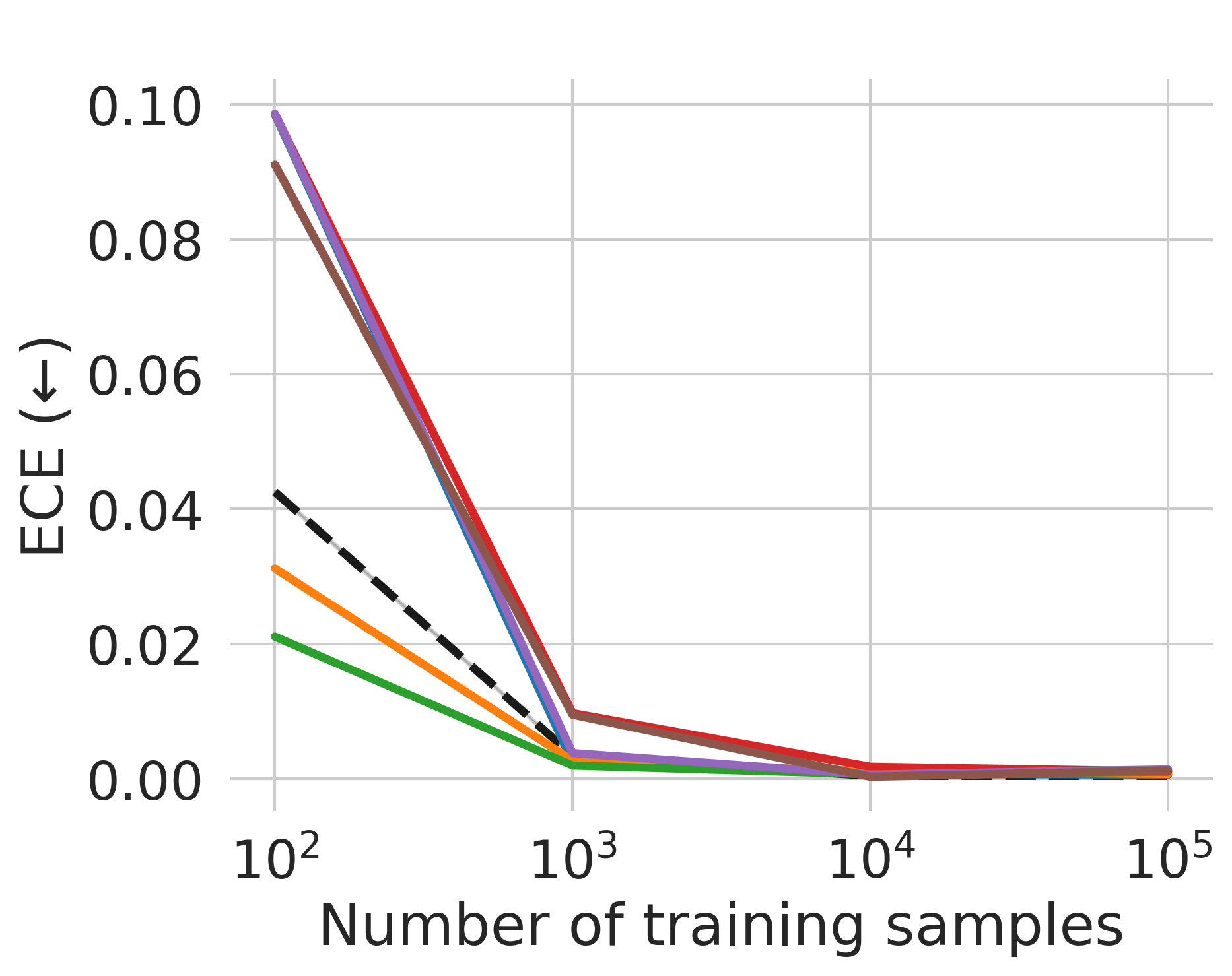}}

    \makebox[\textwidth][l]{
    \includegraphics[height=100px]{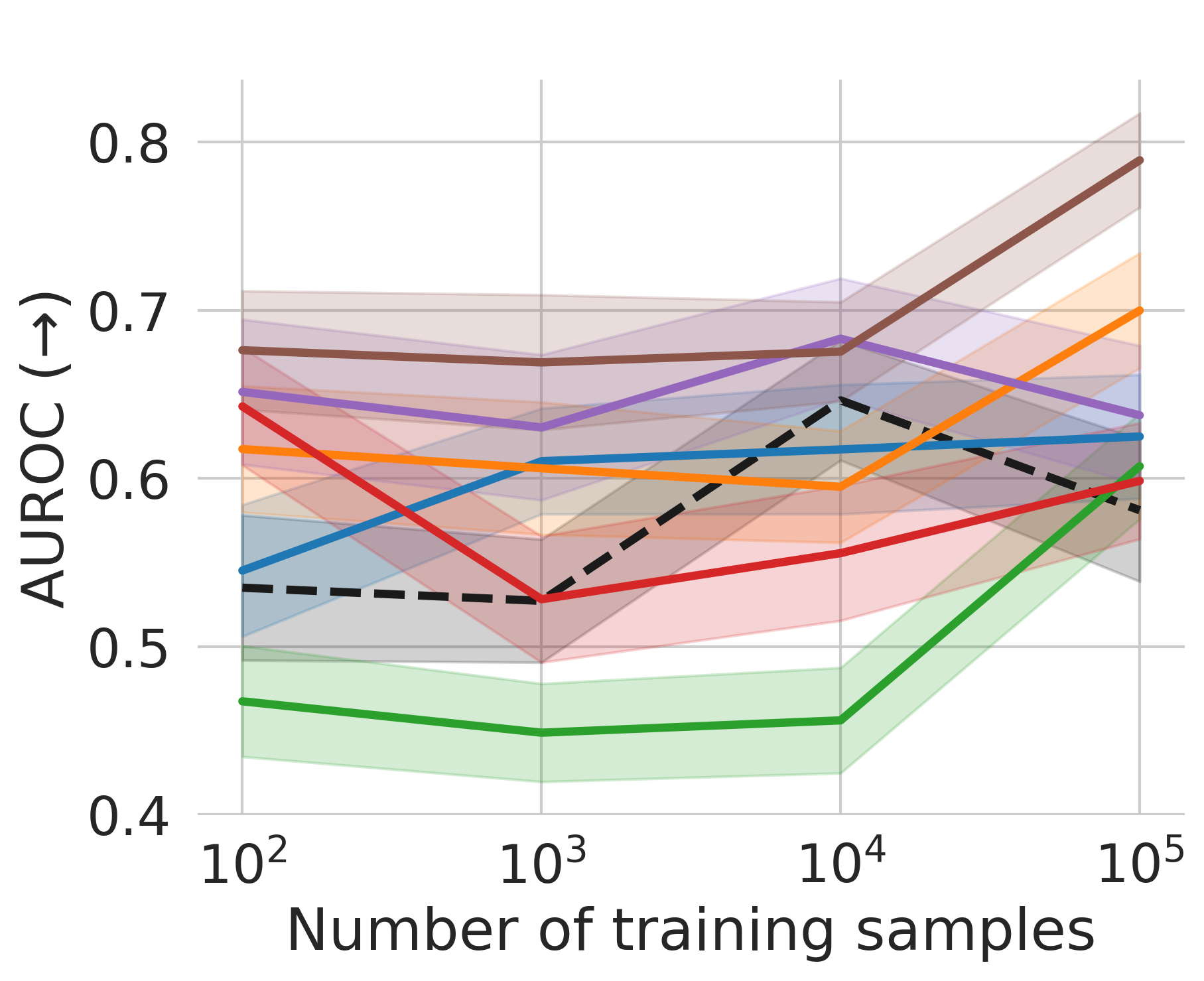}
    \includegraphics[height=100px]{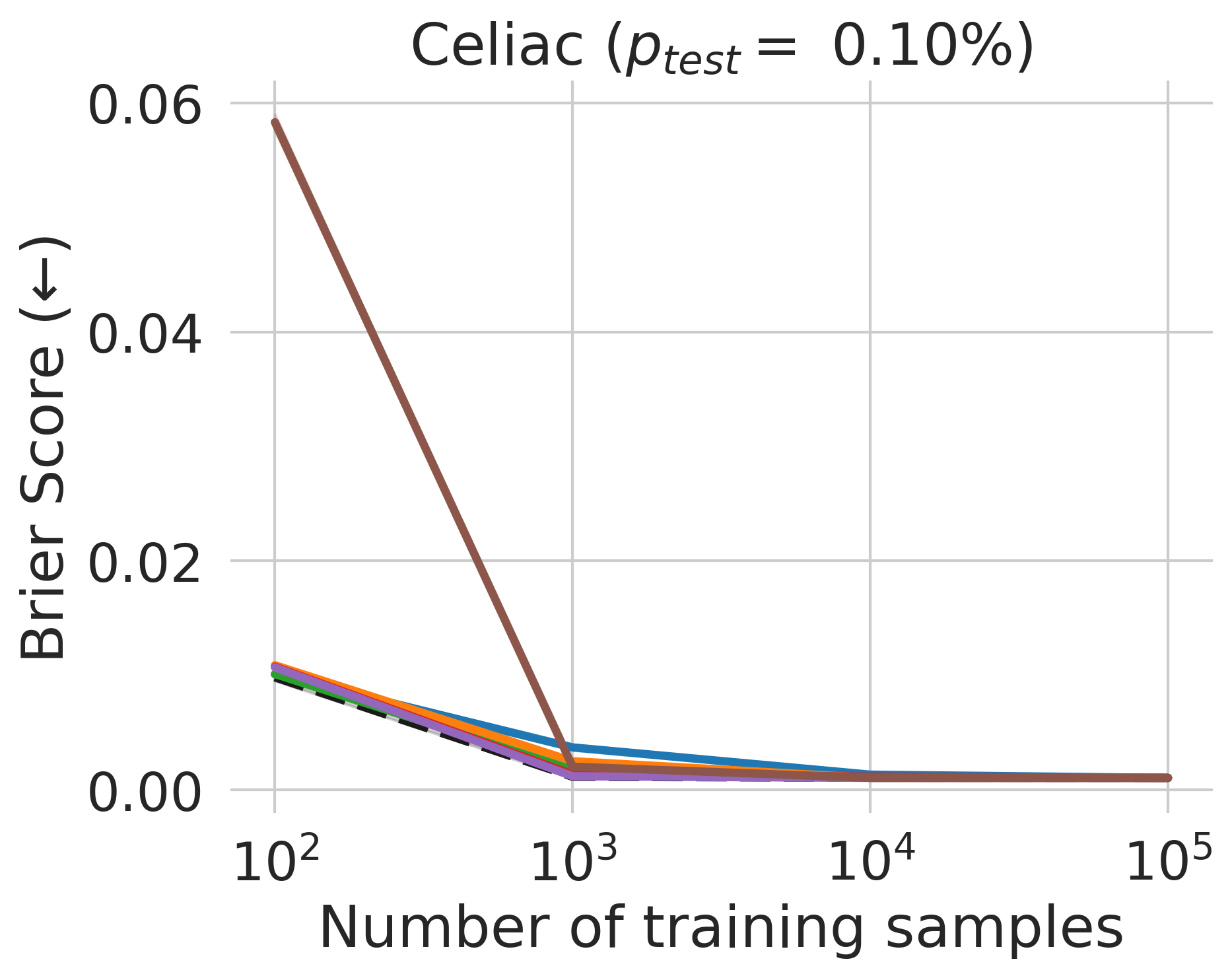}
    \includegraphics[height=100px]{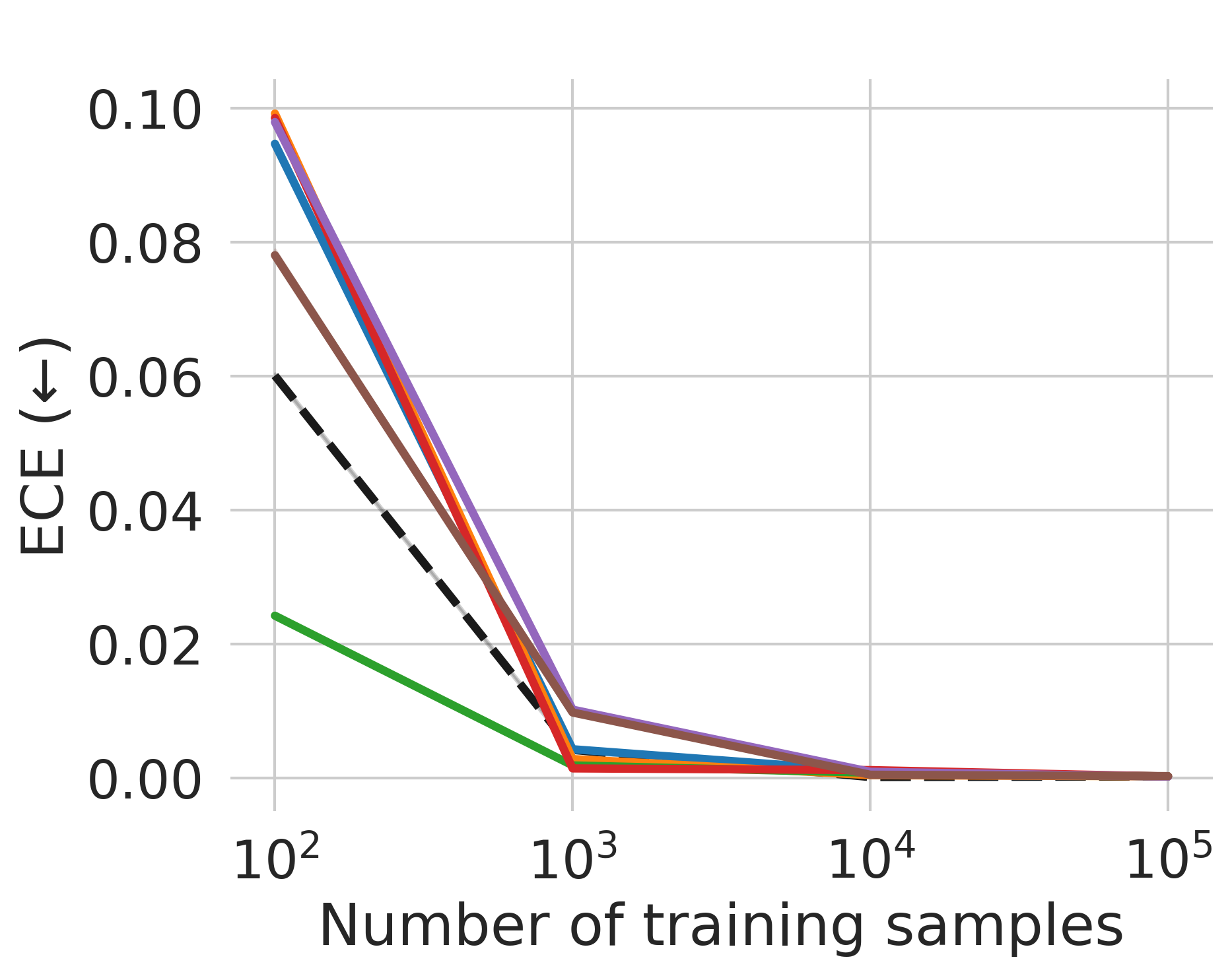}
    \includegraphics[height=100px]{figures/legend.png}}

    \makebox[\textwidth][l]{
    \includegraphics[height=100px]{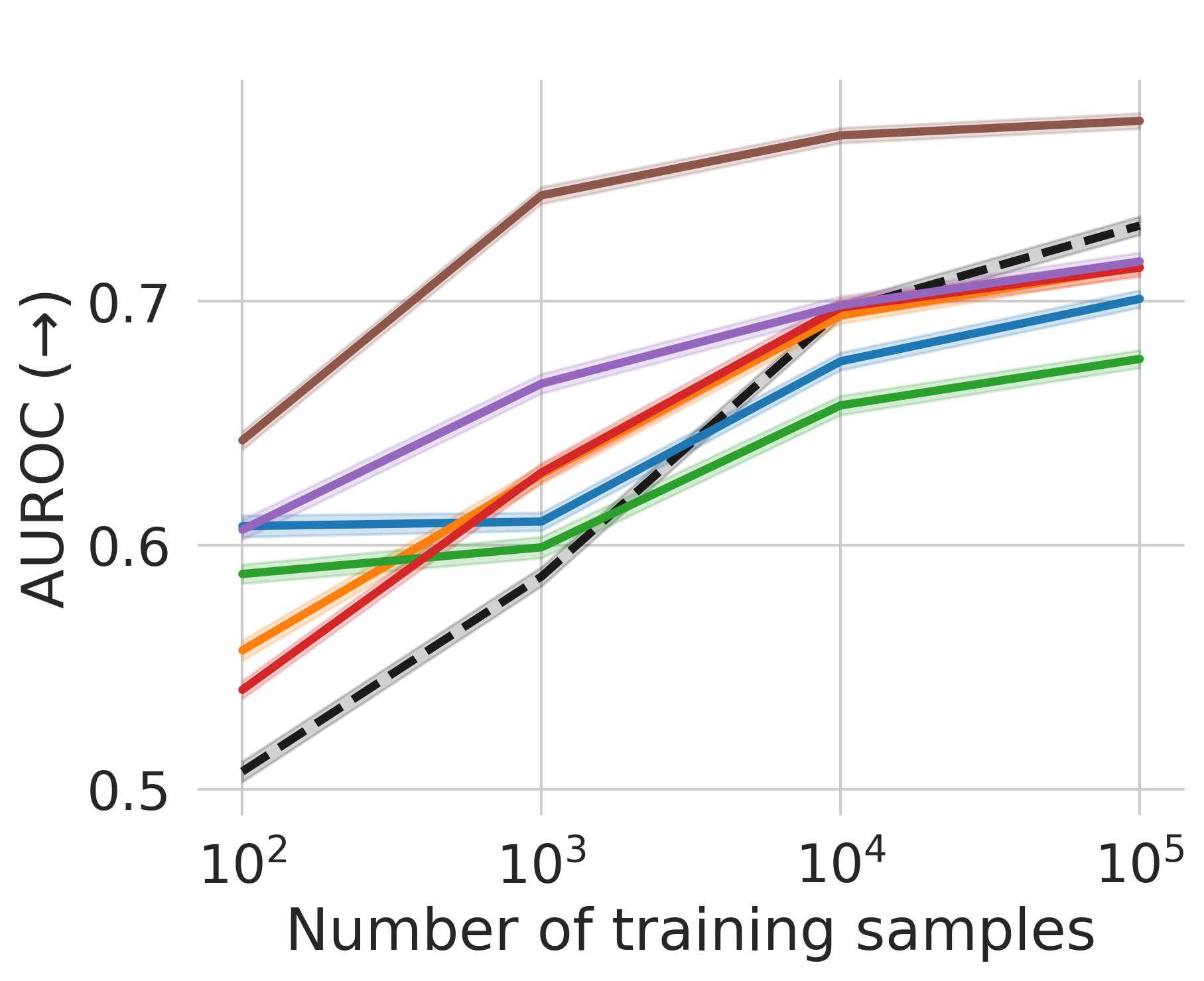}    
    \includegraphics[height=100px]{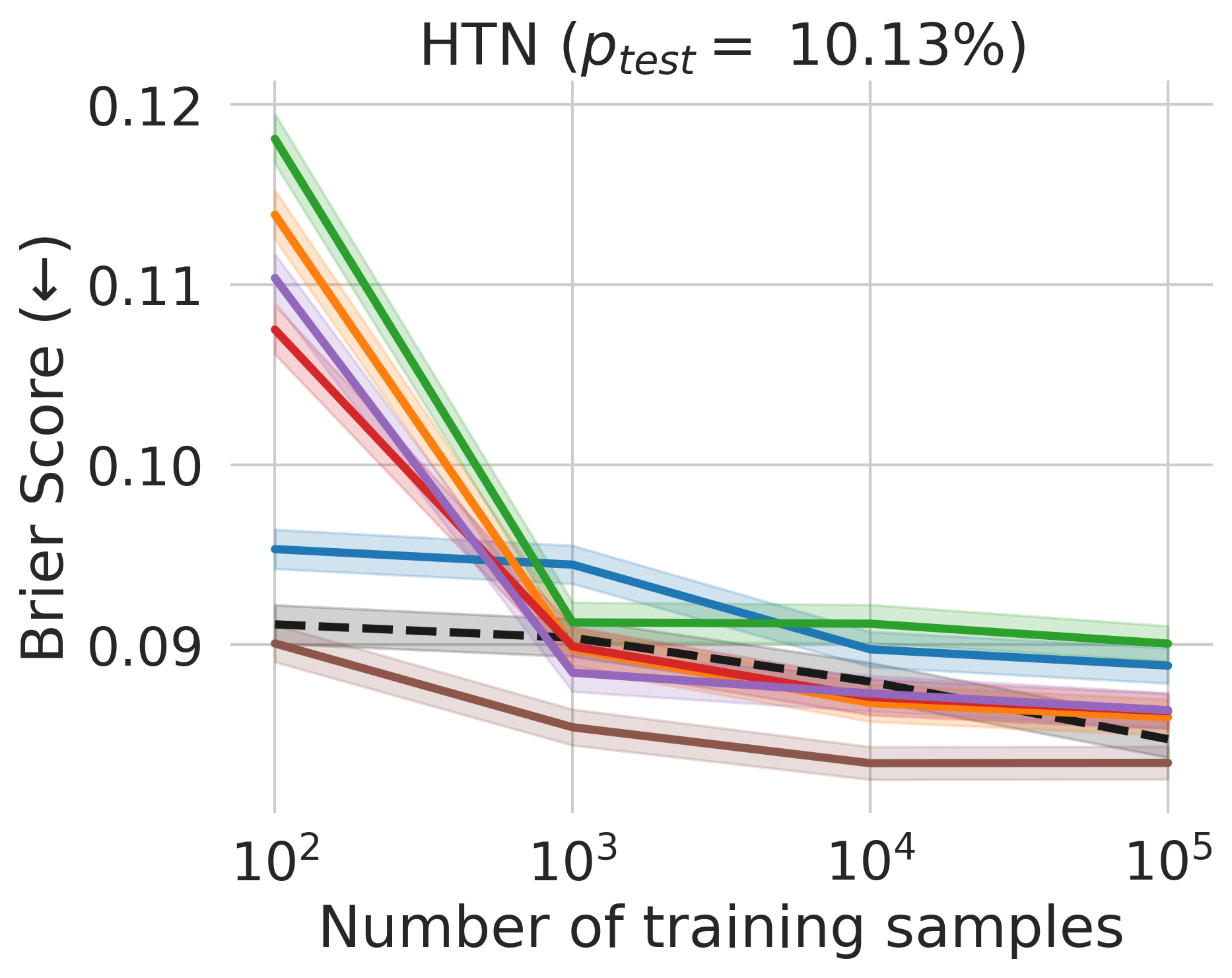}
    \includegraphics[height=100px]{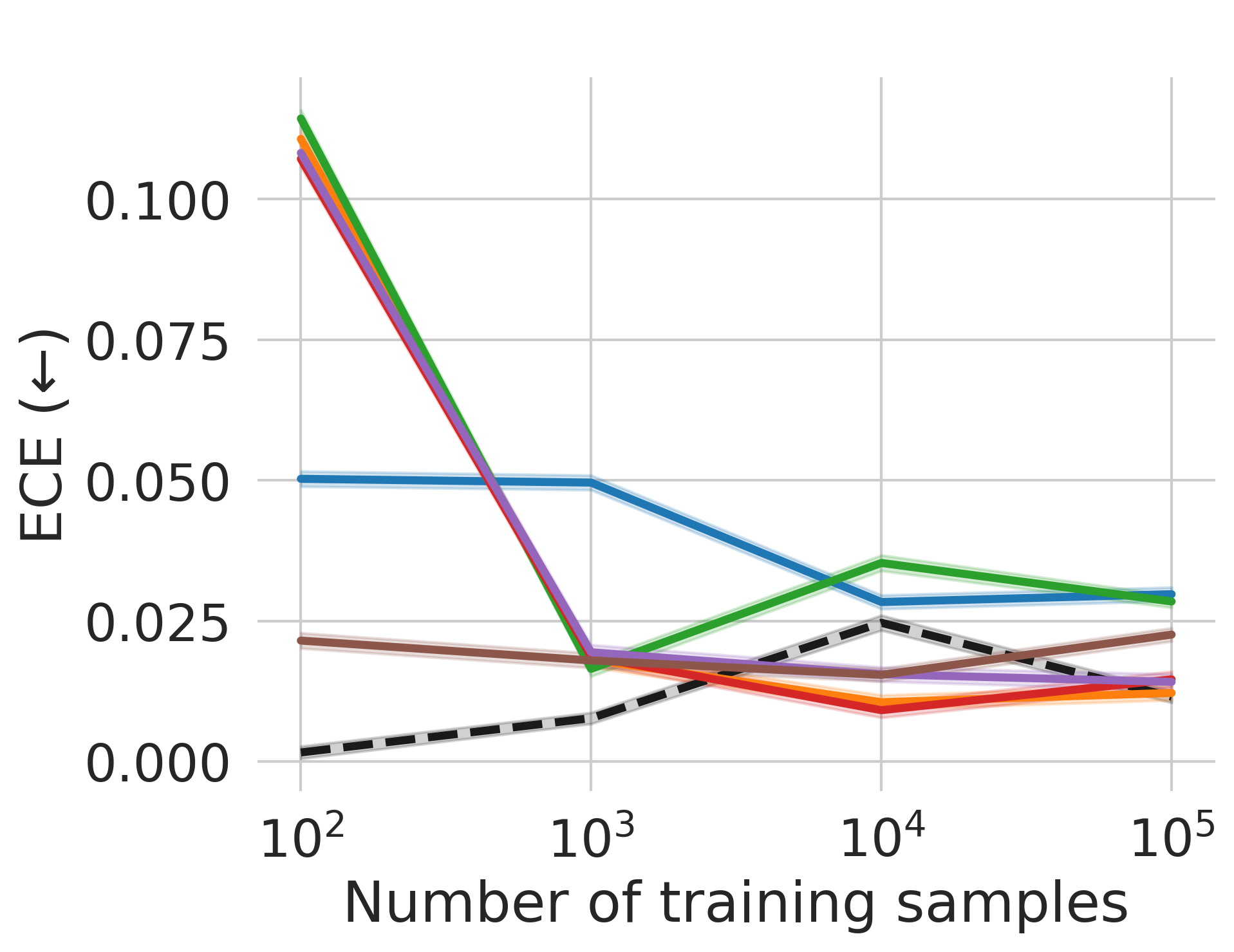}}

    \makebox[\textwidth][l]{
    \includegraphics[height=100px]{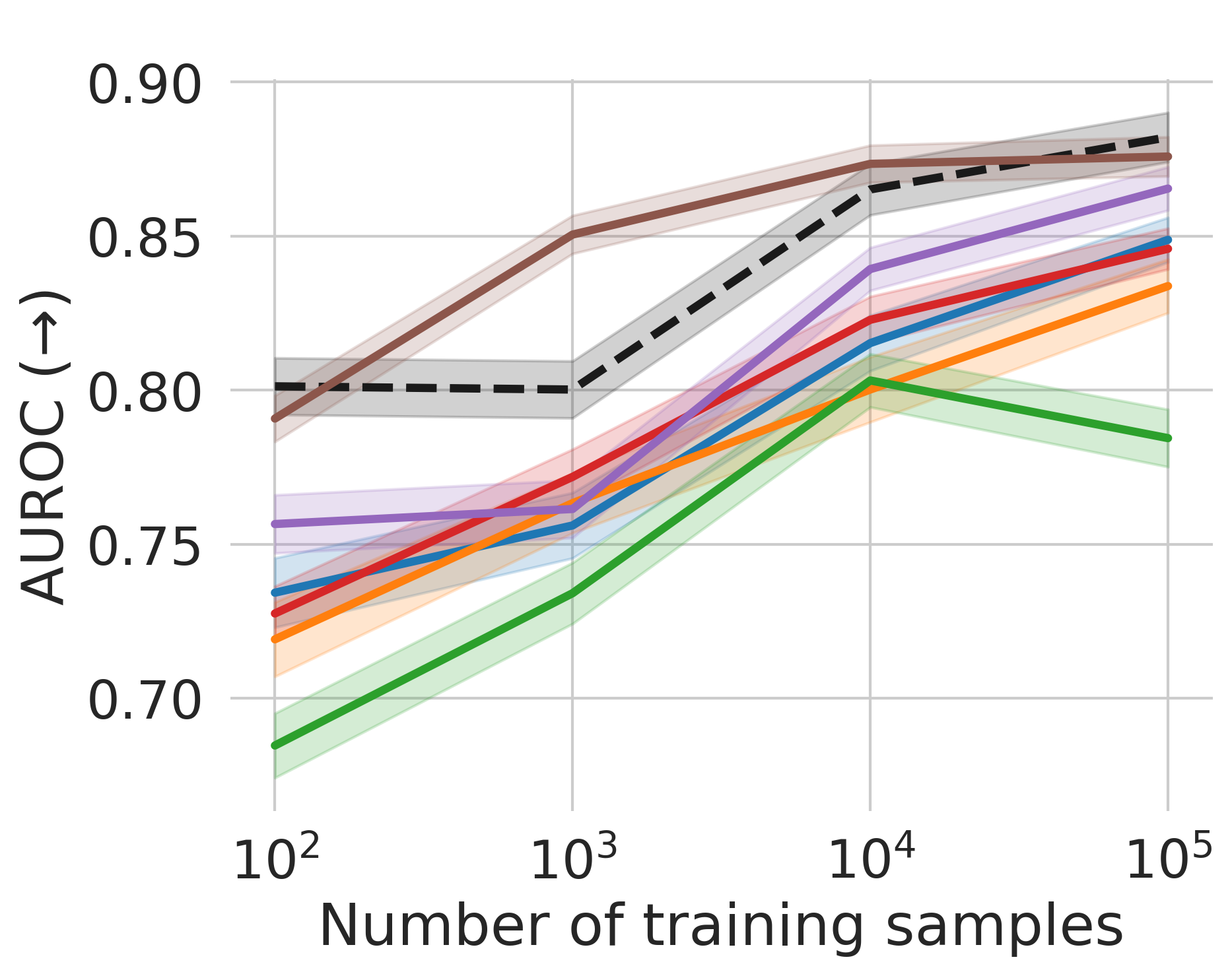}
    \includegraphics[height=100px]{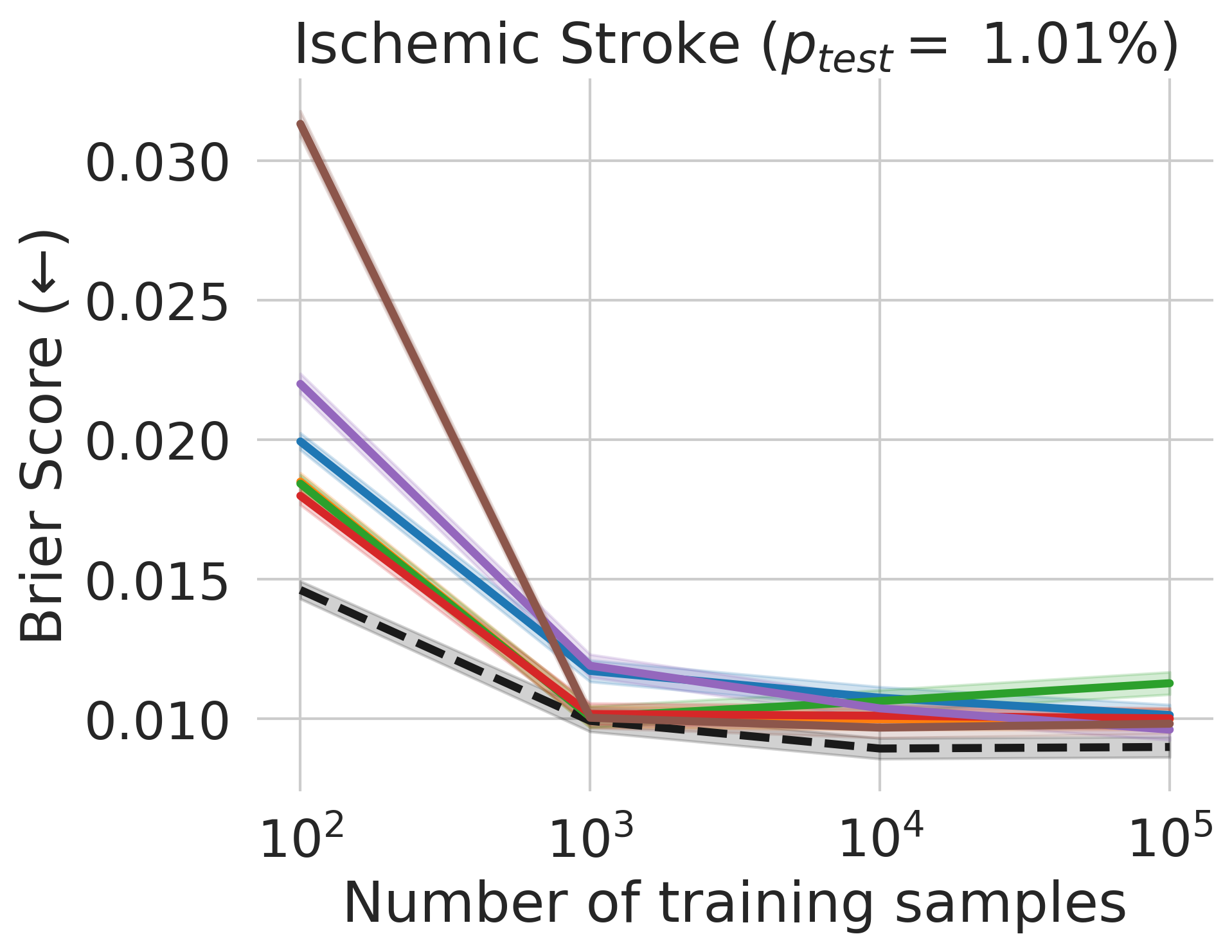}
    \includegraphics[height=100px]{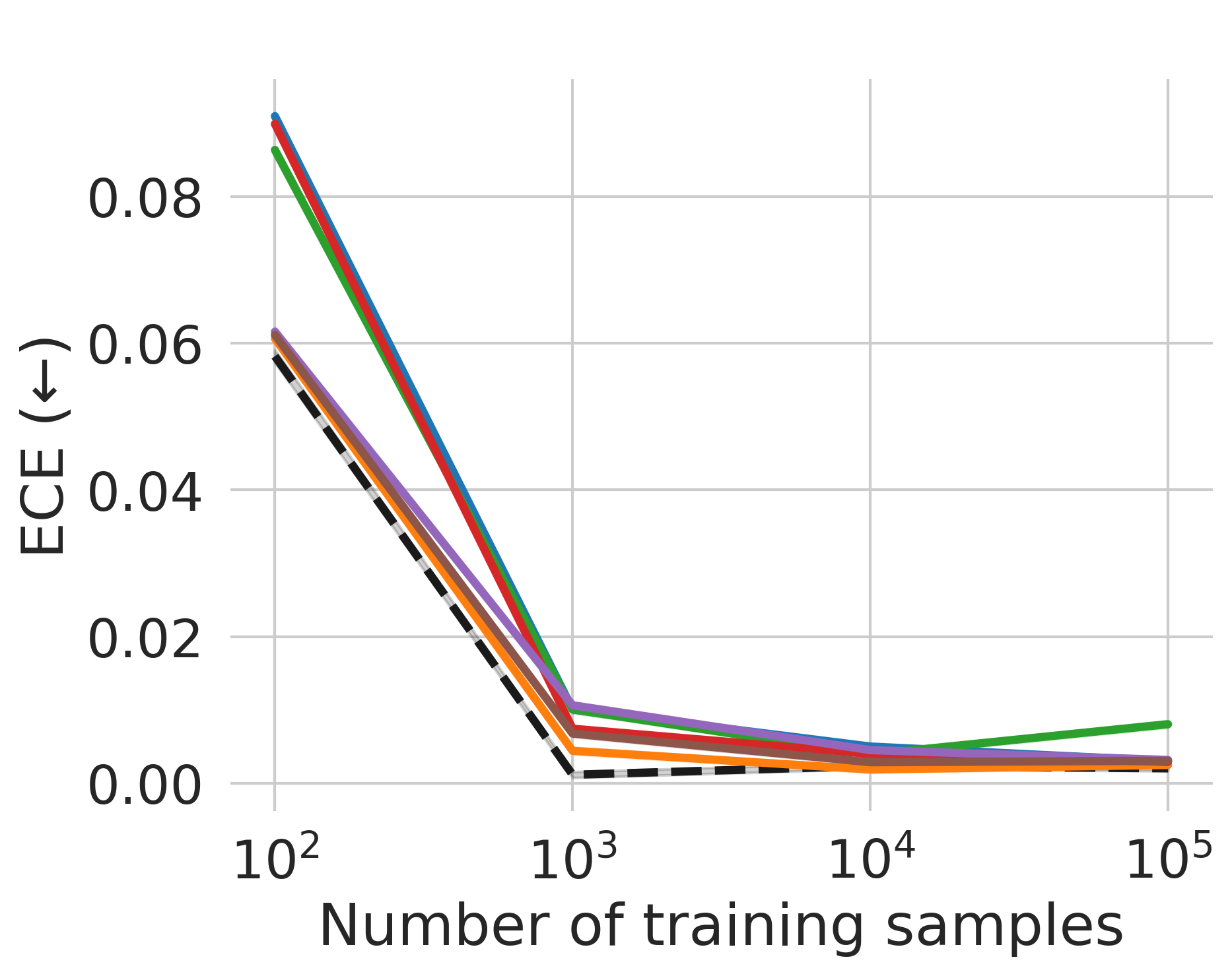}}

    \makebox[\textwidth][l]{
    \includegraphics[height=100px]{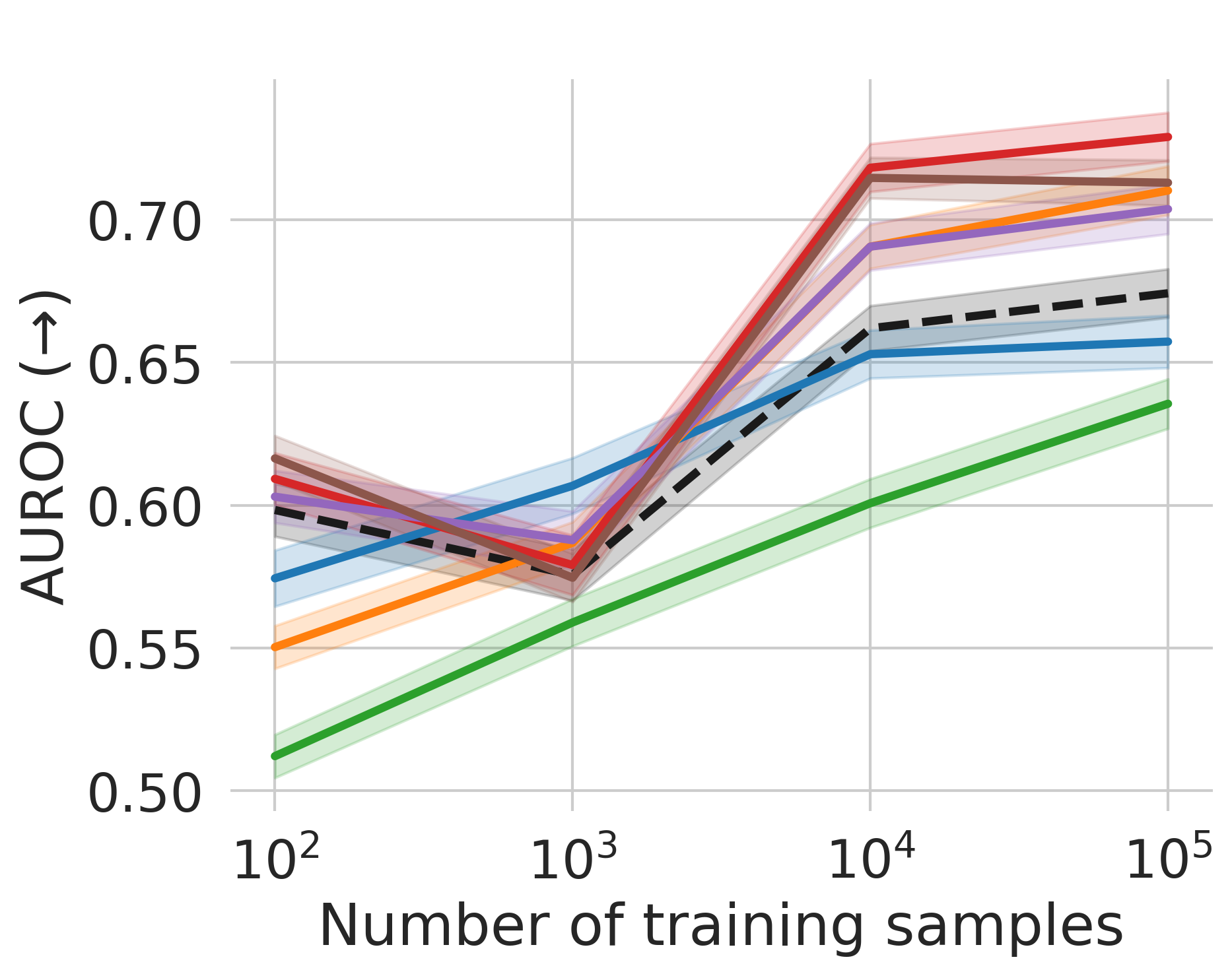}
    \includegraphics[height=100px]{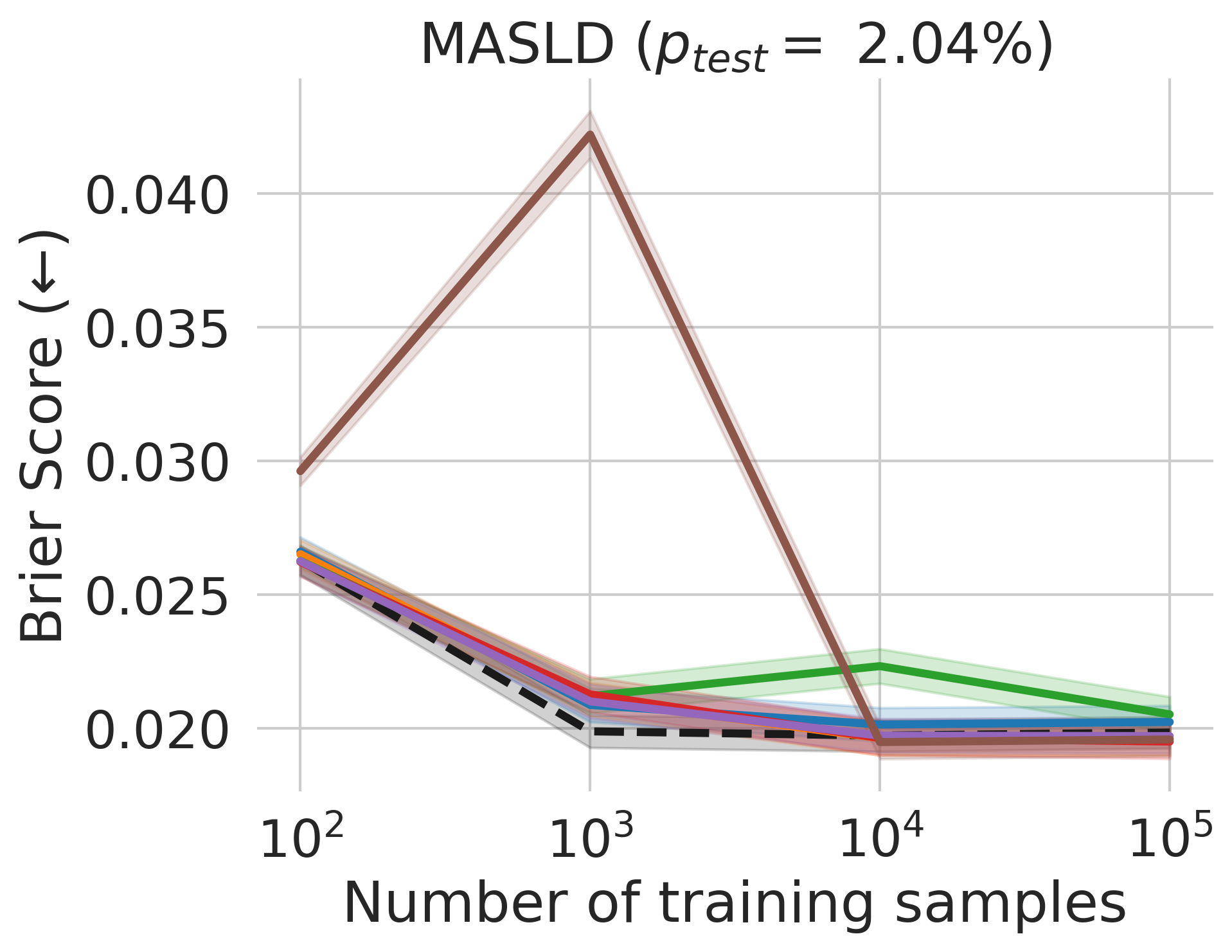}
    \includegraphics[height=100px]{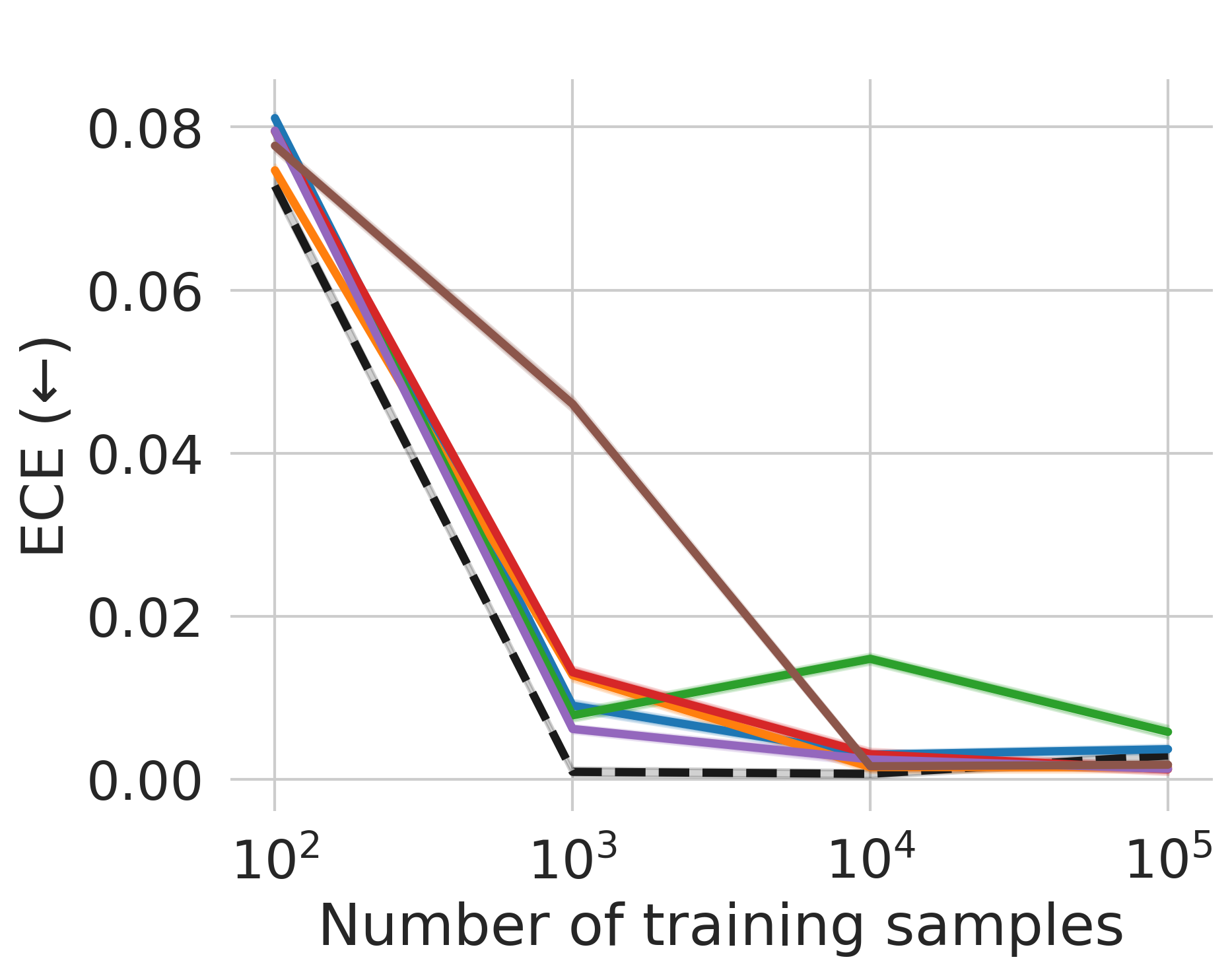}}

    \centering
    \caption{Discriminative performances and calibration results for all phenotypes for an increasing number of training points (Shaded area represents bootstrapped standard deviation).}
    \label{fig:all_phenotypes:cumc}
\end{figure}

\begin{figure}[ht]
    \makebox[\textwidth][l]{
    \includegraphics[height=100px]{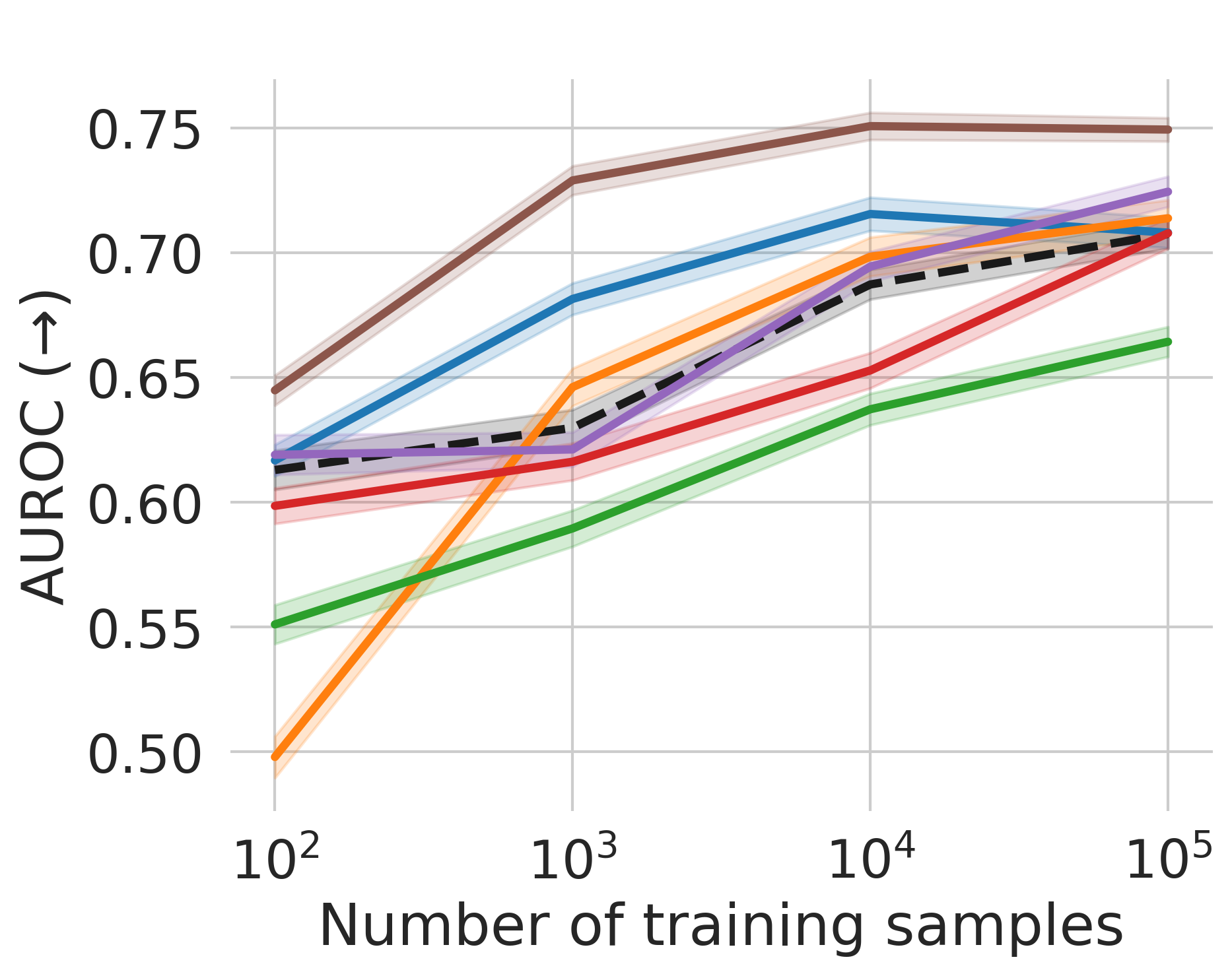}    
    \includegraphics[height=100px]{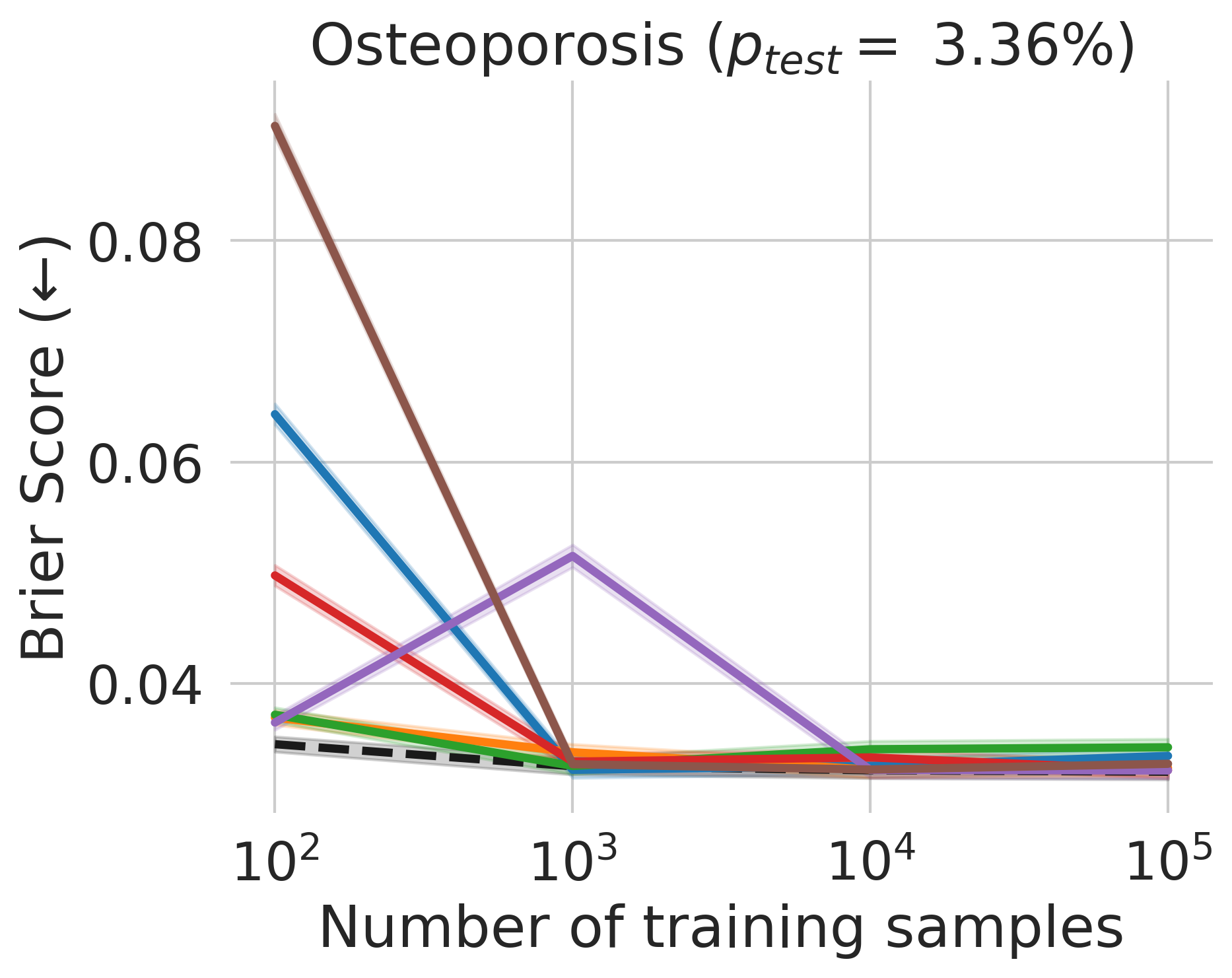}
    \includegraphics[height=100px]{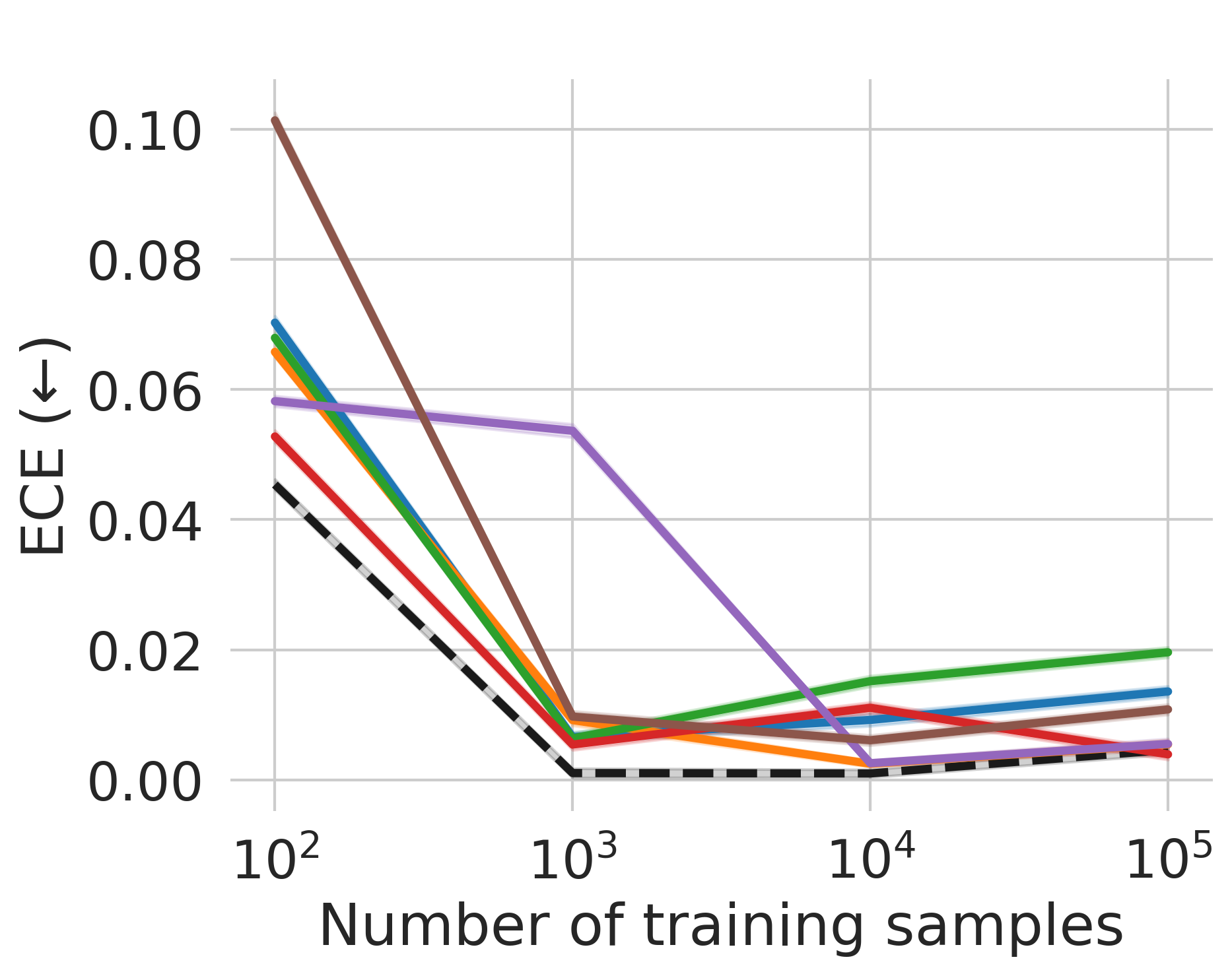}}

    \makebox[\textwidth][l]{
    \includegraphics[height=100px]{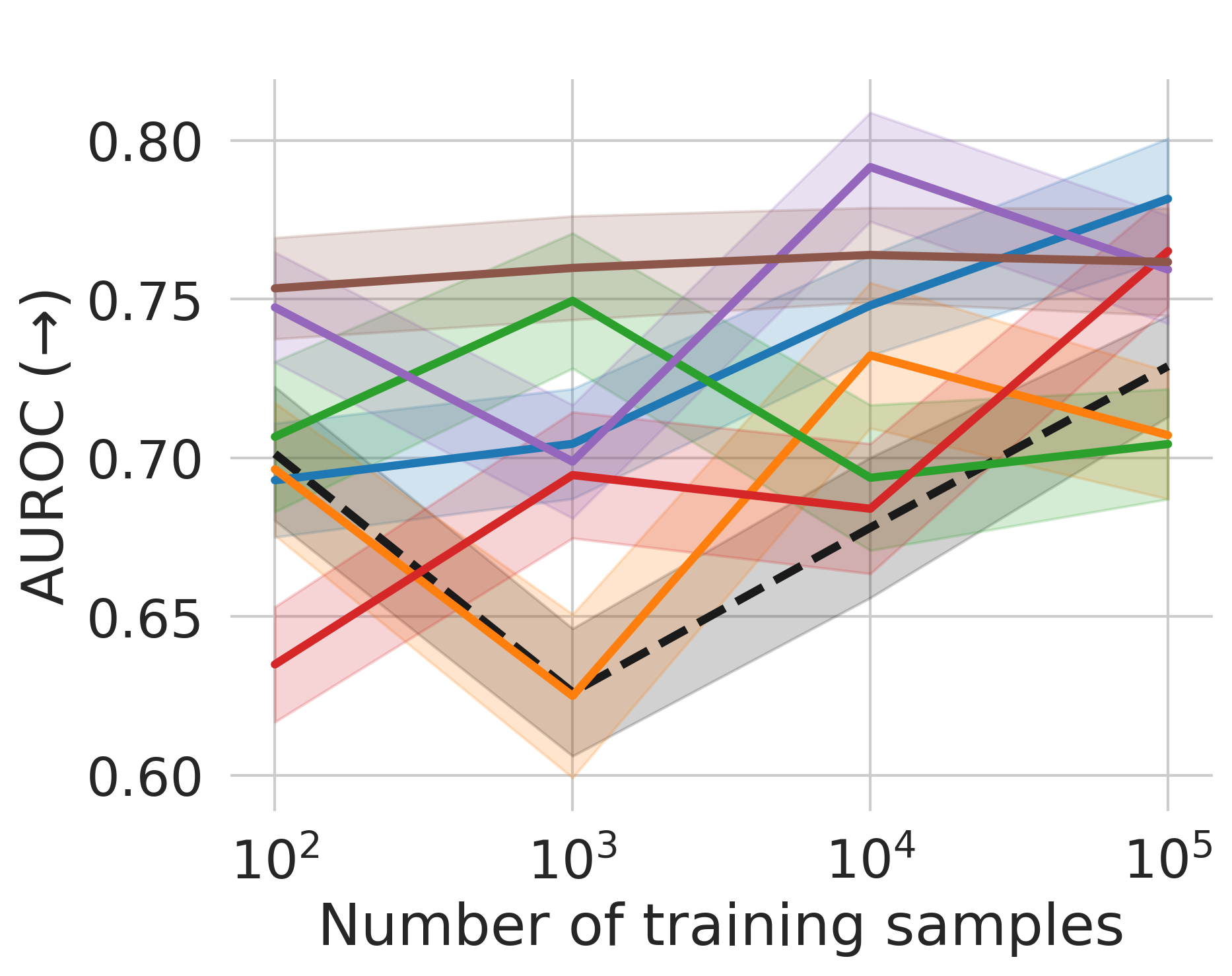}
    \includegraphics[height=100px]{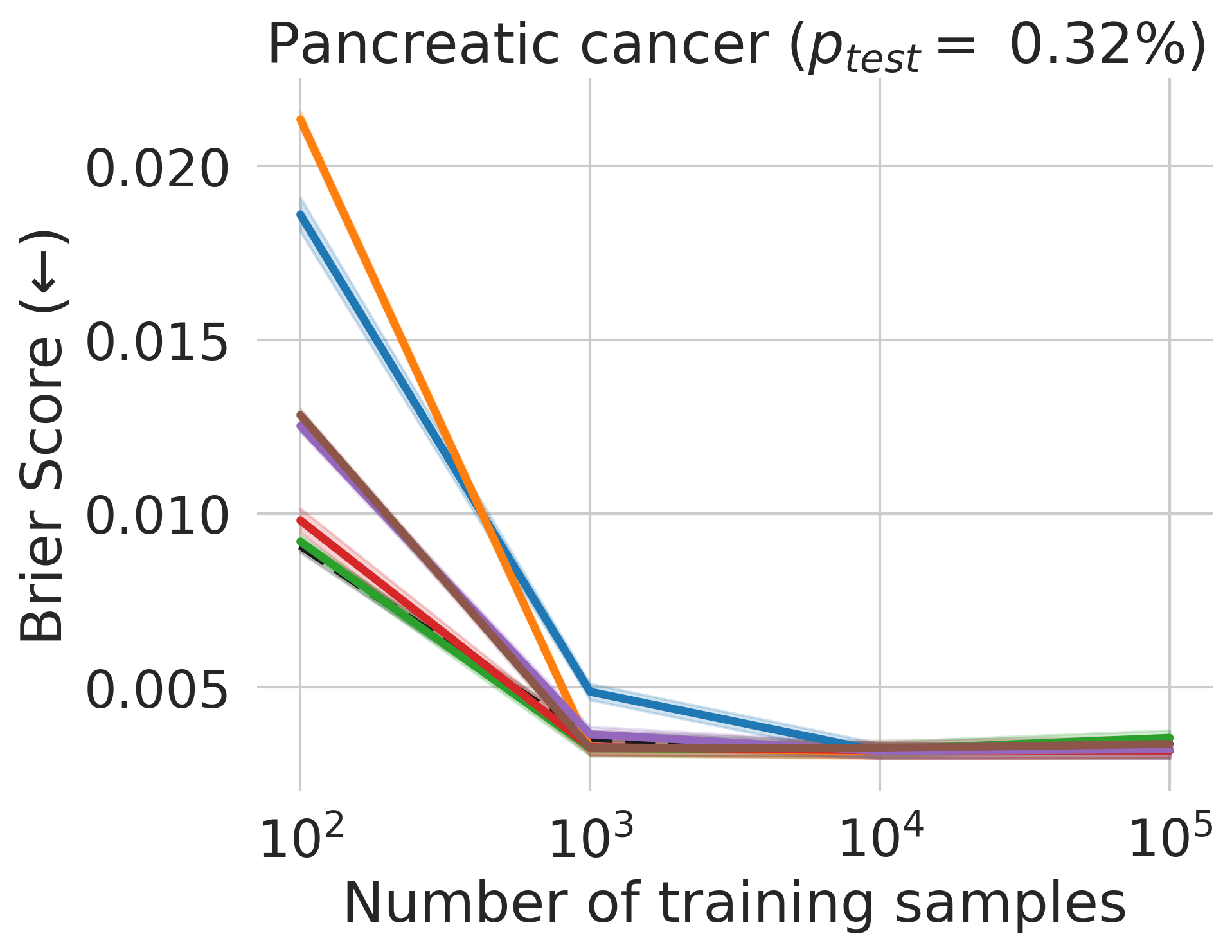}
    \includegraphics[height=100px]{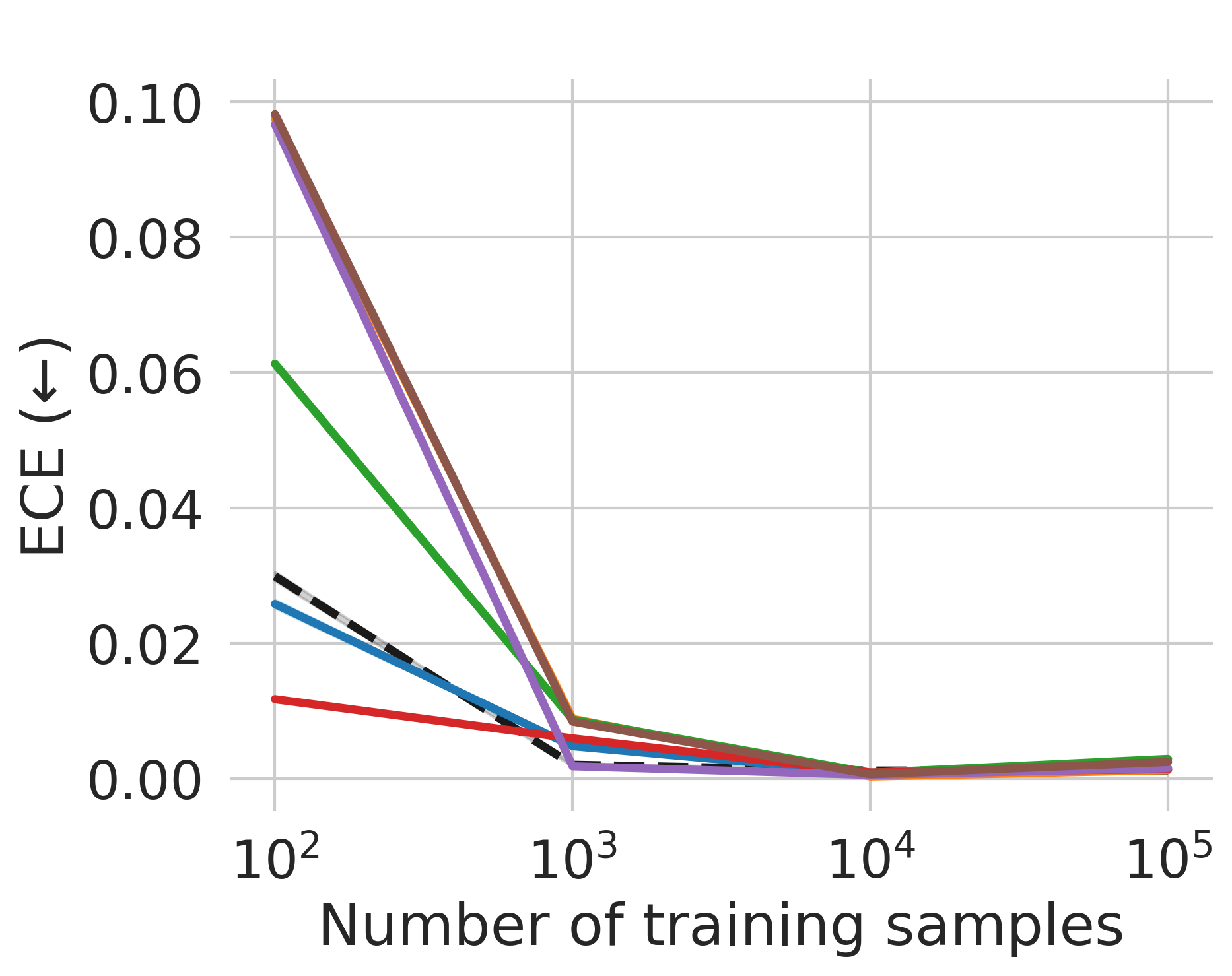}}

    \makebox[\textwidth][l]{
    \includegraphics[height=100px]{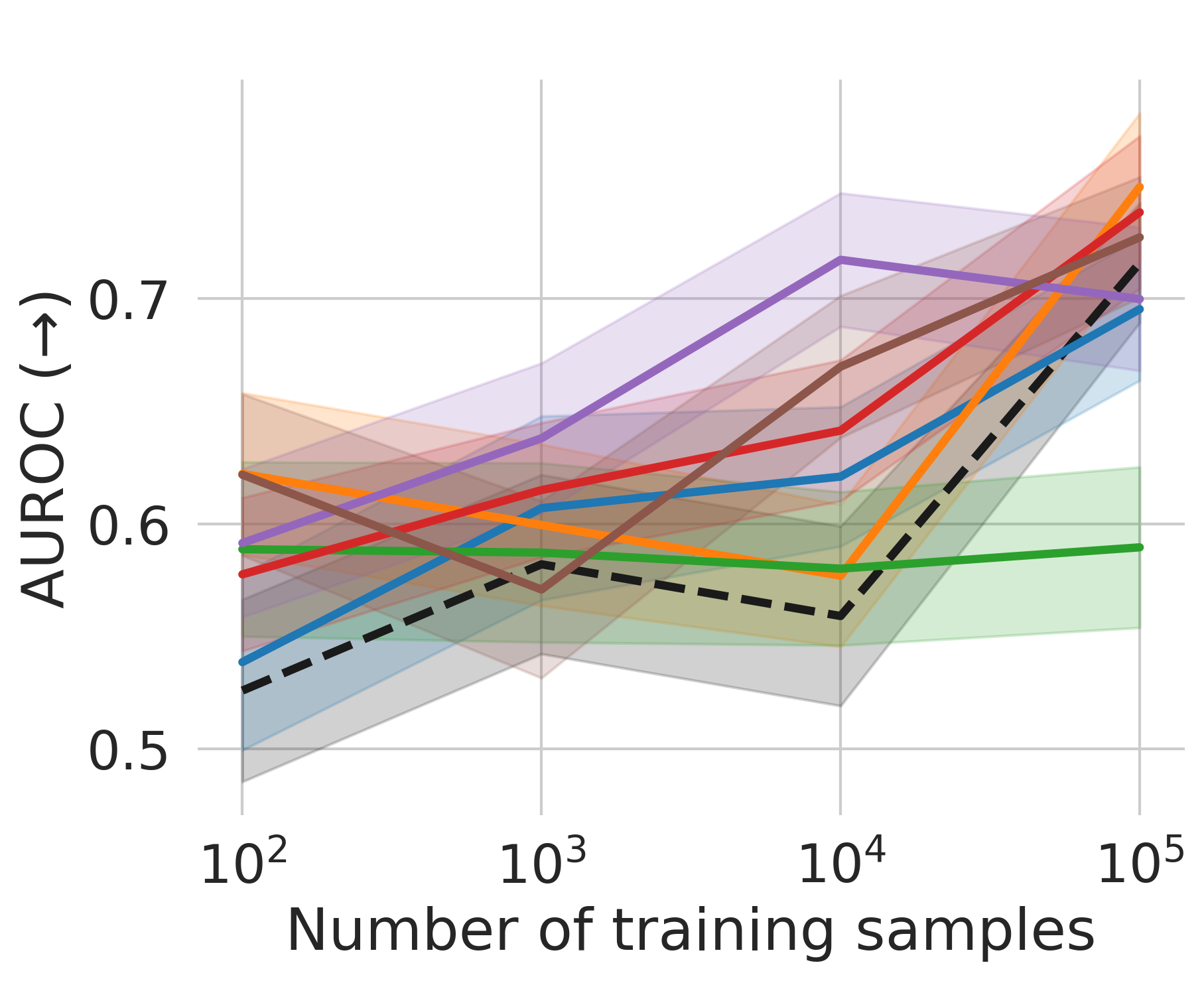}
    \includegraphics[height=100px]{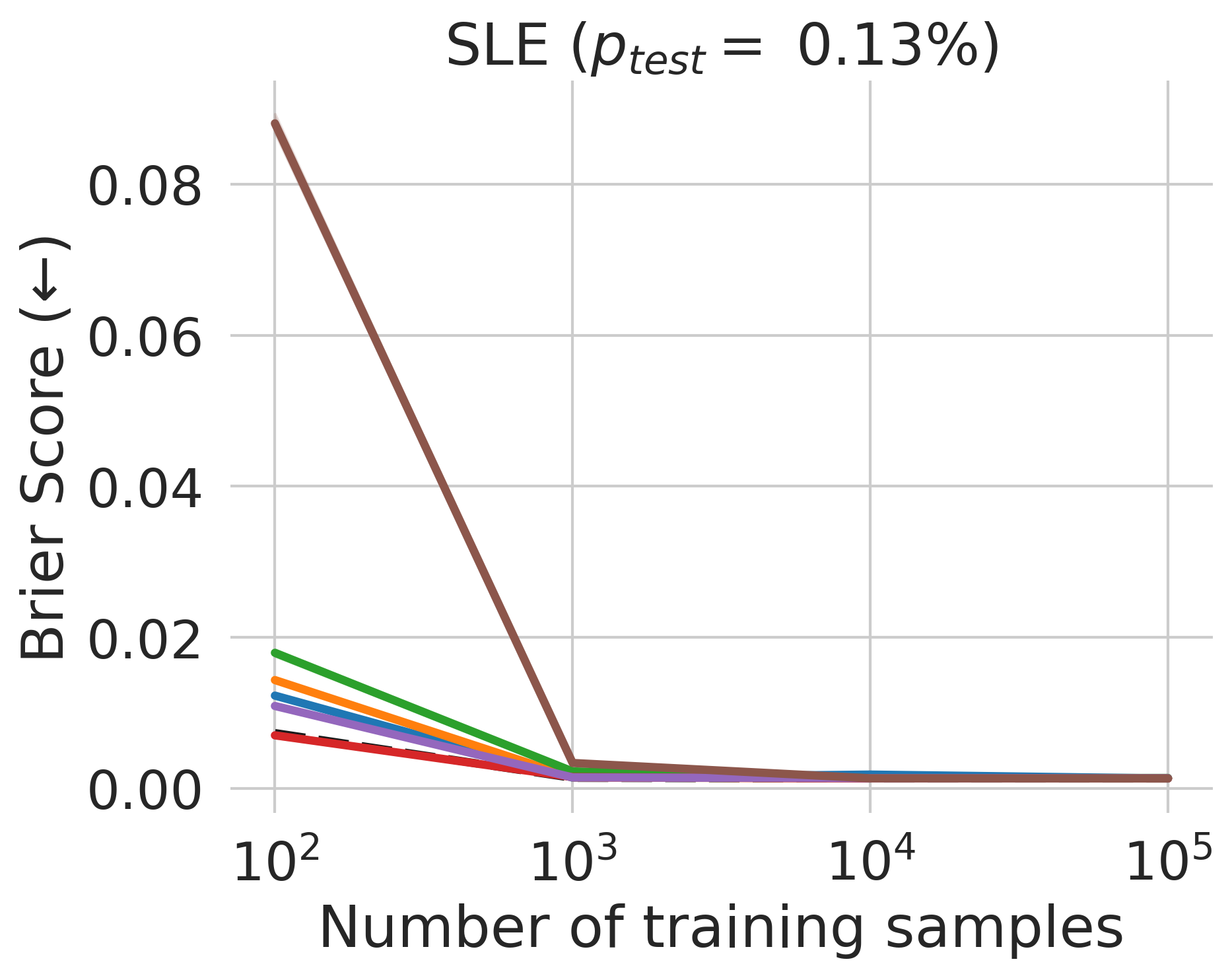}
    \includegraphics[height=100px]{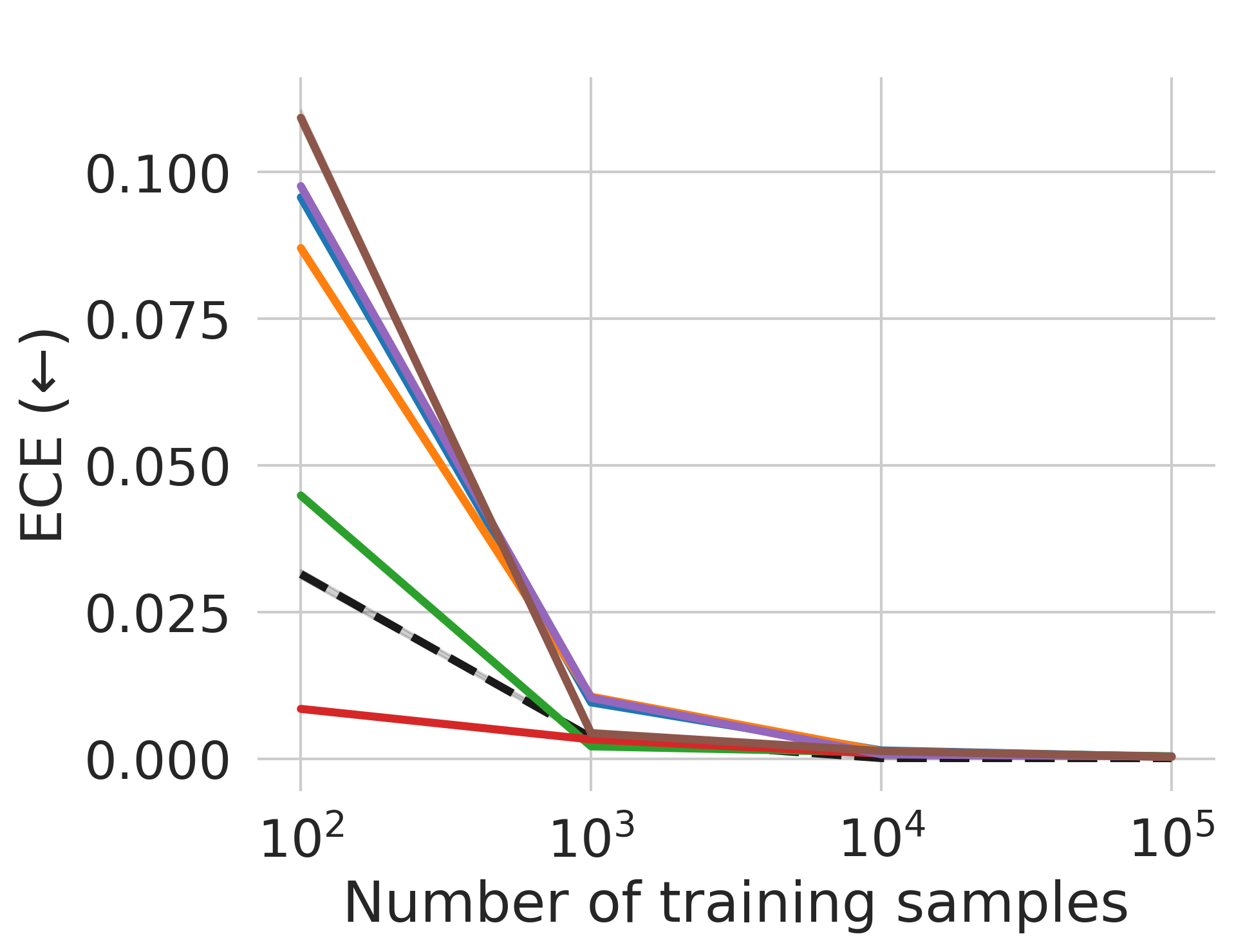}
    \includegraphics[height=100px]{figures/legend.png}}

    \makebox[\textwidth][l]{
    \includegraphics[height=100px]{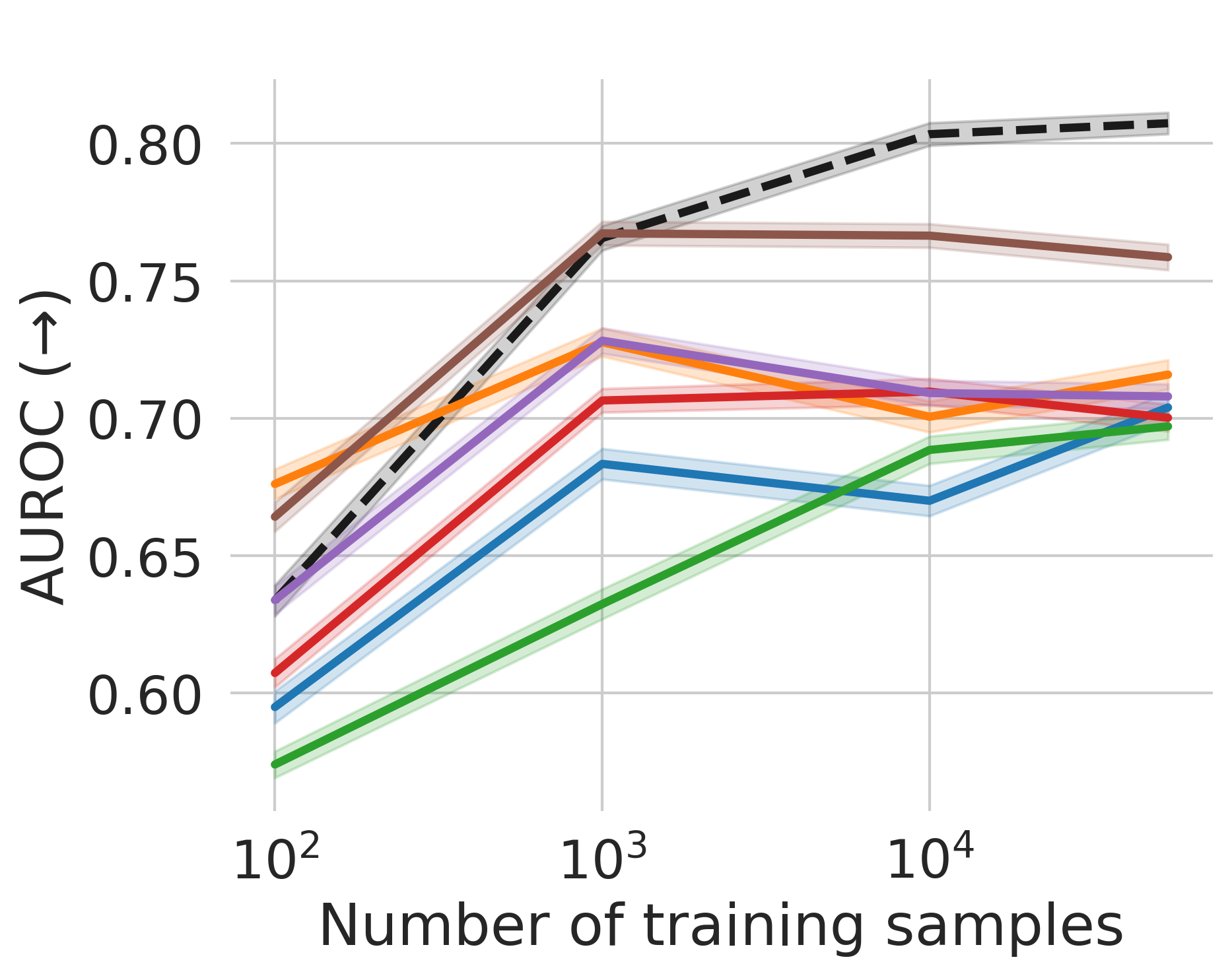}    
    \includegraphics[height=100px]{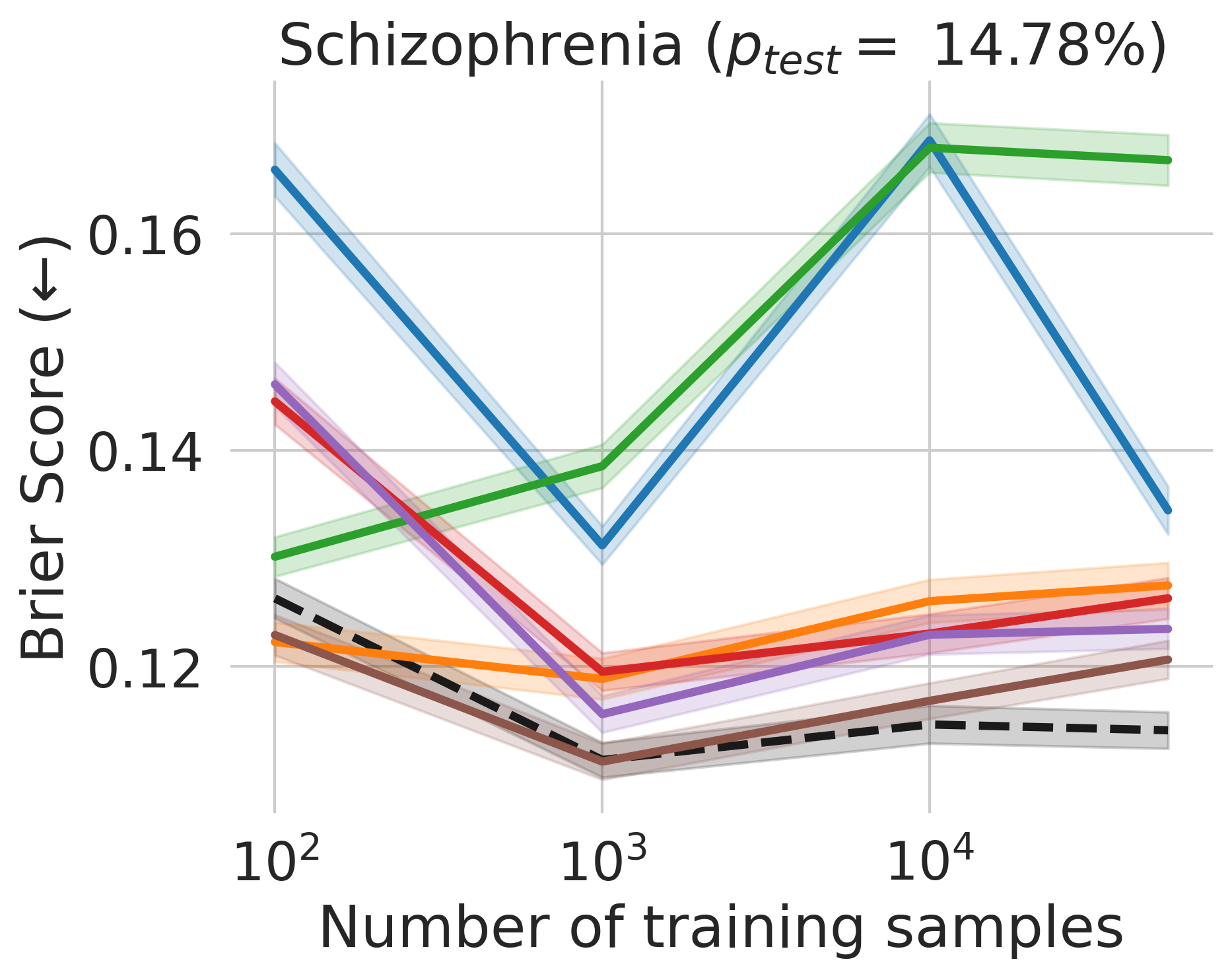}
    \includegraphics[height=100px]{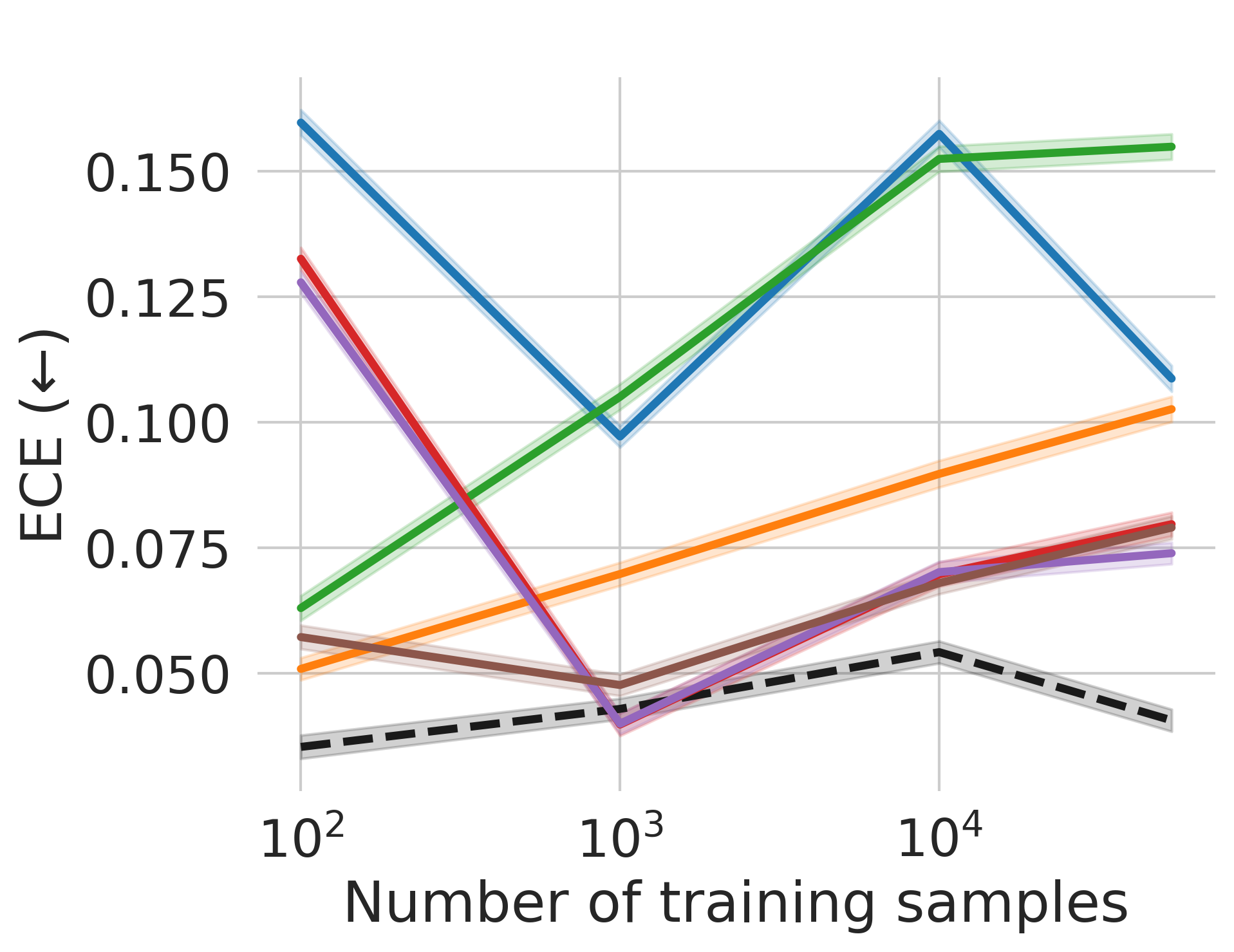}}

    \makebox[\textwidth][l]{
    \includegraphics[height=100px]{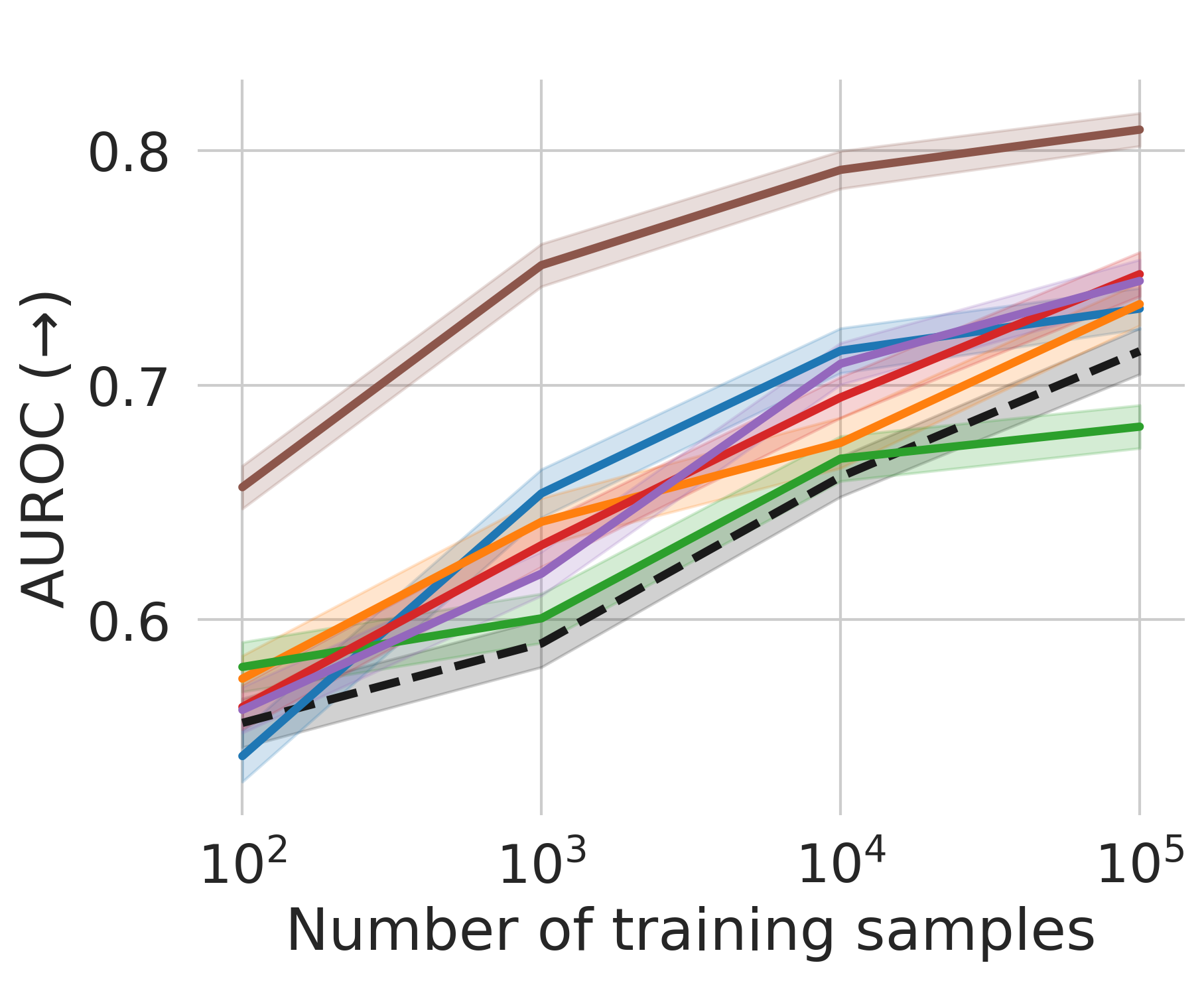}
    \includegraphics[height=100px]{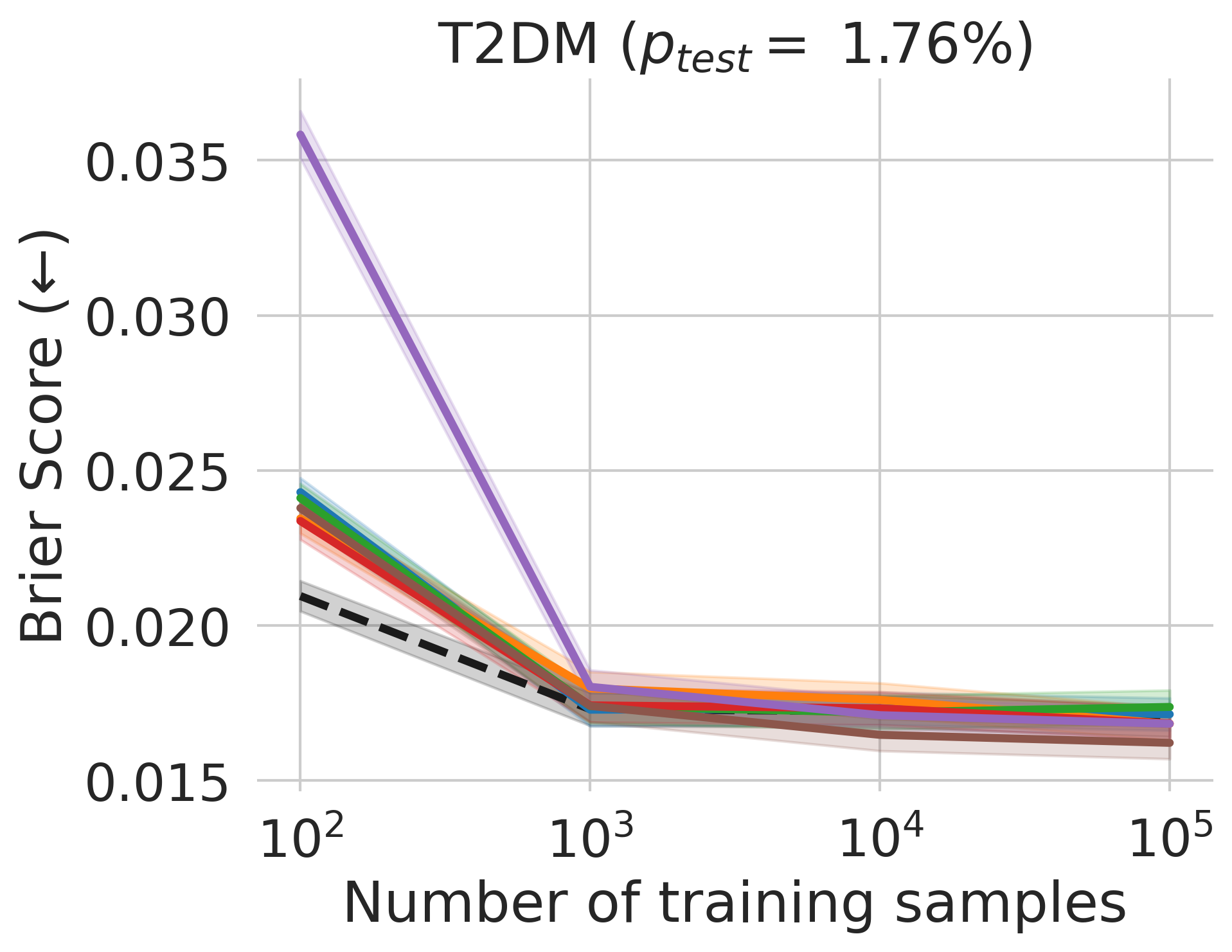}
    \includegraphics[height=100px]{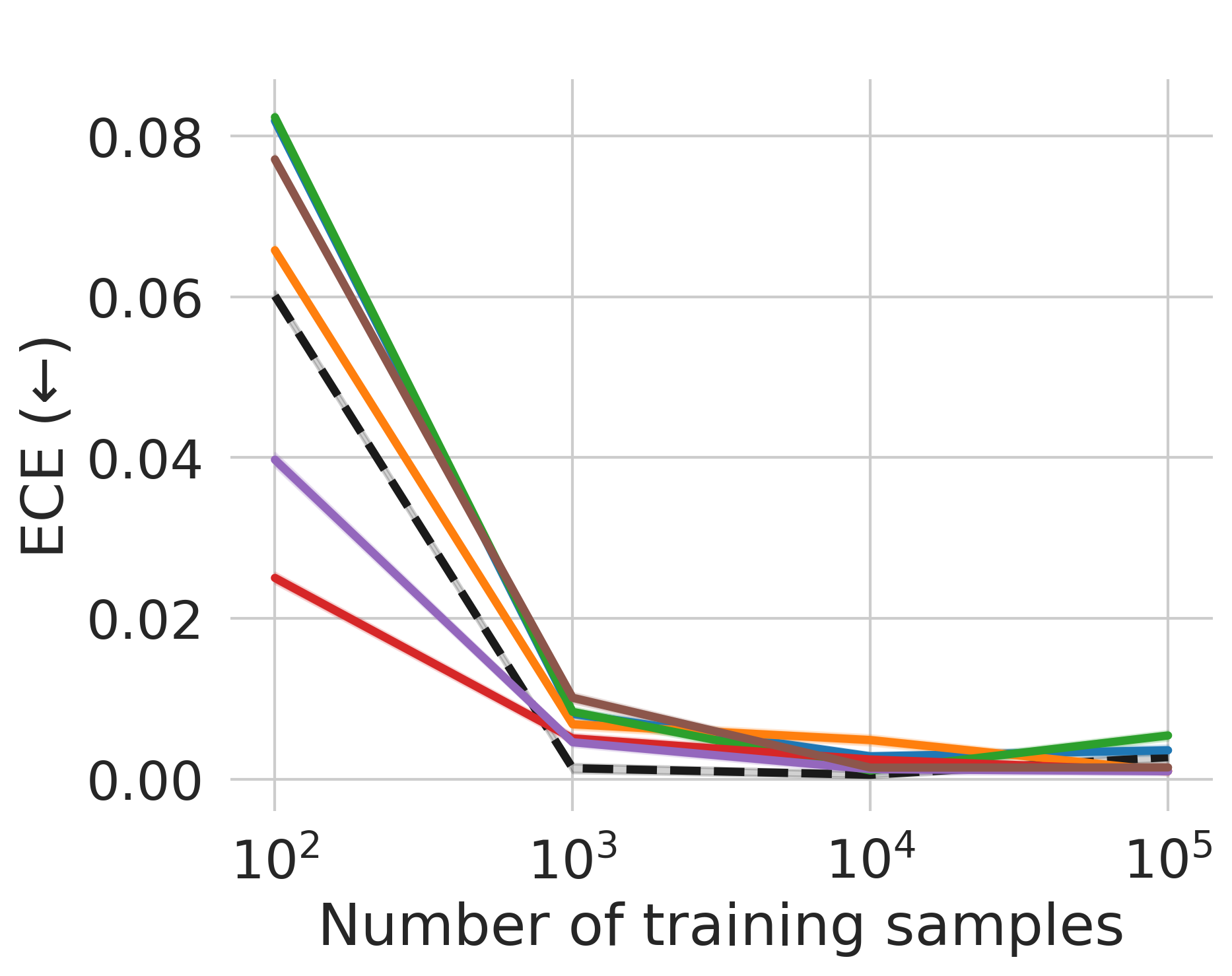}}

    \centering
    \caption{Discriminative performances and calibration results for all phenotypes for an increasing number of training points (Shaded area represents bootstrapped standard deviation) - Continued.}
    \label{fig:all_phenotypes:2}
\end{figure}

\begin{figure}[ht]
    \makebox[\textwidth][l]{
    \includegraphics[height=100px]{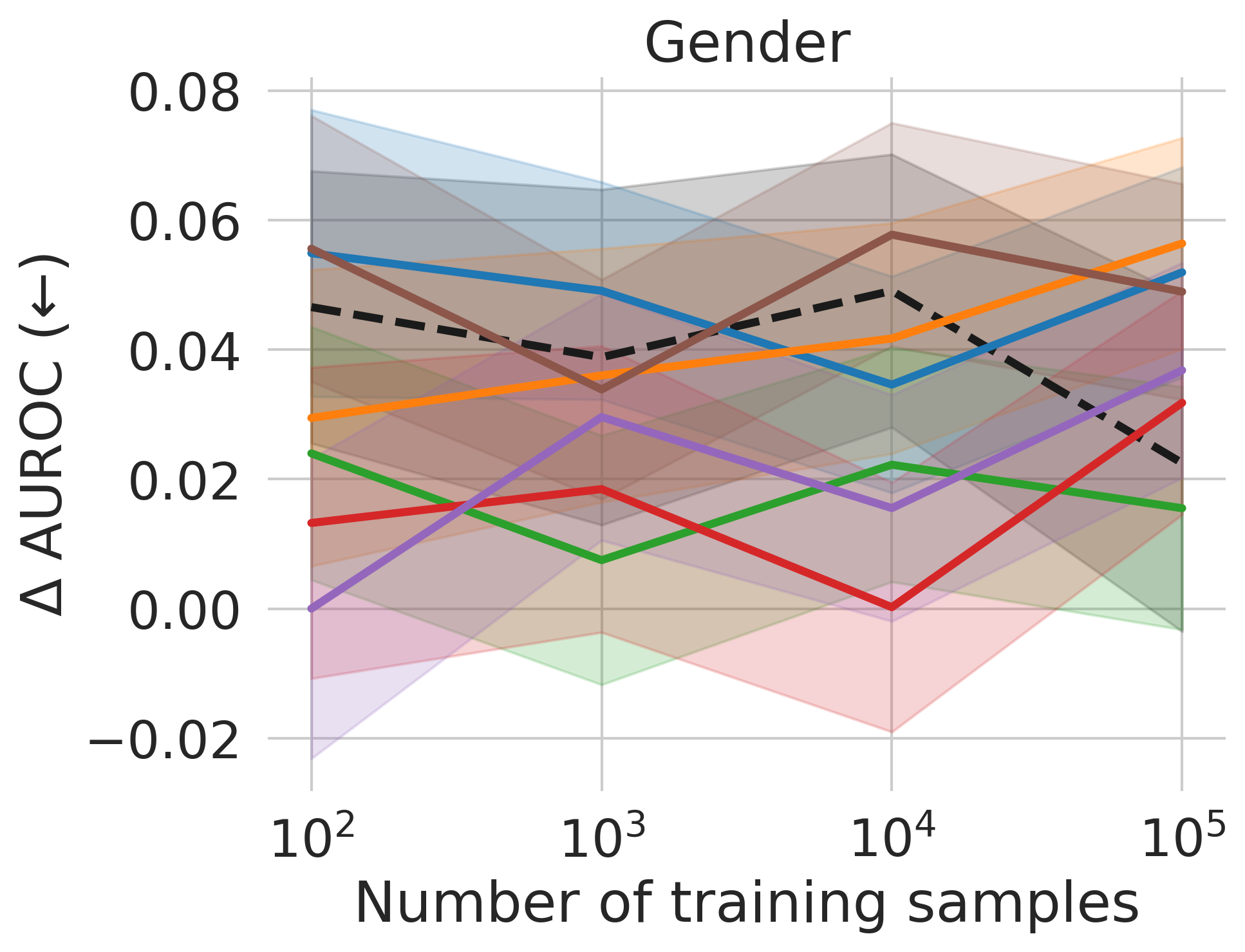}
    \includegraphics[height=100px]{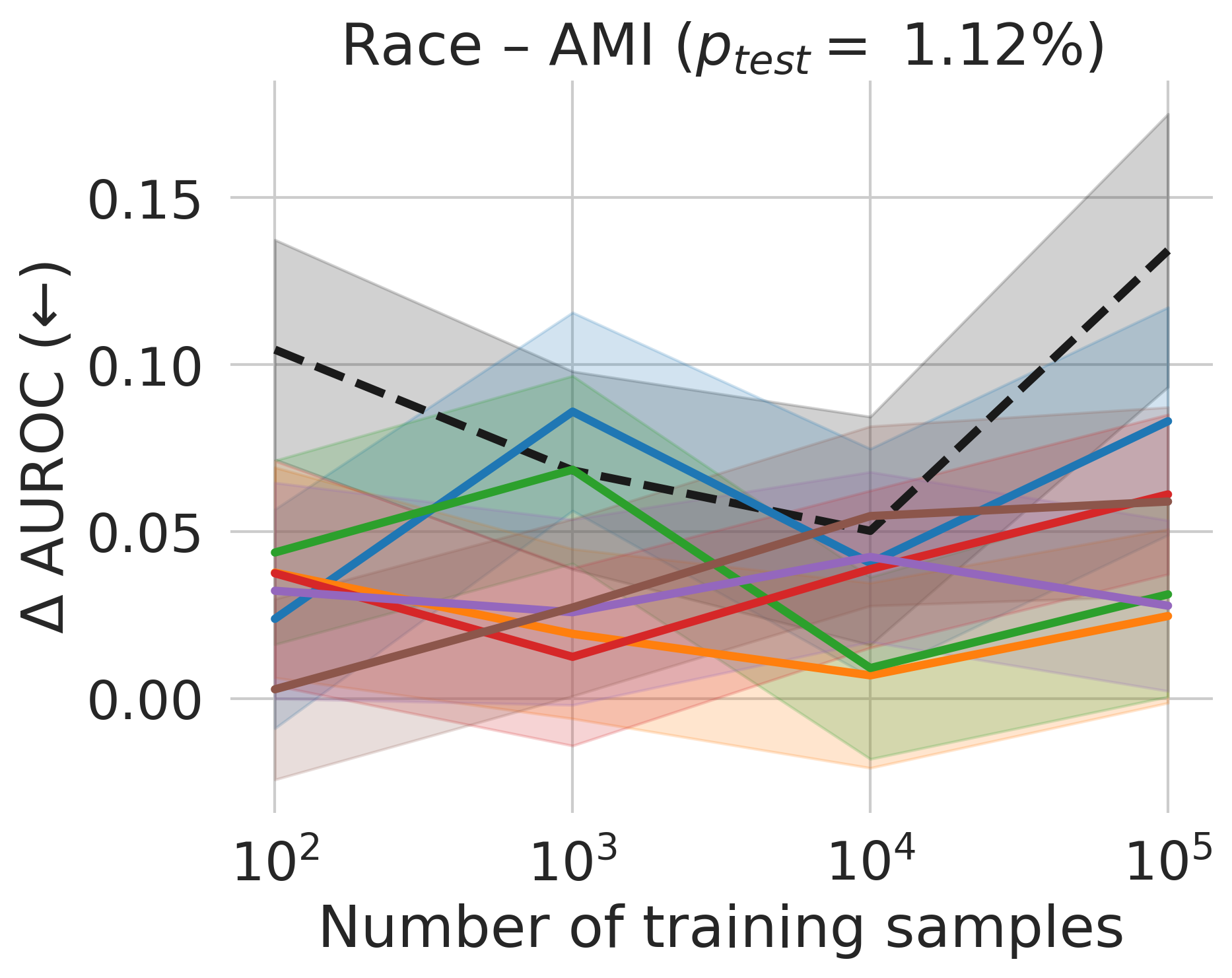}
    \includegraphics[height=100px]{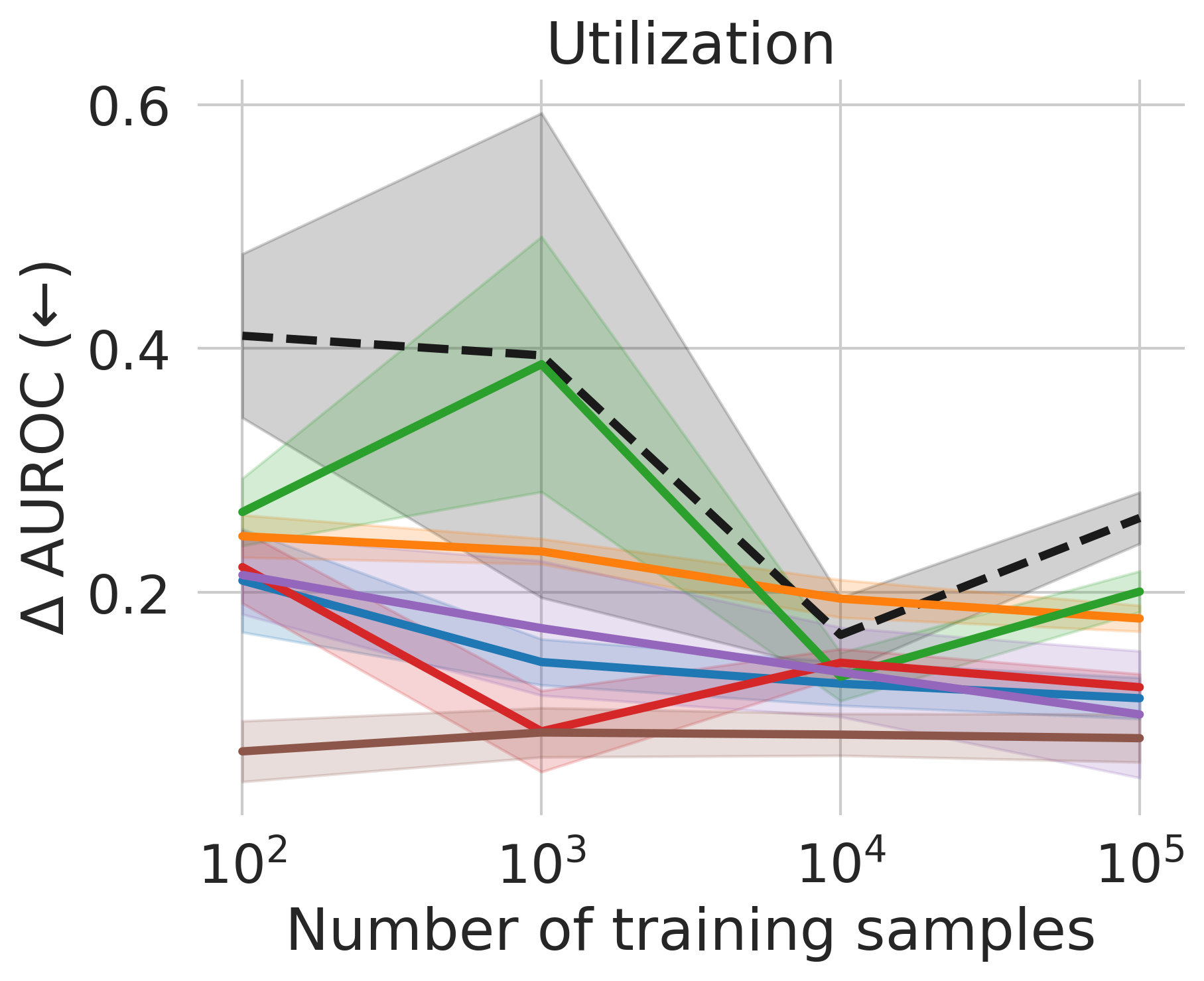}}

    \makebox[\textwidth][l]{
    \includegraphics[height=100px]{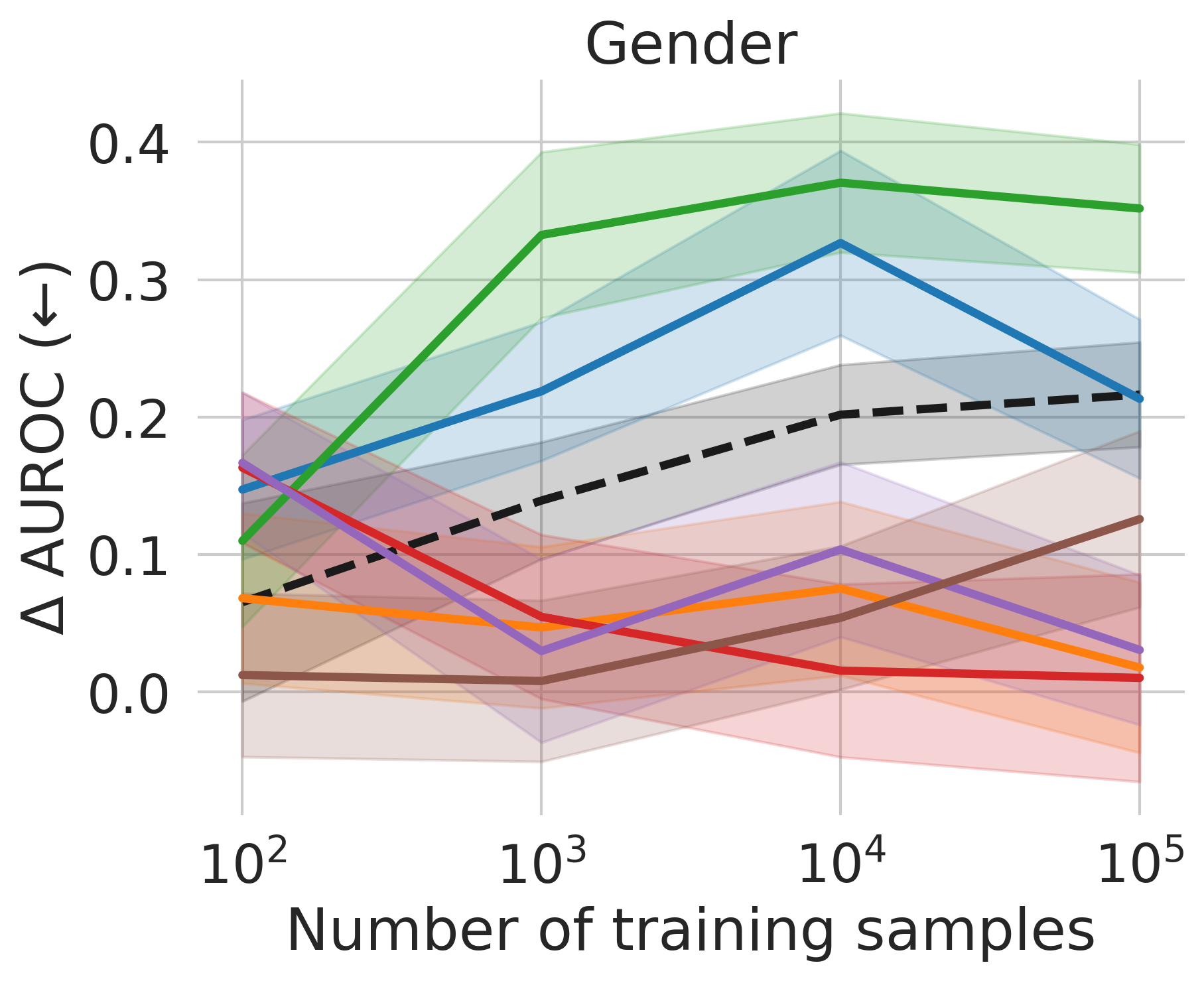}
    \includegraphics[height=100px]{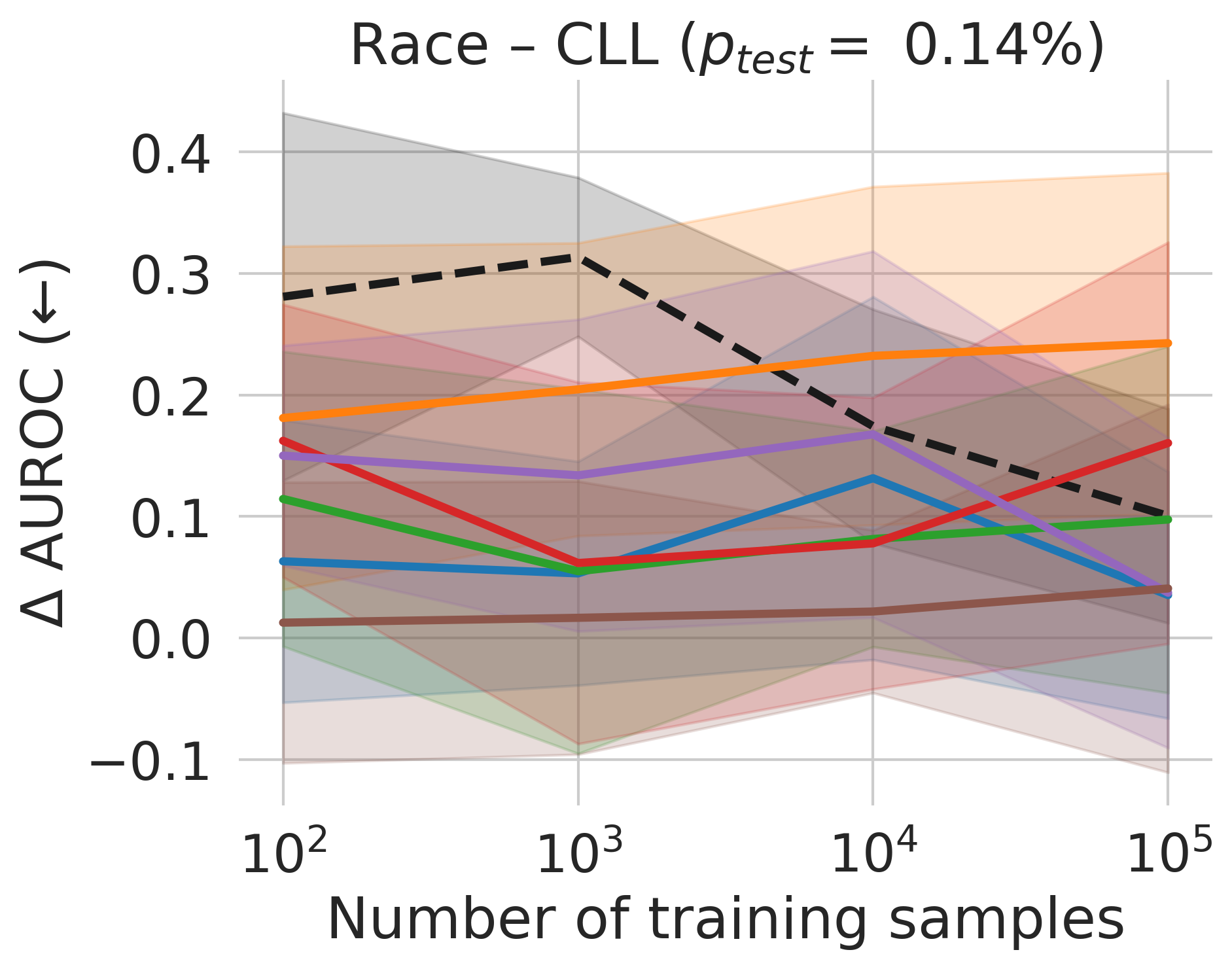}
    \includegraphics[height=100px]{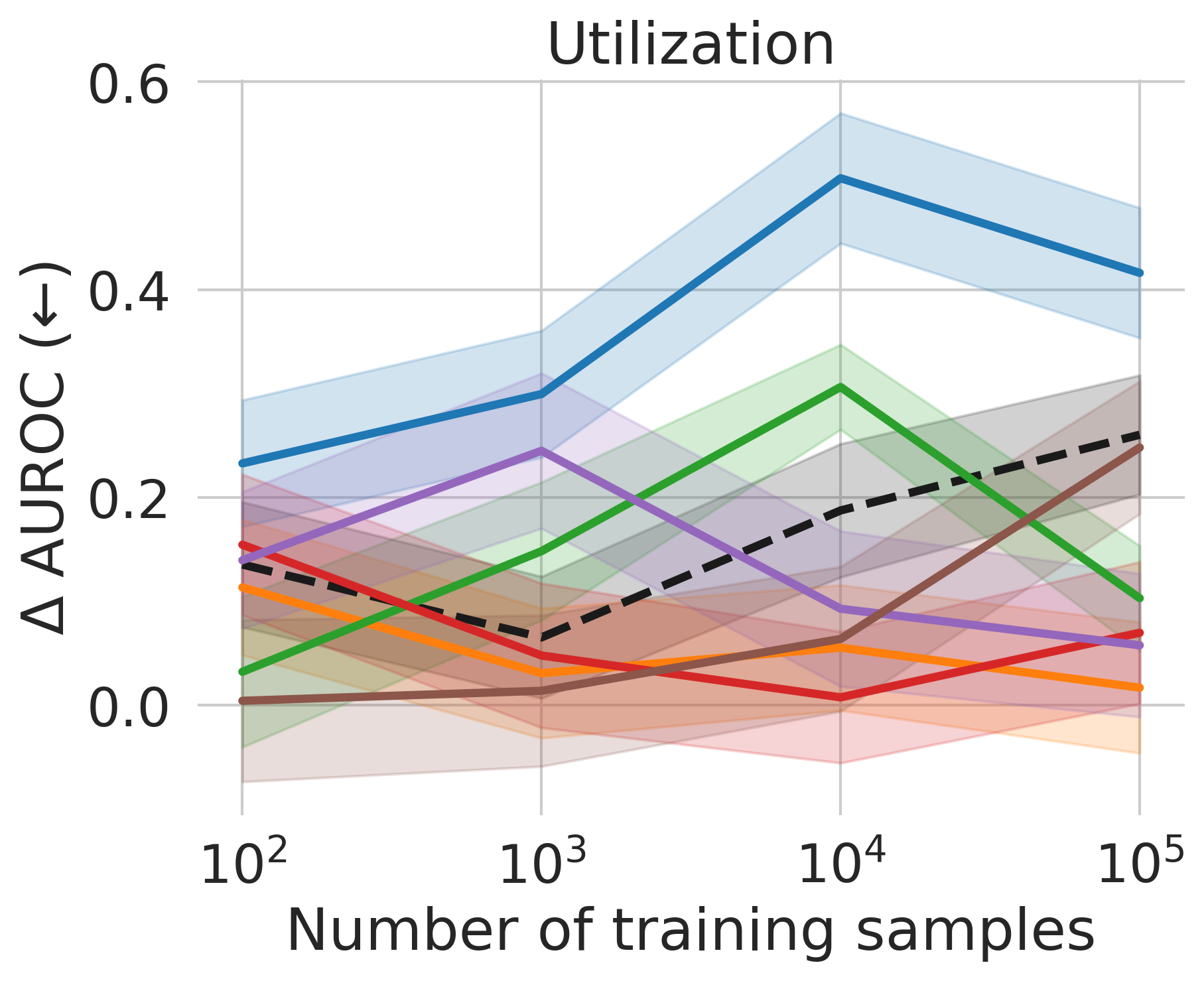}}

    \makebox[\textwidth][l]{
    \includegraphics[height=100px]{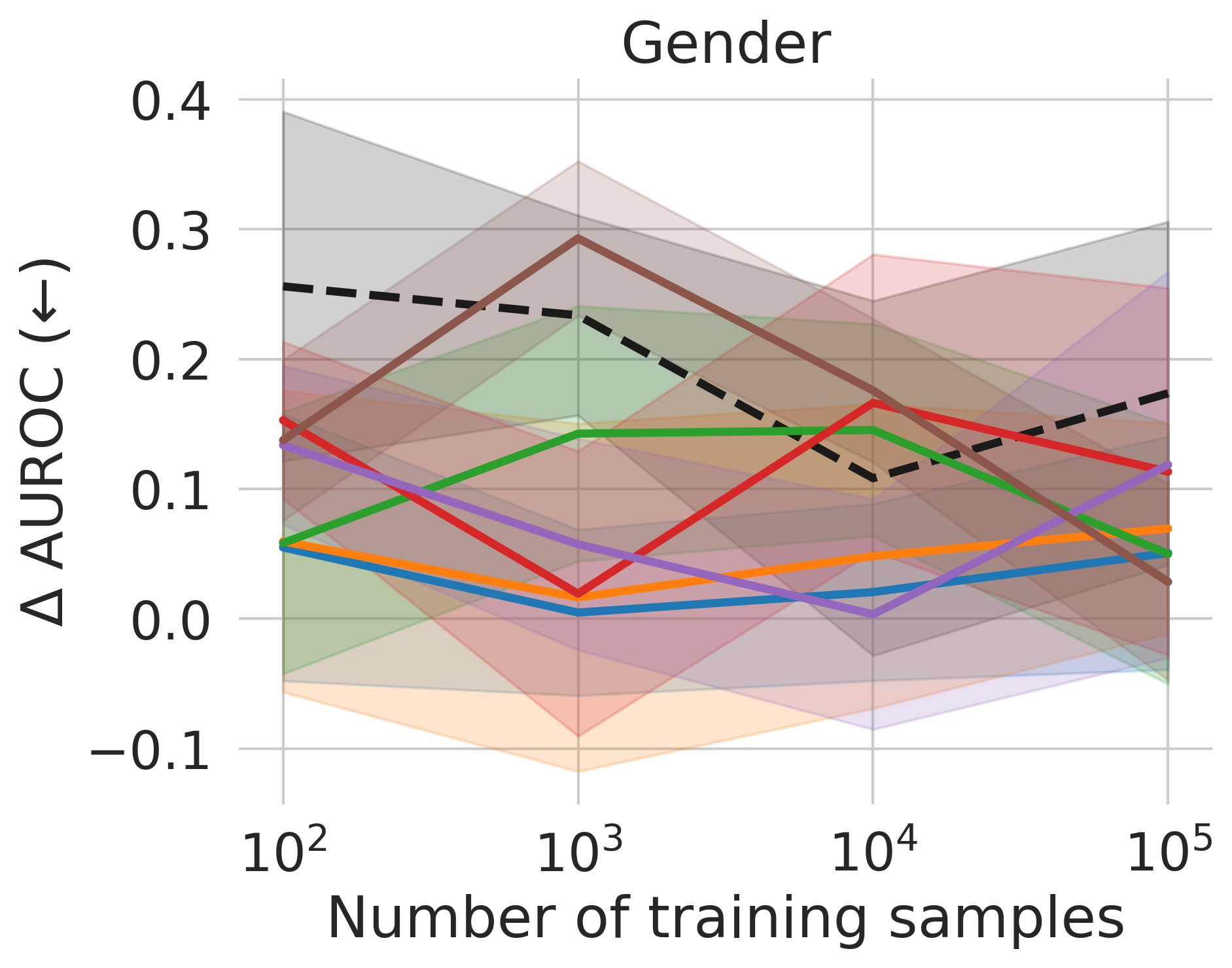}
    \includegraphics[height=100px]{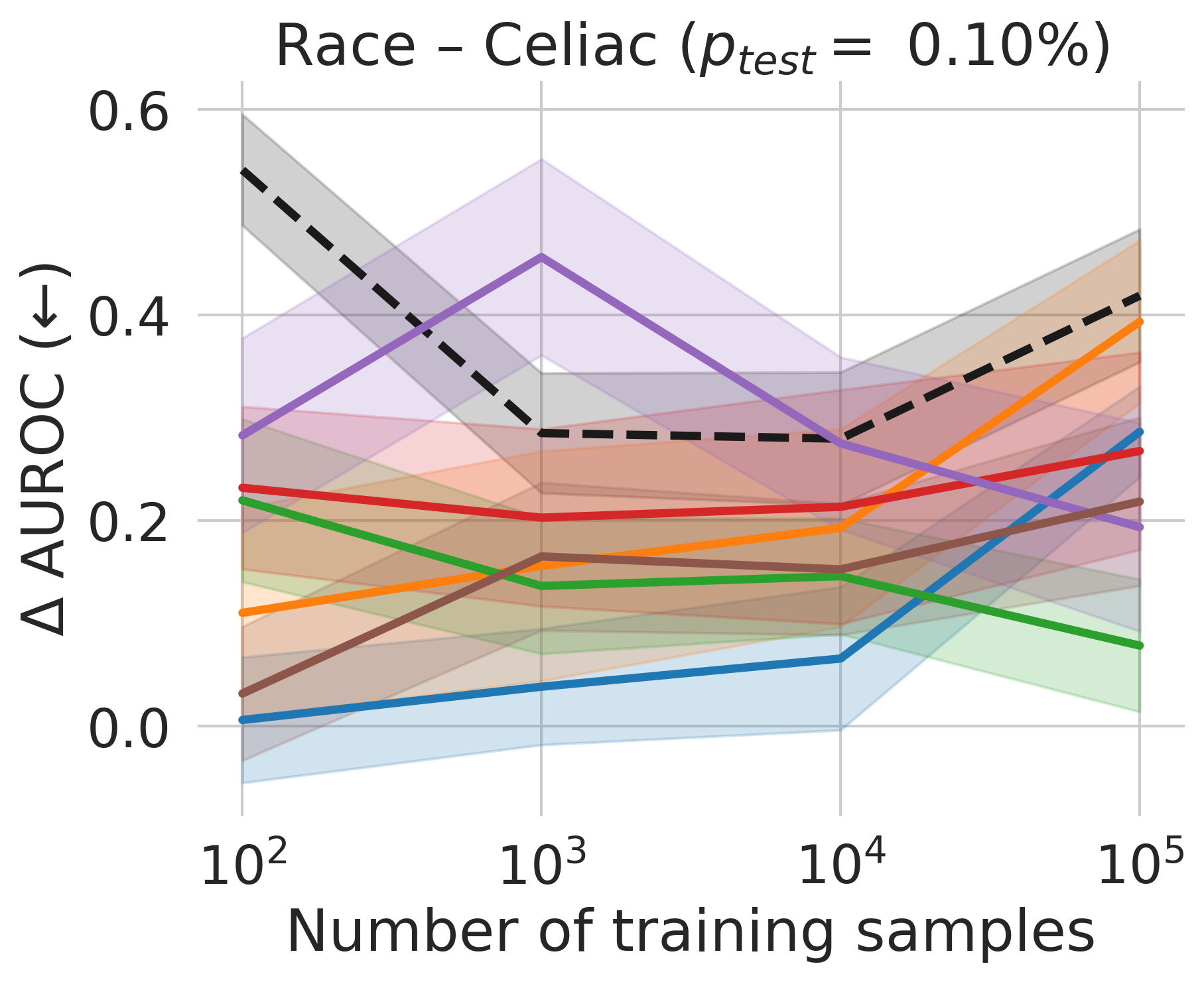}
    \includegraphics[height=100px]{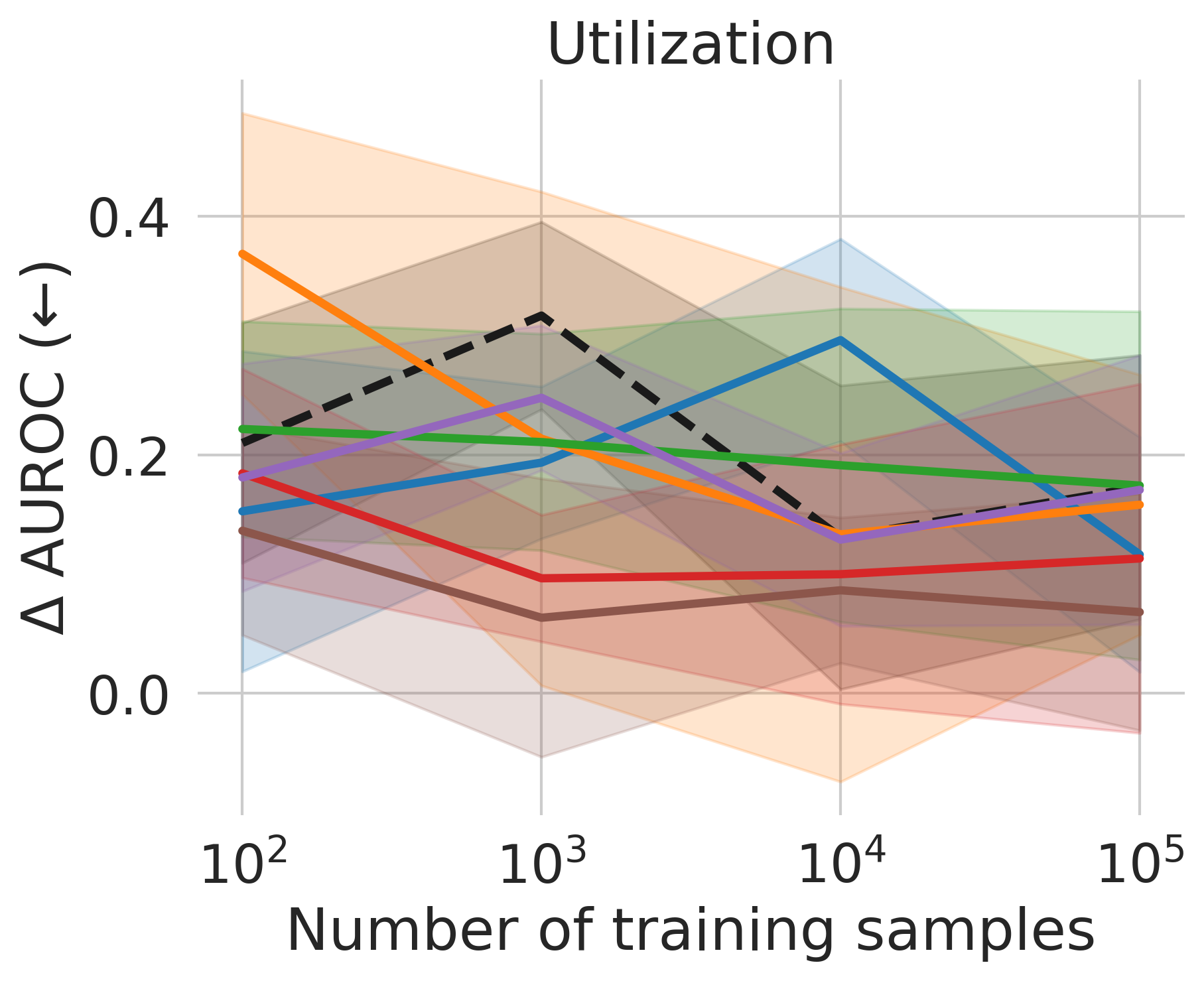}\includegraphics[height=100px]{figures/legend.png}}

    \makebox[\textwidth][l]{
    \includegraphics[height=100px]{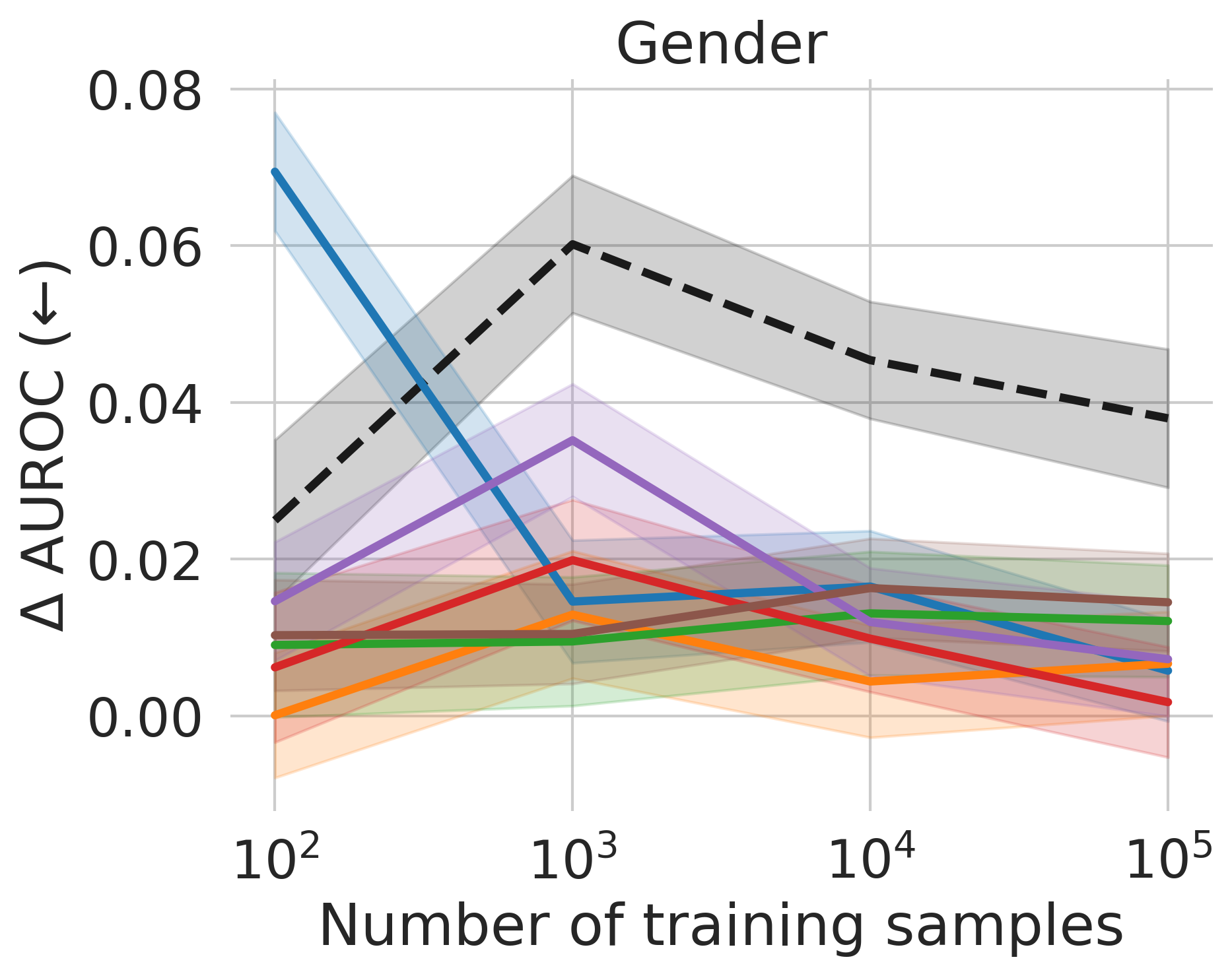}
    \includegraphics[height=100px]{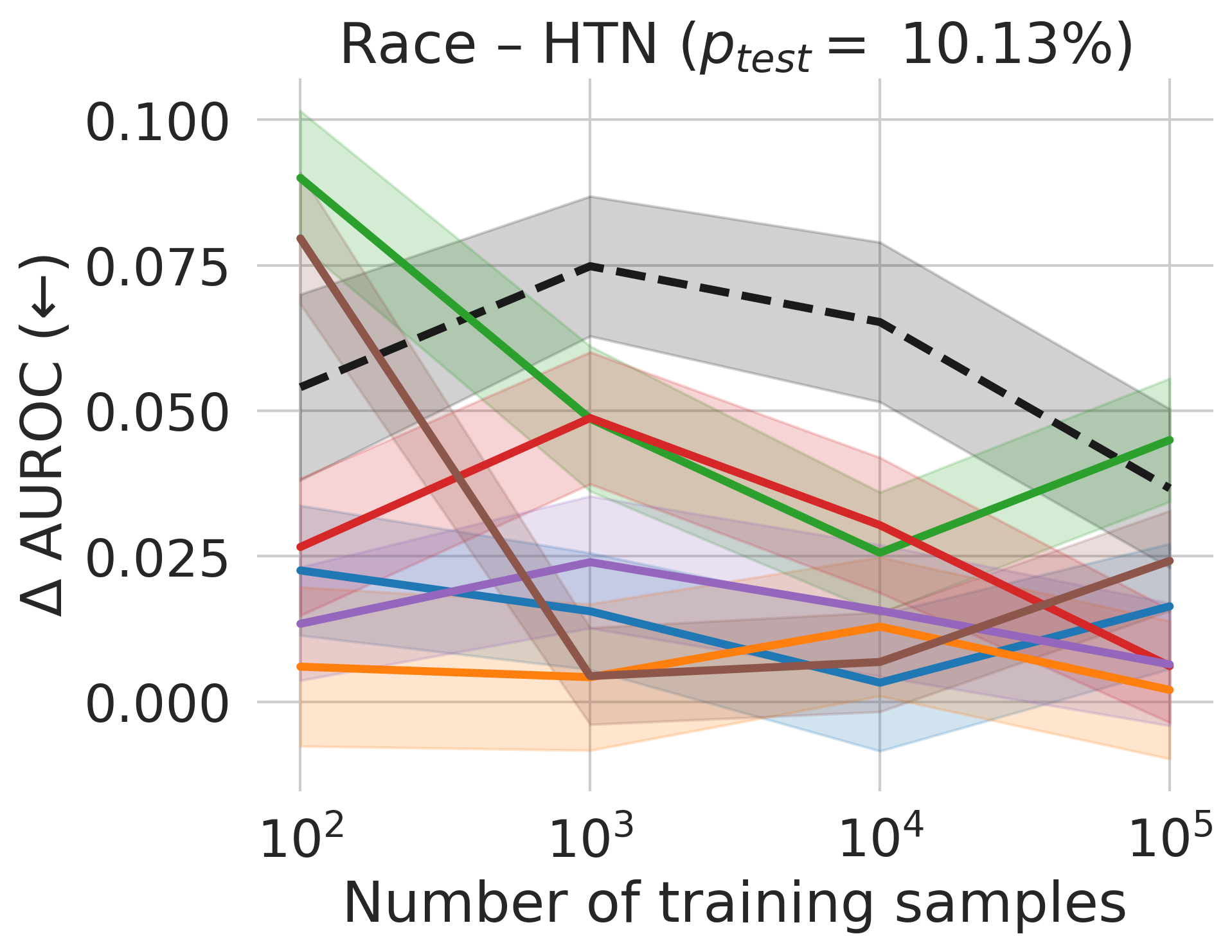}
    \includegraphics[height=100px]{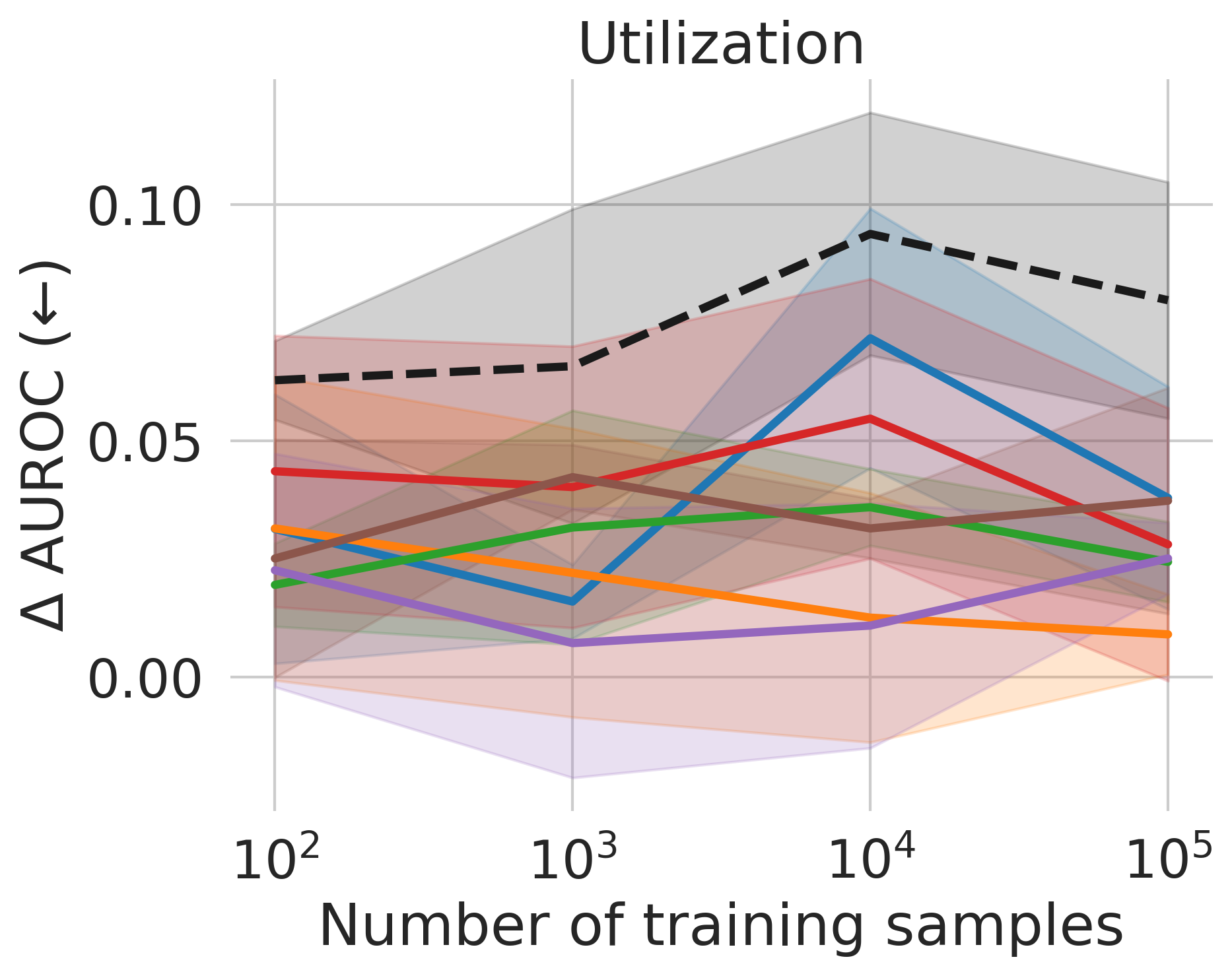}}

    \makebox[\textwidth][l]{
    \includegraphics[height=100px]{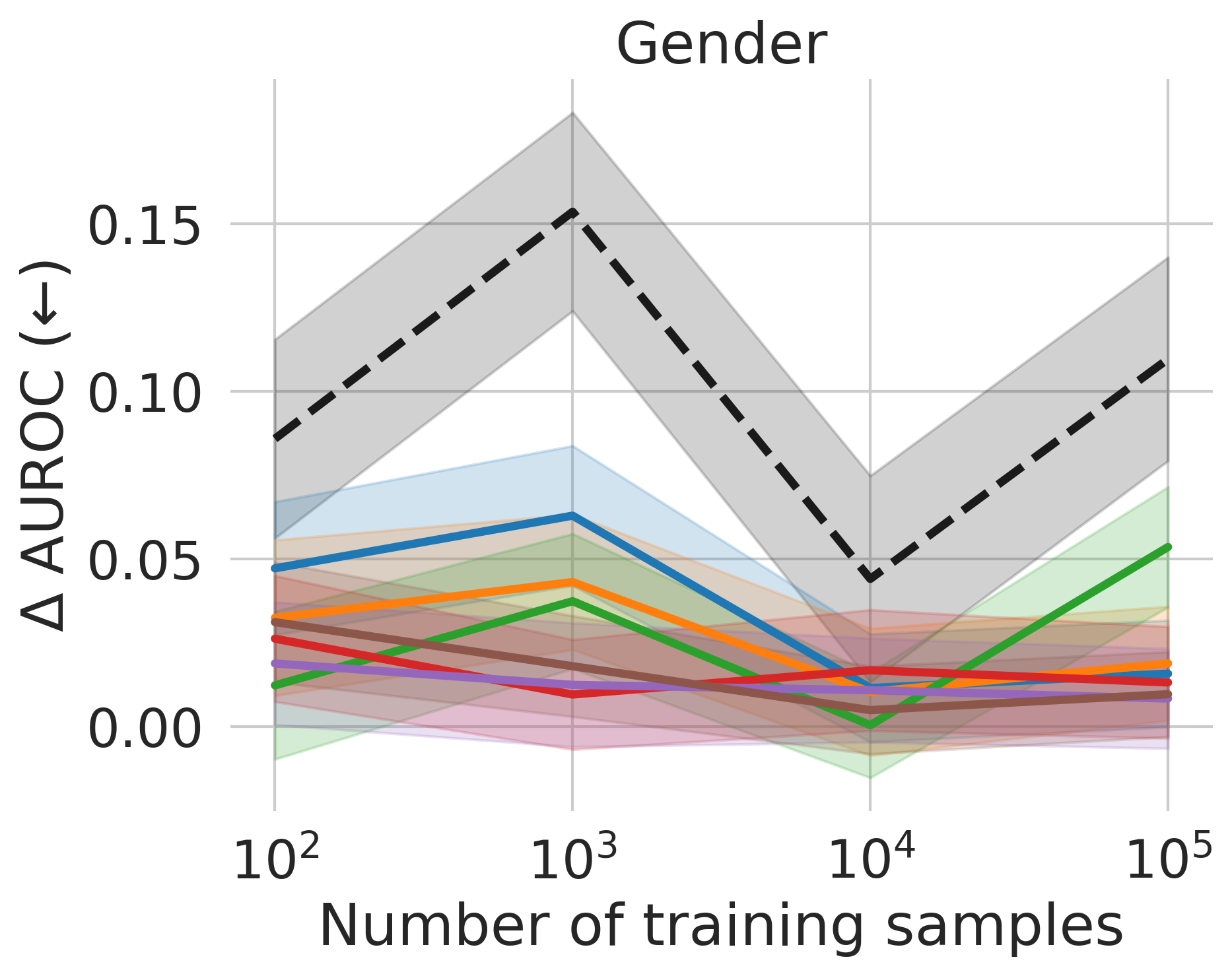}
    \includegraphics[height=100px]{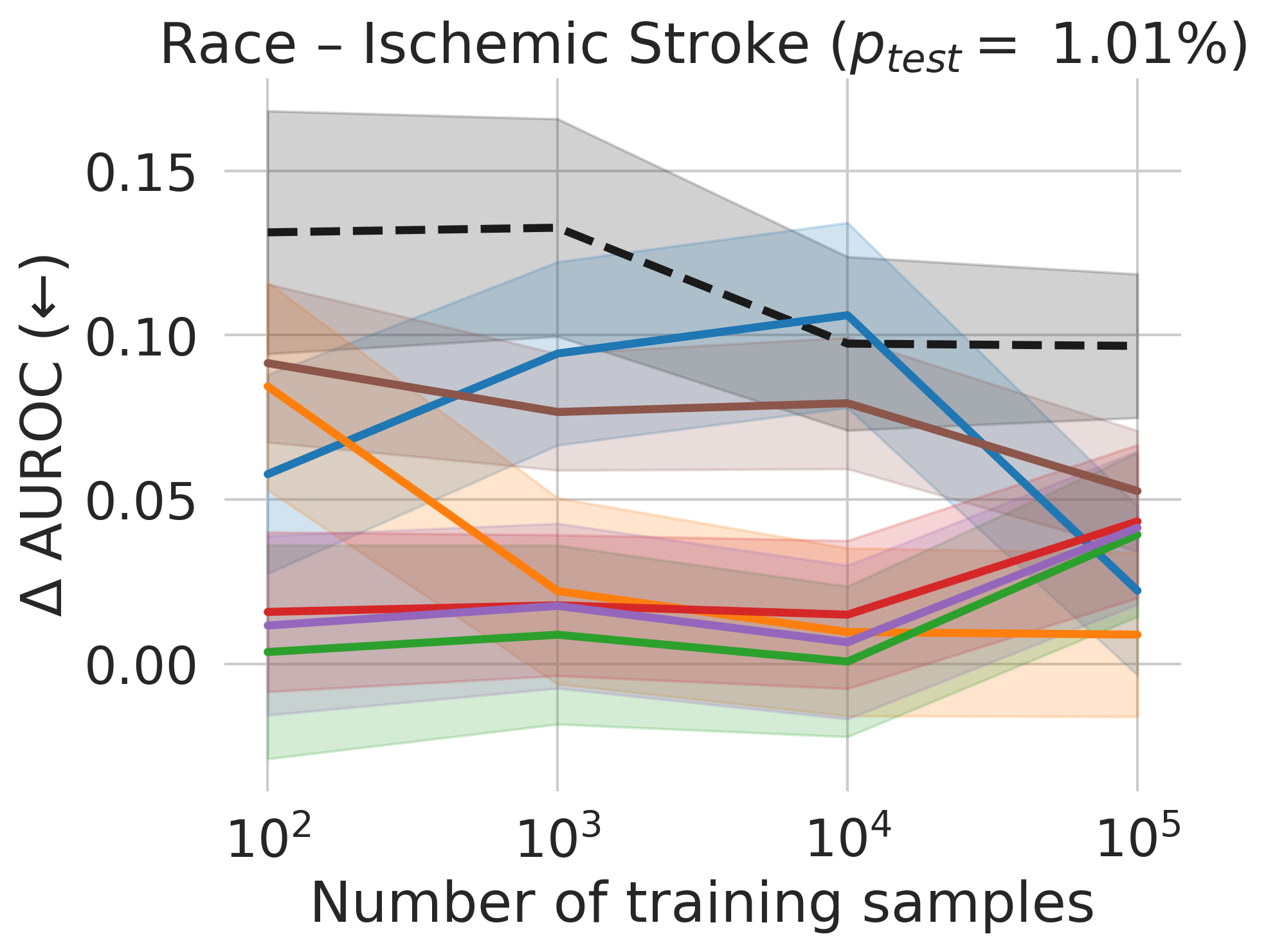}
    \includegraphics[height=100px]{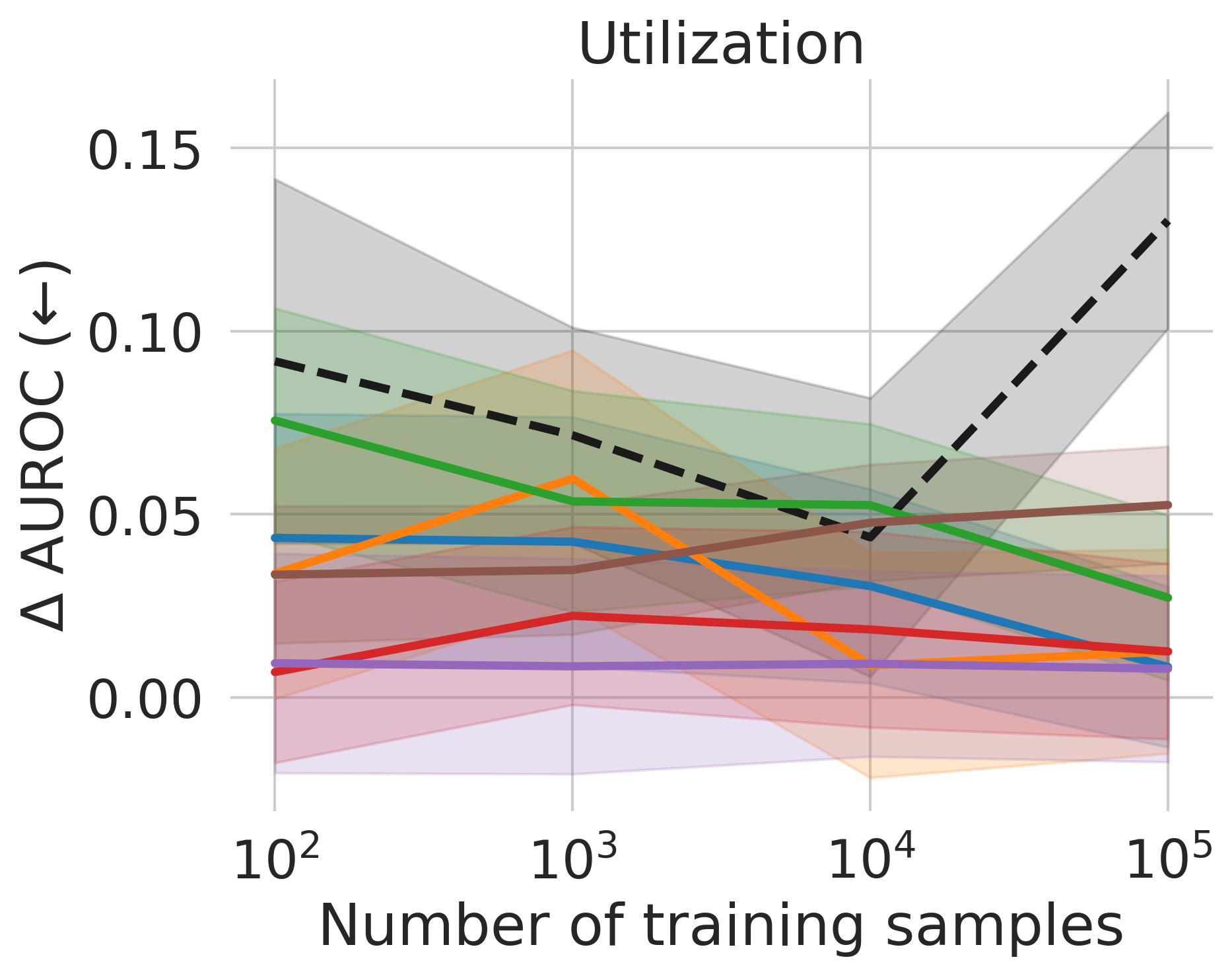}}

    \makebox[\textwidth][l]{
    \includegraphics[height=100px]{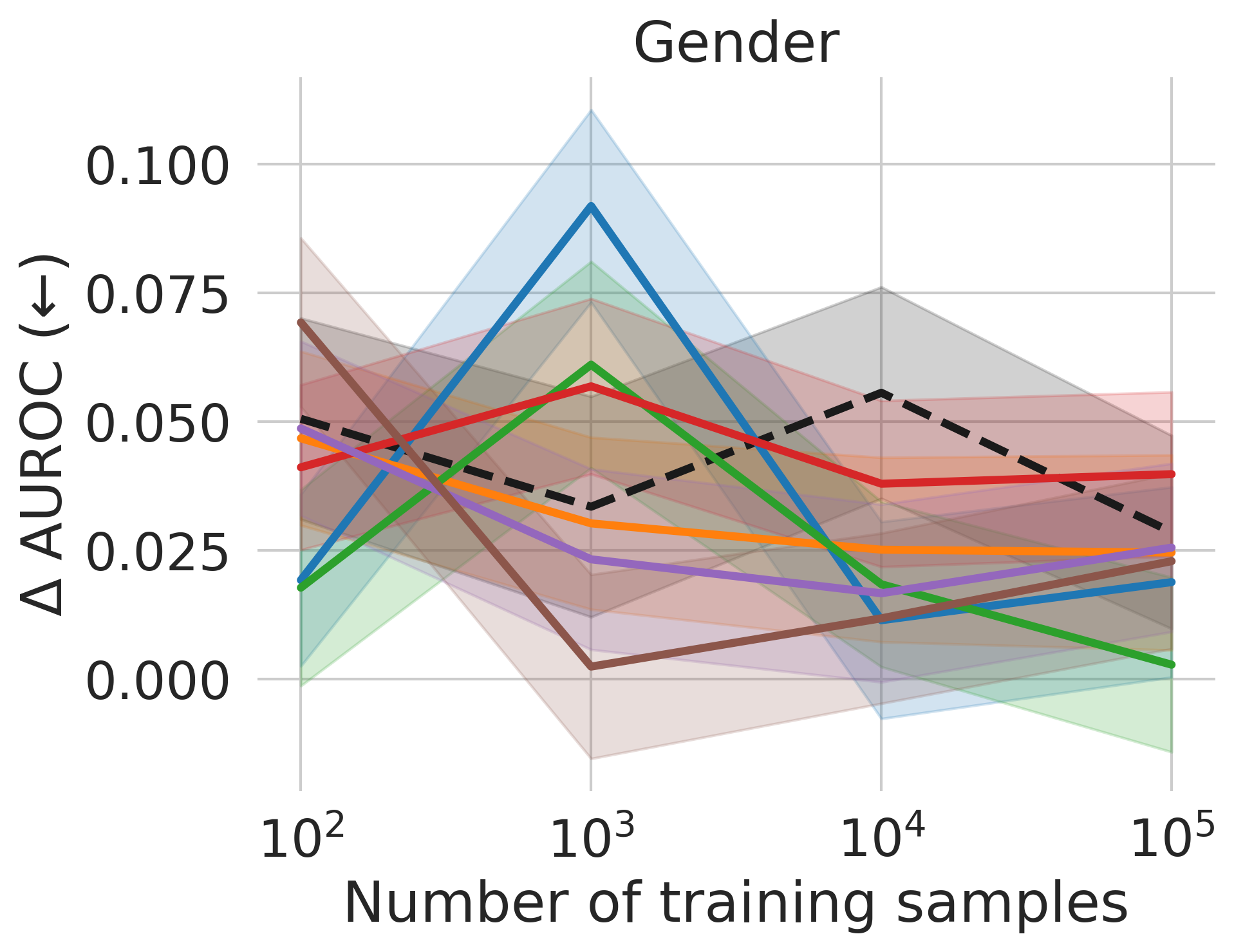}
    \includegraphics[height=100px]{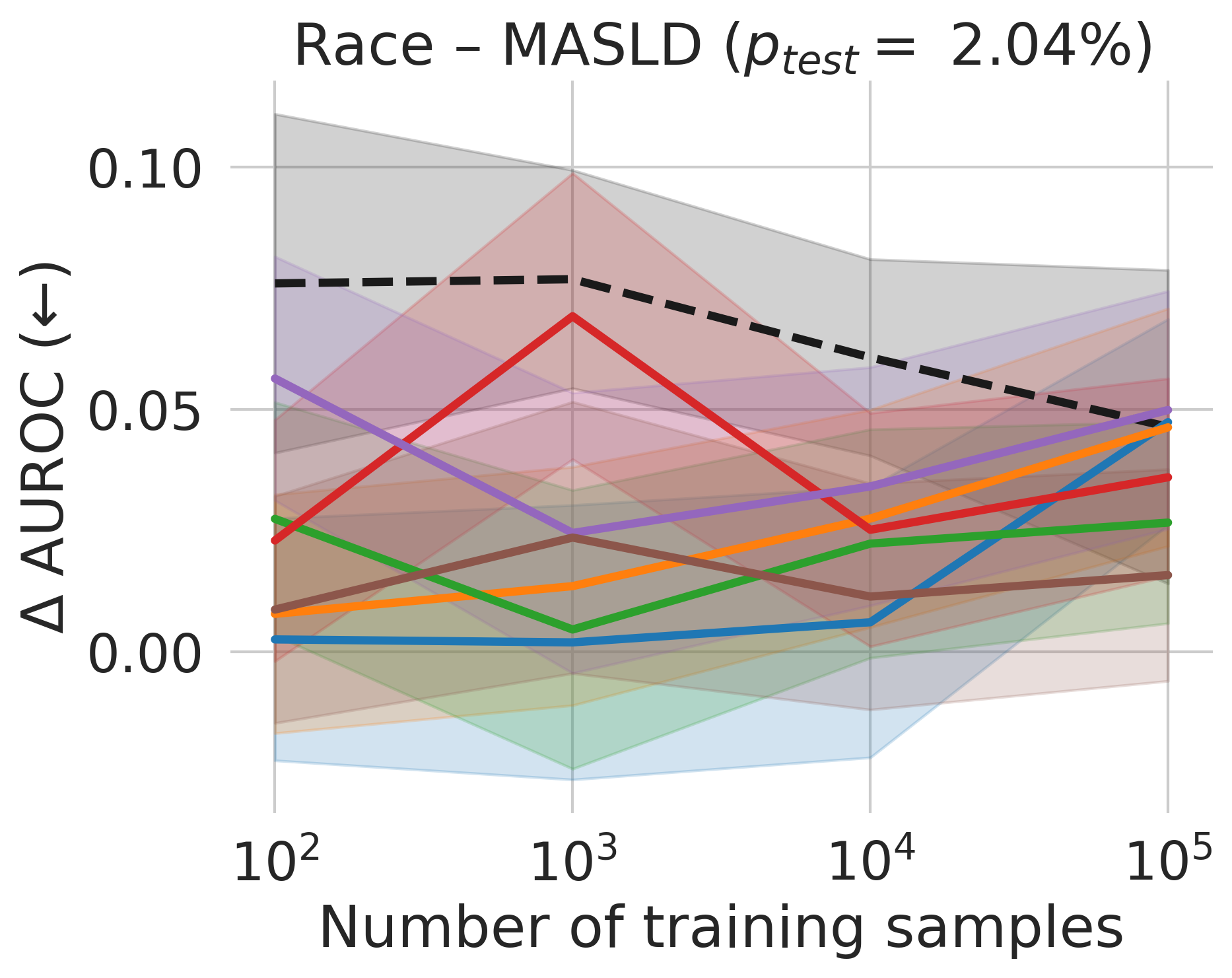}
    \includegraphics[height=100px]{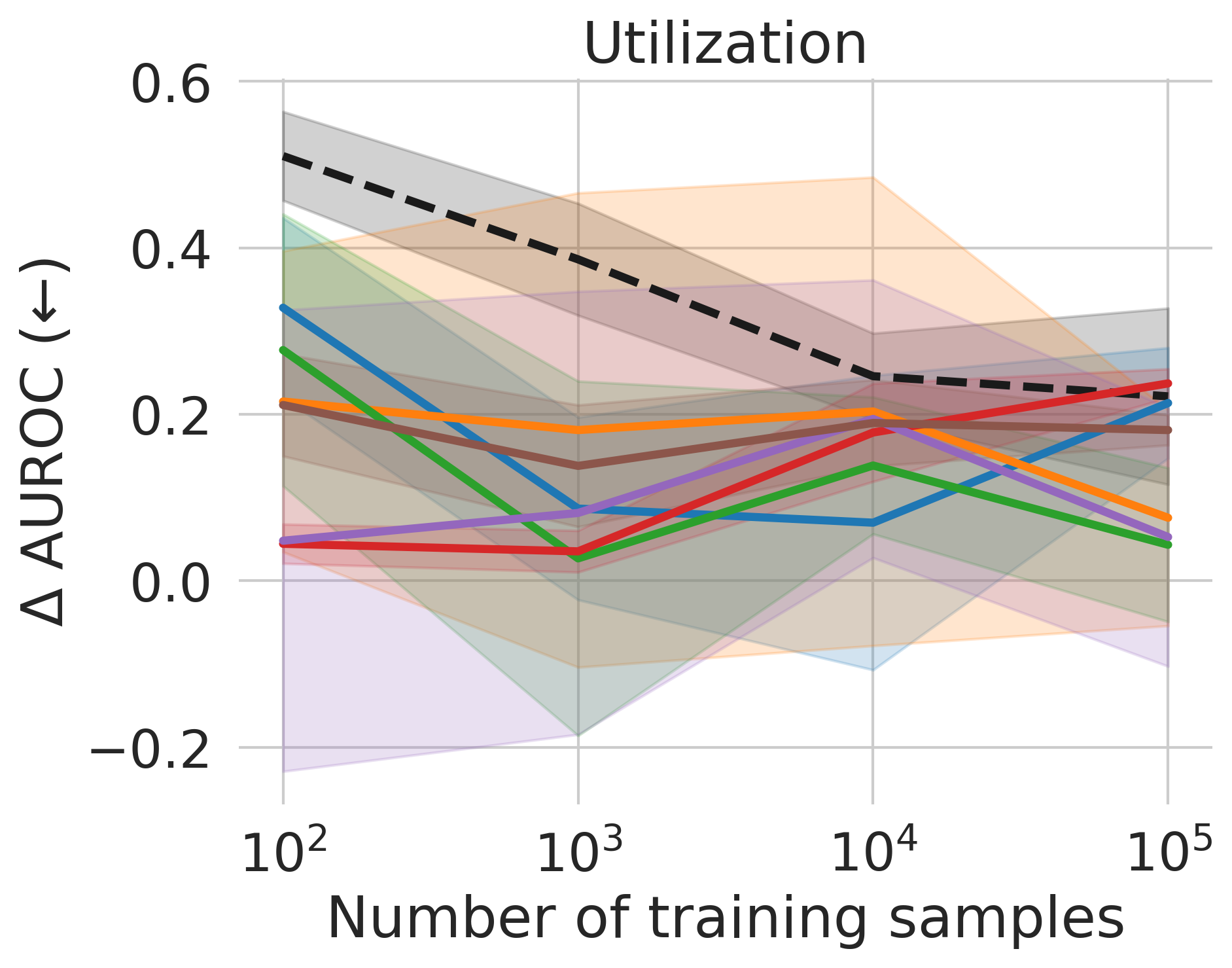}}

    \centering
    \caption{Discriminative performances and calibration results for all phenotypes for an increasing number of training points (Shaded area represents bootstrapped standard deviation).}
    \label{fig:all_phenotypes:cumc:fairness}
\end{figure}

\begin{figure}[ht]
    \makebox[\textwidth][l]{
    \includegraphics[height=100px]{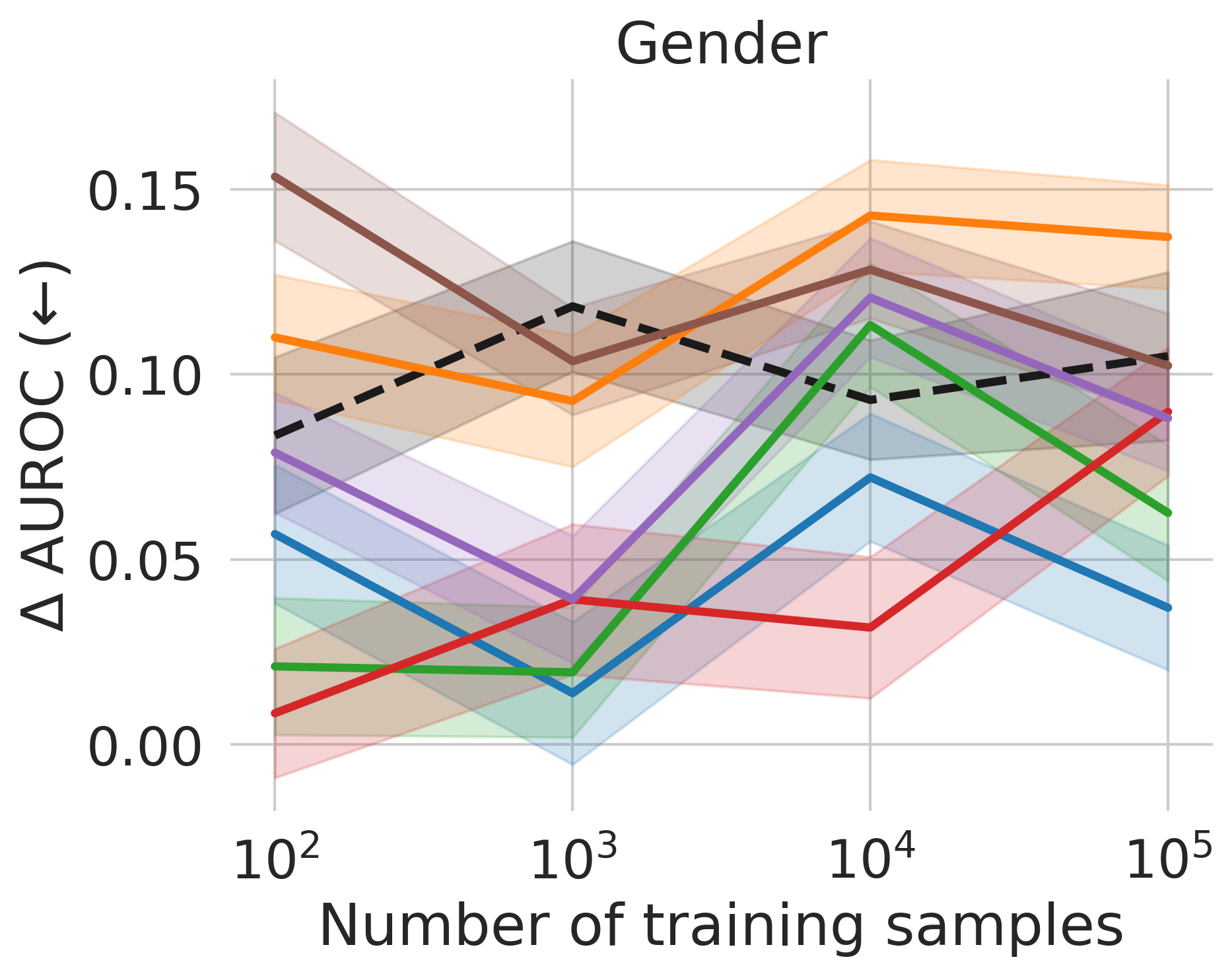}
    \includegraphics[height=100px]{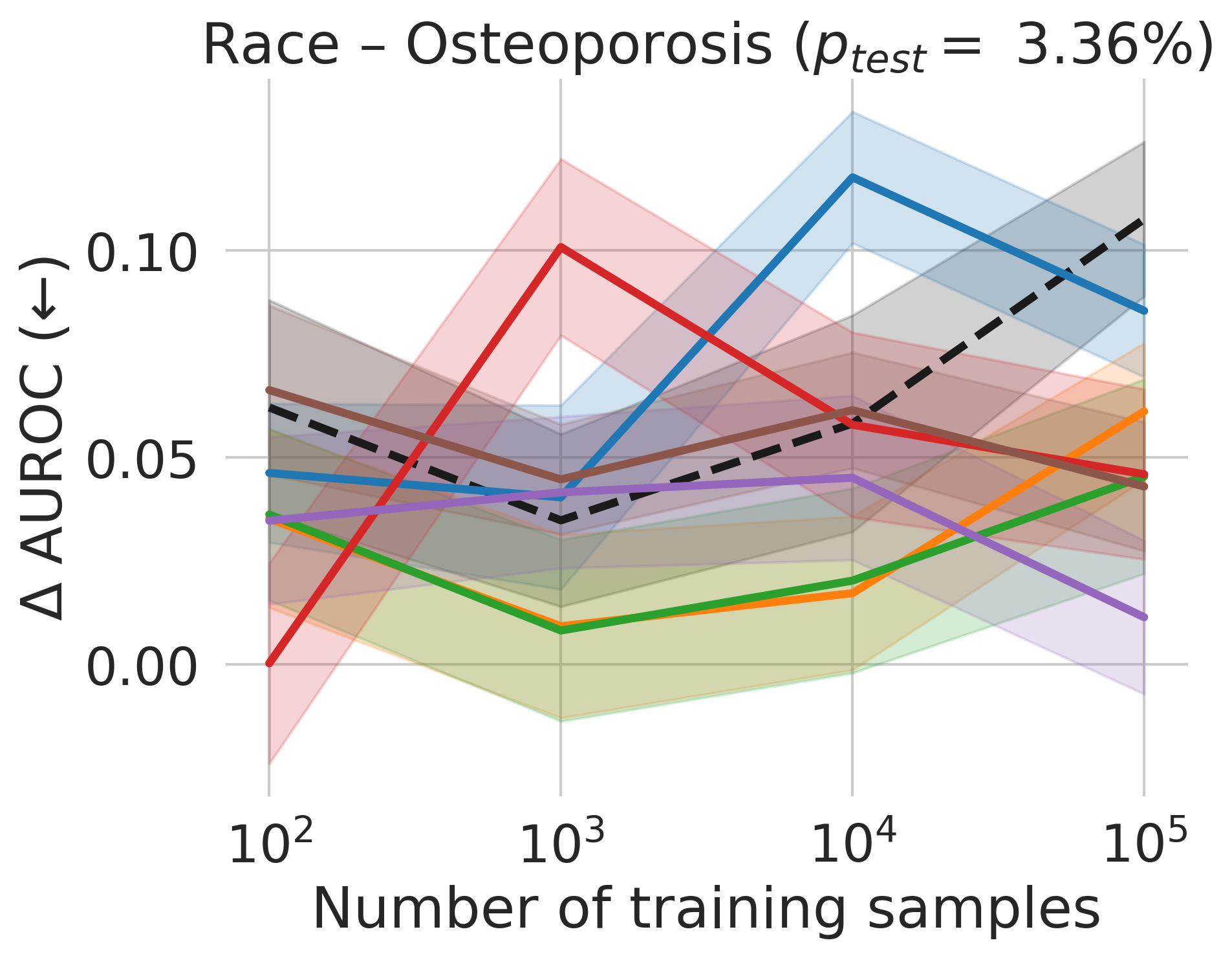}
    \includegraphics[height=100px]{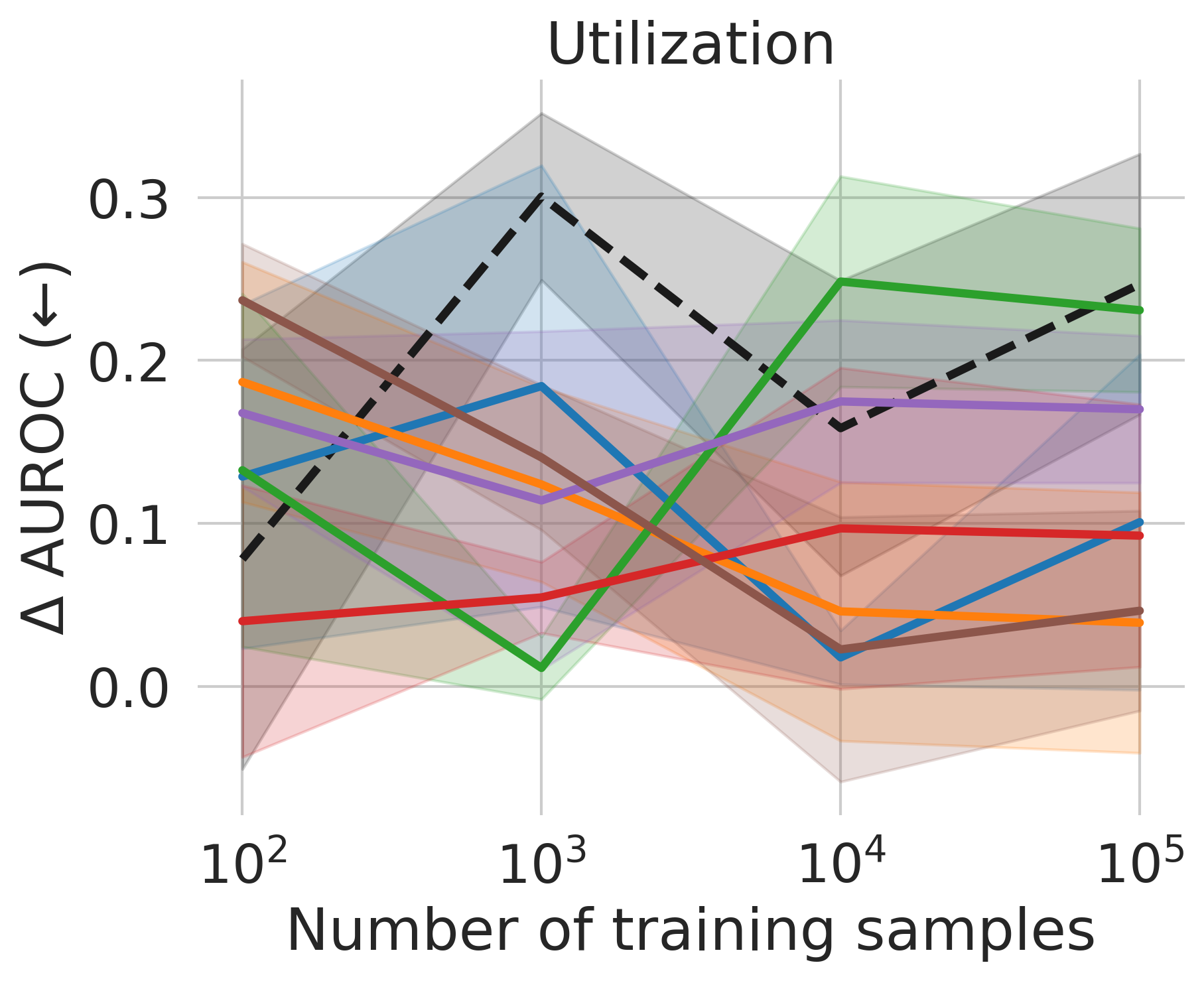}}

    \makebox[\textwidth][l]{
    \includegraphics[height=100px]{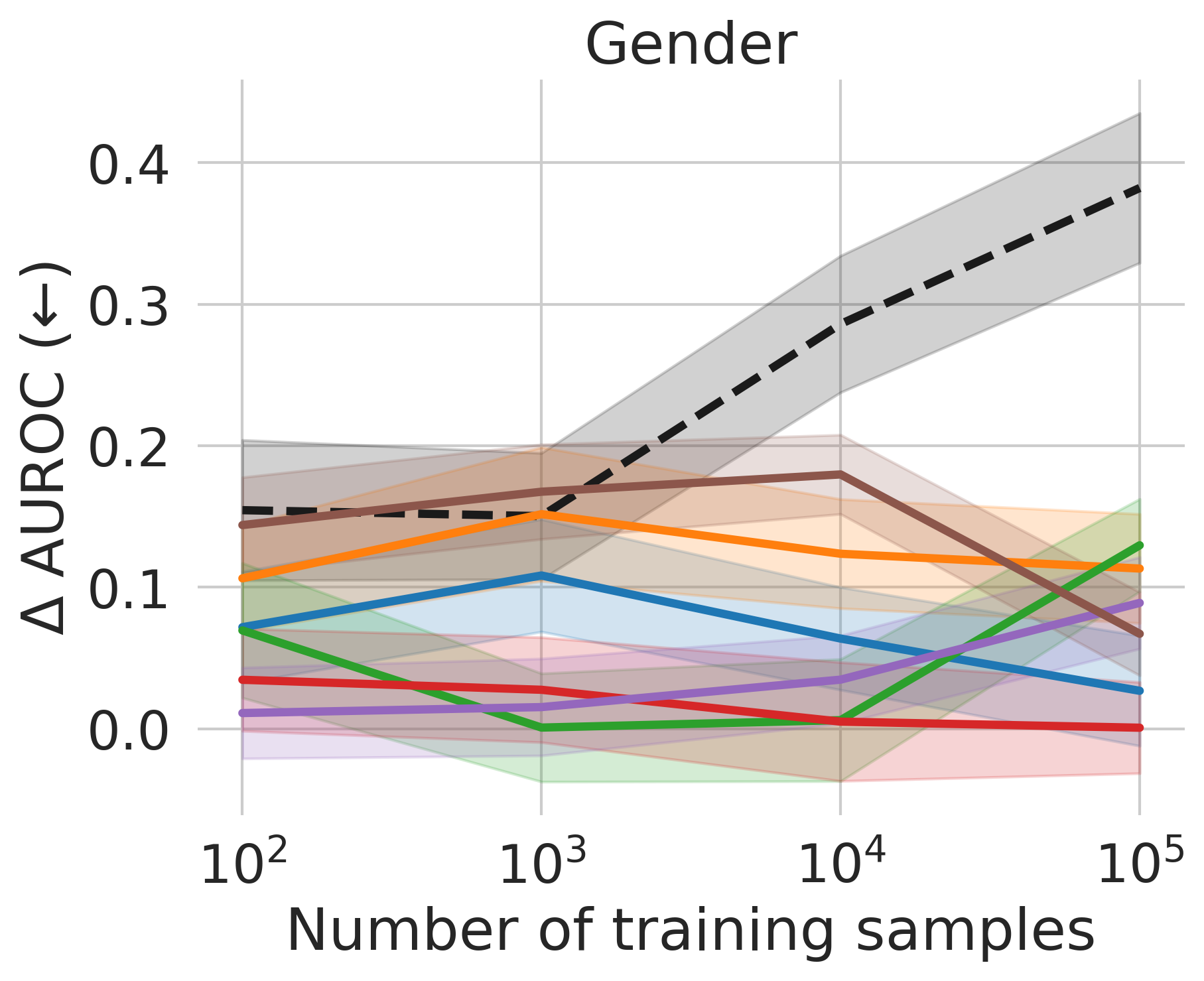}
    \includegraphics[height=100px]{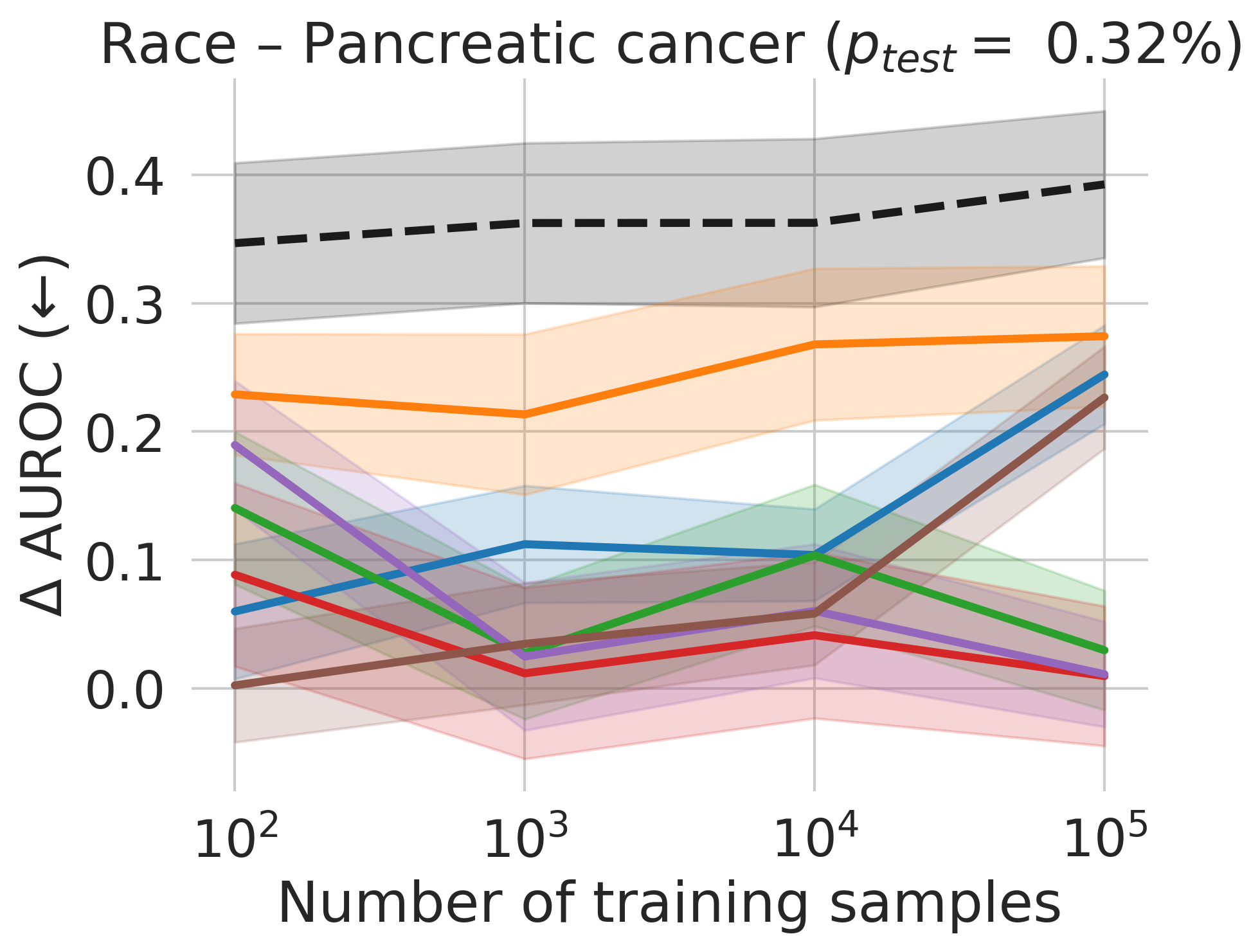}
    \includegraphics[height=100px]{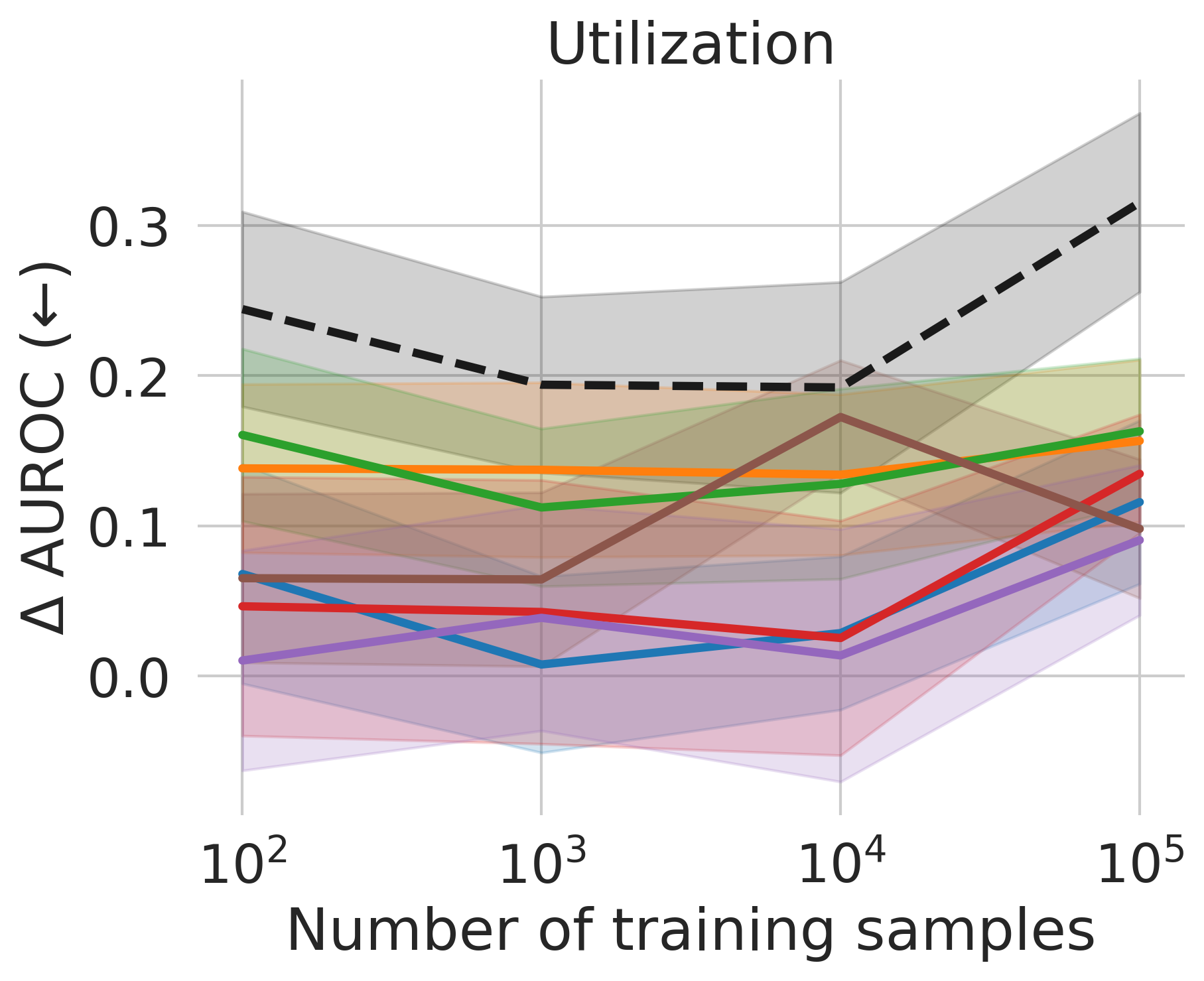}}

    \makebox[\textwidth][l]{
    \includegraphics[height=100px]{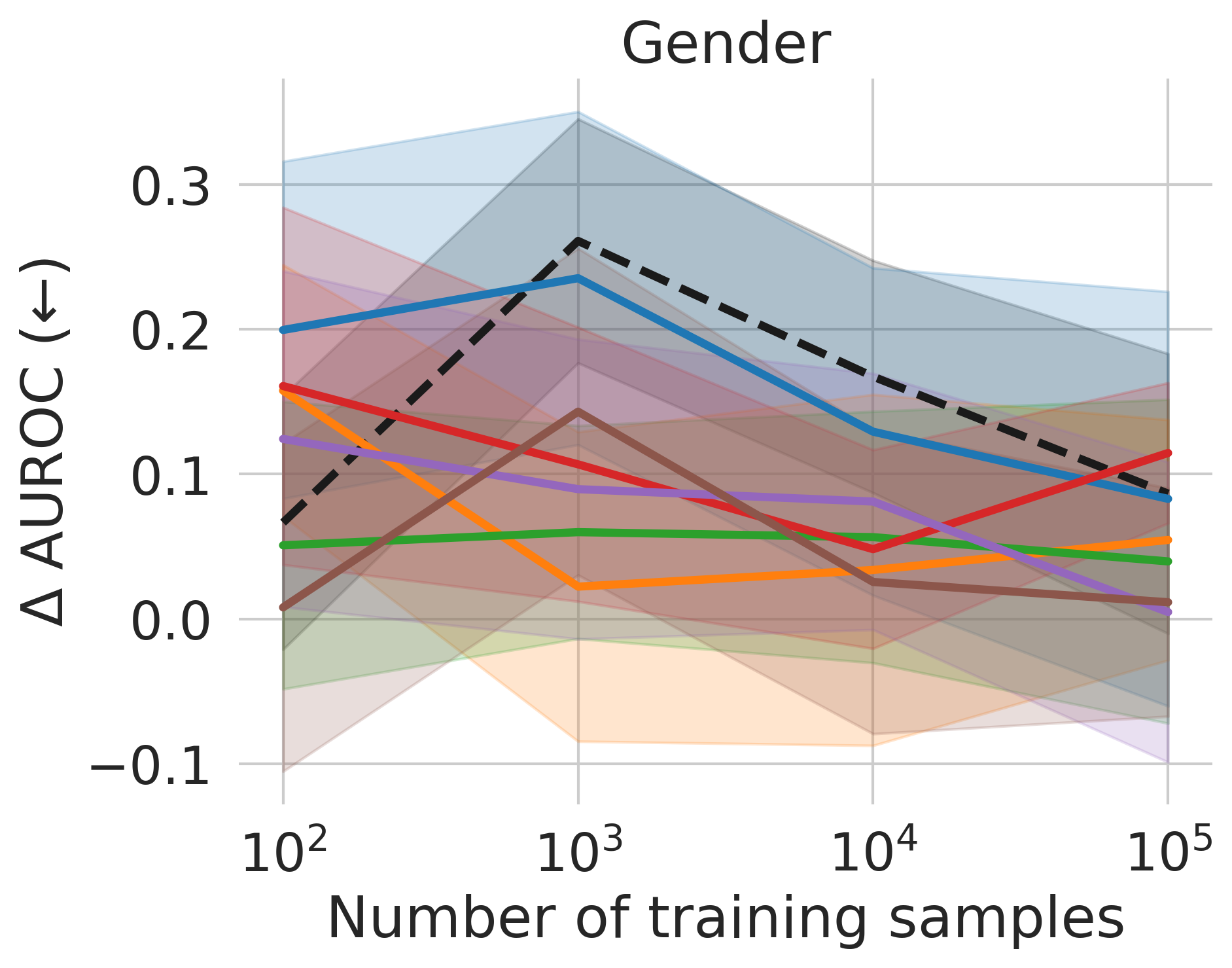}
    \includegraphics[height=100px]{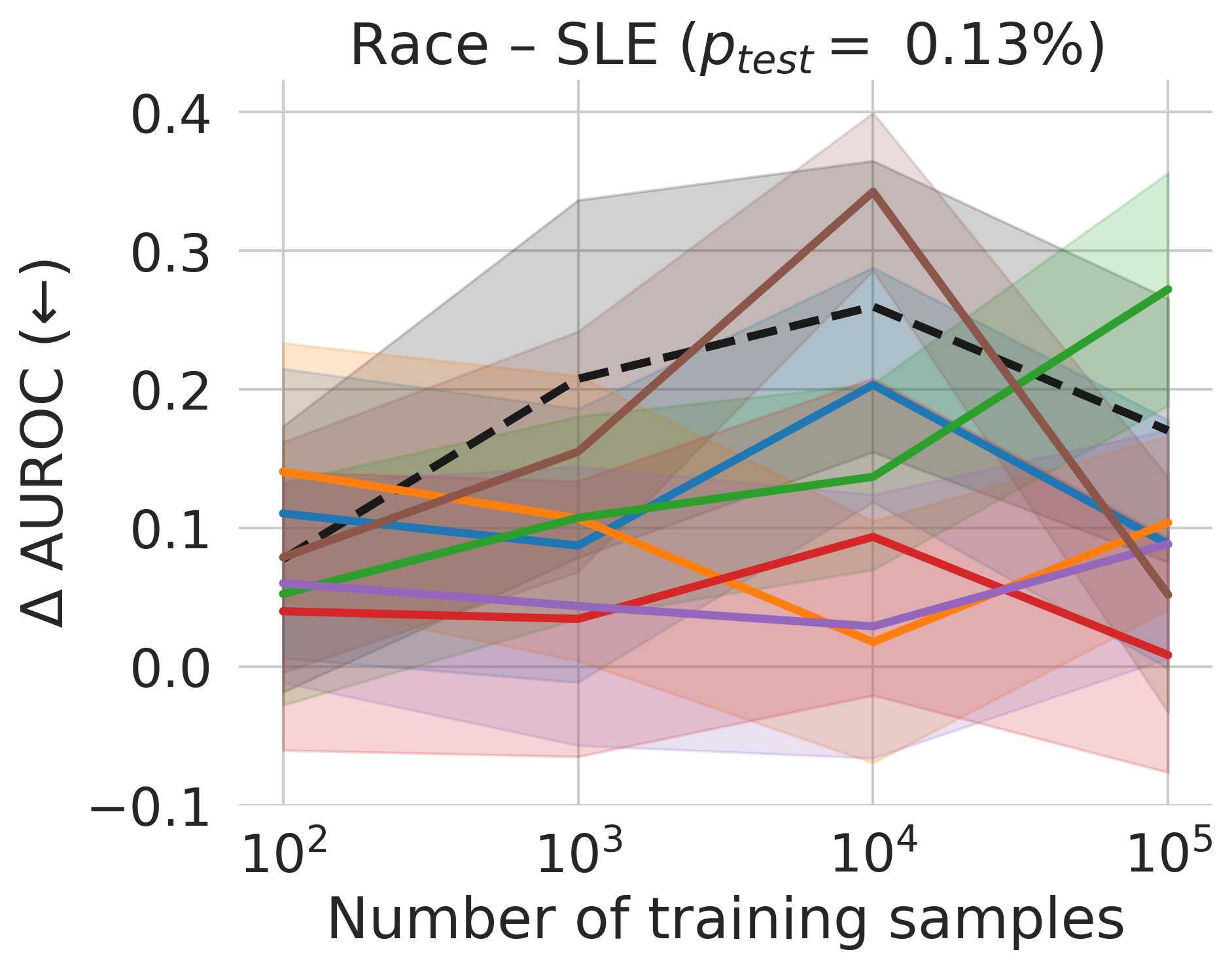}
    \includegraphics[height=100px]{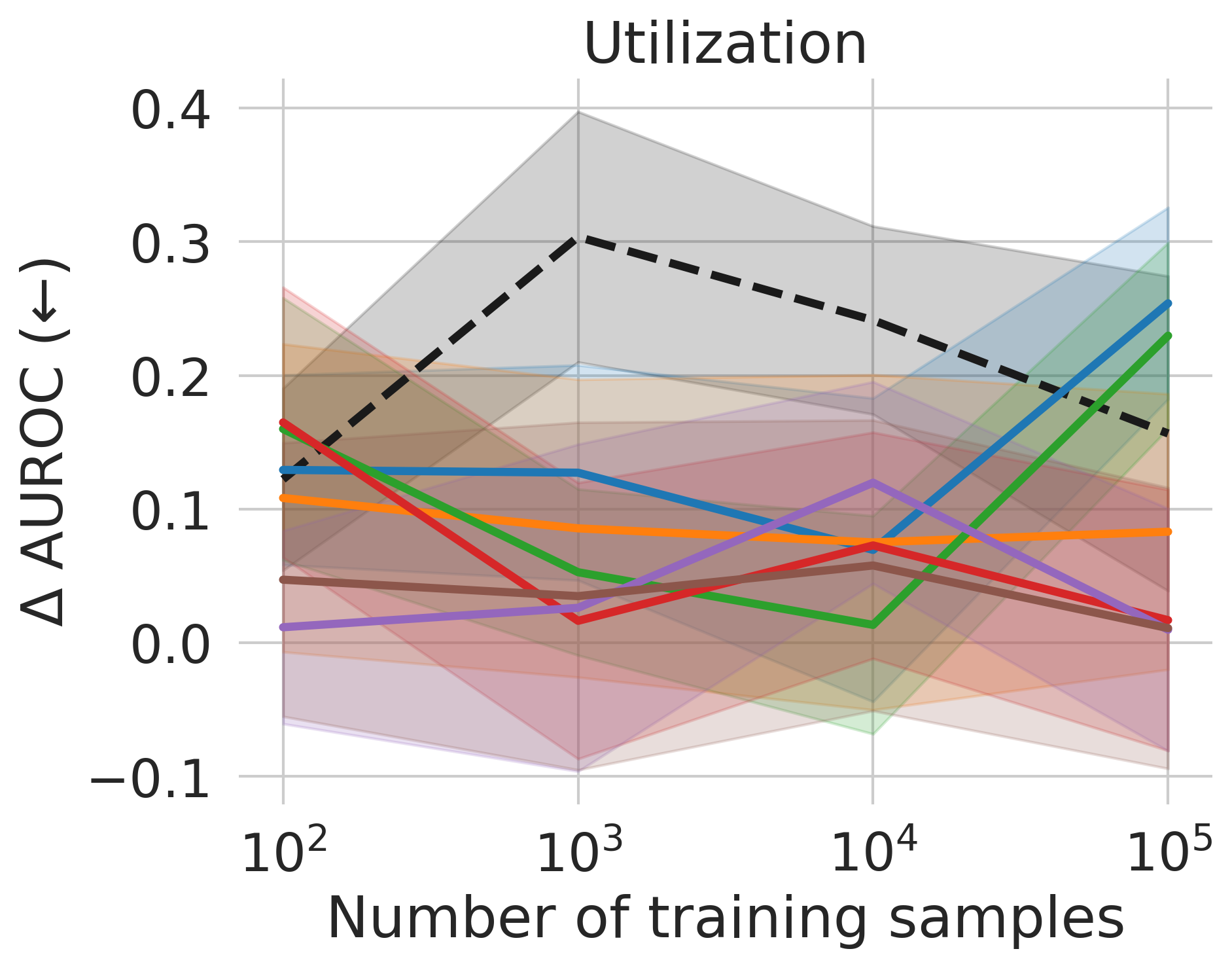}\includegraphics[height=100px]{figures/legend.png}}

    \makebox[\textwidth][l]{
    \includegraphics[height=100px]{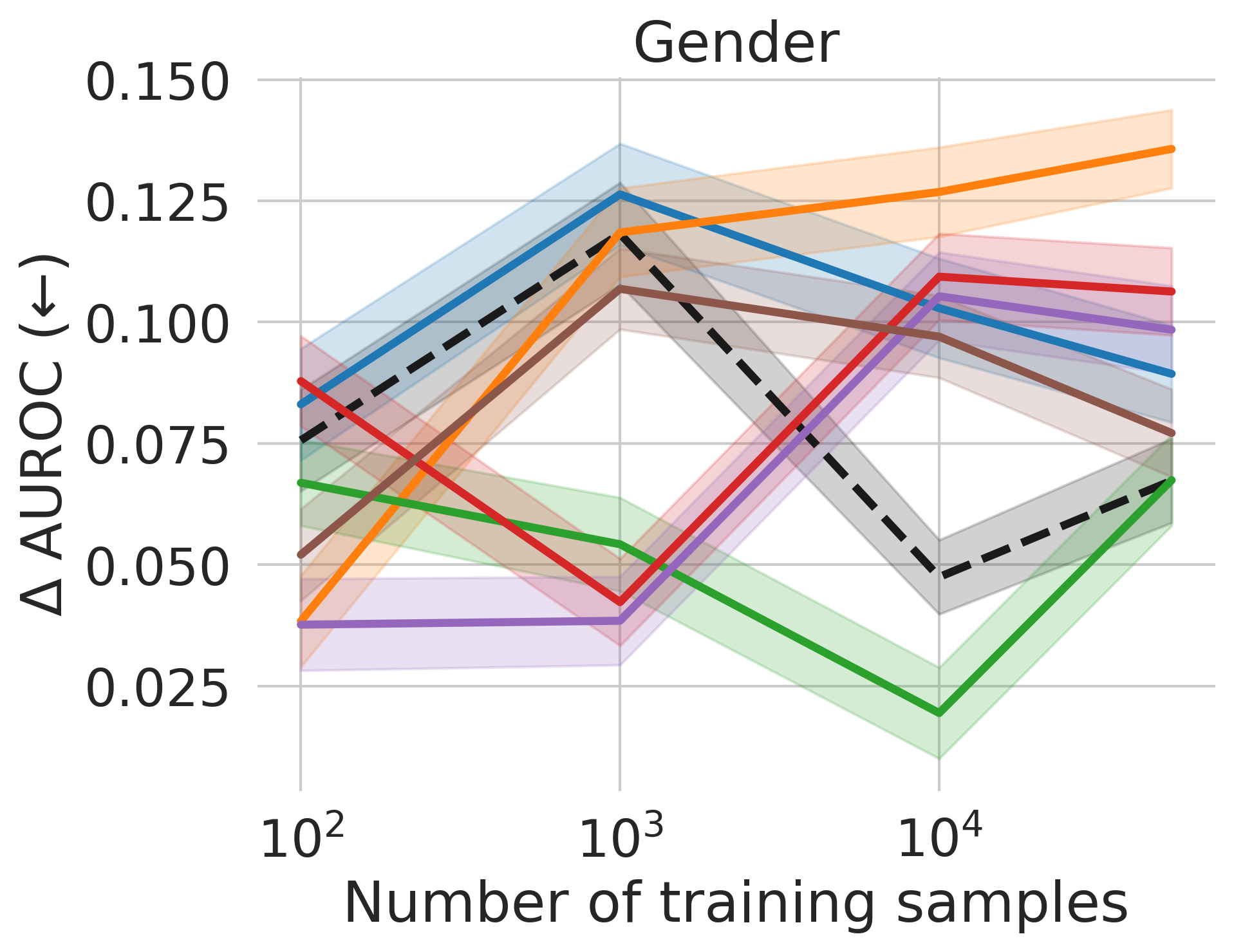}
    \includegraphics[height=100px]{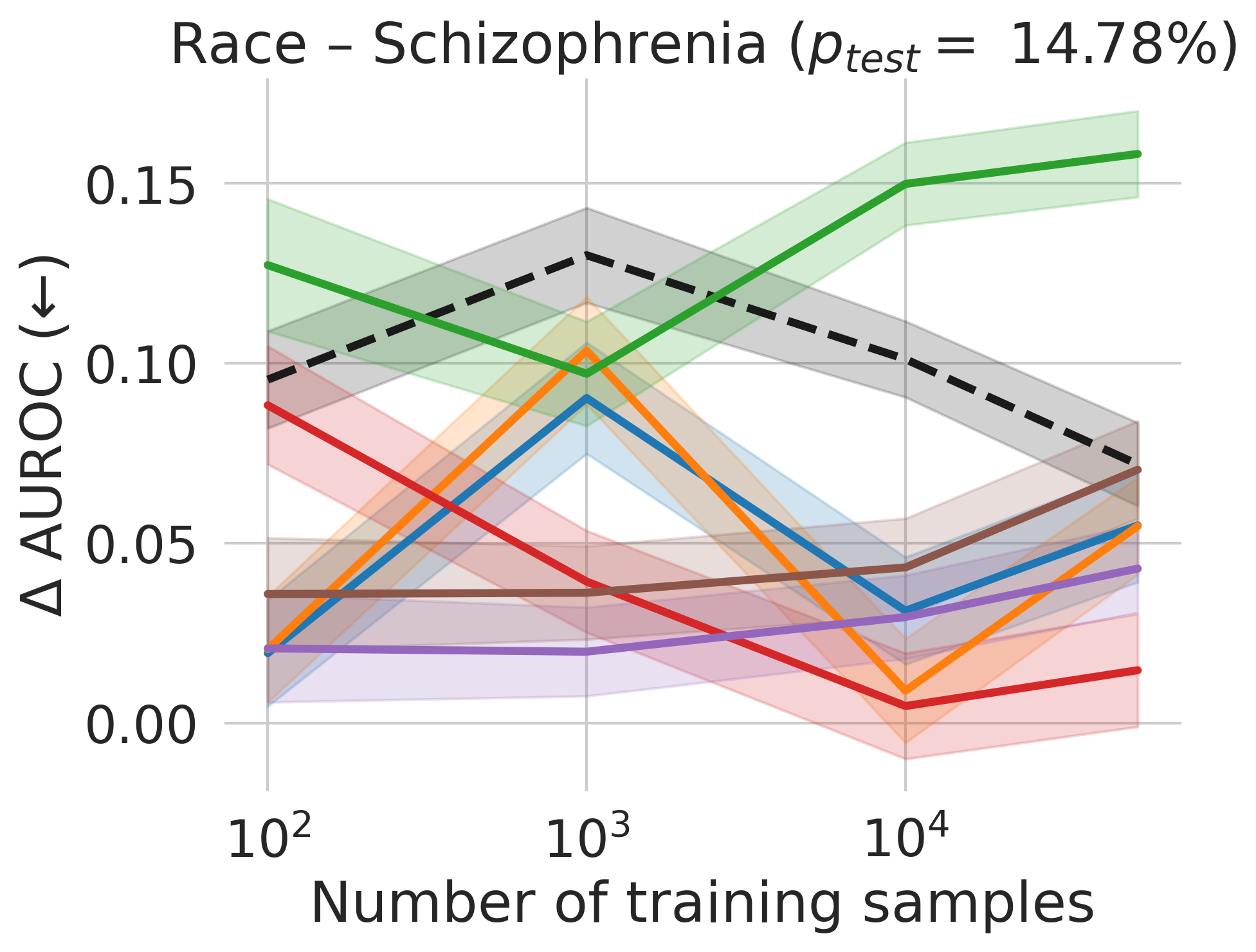}
    \includegraphics[height=100px]{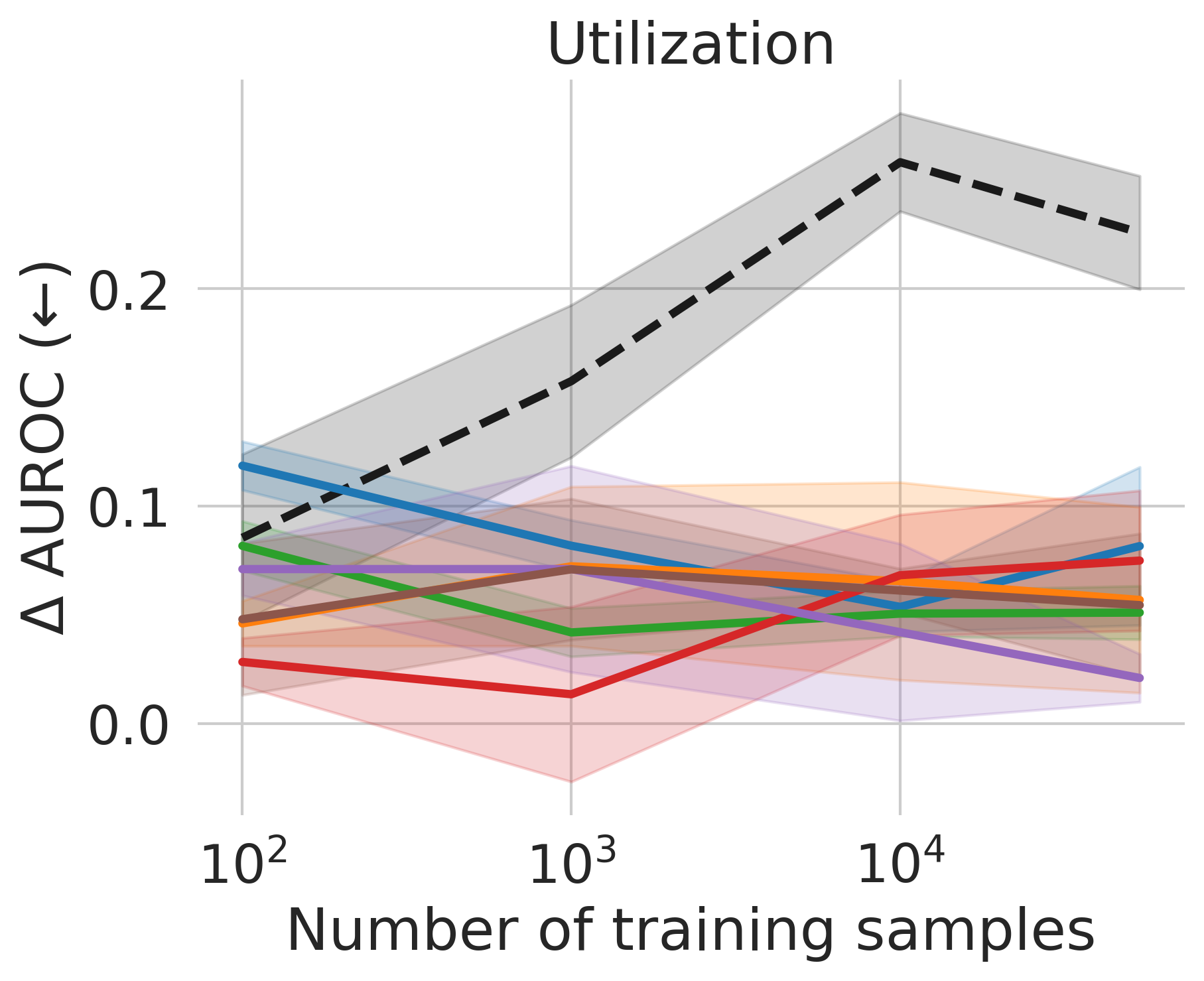}}

    \makebox[\textwidth][l]{
    \includegraphics[height=100px]{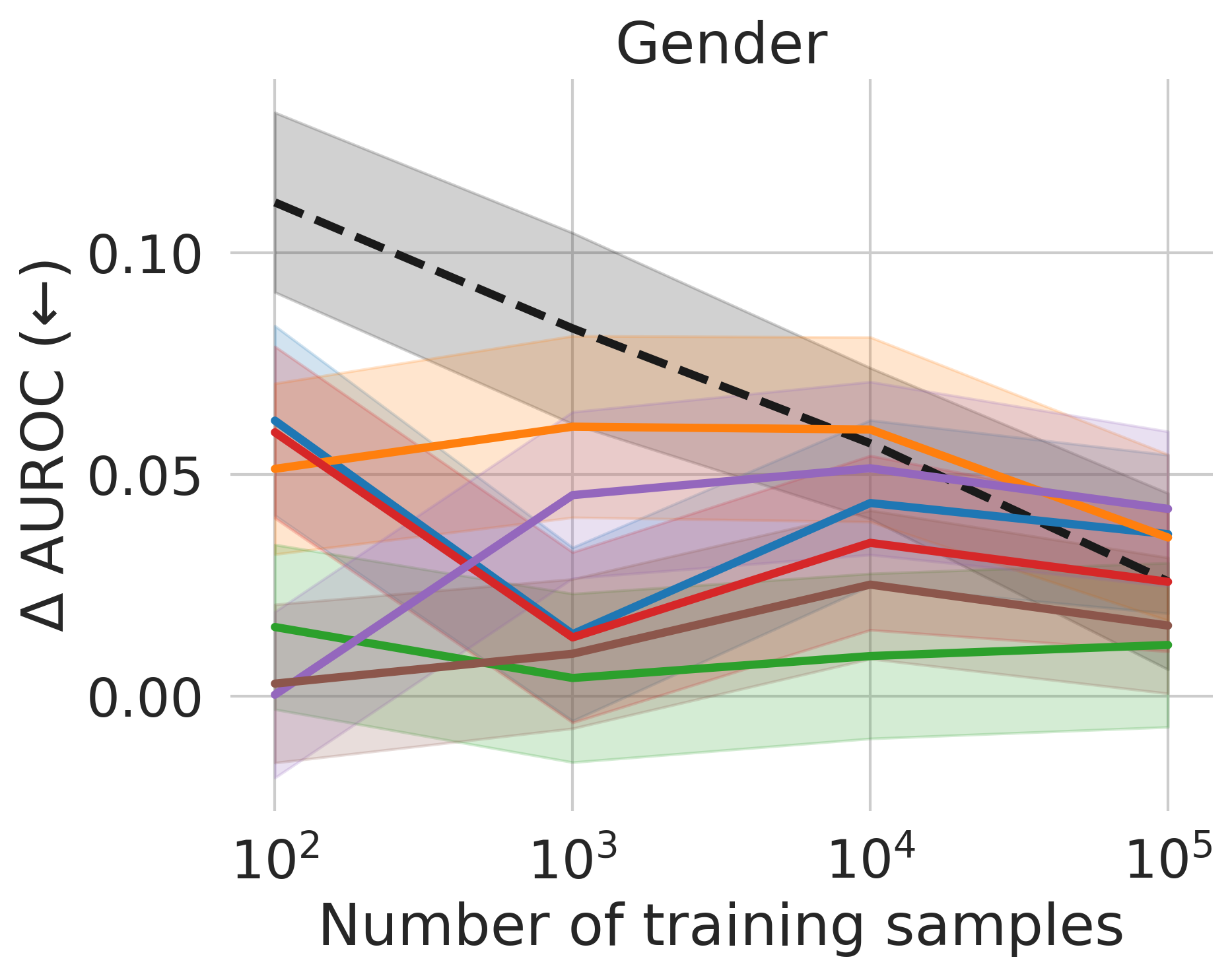}
    \includegraphics[height=100px]{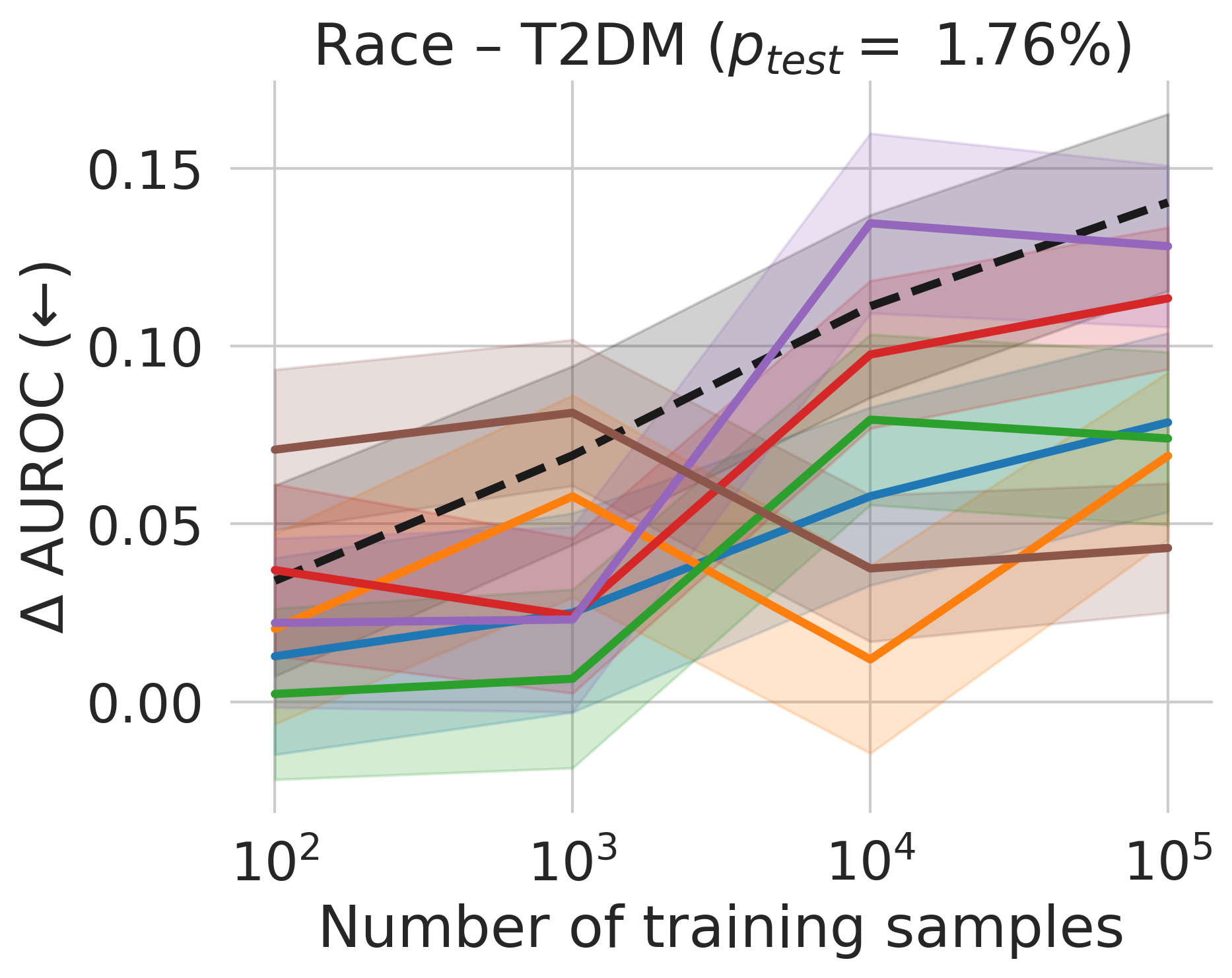}
    \includegraphics[height=100px]{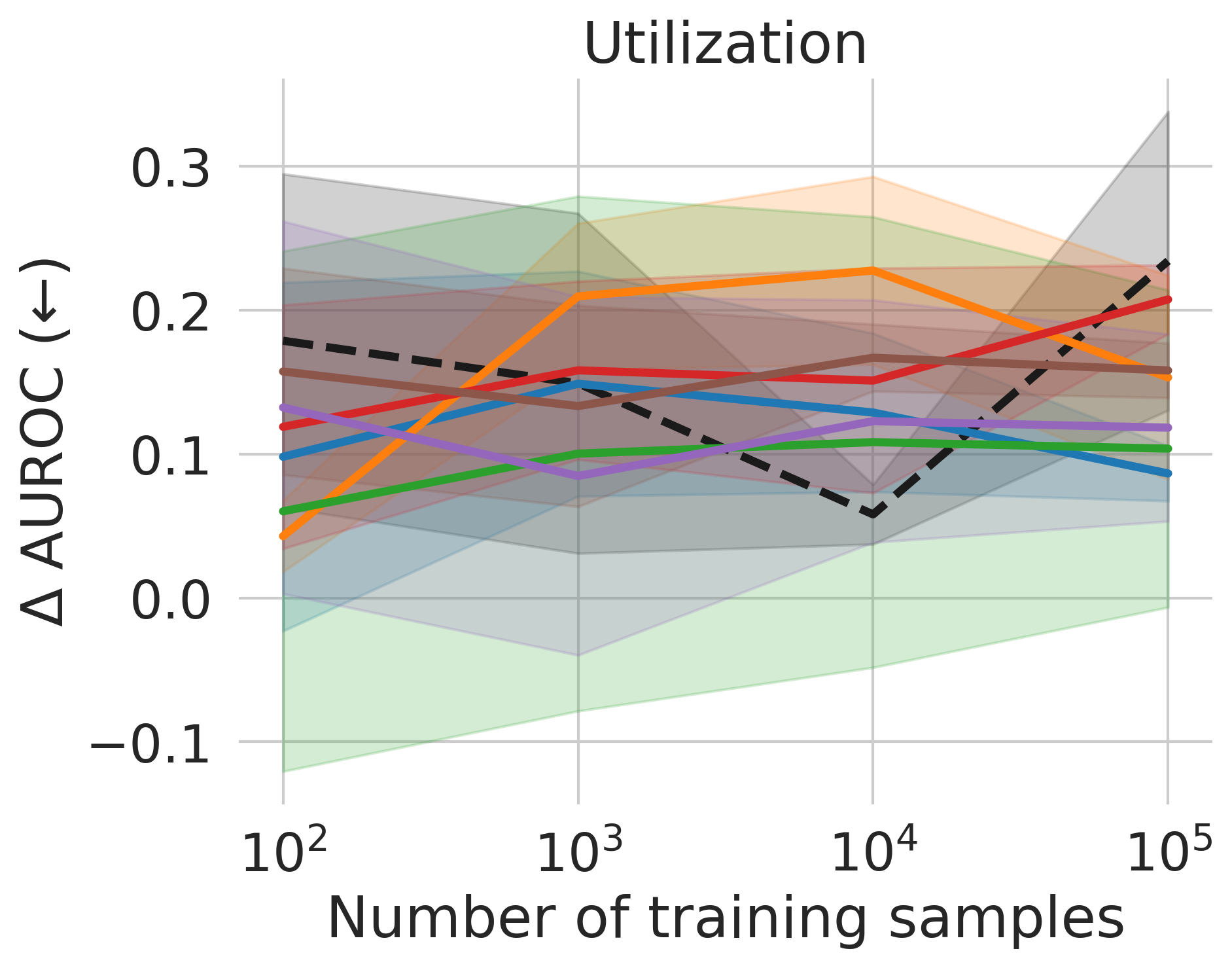}}

    \centering
    \caption{Discriminative performances and calibration results for all phenotypes for an increasing number of training points (Shaded area represents bootstrapped standard deviation) - Continued.}
    \label{fig:all_phenotypes:fairness:2}
\end{figure}

The main text focuses on the ranking of different models. The following results present detailed figures for all outcomes of interest.
Specifically, Figures~\ref{fig:all_outcomes:cumc},~\ref{fig:all_phenotypes:cumc} and~\ref{fig:all_phenotypes:2} present the discriminative and calibration for all tasks and models given increasing amounts of training data, while Figures~\ref{fig:all_outcomes:cumc:fairness},~\ref{fig:all_phenotypes:cumc:fairness} and~\ref{fig:all_phenotypes:fairness:2} present the fairness discriminative gap for all tasks stratified by sex, race\footnote{Due to the granularity of available demographic data, we consider two groups of interest: Black and White patients.}, and healthcare utilization.

\clearpage
\subsection{Transport performance}
\label{app:transport}
\begin{figure}[!ht]
    \makebox[\textwidth][l]{
    \includegraphics[height=100px]{figures/results/CUMC/transport_long_los_roc_auc_score_all.png}    
    \includegraphics[height=100px]{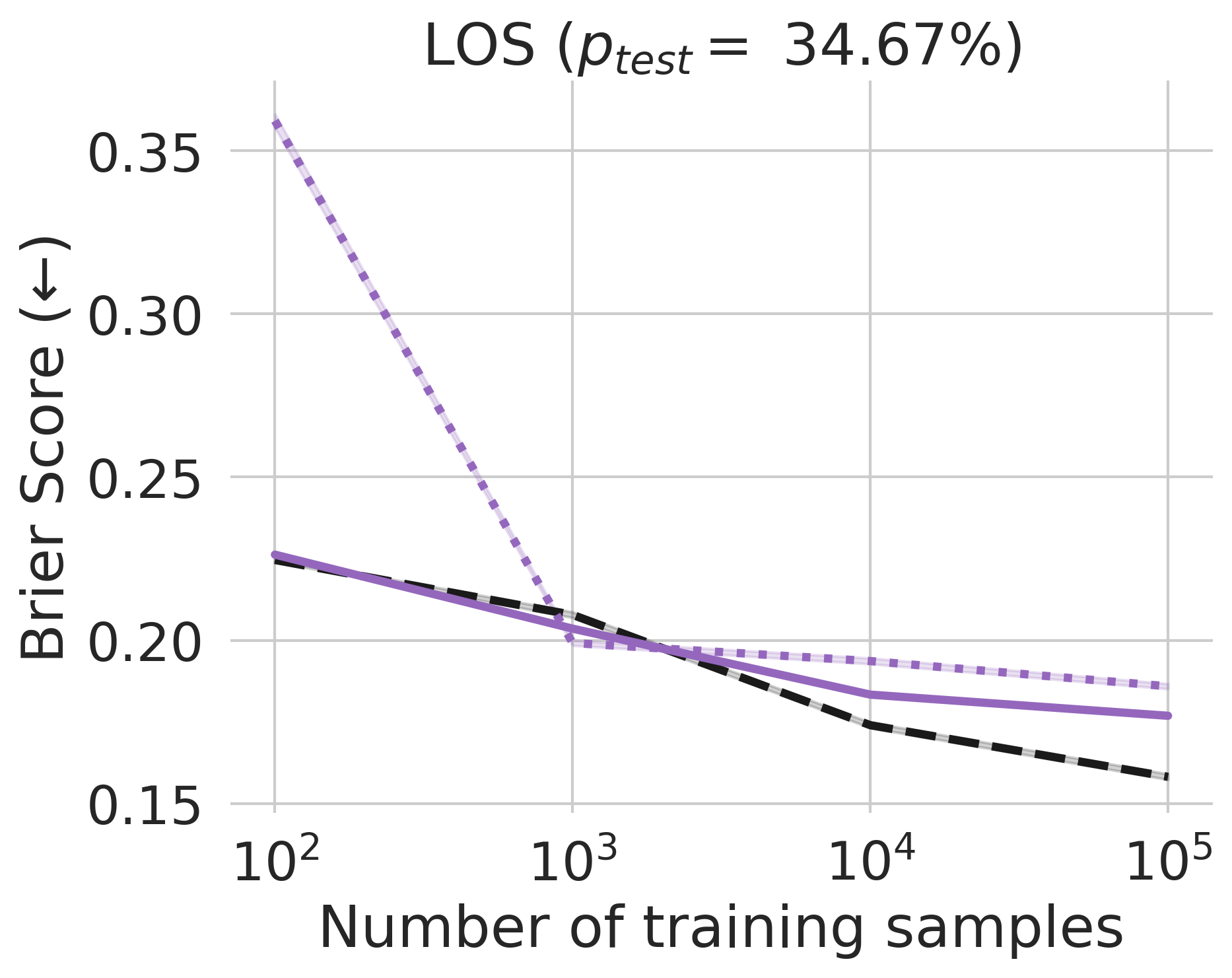}
    \includegraphics[height=100px]{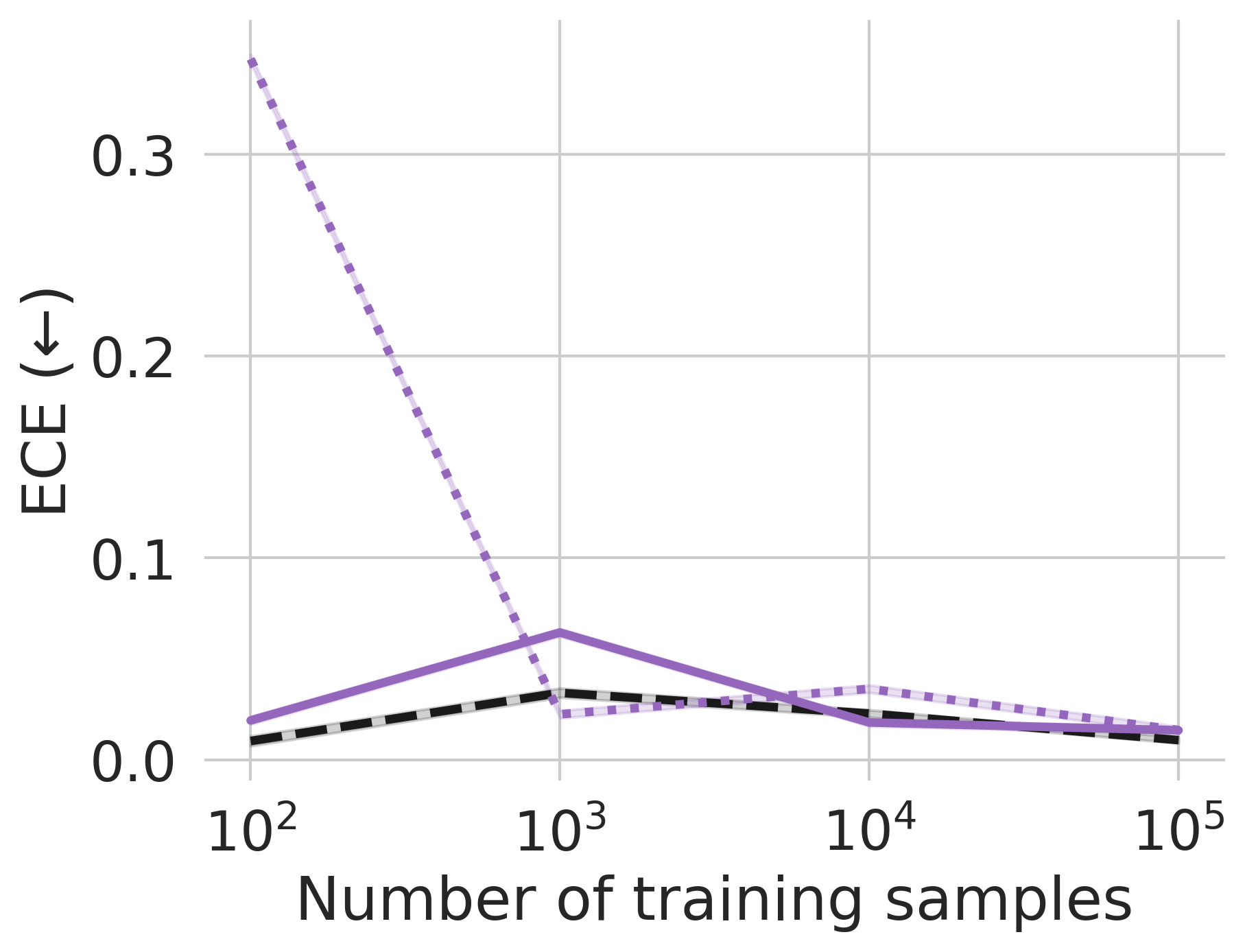}}

    \makebox[\textwidth][l]{
    \includegraphics[height=100px]{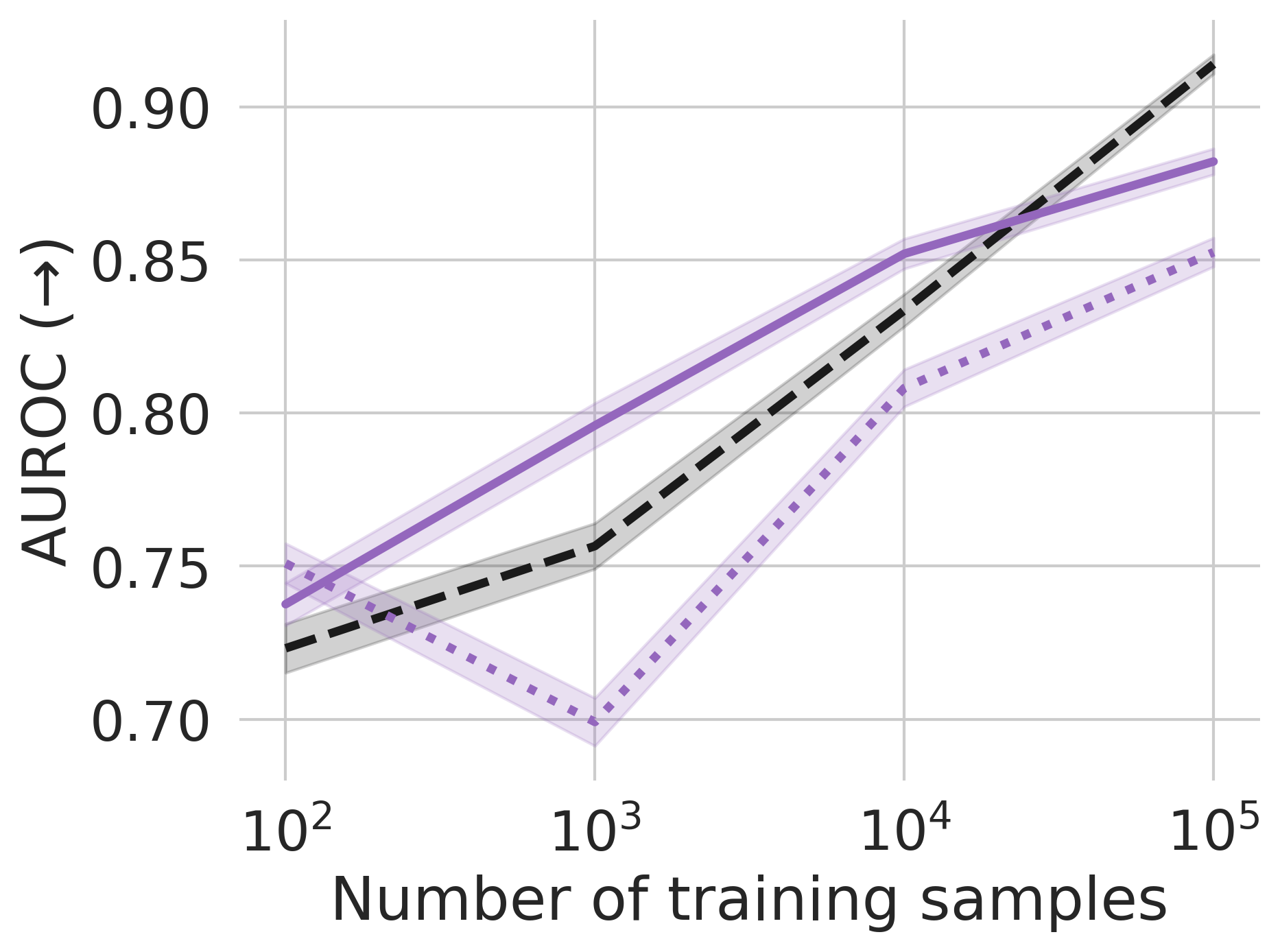}
    \includegraphics[height=100px]{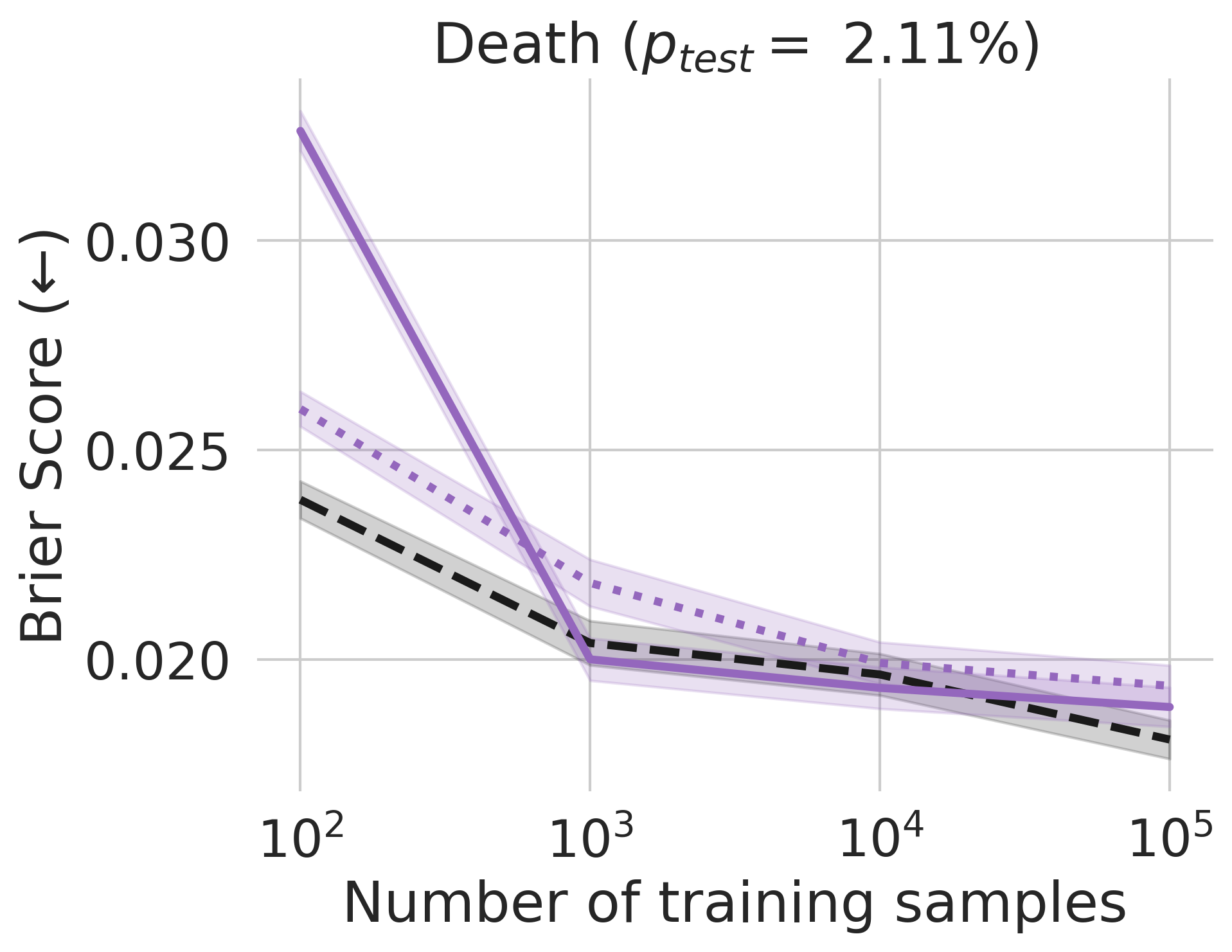}
    \includegraphics[height=100px]{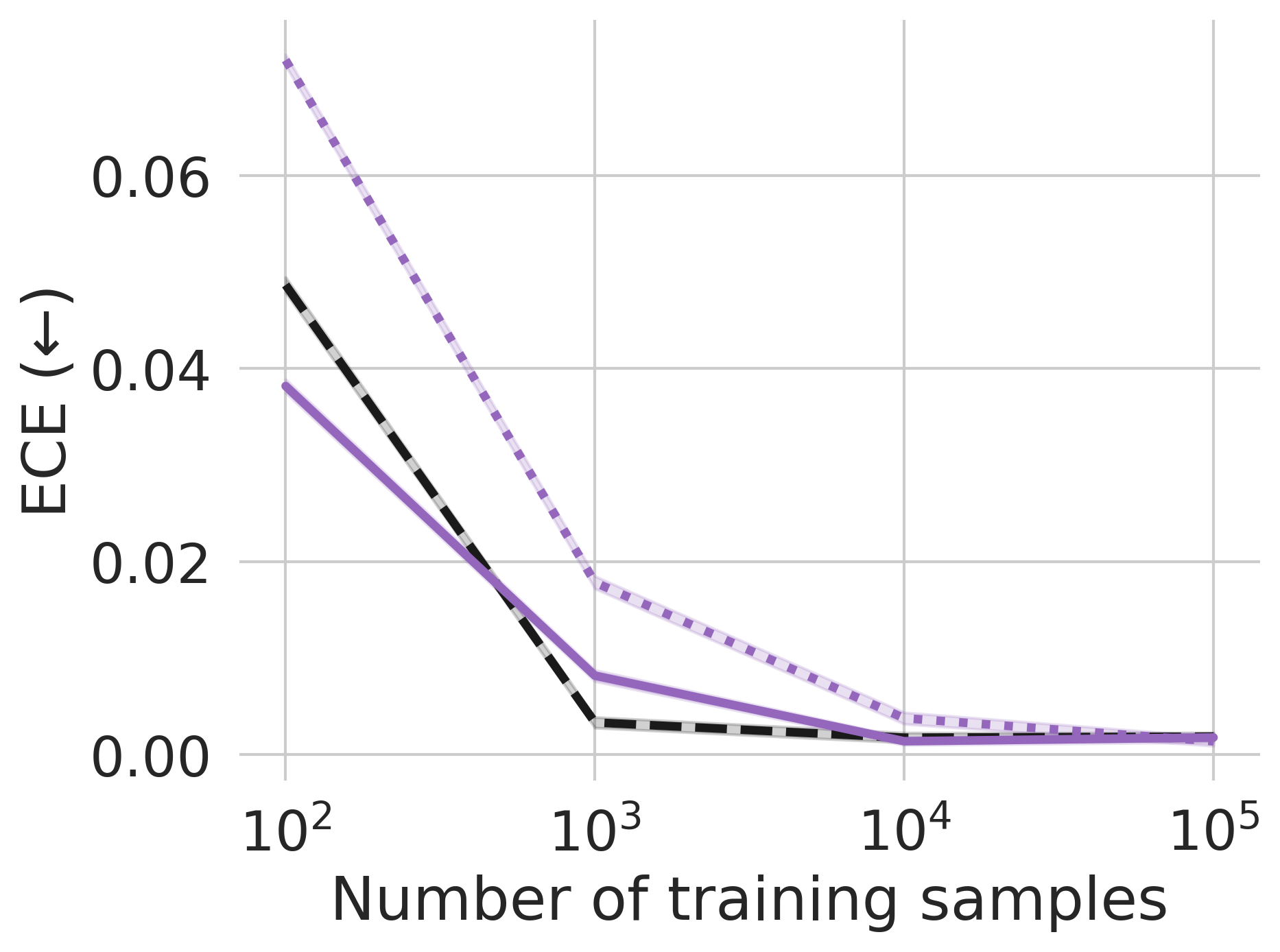}
    \includegraphics[height=100px]{figures/legend_transport.png}}

    \makebox[\textwidth][l]{
    \includegraphics[height=100px]{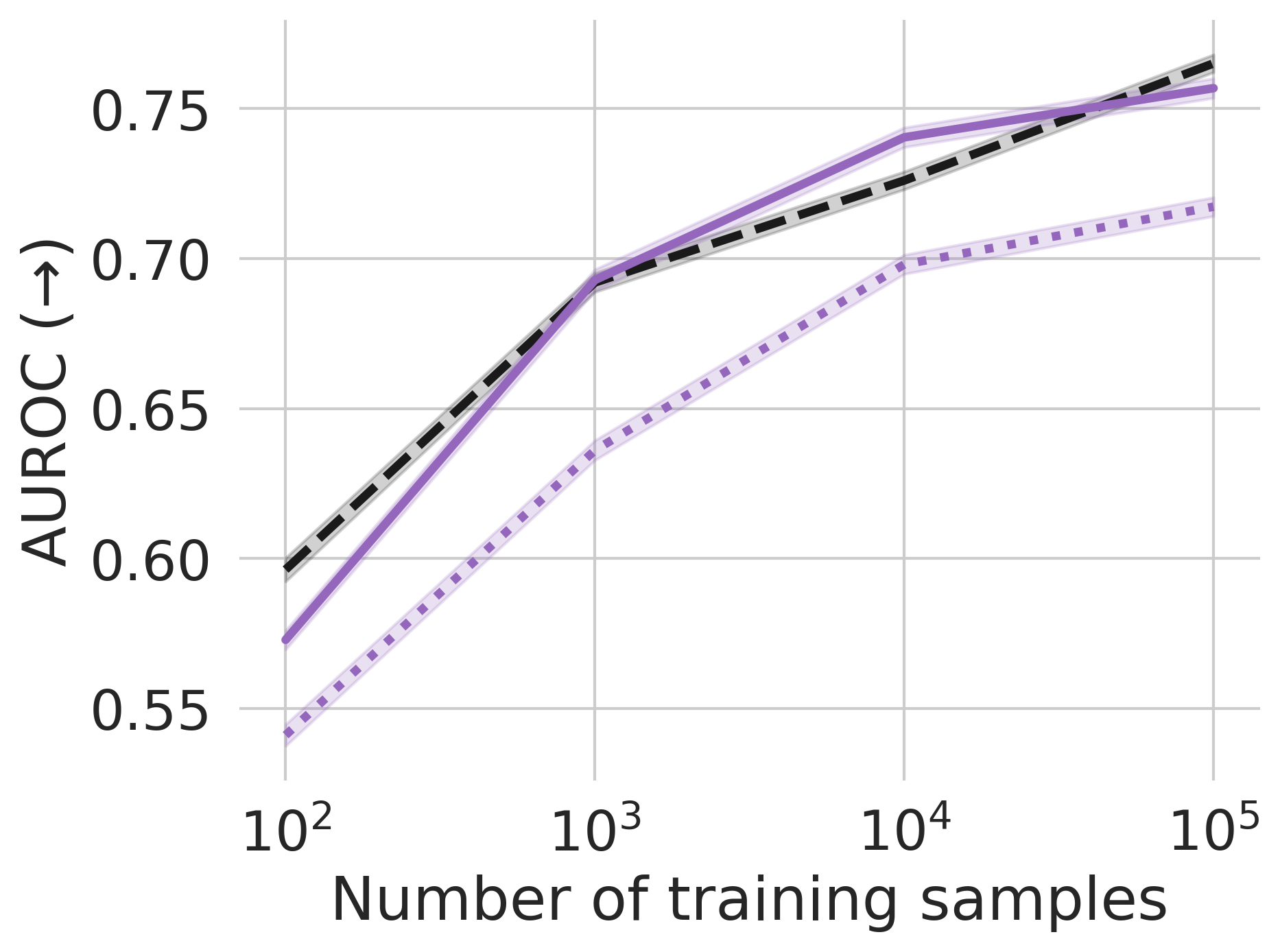}
    \includegraphics[height=100px]{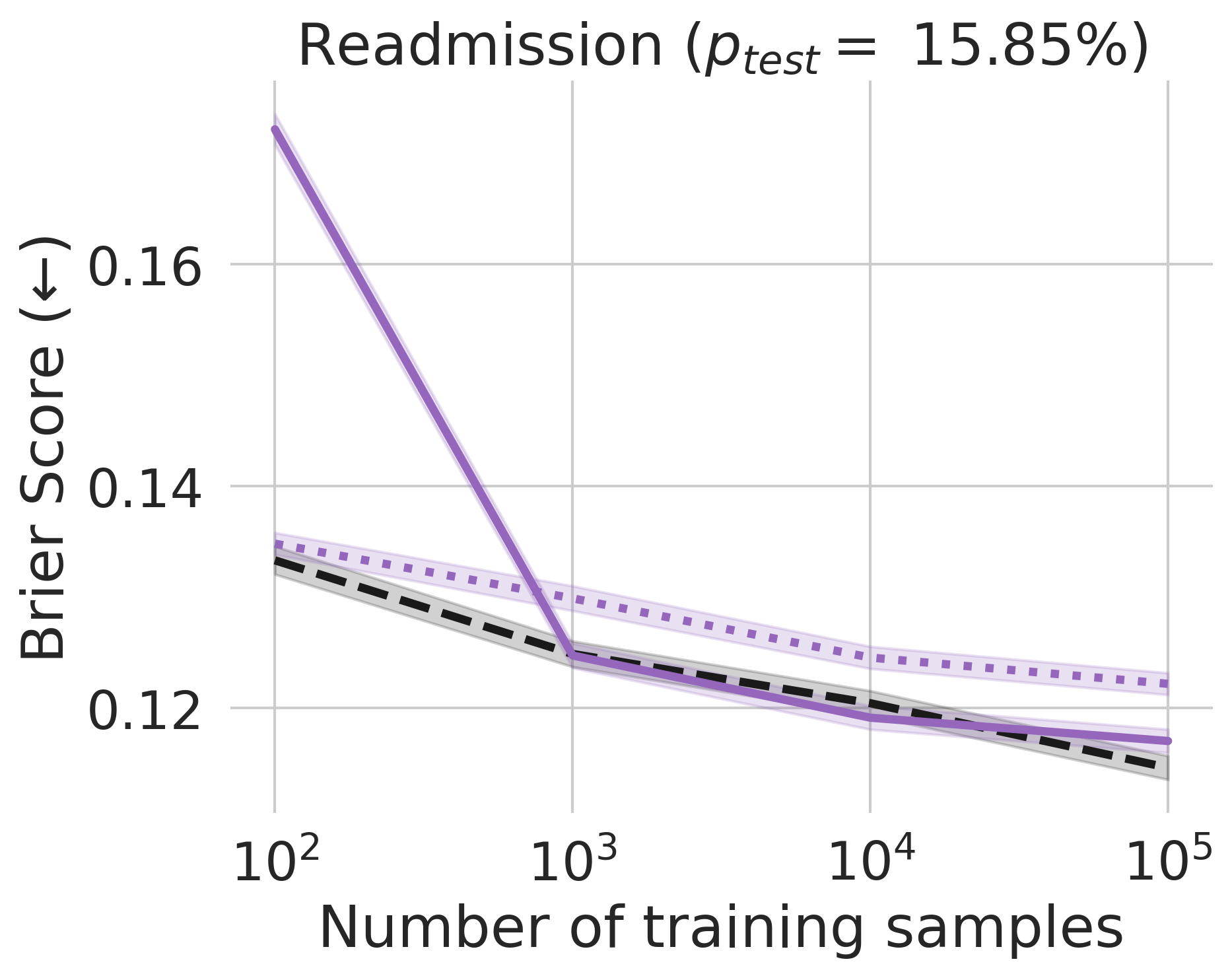}
    \includegraphics[height=100px]{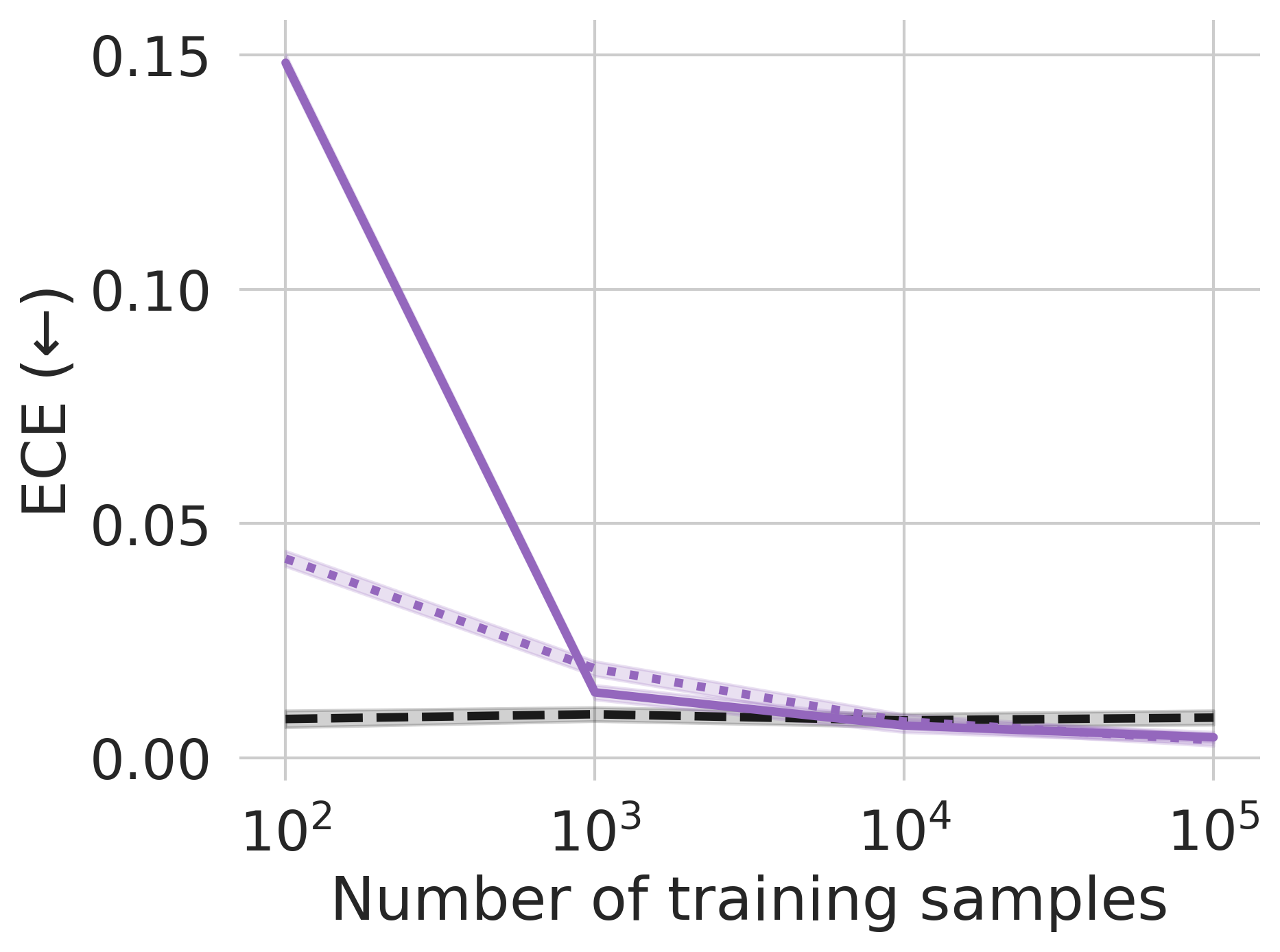}}

    \centering
    \caption{Discriminative performances and calibration results for transported \mamba (Shaded area represents bootstrapped standard deviation).}
    \label{fig:all_outcomes:cumc:transport}
\end{figure}

\begin{figure}[!ht]
    \makebox[\textwidth][l]{
    \includegraphics[height=100px]{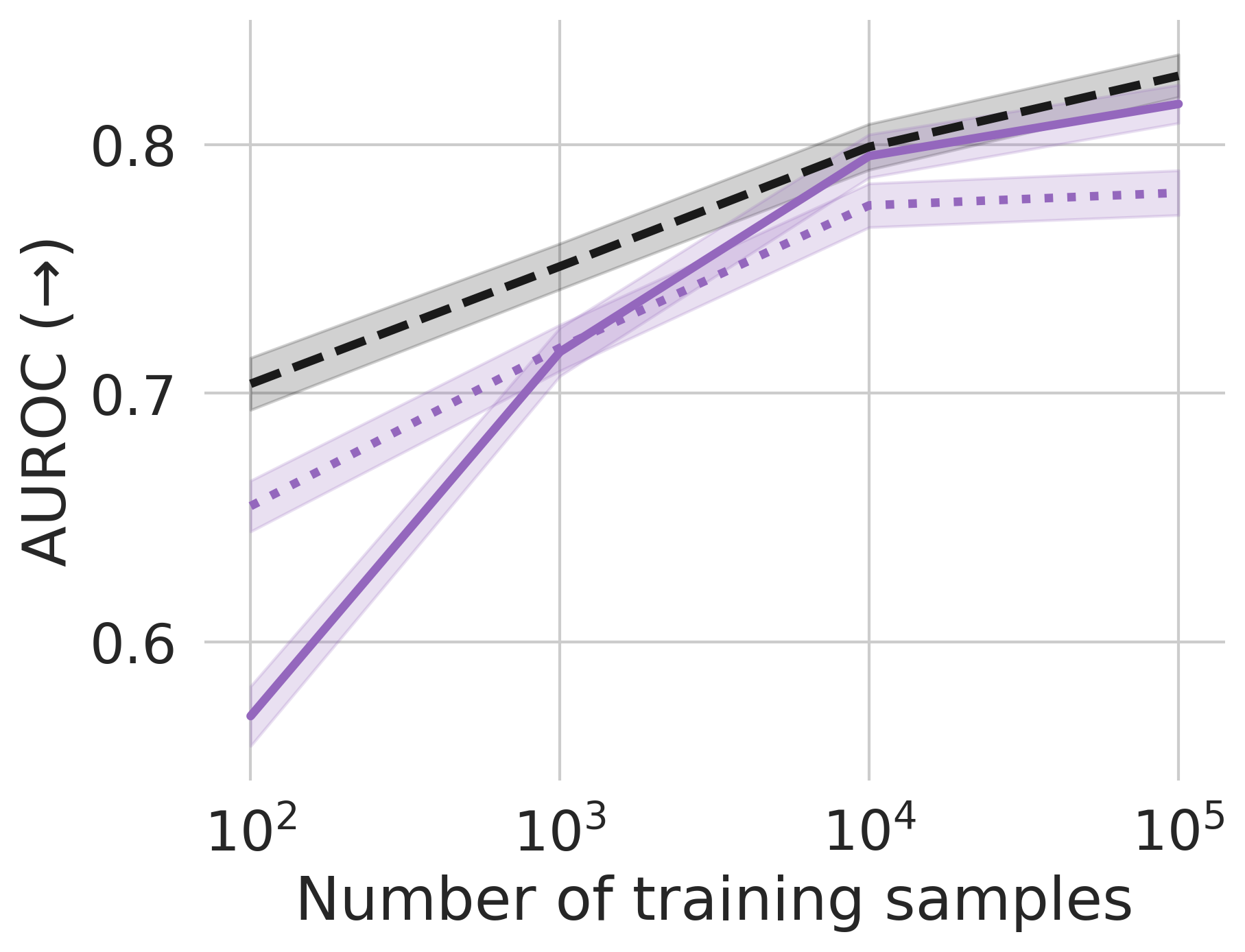}    
    \includegraphics[height=100px]{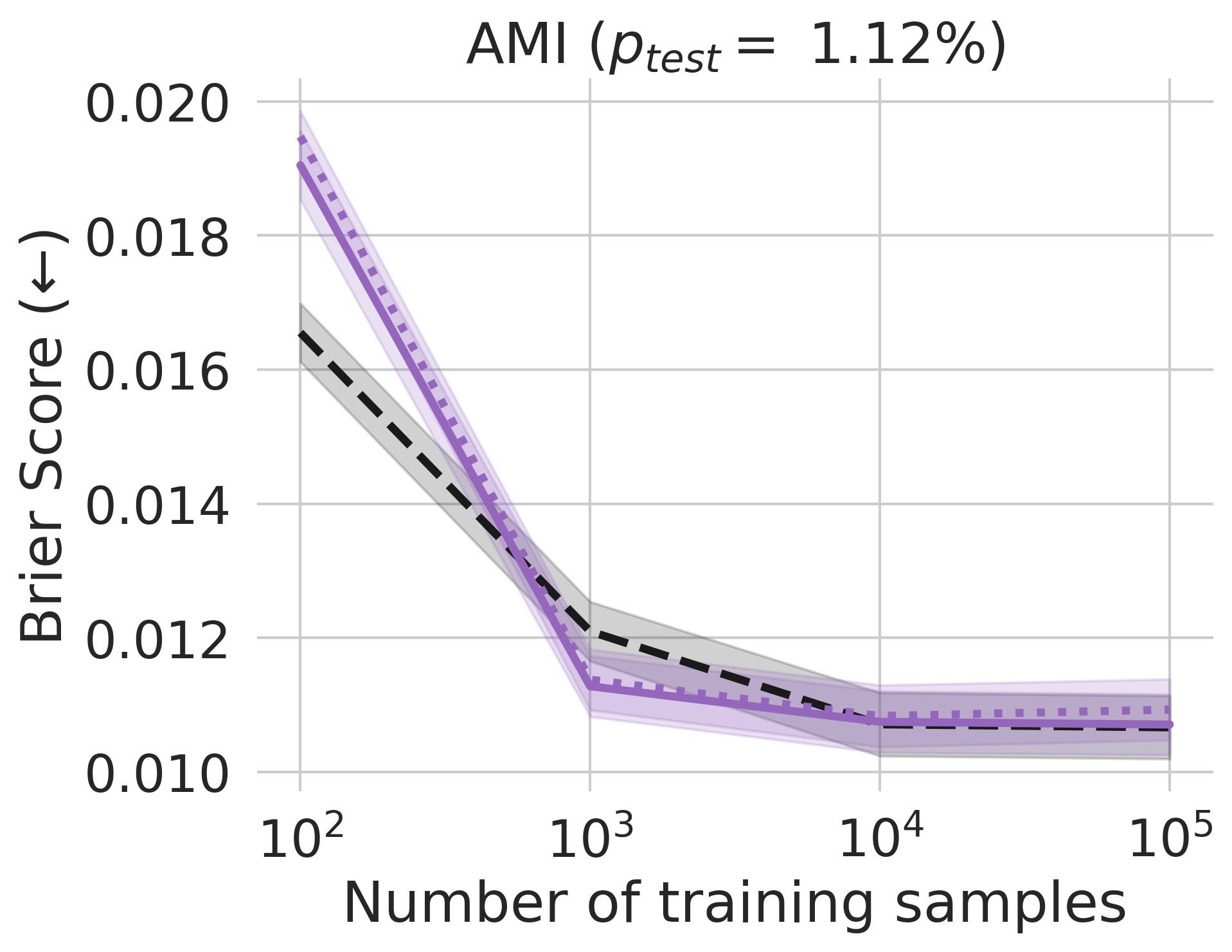}
    \includegraphics[height=100px]{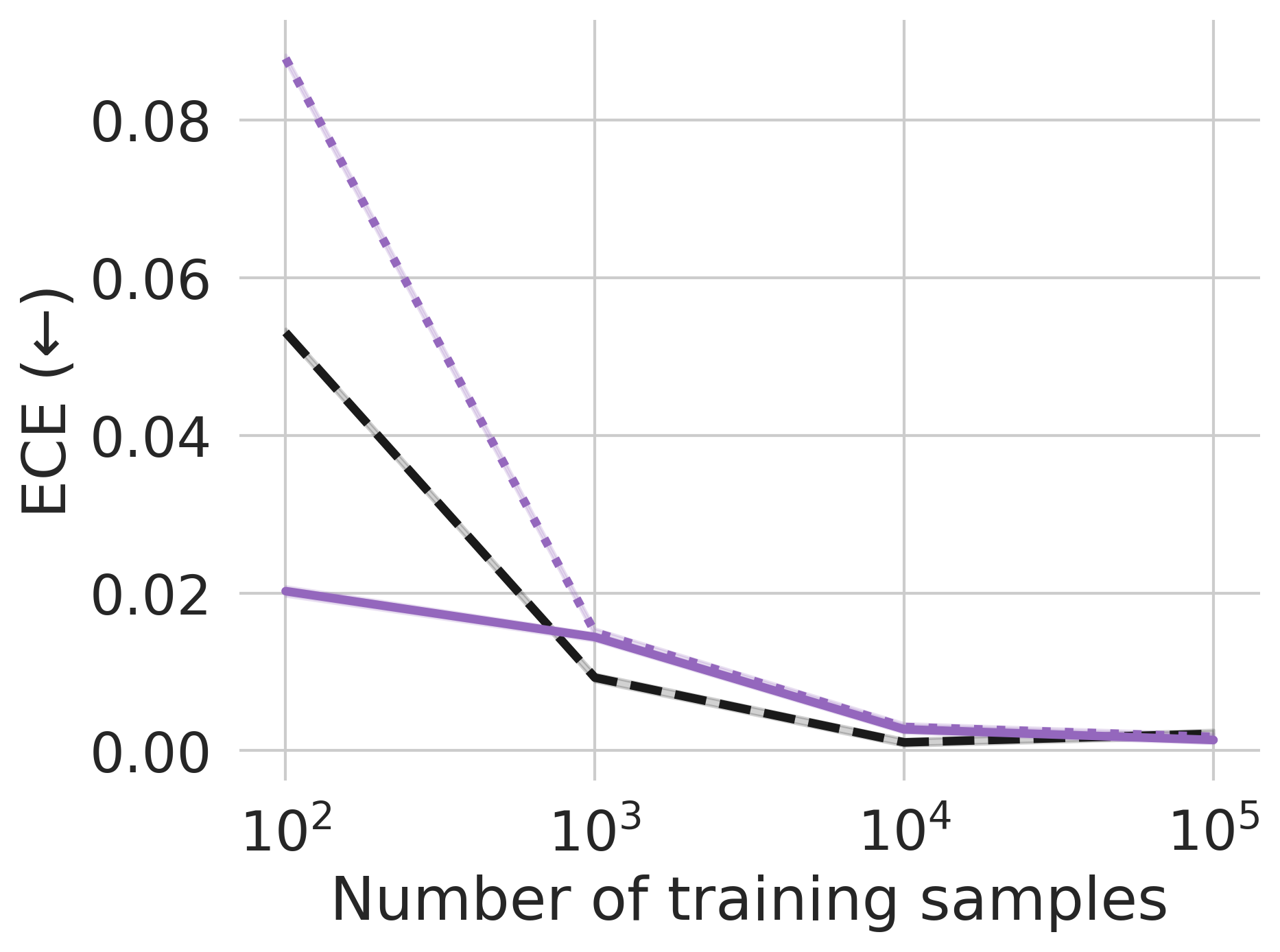}}

    \makebox[\textwidth][l]{
    \includegraphics[height=100px]{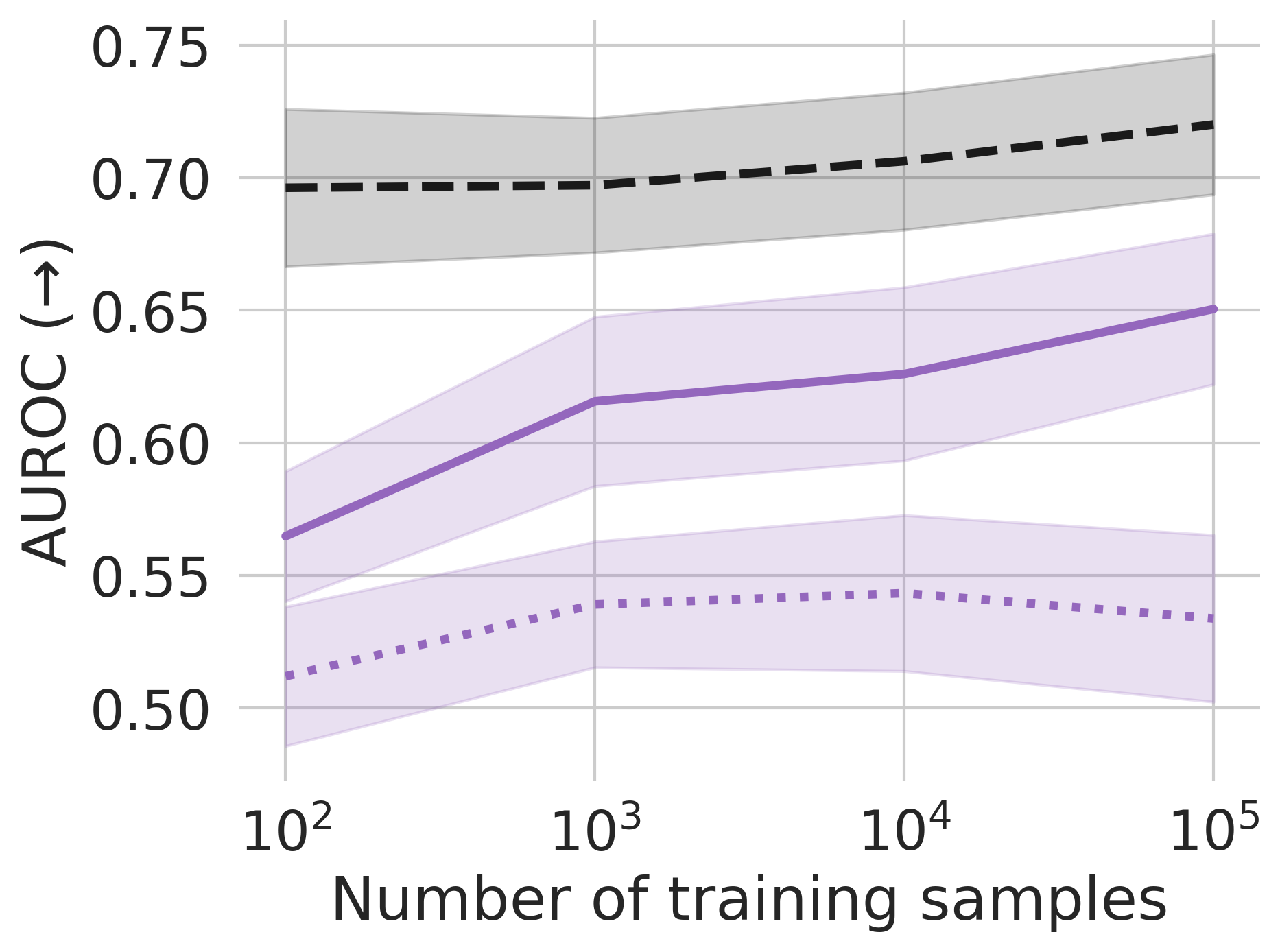}
    \includegraphics[height=100px]{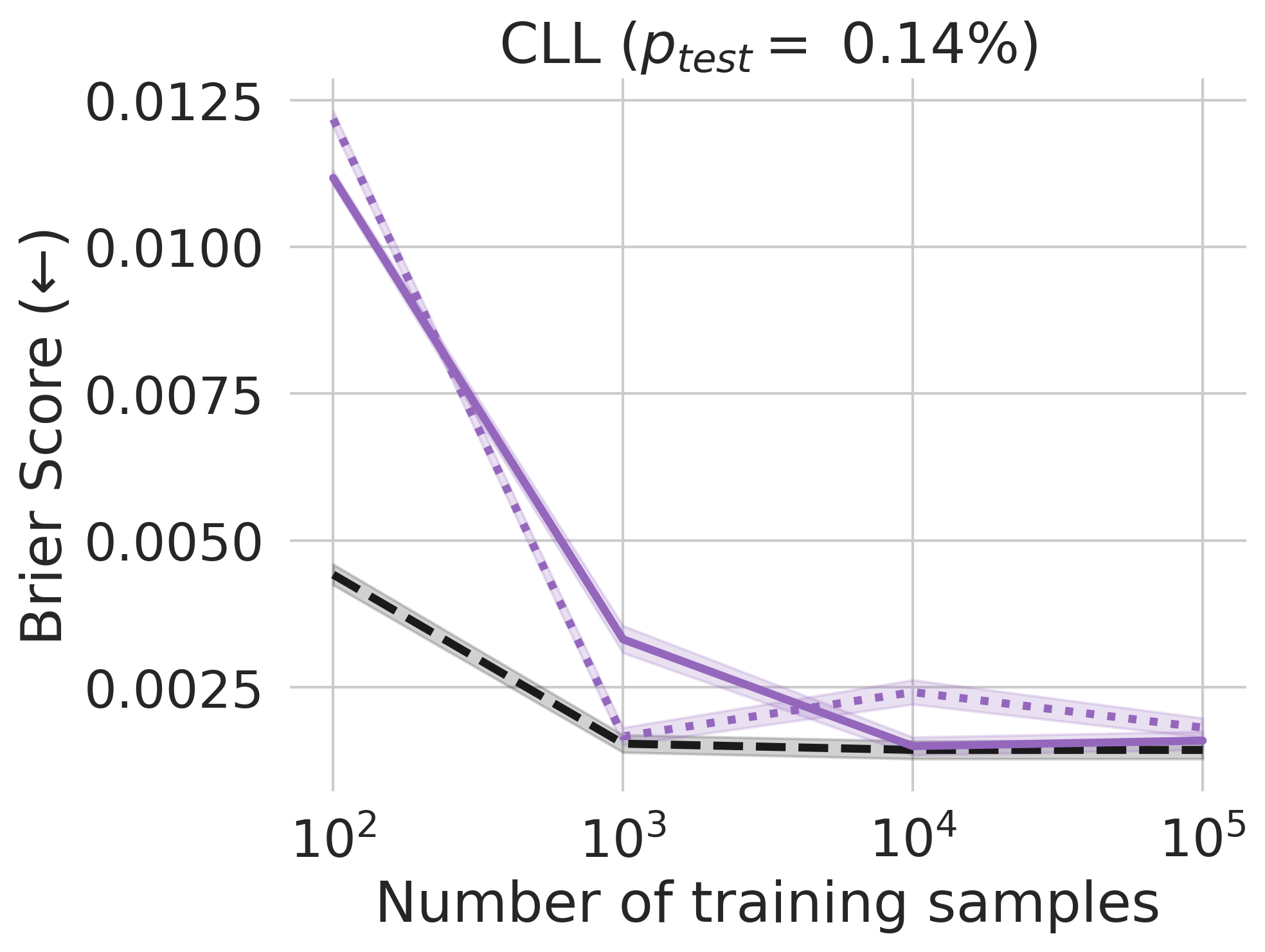}
    \includegraphics[height=100px]{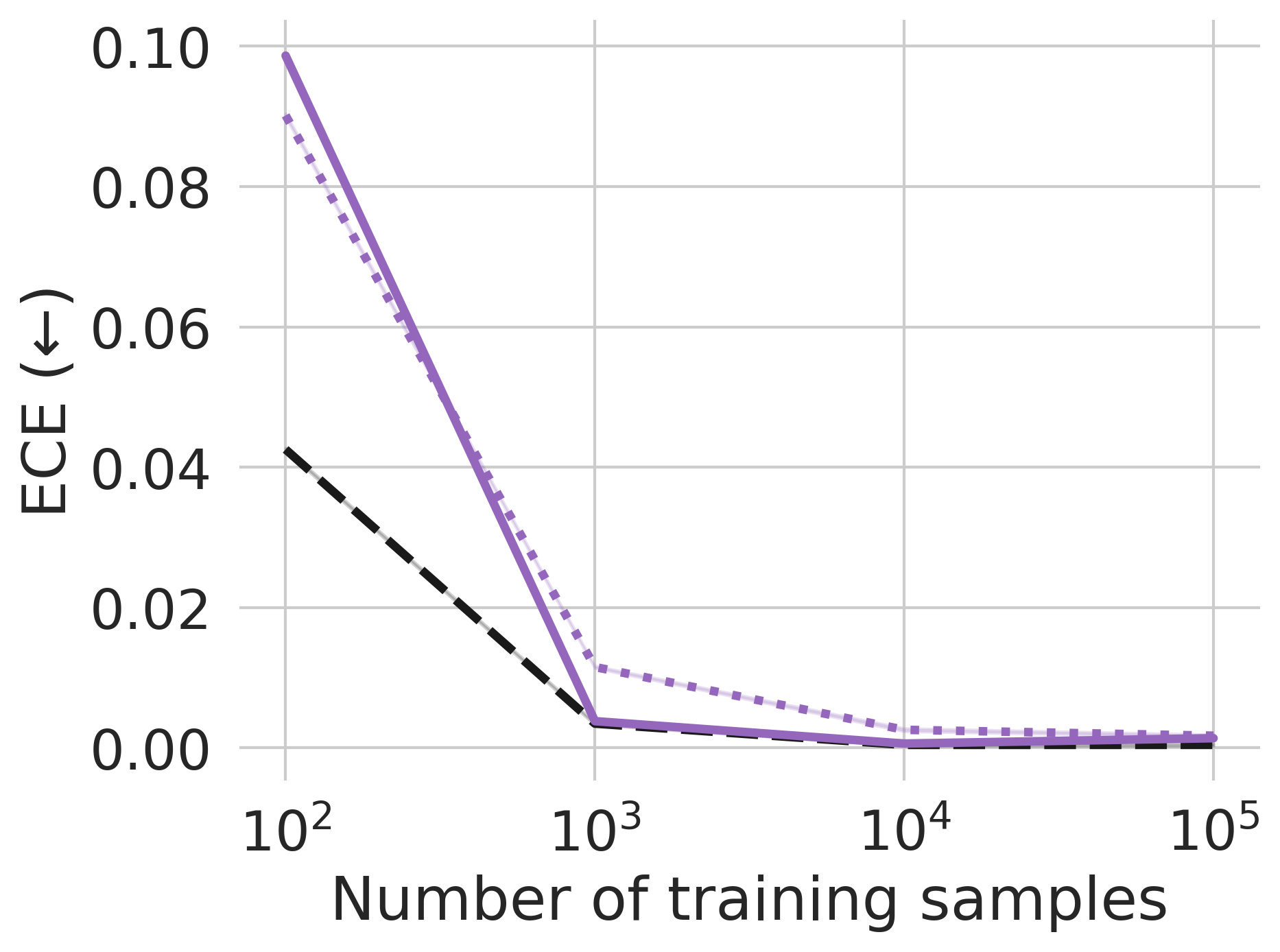}}

    \makebox[\textwidth][l]{
    \includegraphics[height=100px]{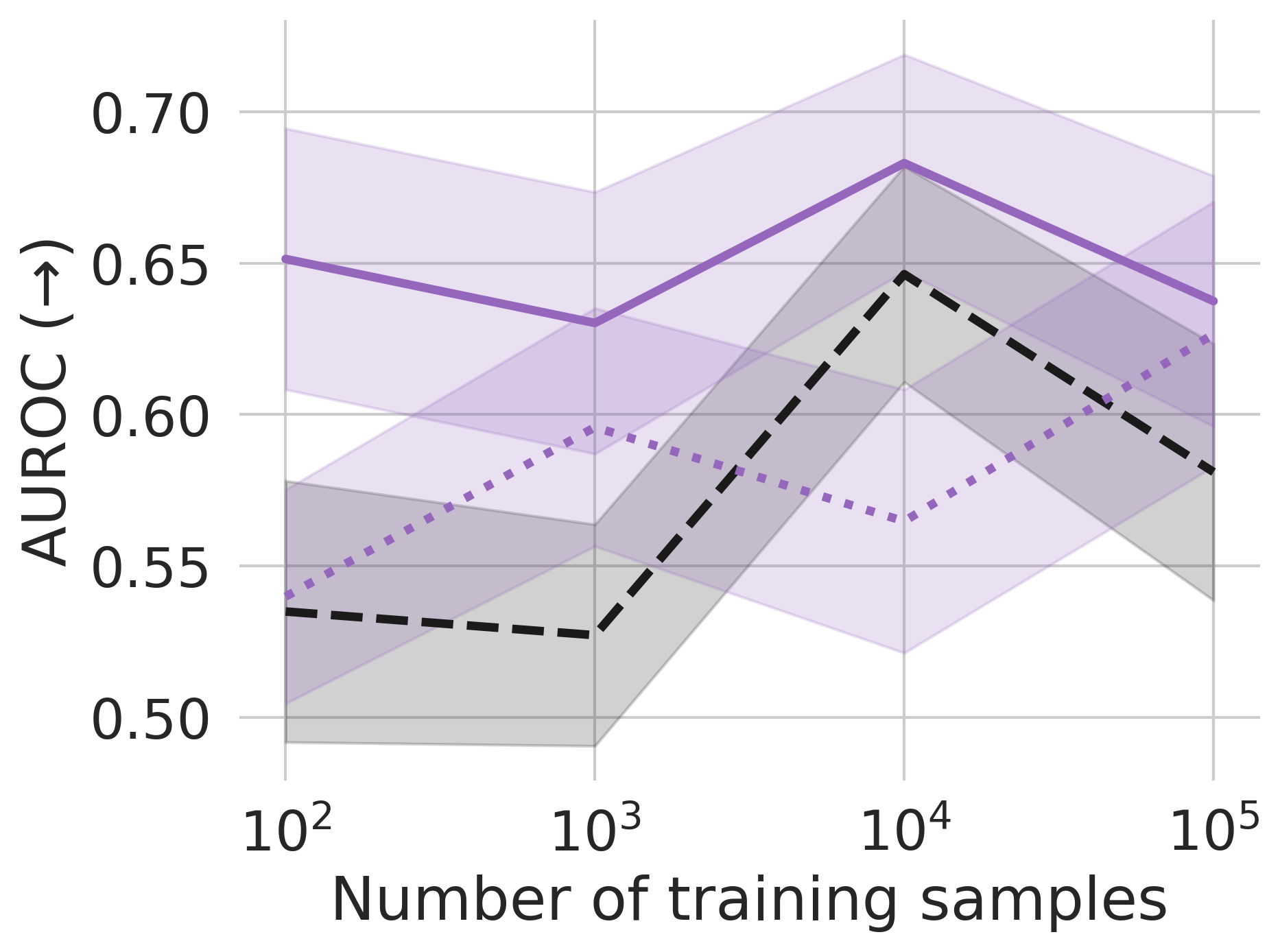}
    \includegraphics[height=100px]{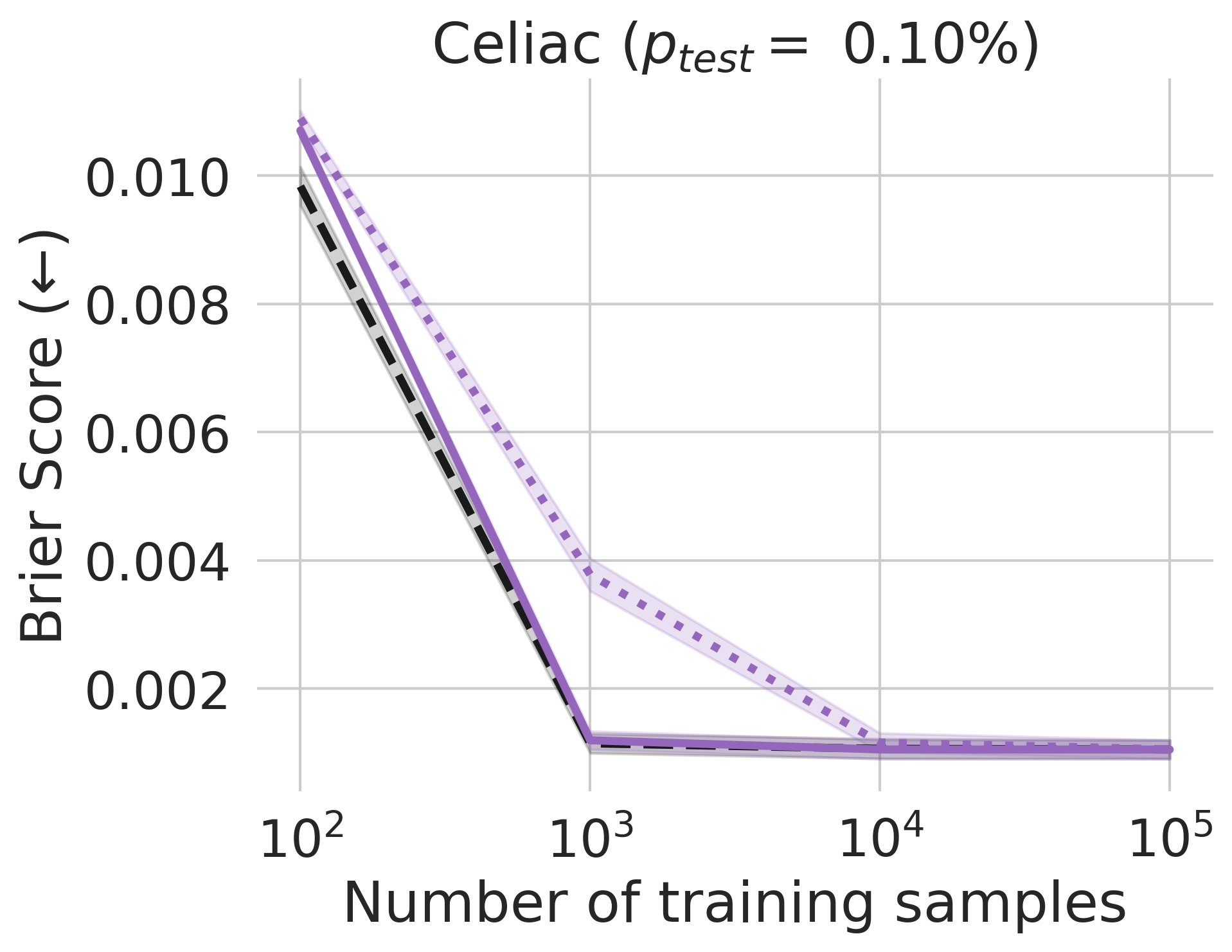}
    \includegraphics[height=100px]{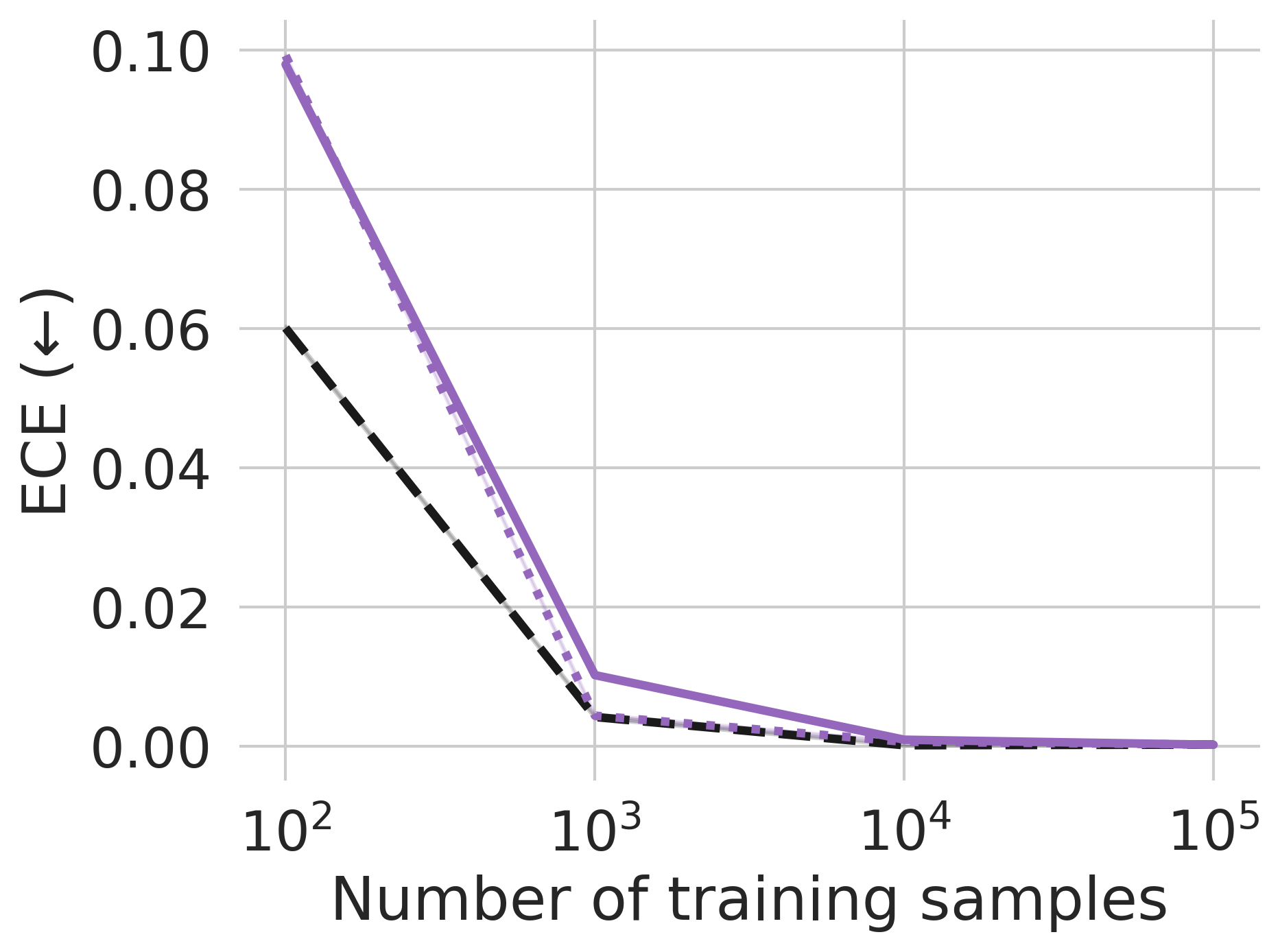}
    \includegraphics[height=100px]{figures/legend_transport.png}}

    \makebox[\textwidth][l]{
    \includegraphics[height=100px]{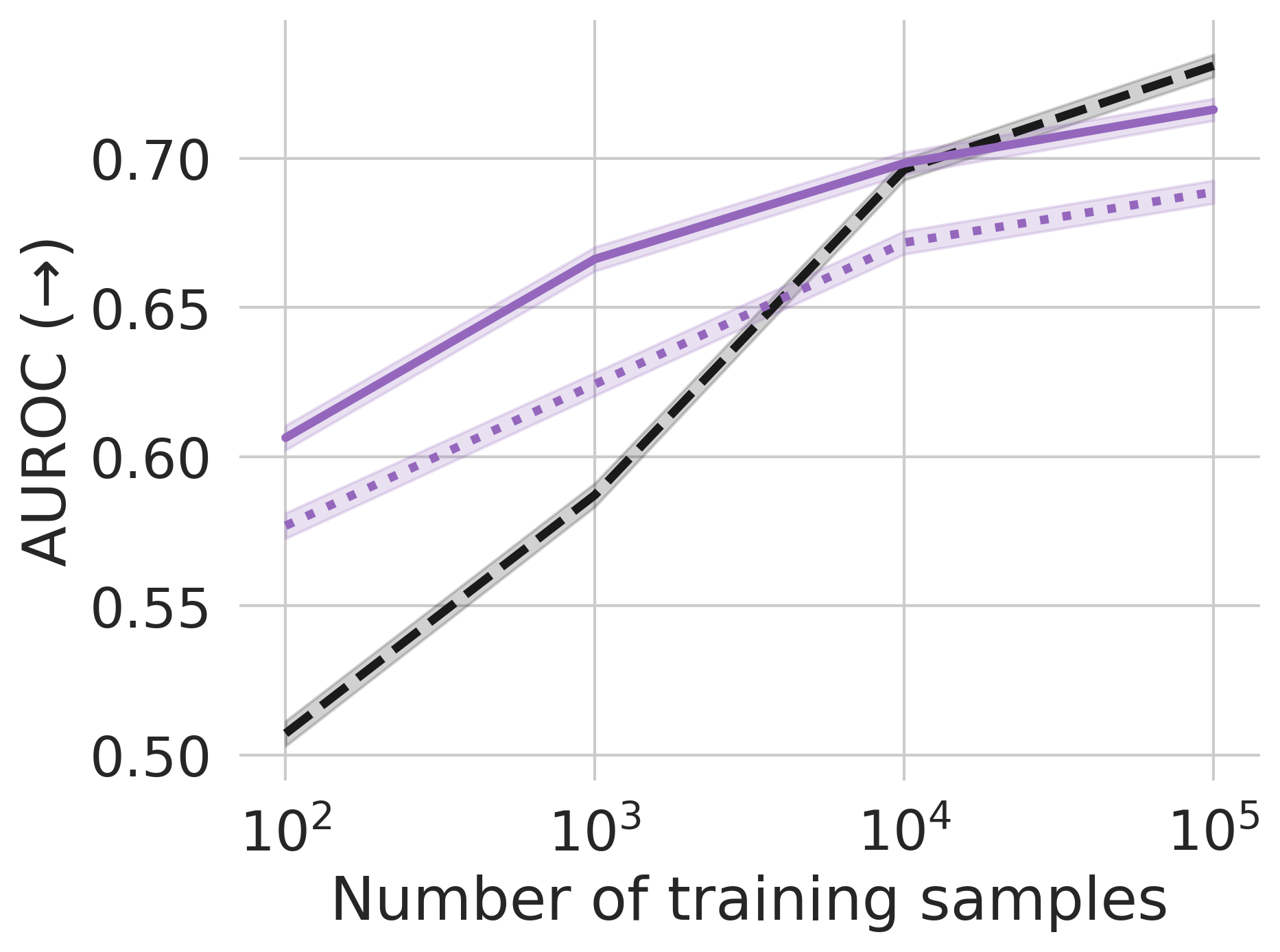}    
    \includegraphics[height=100px]{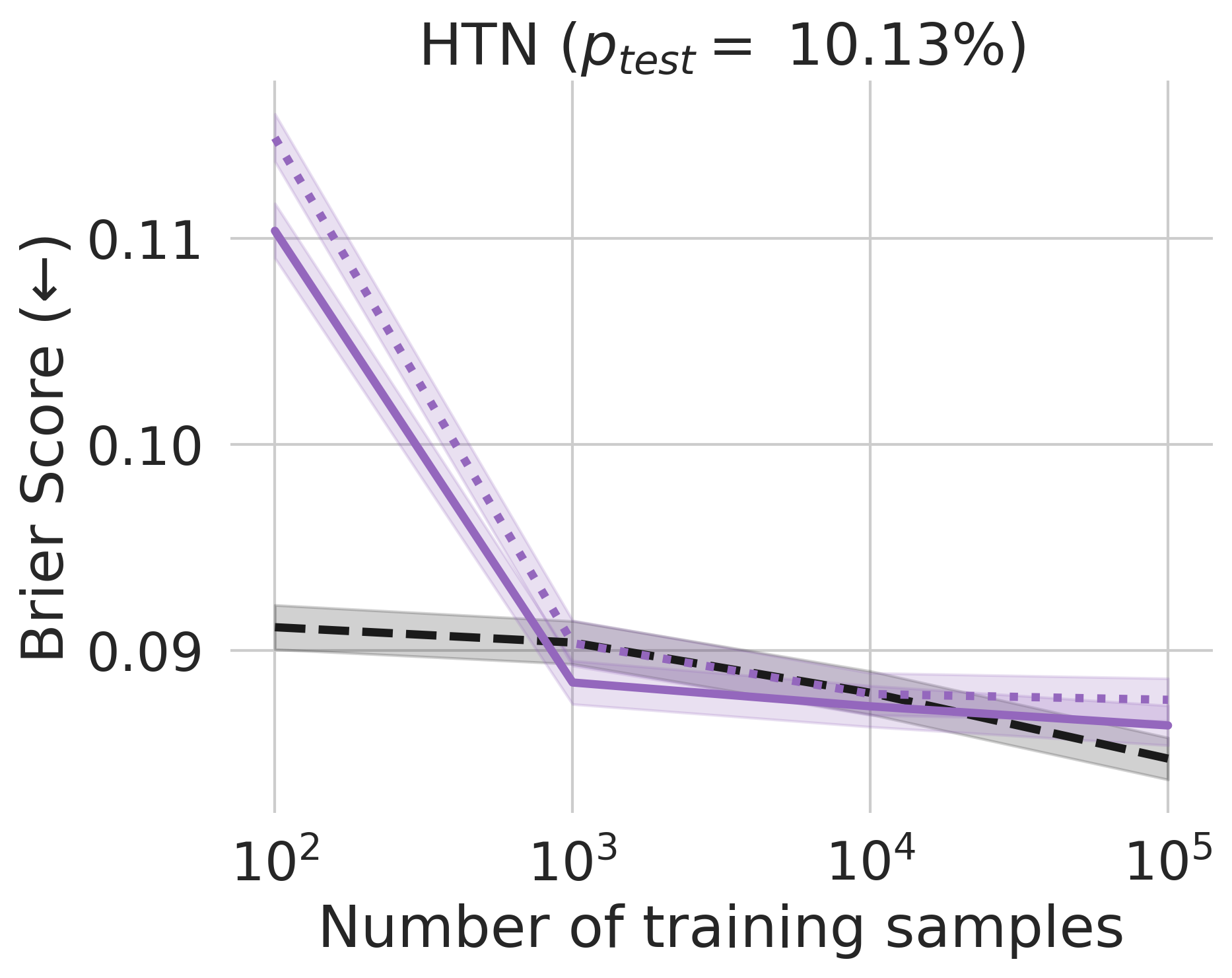}
    \includegraphics[height=100px]{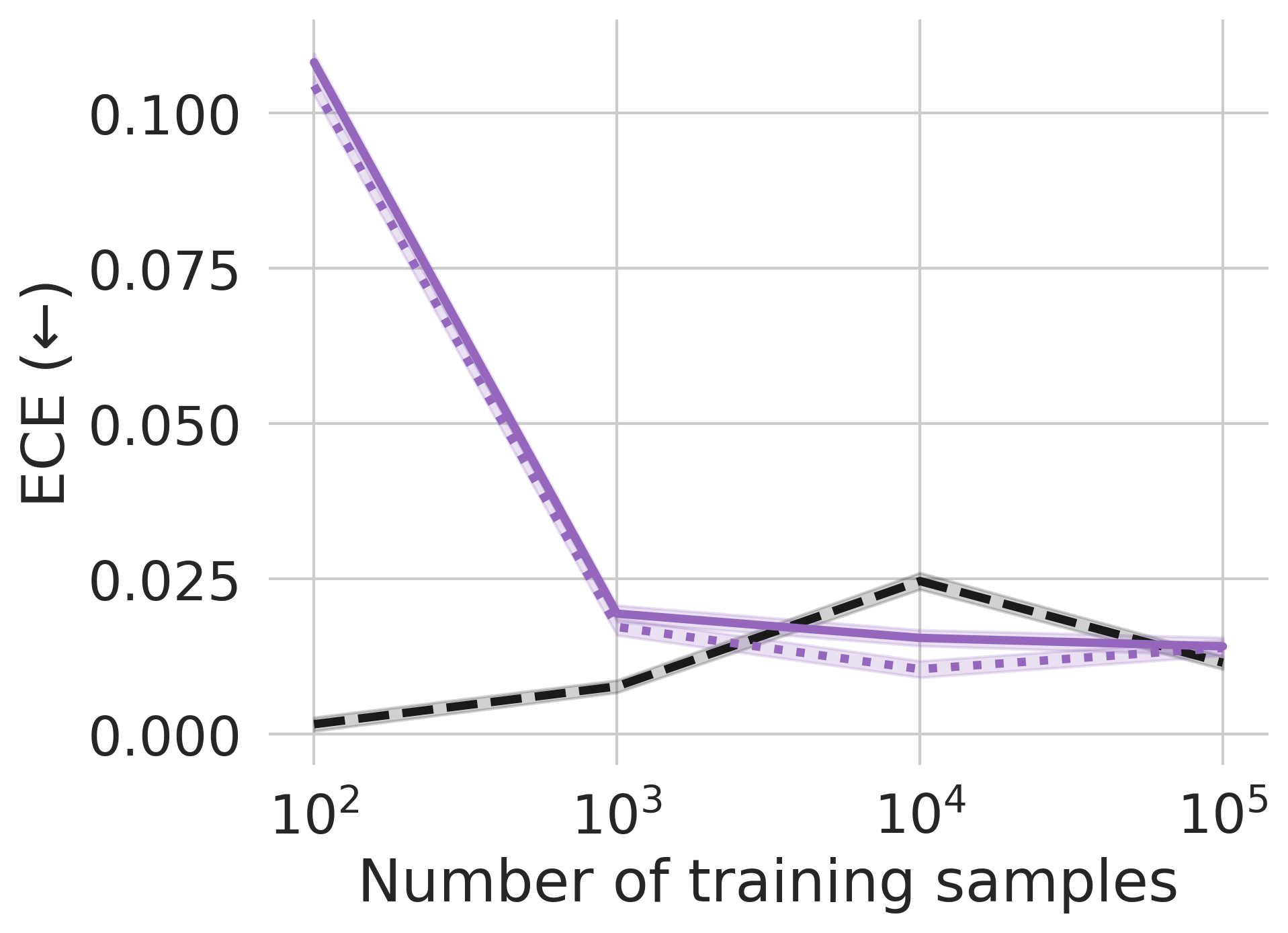}}

    \makebox[\textwidth][l]{
    \includegraphics[height=100px]{figures/results/CUMC/transport_Ischemic_Stroke_roc_auc_score_all.png}
    \includegraphics[height=100px]{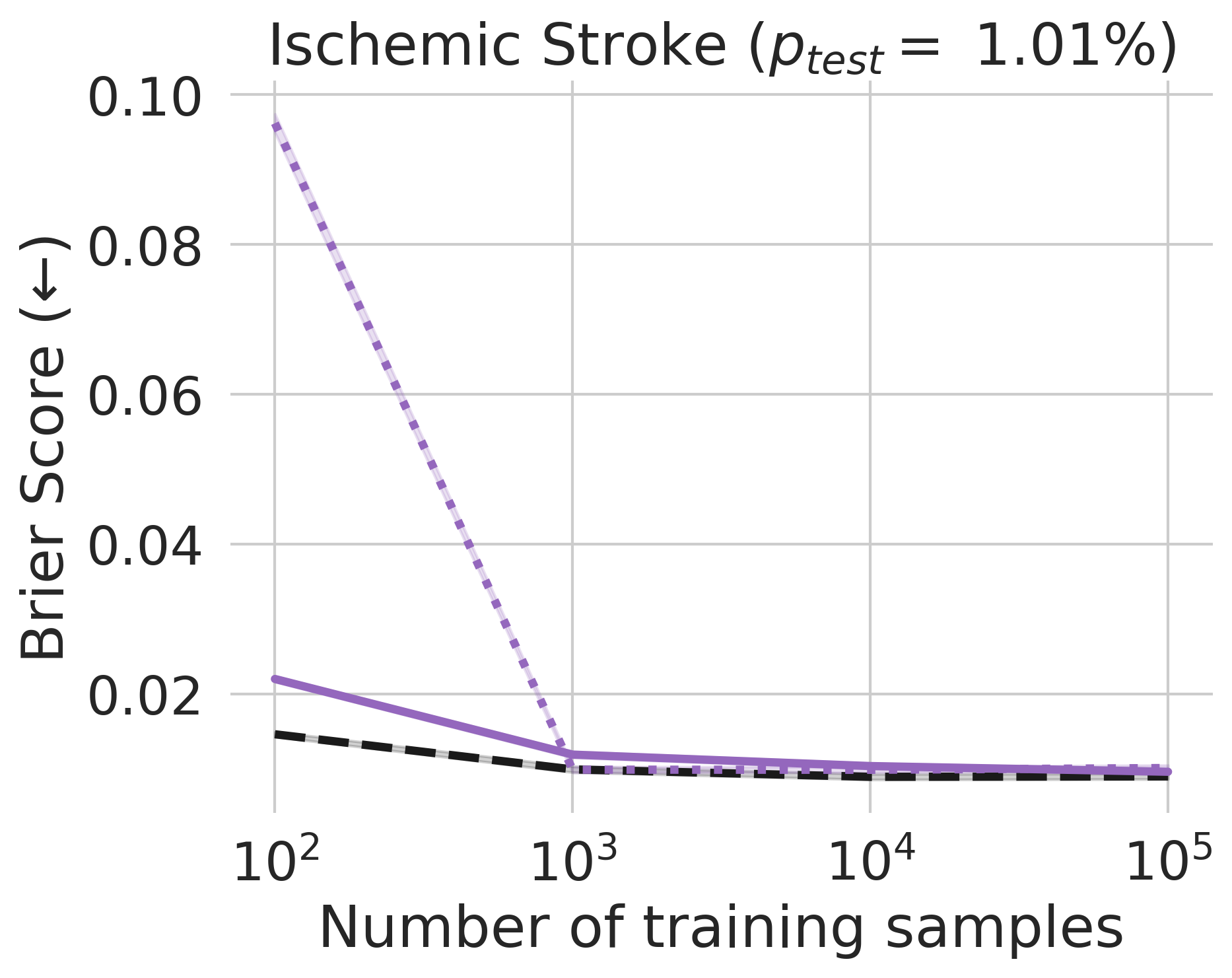}
    \includegraphics[height=100px]{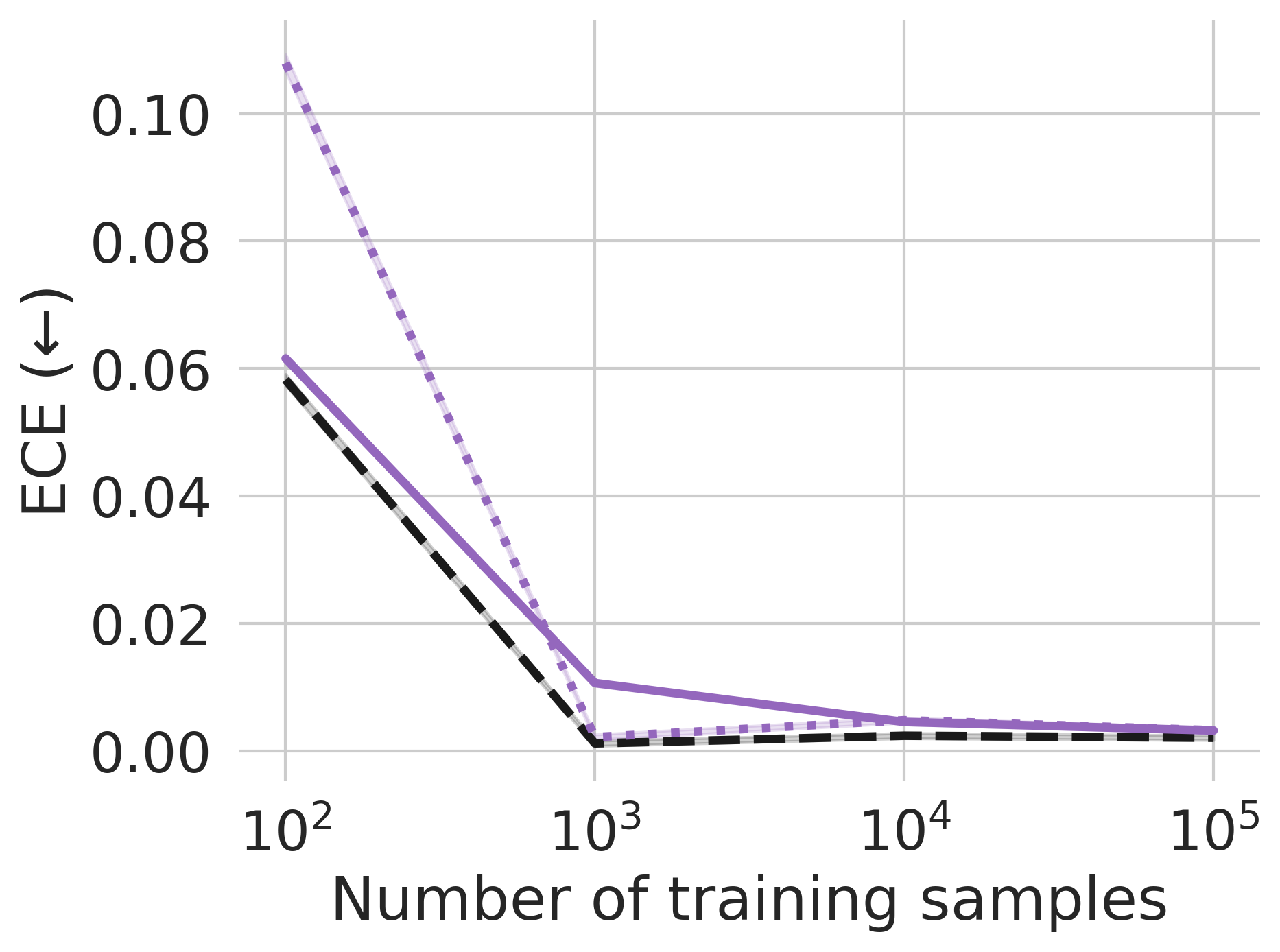}}

    \makebox[\textwidth][l]{
    \includegraphics[height=100px]{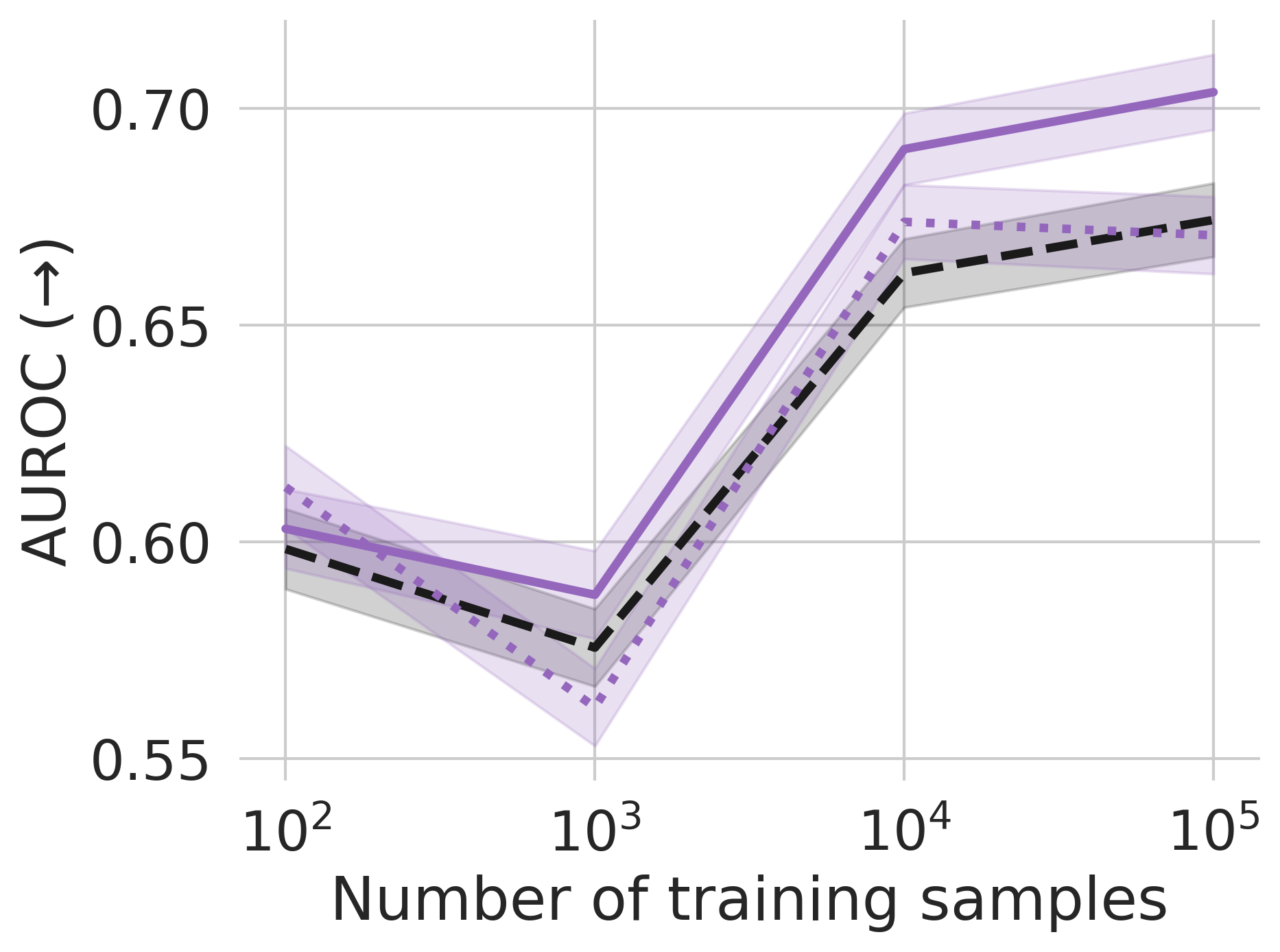}
    \includegraphics[height=100px]{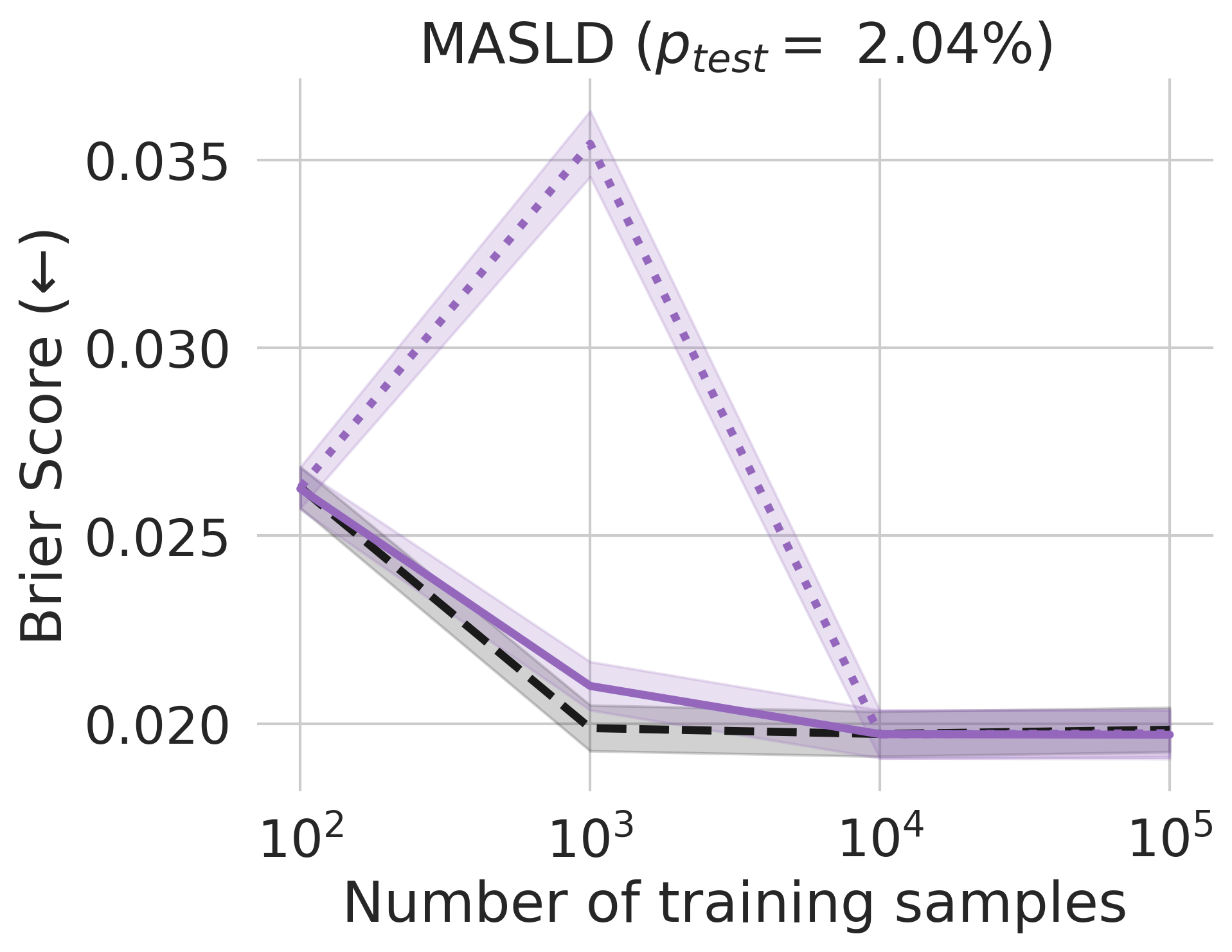}
    \includegraphics[height=100px]{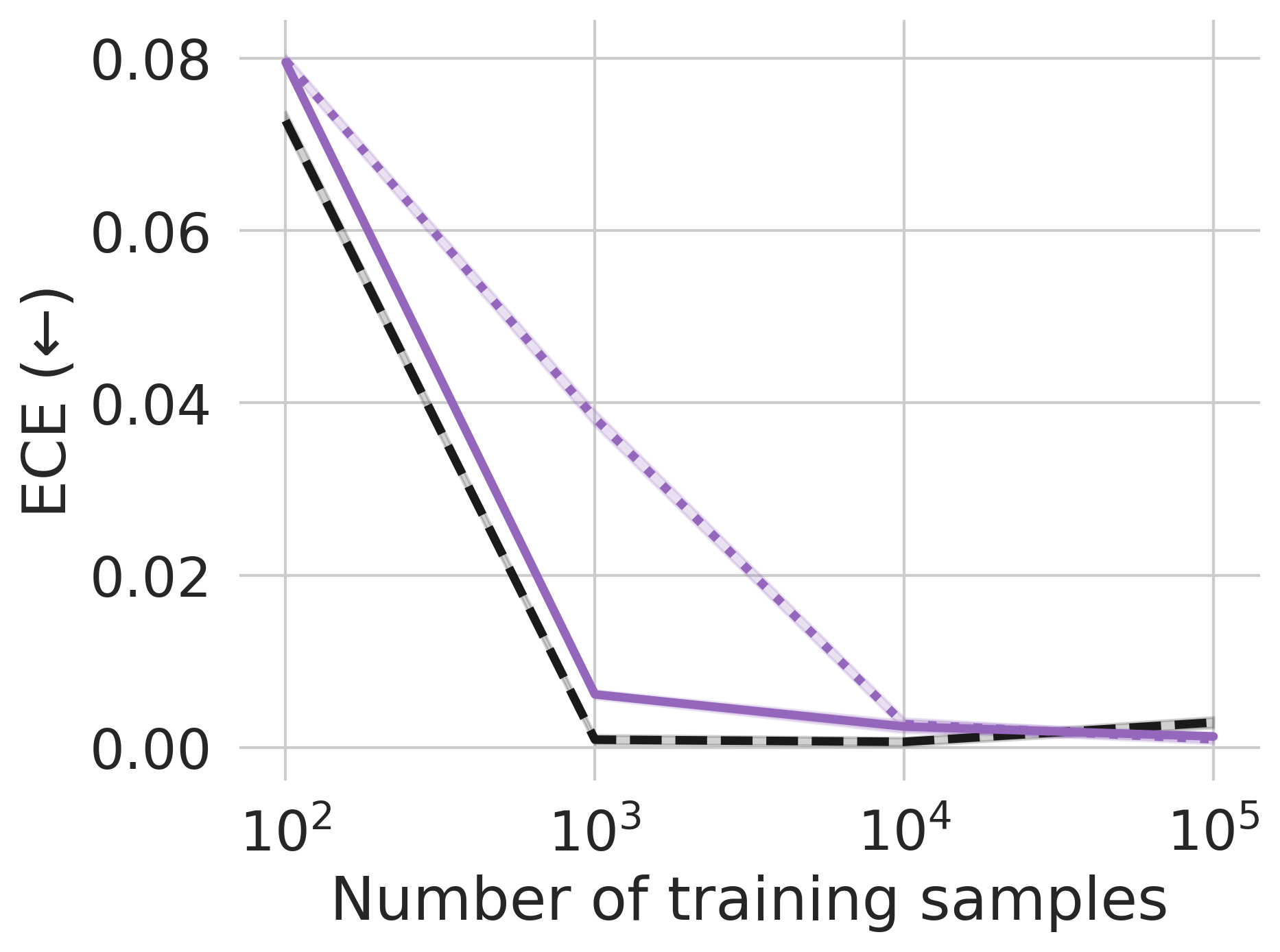}}

    \centering
    \caption{Discriminative performances and calibration results for transported \mamba (Shaded area represents bootstrapped standard deviation).}
    \label{fig:all_phenotypes:cumc:transport}
\end{figure}

\begin{figure}[!ht]
    \makebox[\textwidth][l]{
    \includegraphics[height=100px]{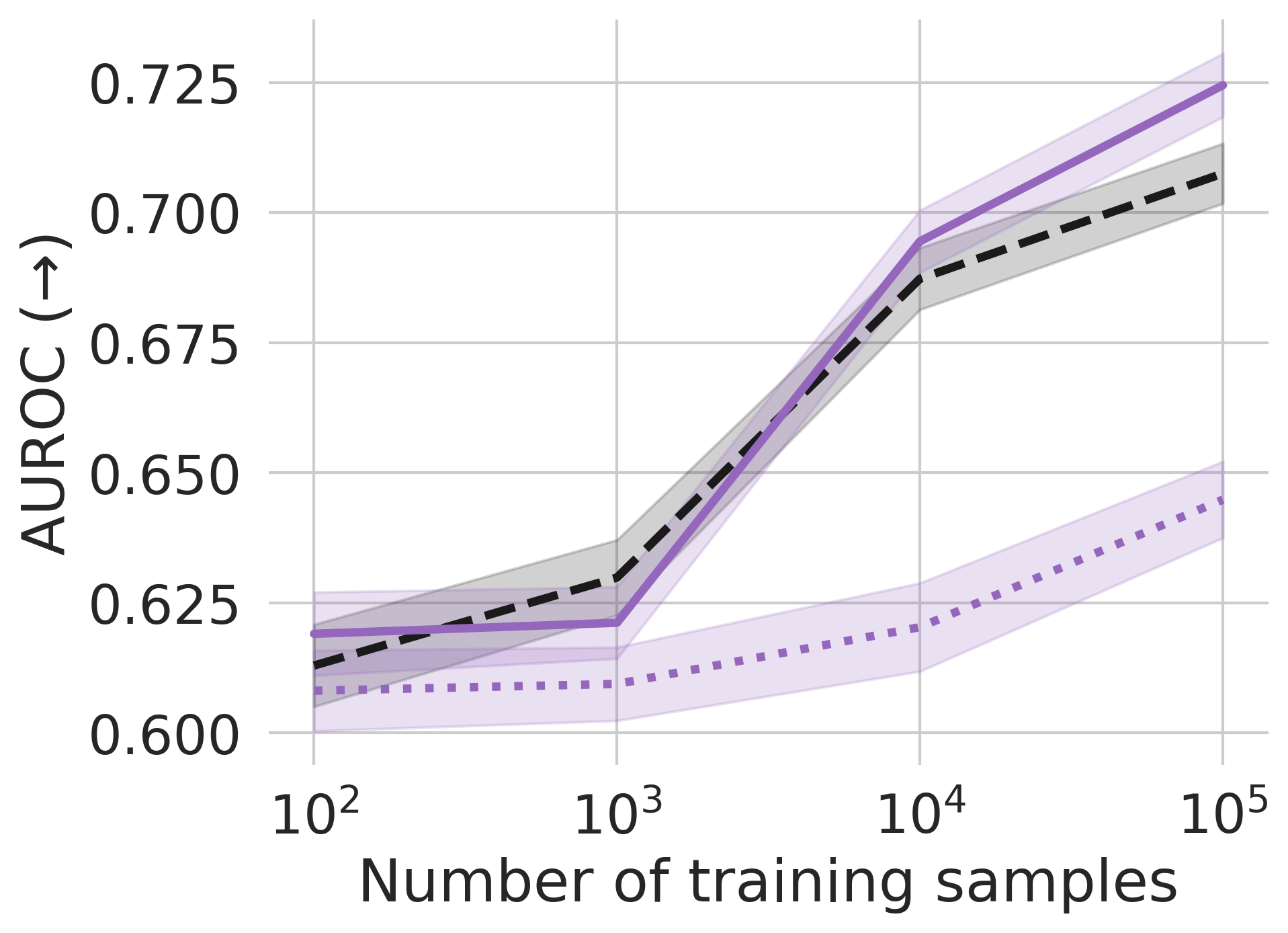}    
    \includegraphics[height=100px]{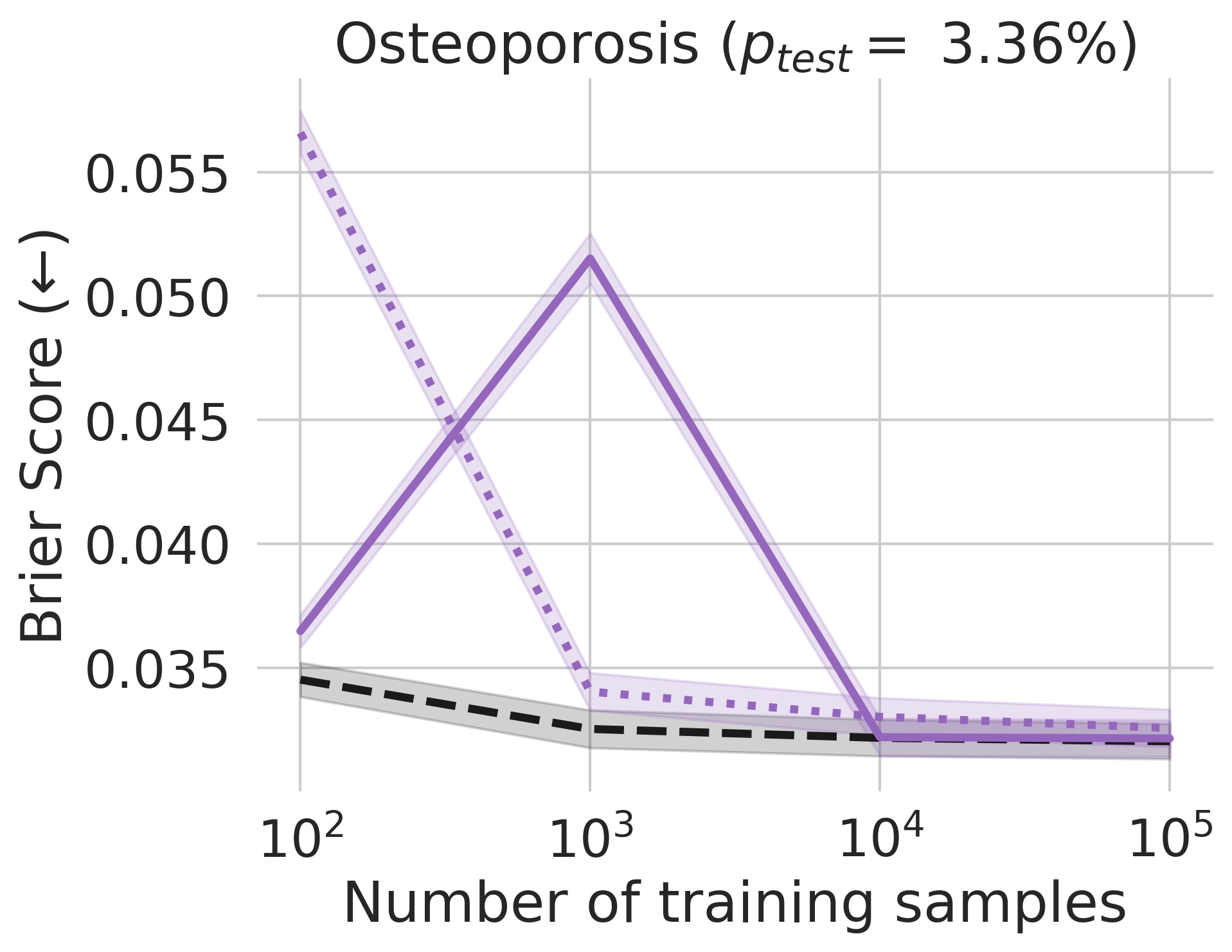}
    \includegraphics[height=100px]{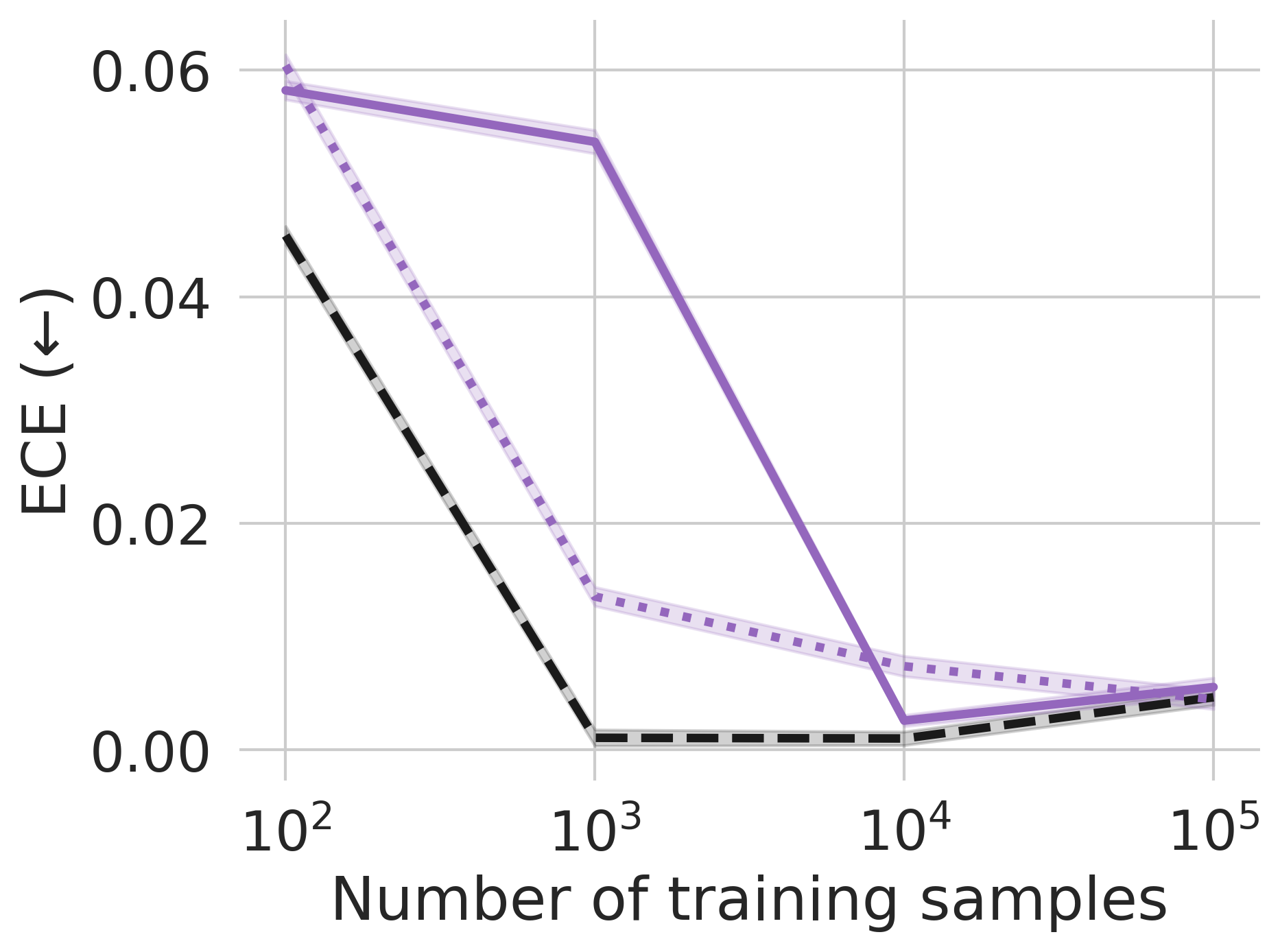}}

    \makebox[\textwidth][l]{
    \includegraphics[height=100px]{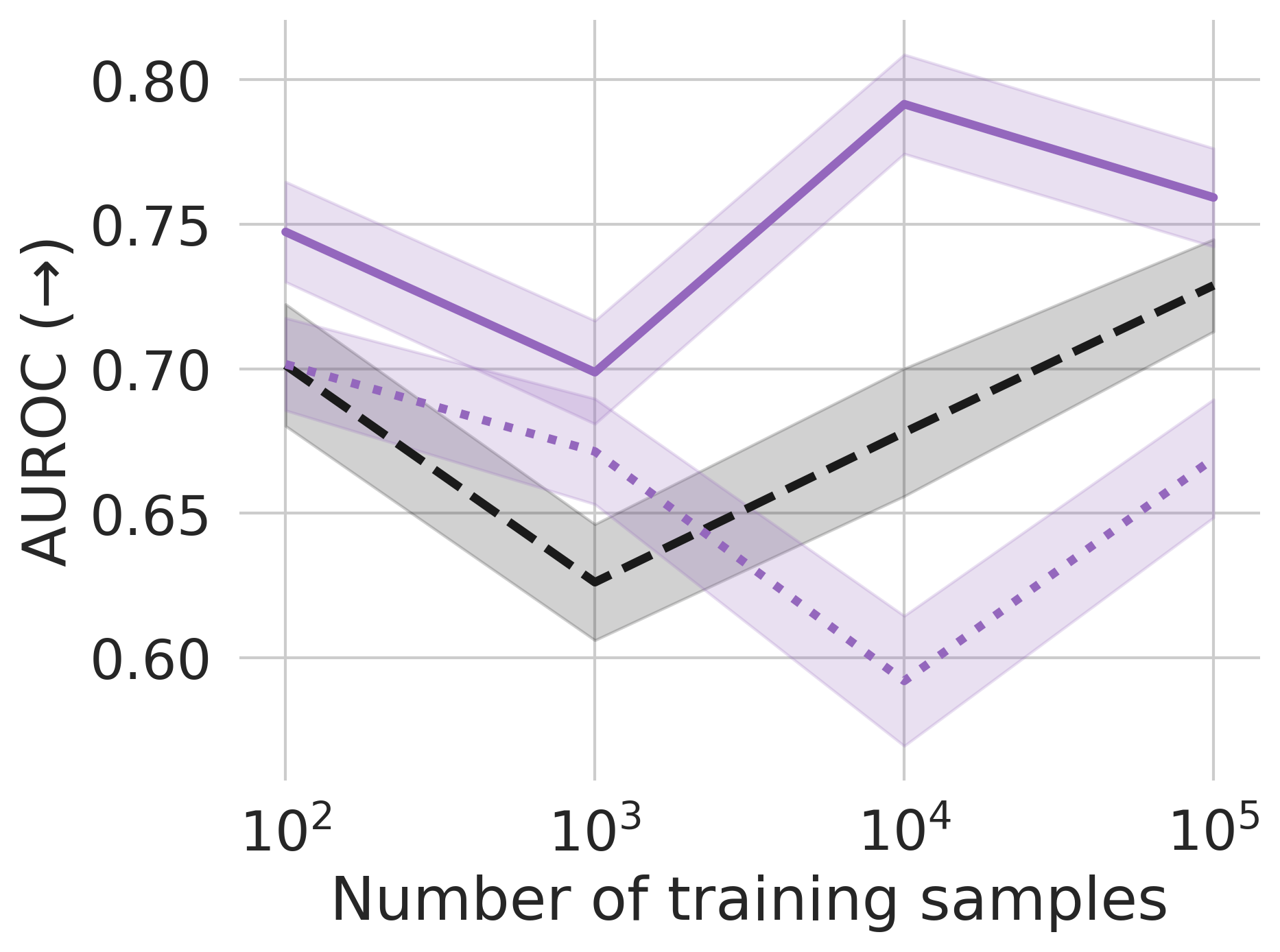}
    \includegraphics[height=100px]{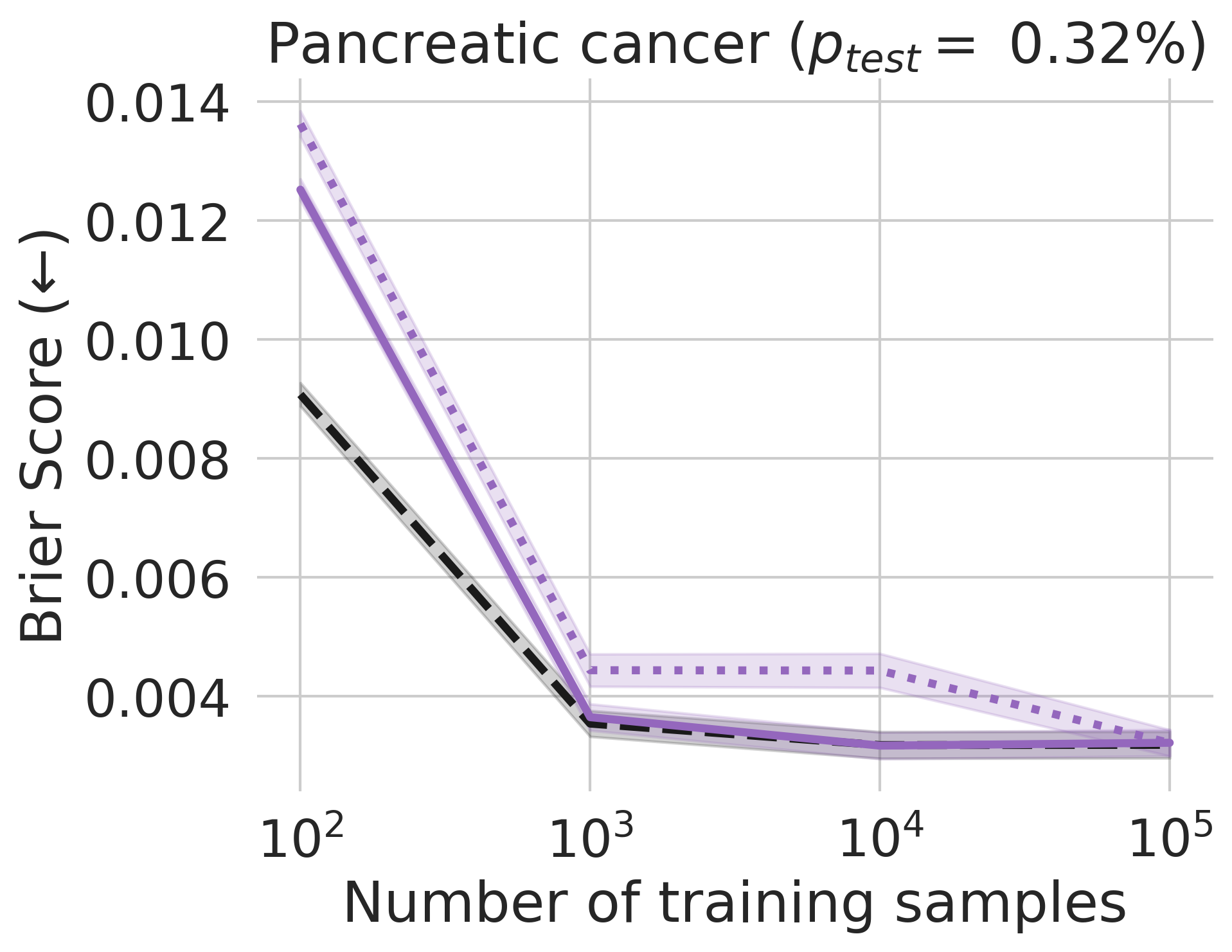}
    \includegraphics[height=100px]{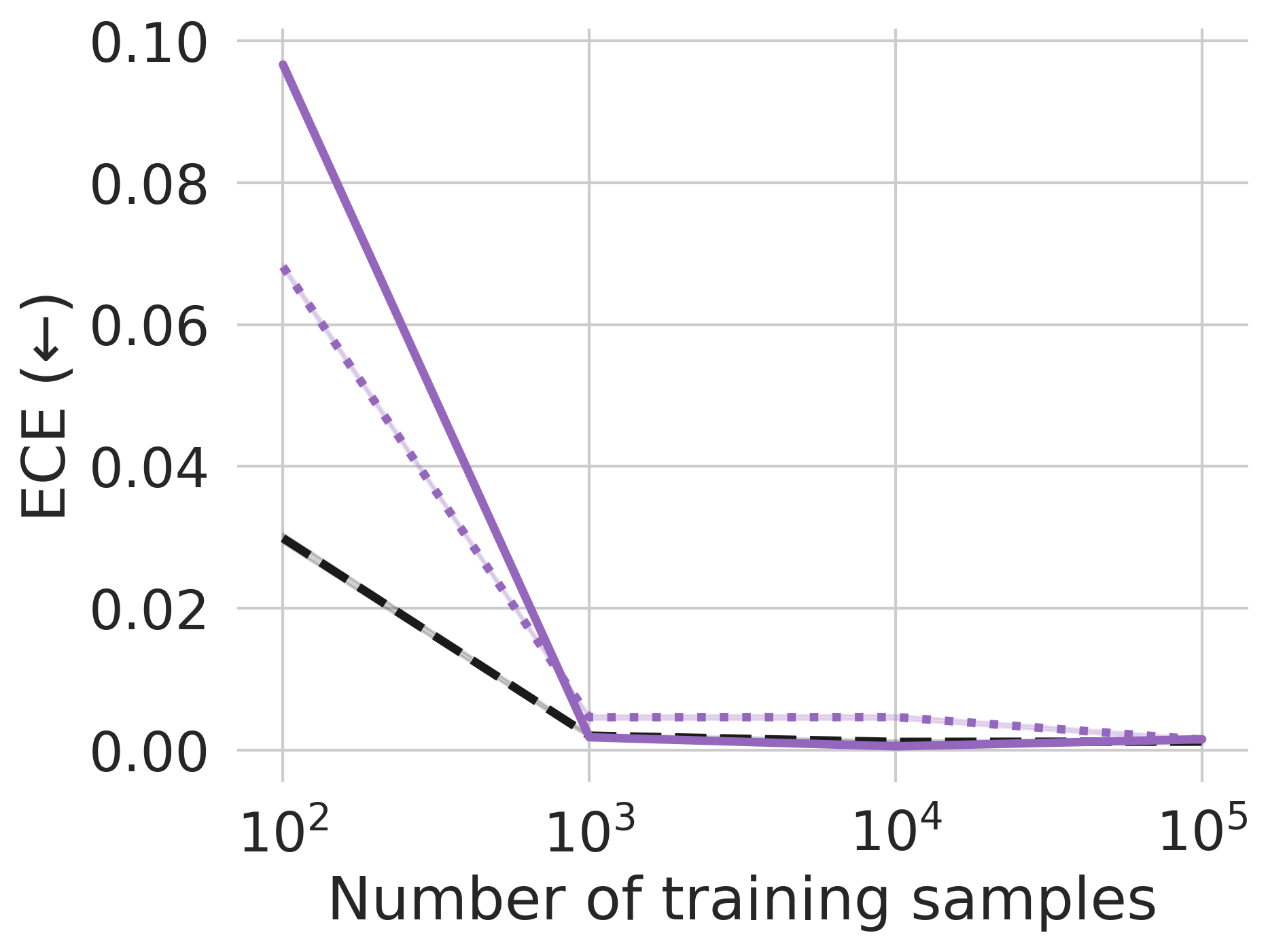}}

    \makebox[\textwidth][l]{
    \includegraphics[height=100px]{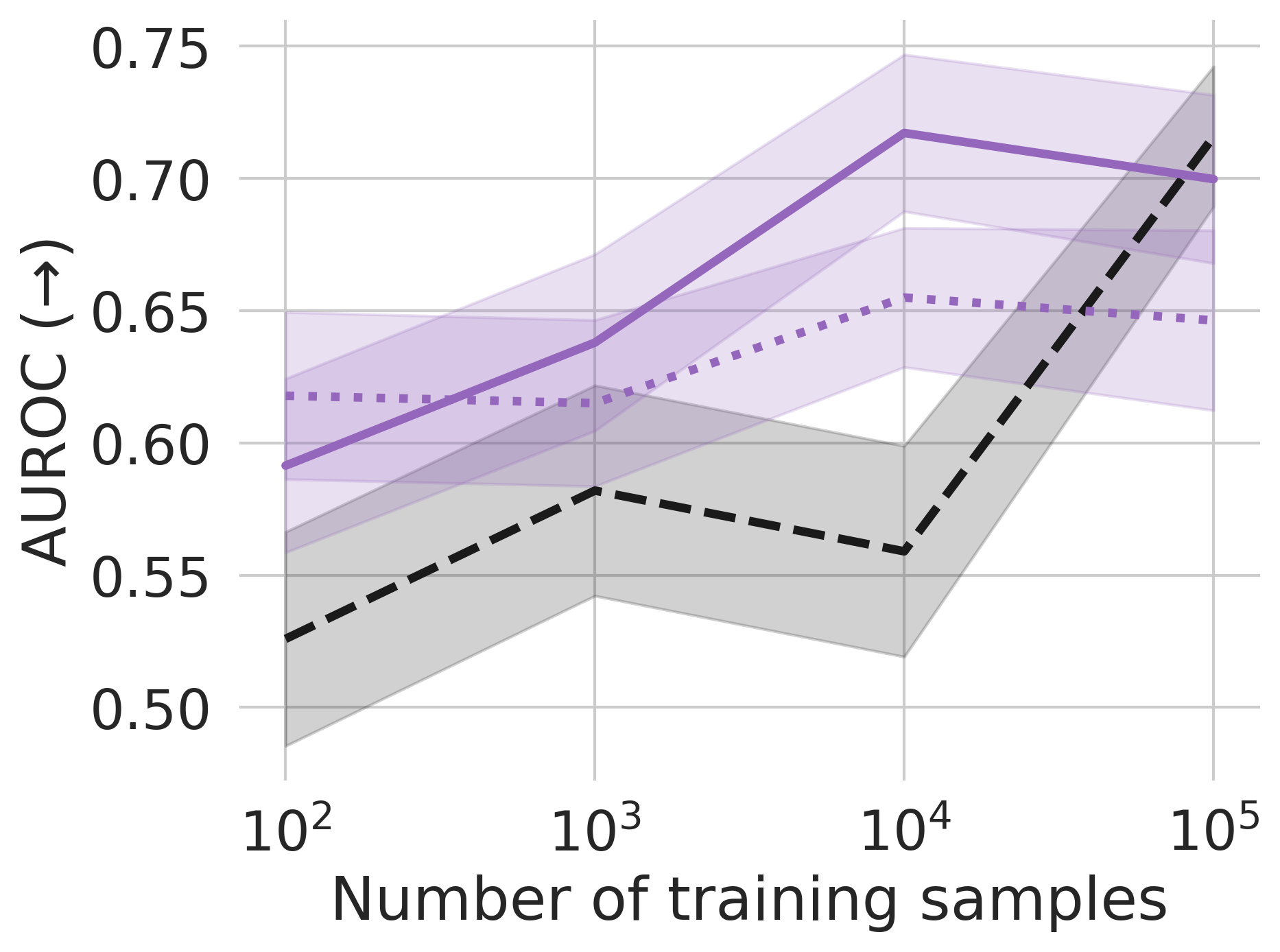}
    \includegraphics[height=100px]{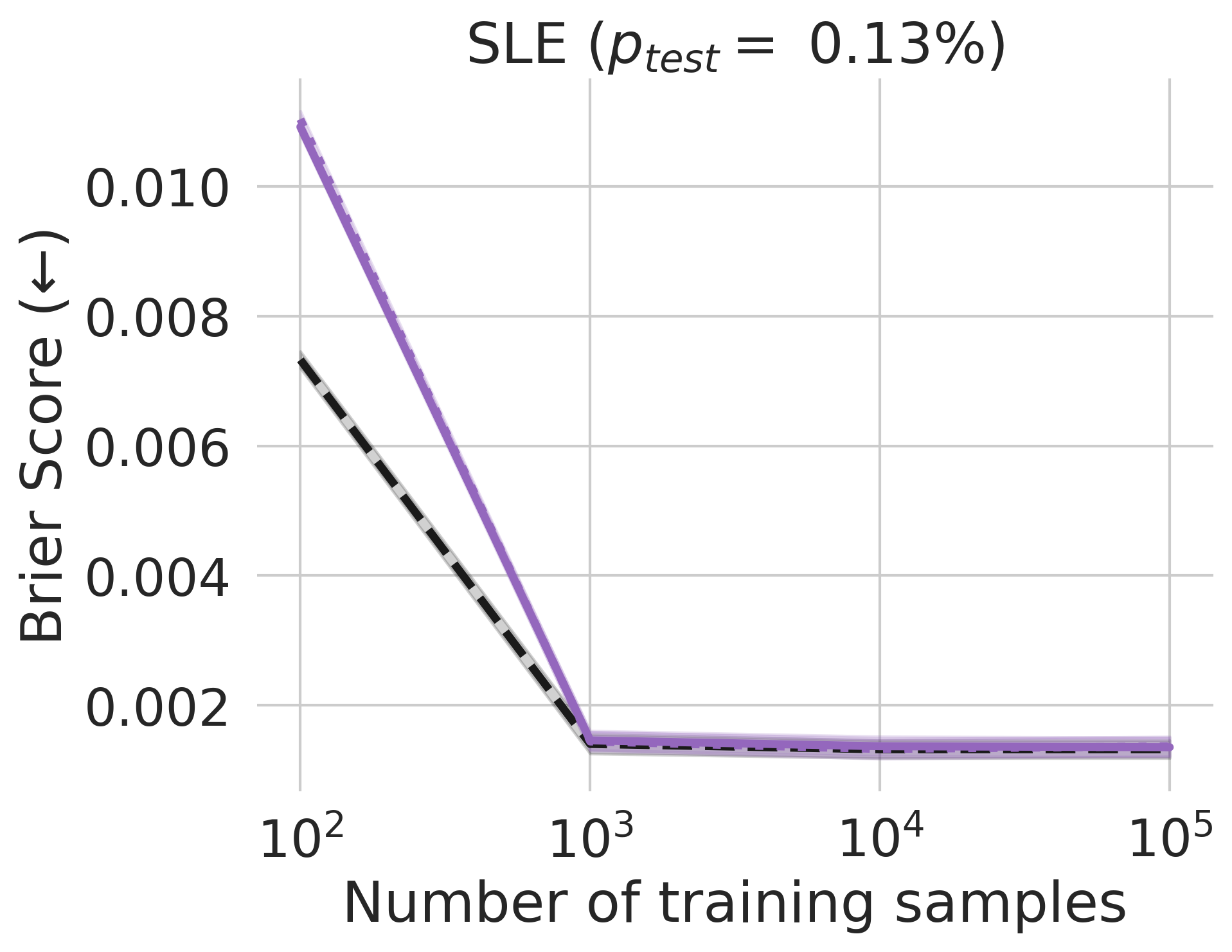}
    \includegraphics[height=100px]{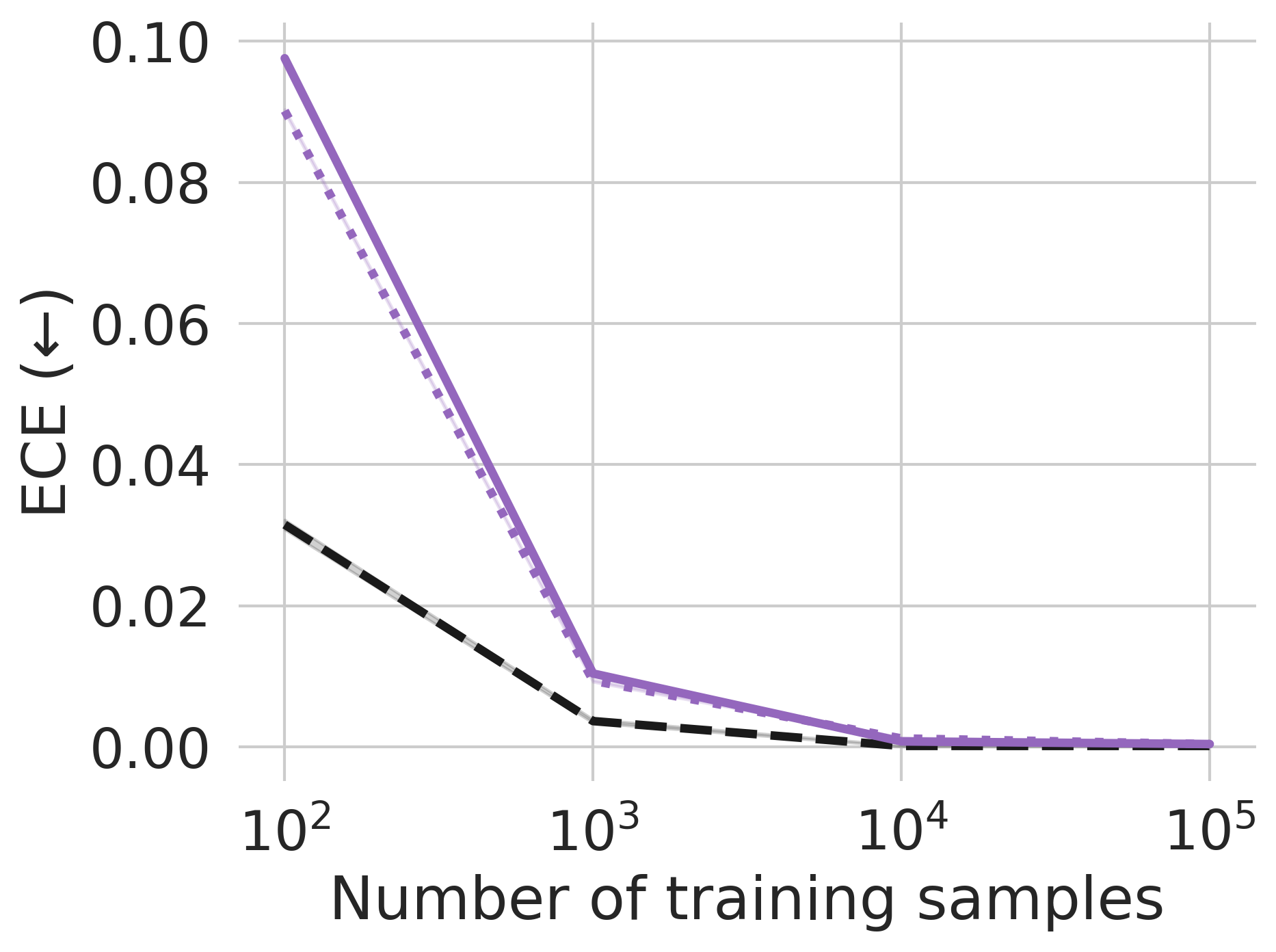}
    \includegraphics[height=100px]{figures/legend_transport.png}}

    \makebox[\textwidth][l]{
    \includegraphics[height=100px]{figures/results/CUMC/transport_Schizophrenia_roc_auc_score_all.png}    
    \includegraphics[height=100px]{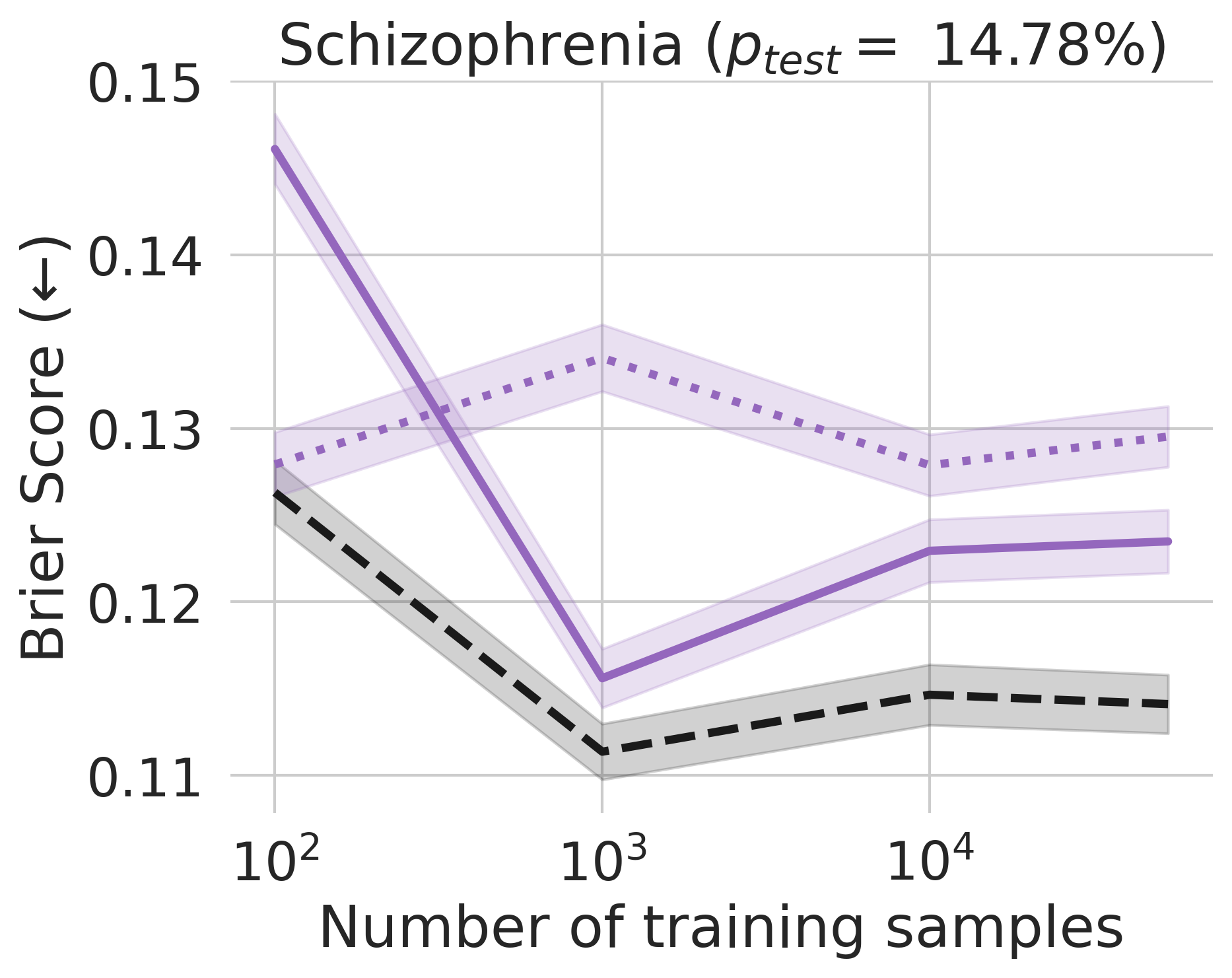}
    \includegraphics[height=100px]{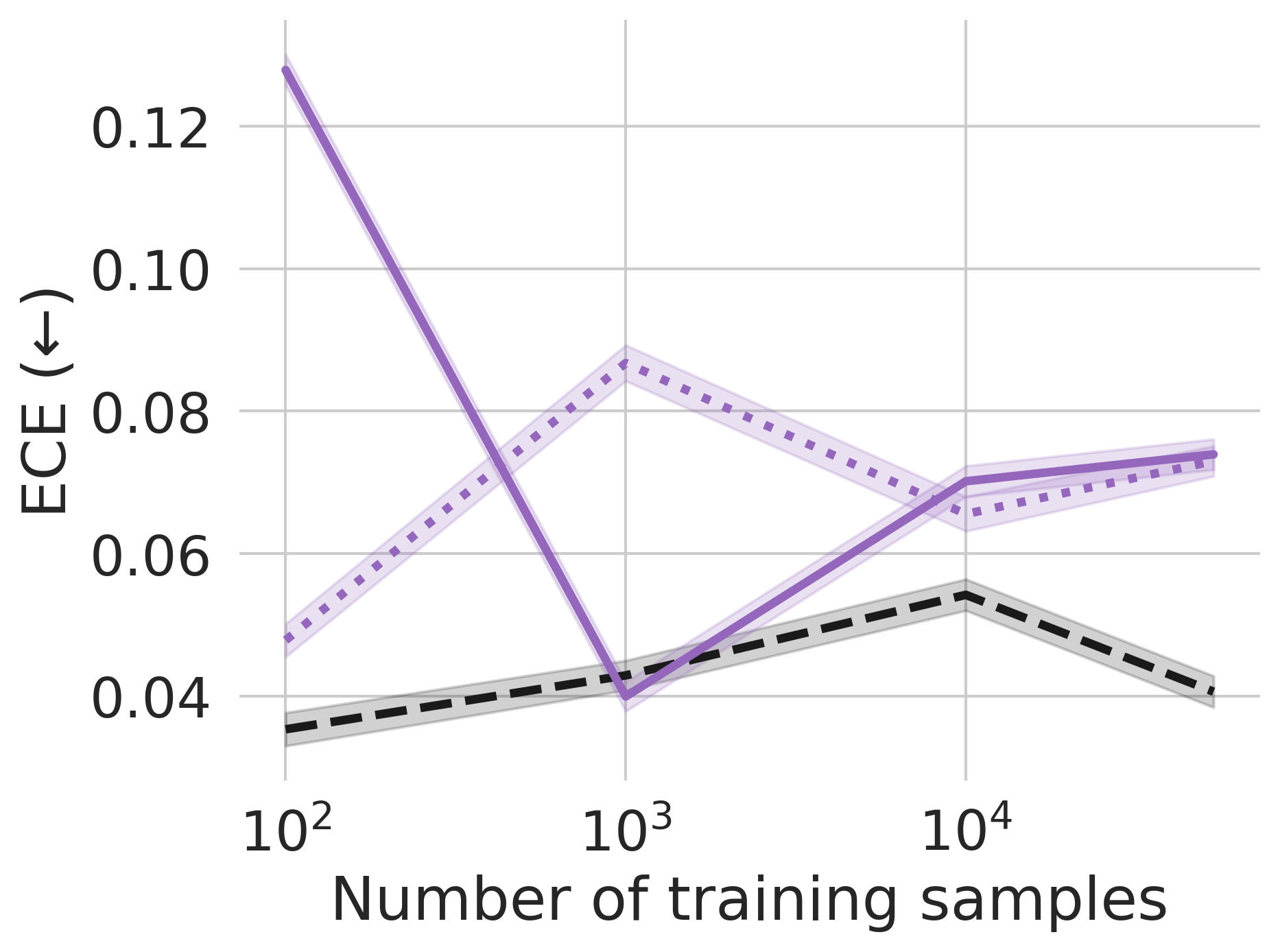}}

    \makebox[\textwidth][l]{
    \includegraphics[height=100px]{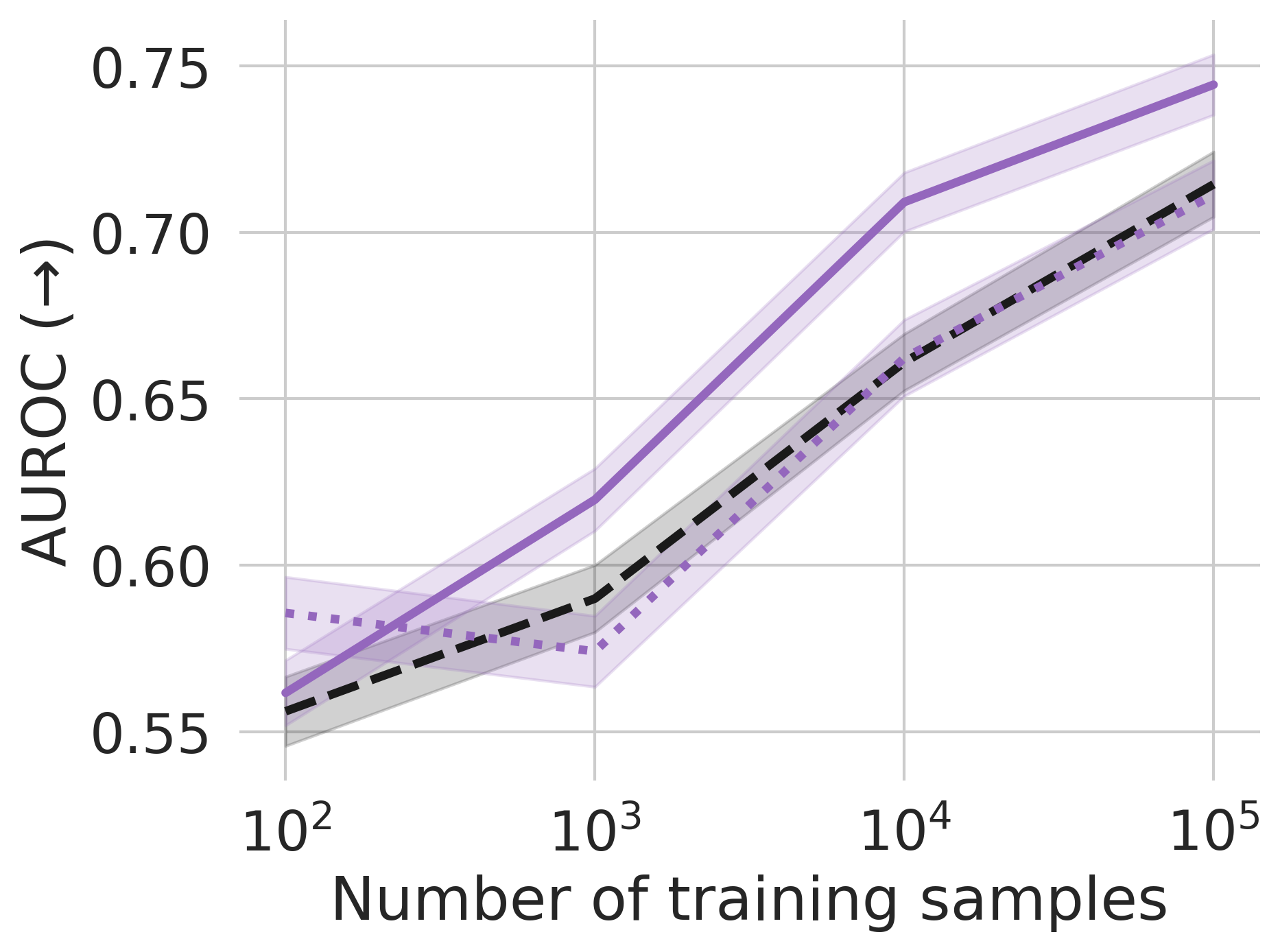}
    \includegraphics[height=100px]{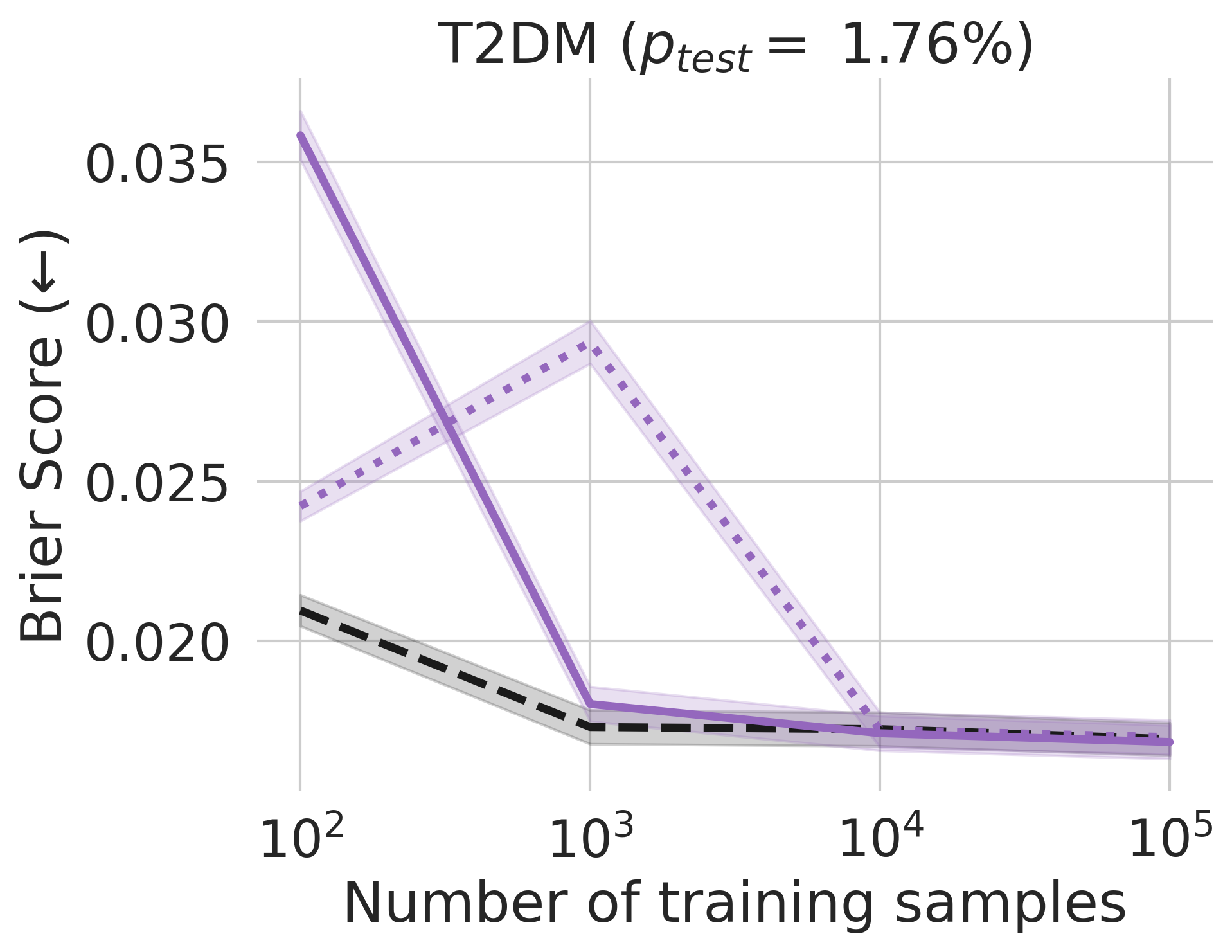}
    \includegraphics[height=100px]{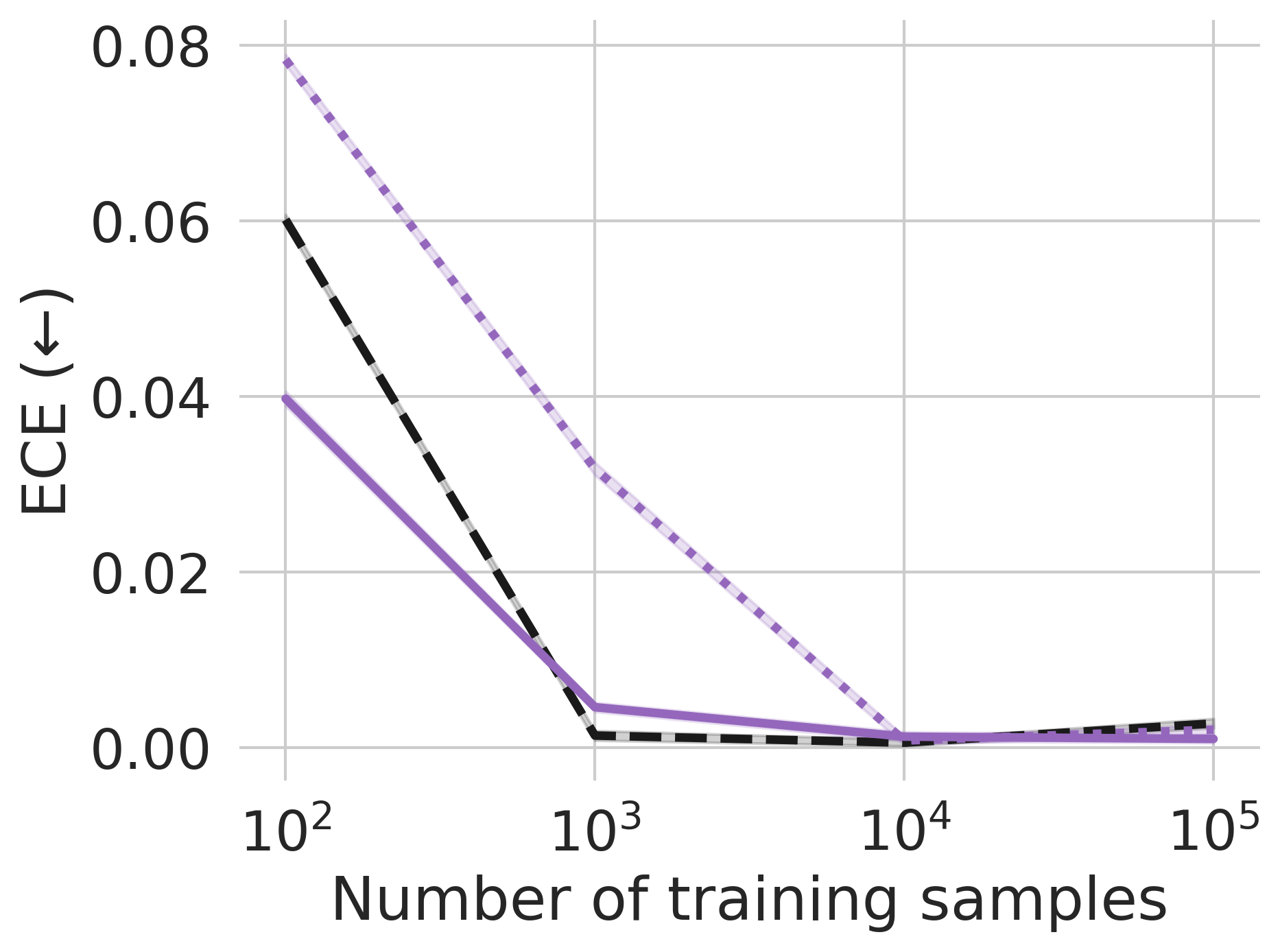}}

    \centering
    \caption{Discriminative performances and calibration results for transported \mamba (Shaded area represents bootstrapped standard deviation).}
    \label{fig:all_phenotypes:2:transport}
\end{figure}

The main text presents transport performance on three representative tasks.  Figure~\ref{fig:all_outcomes:cumc:transport} details the performance across all tasks for patient outcome tasks; Figures~\ref{fig:all_phenotypes:cumc:transport} and~\ref{fig:all_phenotypes:2:transport} show transport performance for chronic and acute phenotypes. 

\clearpage
\subsection{Performances using the largest training set}
Tables~\ref{tab:outcome-overall-performance} and \ref{tab:phenotype-overall-performance} present the discriminative tasks over all considered tasks when using all extracted training points. 
Similarly Tables~\ref{tab:outcome-overall-calibration} and \ref{tab:phenotype-overall-calibration} present the Brier score, and Tables~\ref{tab:outcome-overall-ece} and \ref{tab:phenotype-overall-ece} the ECE. 
Tables~\ref{tab:outcome-sex} and \ref{tab:phenotype-sex}, Tables~\ref{tab:outcome-race} and \ref{tab:phenotype-race} and Tables~\ref{tab:outcome-utilization} and \ref{tab:phenotype-utilization} respectively present the maximal discriminative performances for the sex, ethnicity, and utilization splits.

\begin{table}[ht]
    \footnotesize
    \centering
    \setlength{\tabcolsep}{3pt}
    \setlength\extrarowheight{3pt}
    
    \caption{Outcome discriminative performance results. \textit{Best performances are marked in \textbf{bold}.}}
    \label{tab:outcome-overall-performance}
    \begin{tabular}{cccccccc}
        \toprule
         & \textbf{Best Baseline} & \makecell{\textbf{\textsc{CEHR}}\\\textbf{\textsc{BERT}}} & \makecell{\textbf{\textsc{CEHR}}\\\textbf{\textsc{GPT}}} & \makecell{\textbf{\textsc{Core}}\\\textbf{\textsc{Behrt}}} & 
        \textbf{\llama} & \textbf{\mamba} & \textbf{\motor} \\ \cmidrule{2-8}
            LOS & \makecell { 0.829 \\ (0.002) } & \makecell { 0.787 \\ (0.002) } & \makecell { \textbf{0.870} \\ (0.002) } & \makecell { 0.740 \\ (0.002) } & \makecell { 0.799 \\ (0.002) } & \makecell { 0.784 \\ (0.002) } & \makecell { 0.853 \\ (0.002) } \\
            Mortality & \makecell { 0.914 \\ (0.003) } & \makecell { 0.879 \\ (0.004) } & \makecell { 0.939 \\ (0.003) } & \makecell { 0.817 \\ (0.005) } & \makecell { 0.892 \\ (0.003) } & \makecell { 0.882 \\ (0.004) } & \makecell { \textbf{0.941} \\ (0.003) } \\
            Readmission & \makecell { 0.765 \\ (0.003) } & \makecell { 0.731 \\ (0.003) } & \makecell { 0.776 \\ (0.003) } & \makecell { 0.696 \\ (0.004) } & \makecell { 0.758 \\ (0.003) } & \makecell { 0.757 \\ (0.003) } & \makecell { \textbf{0.780} \\ (0.003) } \\

            \bottomrule
    \end{tabular}
\end{table}

\begin{table}[ht]
    \footnotesize
    \centering
    \setlength{\tabcolsep}{3pt}
    \setlength\extrarowheight{3pt}
    \caption{Phenotype discriminative performance results. \textit{Best performances are marked in \textbf{bold}.}}
    \label{tab:phenotype-overall-performance}
    \begin{tabular}{cc cccccc}
        \toprule
        \textbf{Phenotype} & \textbf{Best Baseline} & \makecell{\textbf{\textsc{CEHR}}\\\textbf{\textsc{BERT}}} & \makecell{\textbf{\textsc{CEHR}}\\\textbf{\textsc{GPT}}} & \makecell{\textbf{\textsc{Core}}\\\textbf{\textsc{Behrt}}} & 
        \textbf{\llama} & \textbf{\mamba} & \textbf{\motor} \\ \cmidrule{2-8}
            AMI & \makecell { 0.828 \\ (0.009) } & \makecell { 0.832 \\ (0.008) } & \makecell { 0.809 \\ (0.008) } & \makecell { 0.776 \\ (0.010) } & \makecell { 0.820 \\ (0.007) } & \makecell { 0.816 \\ (0.008) } & \makecell { \textbf{0.845} \\ (0.007) } \\
            CLL & \makecell { \textbf{0.720} \\ (0.026) } & \makecell { 0.591 \\ (0.027) } & \makecell { 0.599 \\ (0.035) } & \makecell { 0.468 \\ (0.030) } & \makecell { 0.638 \\ (0.038) } & \makecell { 0.650 \\ (0.028) } & \makecell { 0.619 \\ (0.030) } \\
            Celiac & \makecell { 0.581 \\ (0.042) } & \makecell { 0.625 \\ (0.037) } & \makecell { 0.700 \\ (0.034) } & \makecell { 0.607 \\ (0.031) } & \makecell { 0.598 \\ (0.035) } & \makecell { 0.637 \\ (0.041) } & \makecell { \textbf{0.789} \\ (0.028) } \\
            HTN & \makecell { 0.731 \\ (0.004) } & \makecell { 0.701 \\ (0.004) } & \makecell { 0.714 \\ (0.003) } & \makecell { 0.676 \\ (0.004) } & \makecell { 0.714 \\ (0.004) } & \makecell { 0.716 \\ (0.004) } & \makecell { \textbf{0.774} \\ (0.003) } \\
            Ischemic Stroke & \makecell { \textbf{0.882} \\ (0.008) } & \makecell { 0.849 \\ (0.007) } & \makecell { 0.834 \\ (0.009) } & \makecell { 0.784 \\ (0.009) } & \makecell { 0.846 \\ (0.007) } & \makecell { 0.865 \\ (0.007) } & \makecell { 0.876 \\ (0.006) } \\
            MASLD & \makecell { 0.674 \\ (0.008) } & \makecell { 0.657 \\ (0.009) } & \makecell { 0.710 \\ (0.009) } & \makecell { 0.636 \\ (0.009) } & \makecell { \textbf{0.729} \\ (0.009) } & \makecell { 0.704 \\ (0.009) } & \makecell { 0.713 \\ (0.008) } \\
            Osteoporosis & \makecell { 0.707 \\ (0.006) } & \makecell { 0.708 \\ (0.006) } & \makecell { 0.714 \\ (0.007) } & \makecell { 0.664 \\ (0.006) } & \makecell { 0.708 \\ (0.006) } & \makecell { 0.724 \\ (0.006) } & \makecell { \textbf{0.749} \\ (0.005) } \\
            Pancreatic Cancer & \makecell { 0.729 \\ (0.016) } & \makecell { \textbf{0.782} \\ (0.019) } & \makecell { 0.707 \\ (0.020) } & \makecell { 0.704 \\ (0.017) } & \makecell { 0.765 \\ (0.018) } & \makecell { 0.759 \\ (0.017) } & \makecell { 0.762 \\ (0.017) } \\
            SLE & \makecell { 0.716 \\ (0.027) } & \makecell { 0.695 \\ (0.032) } & \makecell { \textbf{0.750} \\ (0.033) } & \makecell { 0.590 \\ (0.036) } & \makecell { 0.738 \\ (0.034) } & \makecell { 0.700 \\ (0.032) } & \makecell { 0.727 \\ (0.027) } \\
            Schizophrenia & \makecell { \textbf{0.807} \\ (0.004) } & \makecell { 0.704 \\ (0.005) } & \makecell { 0.716 \\ (0.005) } & \makecell { 0.697 \\ (0.005) } & \makecell { 0.700 \\ (0.005) } & \makecell { 0.708 \\ (0.005) } & \makecell { 0.759 \\ (0.005) } \\
            T2DM & \makecell { 0.714 \\ (0.010) } & \makecell { 0.733 \\ (0.009) } & \makecell { 0.735 \\ (0.009) } & \makecell { 0.682 \\ (0.009) } & \makecell { 0.747 \\ (0.009) } & \makecell { 0.744 \\ (0.009) } & \makecell { \textbf{0.809} \\ (0.007) } \\

        \bottomrule
    \end{tabular}
    \vspace{2pt}
    \newline
    AMI: Acute myocardial infarction. CLL: Chronic lymphocytic leukemia. HTN: Hypertension. MASLD: Metabolic dysfunction-associated steatotic liver disease. SLE: Systemic lupus erythematosus. T2DM: Type 2 diabetes mellitus.
\end{table}

\newpage
\begin{table}[ht]
    \footnotesize
    \centering
    \setlength{\tabcolsep}{3pt}
    \setlength\extrarowheight{3pt}
    \caption{Outcome Brier score results. \textit{Best performances are marked in \textbf{bold}.}}
    \label{tab:outcome-overall-calibration}
    \begin{tabular}{cccccccc}
        \toprule
         & \textbf{Best Baseline} & \makecell{\textbf{\textsc{CEHR}}\\\textbf{\textsc{BERT}}} & \makecell{\textbf{\textsc{CEHR}}\\\textbf{\textsc{GPT}}} & \makecell{\textbf{\textsc{Core}}\\\textbf{\textsc{Behrt}}} & 
        \textbf{\llama} & \textbf{\mamba} & \textbf{\motor} \\ \cmidrule{2-8}
            LOS & \makecell { 0.158 \\ (0.001) } & \makecell { 0.175 \\ (0.001) } & \makecell { \textbf{0.139} \\ (0.001) } & \makecell { 0.191 \\ (0.001) } & \makecell { 0.172 \\ (0.001) } & \makecell { 0.177 \\ (0.001) } & \makecell { 0.147 \\ (0.001) } \\
            Mortality & \makecell { 0.018 \\ (0.000) } & \makecell { 0.019 \\ (0.000) } & \makecell { 0.017 \\ (0.000) } & \makecell { 0.020 \\ (0.000) } & \makecell { 0.019 \\ (0.000) } & \makecell { 0.019 \\ (0.000) } & \makecell { \textbf{0.017} \\ (0.000) } \\
            Readmission & \makecell { 0.115 \\ (0.001) } & \makecell { 0.120 \\ (0.001) } & \makecell { 0.113 \\ (0.001) } & \makecell { 0.124 \\ (0.001) } & \makecell { 0.116 \\ (0.001) } & \makecell { 0.117 \\ (0.001) } & \makecell { \textbf{0.113} \\ (0.001) } \\
        \bottomrule
    \end{tabular}
\end{table}

\begin{table}[ht]
    \footnotesize
    \centering
    \setlength{\tabcolsep}{3pt}
    \setlength\extrarowheight{3pt}
    \caption{Phenotype Brier score results. \textit{Best performances are marked in \textbf{bold}.}}
    \label{tab:phenotype-overall-calibration}
    \begin{tabular}{cc cccccc}
        \toprule
        \textbf{Phenotype} & \textbf{Best Baseline} & \makecell{\textbf{\textsc{CEHR}}\\\textbf{\textsc{BERT}}} & \makecell{\textbf{\textsc{CEHR}}\\\textbf{\textsc{GPT}}} & \makecell{\textbf{\textsc{Core}}\\\textbf{\textsc{Behrt}}} & 
        \textbf{\llama} & \textbf{\mamba} & \textbf{\motor} \\ \cmidrule{2-8}
            AMI & \makecell { \textbf{0.011} \\ (0.000) } & \makecell { 0.011 \\ (0.000) } & \makecell { 0.011 \\ (0.000) } & \makecell { 0.011 \\ (0.000) } & \makecell { 0.011 \\ (0.000) } & \makecell { 0.011 \\ (0.000) } & \makecell { 0.011 \\ (0.000) } \\
            CLL & \makecell { 0.001 \\ (0.000) } & \makecell { 0.001 \\ (0.000) } & \makecell { \textbf{0.001} \\ (0.000) } & \makecell { 0.001 \\ (0.000) } & \makecell { 0.001 \\ (0.000) } & \makecell { 0.002 \\ (0.000) } & \makecell { 0.001 \\ (0.000) } \\
            Celiac & \makecell { 0.001 \\ (0.000) } & \makecell { 0.001 \\ (0.000) } & \makecell { 0.001 \\ (0.000) } & \makecell { 0.001 \\ (0.000) } & \makecell { 0.001 \\ (0.000) } & \makecell { 0.001 \\ (0.000) } & \makecell { \textbf{0.001} \\ (0.000) } \\
            HTN & \makecell { 0.085 \\ (0.001) } & \makecell { 0.089 \\ (0.001) } & \makecell { 0.086 \\ (0.001) } & \makecell { 0.090 \\ (0.001) } & \makecell { 0.086 \\ (0.001) } & \makecell { 0.086 \\ (0.001) } & \makecell { \textbf{0.083} \\ (0.001) } \\
            Ischemic Stroke & \makecell { \textbf{0.009} \\ (0.000) } & \makecell { 0.010 \\ (0.000) } & \makecell { 0.010 \\ (0.000) } & \makecell { 0.011 \\ (0.000) } & \makecell { 0.010 \\ (0.000) } & \makecell { 0.010 \\ (0.000) } & \makecell { 0.010 \\ (0.000) } \\
            MASLD & \makecell { 0.020 \\ (0.001) } & \makecell { 0.020 \\ (0.001) } & \makecell { 0.020 \\ (0.001) } & \makecell { 0.021 \\ (0.001) } & \makecell { \textbf{0.019} \\ (0.001) } & \makecell { 0.020 \\ (0.001) } & \makecell { 0.020 \\ (0.001) } \\
            Osteoporosis & \makecell { \textbf{0.032} \\ (0.001) } & \makecell { 0.033 \\ (0.001) } & \makecell { 0.032 \\ (0.001) } & \makecell { 0.034 \\ (0.001) } & \makecell { 0.032 \\ (0.001) } & \makecell { 0.032 \\ (0.001) } & \makecell { 0.033 \\ (0.001) } \\
            Pancreatic Cancer & \makecell { 0.003 \\ (0.000) } & \makecell { \textbf{0.003} \\ (0.000) } & \makecell { 0.003 \\ (0.000) } & \makecell { 0.004 \\ (0.000) } & \makecell { 0.003 \\ (0.000) } & \makecell { 0.003 \\ (0.000) } & \makecell { 0.003 \\ (0.000) } \\
            SLE & \makecell { \textbf{0.001} \\ (0.000) } & \makecell { 0.001 \\ (0.000) } & \makecell { 0.001 \\ (0.000) } & \makecell { 0.001 \\ (0.000) } & \makecell { 0.001 \\ (0.000) } & \makecell { 0.001 \\ (0.000) } & \makecell { 0.001 \\ (0.000) } \\
            Schizophrenia & \makecell { \textbf{0.114} \\ (0.002) } & \makecell { 0.134 \\ (0.002) } & \makecell { 0.128 \\ (0.002) } & \makecell { 0.167 \\ (0.002) } & \makecell { 0.126 \\ (0.002) } & \makecell { 0.123 \\ (0.002) } & \makecell { 0.121 \\ (0.002) } \\
            T2DM & \makecell { 0.017 \\ (0.001) } & \makecell { 0.017 \\ (0.001) } & \makecell { 0.017 \\ (0.001) } & \makecell { 0.017 \\ (0.001) } & \makecell { 0.017 \\ (0.001) } & \makecell { 0.017 \\ (0.001) } & \makecell { \textbf{0.016} \\ (0.001) } \\

        \bottomrule
    \end{tabular}
    \vspace{2pt}
    \newline
    AMI: Acute myocardial infarction. CLL: Chronic lymphocytic leukemia. HTN: Hypertension. MASLD: Metabolic dysfunction-associated steatotic liver disease. SLE: Systemic lupus erythematosus. T2DM: Type 2 diabetes mellitus.
\end{table}

\newpage
\begin{table}[ht]
    \footnotesize
    \centering
    \setlength{\tabcolsep}{3pt}
    \setlength\extrarowheight{3pt}
    \caption{Outcome ECE results. \textit{Best performances are marked in \textbf{bold}.}}
    \label{tab:outcome-overall-ece}
    \begin{tabular}{cccccccc}
        \toprule
         & \textbf{Best Baseline} & \makecell{\textbf{\textsc{CEHR}}\\\textbf{\textsc{BERT}}} & \makecell{\textbf{\textsc{CEHR}}\\\textbf{\textsc{GPT}}} & \makecell{\textbf{\textsc{Core}}\\\textbf{\textsc{Behrt}}} & 
        \textbf{\llama} & \textbf{\mamba} & \textbf{\motor} \\ \cmidrule{2-8}
            LOS & \makecell { 0.010 \\ (0.001) } & \makecell { 0.011 \\ (0.002) } & \makecell { \textbf{0.007} \\ (0.001) } & \makecell { 0.017 \\ (0.002) } & \makecell { 0.011 \\ (0.001) } & \makecell { 0.015 \\ (0.002) } & \makecell { 0.009 \\ (0.002) } \\
            Mortality & \makecell { 0.002 \\ (0.000) } & \makecell { 0.002 \\ (0.001) } & \makecell { 0.002 \\ (0.000) } & \makecell { 0.003 \\ (0.001) } & \makecell { \textbf{0.001} \\ (0.000) } & \makecell { 0.002 \\ (0.000) } & \makecell { 0.002 \\ (0.000) } \\
            Readmission & \makecell { 0.009 \\ (0.002) } & \makecell { 0.006 \\ (0.001) } & \makecell { 0.005 \\ (0.001) } & \makecell { 0.006 \\ (0.001) } & \makecell { \textbf{0.004} \\ (0.001) } & \makecell { 0.004 \\ (0.001) } & \makecell { 0.008 \\ (0.001) } \\

        \bottomrule
    \end{tabular}
\end{table}

\begin{table}[ht]
    \footnotesize
    \centering
    \setlength{\tabcolsep}{3pt}
    \setlength\extrarowheight{3pt}
    \caption{Phenotype ECE results. \textit{Best performances are marked in \textbf{bold}.}}
    \label{tab:phenotype-overall-ece}
    \begin{tabular}{cc cccccc}
        \toprule
        \textbf{Phenotype} & \textbf{Best Baseline} & \makecell{\textbf{\textsc{CEHR}}\\\textbf{\textsc{BERT}}} & \makecell{\textbf{\textsc{CEHR}}\\\textbf{\textsc{GPT}}} & \makecell{\textbf{\textsc{Core}}\\\textbf{\textsc{Behrt}}} & 
        \textbf{\llama} & \textbf{\mamba} & \textbf{\motor} \\ \cmidrule{2-8}
            AMI & \makecell { 0.002 \\ (0.001) } & \makecell { 0.002 \\ (0.000) } & \makecell { 0.002 \\ (0.000) } & \makecell { 0.004 \\ (0.000) } & \makecell { 0.002 \\ (0.000) } & \makecell { \textbf{0.001} \\ (0.000) } & \makecell { 0.002 \\ (0.000) } \\
            CLL & \makecell { \textbf{0.000} \\ (0.000) } & \makecell { 0.001 \\ (0.000) } & \makecell { 0.001 \\ (0.000) } & \makecell { 0.001 \\ (0.000) } & \makecell { 0.001 \\ (0.000) } & \makecell { 0.001 \\ (0.000) } & \makecell { 0.001 \\ (0.000) } \\
            Celiac & \makecell { 0.000 \\ (0.000) } & \makecell { 0.000 \\ (0.000) } & \makecell { 0.000 \\ (0.000) } & \makecell { 0.000 \\ (0.000) } & \makecell { 0.000 \\ (0.000) } & \makecell { \textbf{0.000} \\ (0.000) } & \makecell { 0.000 \\ (0.000) } \\
            HTN & \makecell { \textbf{0.012} \\ (0.001) } & \makecell { 0.030 \\ (0.001) } & \makecell { 0.012 \\ (0.001) } & \makecell { 0.028 \\ (0.001) } & \makecell { 0.015 \\ (0.001) } & \makecell { 0.014 \\ (0.001) } & \makecell { 0.023 \\ (0.001) } \\
            Ischemic Stroke & \makecell { \textbf{0.002} \\ (0.000) } & \makecell { 0.003 \\ (0.000) } & \makecell { 0.002 \\ (0.000) } & \makecell { 0.008 \\ (0.000) } & \makecell { 0.003 \\ (0.000) } & \makecell { 0.003 \\ (0.000) } & \makecell { 0.003 \\ (0.000) } \\
            MASLD & \makecell { 0.003 \\ (0.001) } & \makecell { 0.004 \\ (0.001) } & \makecell { 0.001 \\ (0.001) } & \makecell { 0.006 \\ (0.001) } & \makecell { \textbf{0.001} \\ (0.001) } & \makecell { 0.001 \\ (0.000) } & \makecell { 0.002 \\ (0.001) } \\
            Osteoporosis & \makecell { 0.005 \\ (0.001) } & \makecell { 0.014 \\ (0.001) } & \makecell { 0.005 \\ (0.001) } & \makecell { 0.020 \\ (0.001) } & \makecell { \textbf{0.004} \\ (0.001) } & \makecell { 0.006 \\ (0.001) } & \makecell { 0.011 \\ (0.001) } \\
            Pancreatic Cancer & \makecell { \textbf{0.001} \\ (0.000) } & \makecell { 0.001 \\ (0.000) } & \makecell { 0.001 \\ (0.000) } & \makecell { 0.003 \\ (0.000) } & \makecell { 0.001 \\ (0.000) } & \makecell { 0.002 \\ (0.000) } & \makecell { 0.002 \\ (0.000) } \\
            SLE & \makecell { \textbf{0.000} \\ (0.000) } & \makecell { 0.000 \\ (0.000) } & \makecell { 0.000 \\ (0.000) } & \makecell { 0.000 \\ (0.000) } & \makecell { 0.000 \\ (0.000) } & \makecell { 0.000 \\ (0.000) } & \makecell { 0.000 \\ (0.000) } \\
            Schizophrenia & \makecell { \textbf{0.041} \\ (0.002) } & \makecell { 0.109 \\ (0.003) } & \makecell { 0.103 \\ (0.003) } & \makecell { 0.155 \\ (0.003) } & \makecell { 0.080 \\ (0.002) } & \makecell { 0.074 \\ (0.002) } & \makecell { 0.079 \\ (0.002) } \\
            T2DM & \makecell { 0.003 \\ (0.001) } & \makecell { 0.004 \\ (0.001) } & \makecell { 0.001 \\ (0.000) } & \makecell { 0.005 \\ (0.001) } & \makecell { 0.001 \\ (0.000) } & \makecell { \textbf{0.001} \\ (0.000) } & \makecell { 0.001 \\ (0.001) } \\
        \bottomrule
    \end{tabular}
    \vspace{2pt}
    \newline
    AMI: Acute myocardial infarction. CLL: Chronic lymphocytic leukemia. HTN: Hypertension. MASLD: Metabolic dysfunction-associated steatotic liver disease. SLE: Systemic lupus erythematosus. T2DM: Type 2 diabetes mellitus.
\end{table}

\clearpage
\begin{table}[ht]
    \footnotesize
    \centering
    \setlength{\tabcolsep}{3pt}
    \setlength\extrarowheight{3pt}
    \caption{Outcome sex gap in discriminative performance. \textit{Best performances are marked in \textbf{bold}.}}
    \label{tab:outcome-sex}
    \begin{tabular}{cccccccc}
        \toprule
         & \textbf{Best Baseline} & \makecell{\textbf{\textsc{CEHR}}\\\textbf{\textsc{BERT}}} & \makecell{\textbf{\textsc{CEHR}}\\\textbf{\textsc{GPT}}} & \makecell{\textbf{\textsc{Core}}\\\textbf{\textsc{Behrt}}} & 
        \textbf{\llama} & \textbf{\mamba} & \textbf{\motor} \\ \cmidrule{2-8}
            OS & \makecell { 0.101 \\ (0.005) } & \makecell { 0.086 \\ (0.005) } & \makecell { \textbf{0.053} \\ (0.003) } & \makecell { 0.108 \\ (0.005) } & \makecell { 0.088 \\ (0.004) } & \makecell { 0.103 \\ (0.004) } & \makecell { 0.058 \\ (0.004) } \\
            Mortality & \makecell { 0.048 \\ (0.007) } & \makecell { 0.031 \\ (0.008) } & \makecell { \textbf{0.027} \\ (0.005) } & \makecell { 0.056 \\ (0.010) } & \makecell { 0.047 \\ (0.009) } & \makecell { 0.053 \\ (0.010) } & \makecell { 0.028 \\ (0.005) } \\
            Readmission & \makecell { 0.074 \\ (0.006) } & \makecell { \textbf{0.049} \\ (0.006) } & \makecell { 0.070 \\ (0.006) } & \makecell { 0.056 \\ (0.006) } & \makecell { 0.082 \\ (0.006) } & \makecell { 0.081 \\ (0.006) } & \makecell { 0.068 \\ (0.006) } \\

        \bottomrule
    \end{tabular}
\end{table}

\begin{table}[ht]
    \footnotesize
    \centering
    \setlength{\tabcolsep}{3pt}
    \setlength\extrarowheight{3pt}
    \caption{Phenotype sex gap in discriminative performance.\textit{Best performances are marked in \textbf{bold}.}}
    \label{tab:phenotype-sex}
    \begin{tabular}{cc cccccc}
        \toprule
        \textbf{Phenotype} & \textbf{Best Baseline} & \makecell{\textbf{\textsc{CEHR}}\\\textbf{\textsc{BERT}}} & \makecell{\textbf{\textsc{CEHR}}\\\textbf{\textsc{GPT}}} & \makecell{\textbf{\textsc{Core}}\\\textbf{\textsc{Behrt}}} & 
        \textbf{\llama} & \textbf{\mamba} & \textbf{\motor} \\ \cmidrule{2-8}
            AMI & \makecell { 0.022 \\ (0.026) } & \makecell { 0.052 \\ (0.016) } & \makecell { 0.056 \\ (0.016) } & \makecell { \textbf{0.016} \\ (0.019) } & \makecell { 0.032 \\ (0.017) } & \makecell { 0.037 \\ (0.017) } & \makecell { 0.049 \\ (0.017) } \\
CLL & \makecell { 0.216 \\ (0.038) } & \makecell { 0.213 \\ (0.058) } & \makecell { 0.018 \\ (0.062) } & \makecell { 0.352 \\ (0.047) } & \makecell { \textbf{0.010} \\ (0.076) } & \makecell { 0.031 \\ (0.054) } & \makecell { 0.126 \\ (0.064) } \\
Celiac & \makecell { 0.174 \\ (0.132) } & \makecell { 0.051 \\ (0.090) } & \makecell { 0.069 \\ (0.081) } & \makecell { 0.050 \\ (0.100) } & \makecell { 0.113 \\ (0.141) } & \makecell { 0.119 \\ (0.149) } & \makecell { \textbf{0.028} \\ (0.076) } \\
HTN & \makecell { 0.038 \\ (0.009) } & \makecell { 0.006 \\ (0.006) } & \makecell { 0.007 \\ (0.007) } & \makecell { 0.012 \\ (0.007) } & \makecell { \textbf{0.002} \\ (0.007) } & \makecell { 0.007 \\ (0.007) } & \makecell { 0.014 \\ (0.006) } \\
Ischemic Stroke & \makecell { 0.110 \\ (0.030) } & \makecell { 0.016 \\ (0.016) } & \makecell { 0.019 \\ (0.017) } & \makecell { 0.053 \\ (0.018) } & \makecell { 0.013 \\ (0.017) } & \makecell { \textbf{0.008} \\ (0.015) } & \makecell { 0.010 \\ (0.013) } \\
MASLD & \makecell { 0.029 \\ (0.019) } & \makecell { 0.019 \\ (0.018) } & \makecell { 0.025 \\ (0.019) } & \makecell { \textbf{0.003} \\ (0.017) } & \makecell { 0.040 \\ (0.016) } & \makecell { 0.026 \\ (0.016) } & \makecell { 0.023 \\ (0.017) } \\
Osteoporosis & \makecell { 0.105 \\ (0.023) } & \makecell { \textbf{0.037} \\ (0.017) } & \makecell { 0.137 \\ (0.014) } & \makecell { 0.063 \\ (0.018) } & \makecell { 0.090 \\ (0.017) } & \makecell { 0.088 \\ (0.014) } & \makecell { 0.102 \\ (0.014) } \\
Pancreatic Cancer & \makecell { 0.382 \\ (0.053) } & \makecell { 0.027 \\ (0.039) } & \makecell { 0.113 \\ (0.039) } & \makecell { 0.130 \\ (0.033) } & \makecell { \textbf{0.001} \\ (0.032) } & \makecell { 0.089 \\ (0.032) } & \makecell { 0.067 \\ (0.029) } \\
SLE & \makecell { 0.086 \\ (0.097) } & \makecell { 0.083 \\ (0.143) } & \makecell { 0.054 \\ (0.083) } & \makecell { 0.040 \\ (0.112) } & \makecell { 0.115 \\ (0.048) } & \makecell { \textbf{0.005} \\ (0.104) } & \makecell { 0.011 \\ (0.079) } \\
Schizophrenia & \makecell { \textbf{0.067} \\ (0.009) } & \makecell { 0.089 \\ (0.010) } & \makecell { 0.136 \\ (0.008) } & \makecell { 0.067 \\ (0.009) } & \makecell { 0.106 \\ (0.009) } & \makecell { 0.098 \\ (0.009) } & \makecell { 0.077 \\ (0.009) } \\
T2DM & \makecell { 0.026 \\ (0.020) } & \makecell { 0.037 \\ (0.018) } & \makecell { 0.036 \\ (0.019) } & \makecell { \textbf{0.012} \\ (0.019) } & \makecell { 0.026 \\ (0.016) } & \makecell { 0.042 \\ (0.017) } & \makecell { 0.016 \\ (0.015) } \\
        \bottomrule
    \end{tabular}
    \vspace{2pt}
    \newline
    AMI: Acute myocardial infarction. CLL: Chronic lymphocytic leukemia. HTN: Hypertension. MASLD: Metabolic dysfunction-associated steatotic liver disease. SLE: Systemic lupus erythematosus. T2DM: Type 2 diabetes mellitus.
\end{table}

\clearpage
\begin{table}[ht]
    \footnotesize
    \centering
    \setlength{\tabcolsep}{3pt}
    \setlength\extrarowheight{3pt}
    \caption{Outcome racial gap in discriminative performance. \textit{Best performances are marked in \textbf{bold}.}}
    \label{tab:outcome-race}
    \begin{tabular}{cccccccc}
        \toprule
         & \textbf{Best Baseline} & \makecell{\textbf{\textsc{CEHR}}\\\textbf{\textsc{BERT}}} & \makecell{\textbf{\textsc{CEHR}}\\\textbf{\textsc{GPT}}} & \makecell{\textbf{\textsc{Core}}\\\textbf{\textsc{Behrt}}} & 
        \textbf{\llama} & \textbf{\mamba} & \textbf{\motor} \\ \cmidrule{2-8}
LOS & \makecell { 0.029 \\ (0.006) } & \makecell { 0.017 \\ (0.006) } & \makecell { 0.026 \\ (0.005) } & \makecell { \textbf{0.007} \\ (0.007) } & \makecell { 0.022 \\ (0.006) } & \makecell { 0.010 \\ (0.006) } & \makecell { 0.017 \\ (0.005) } \\
Mortality & \makecell { 0.034 \\ (0.023) } & \makecell { 0.006 \\ (0.014) } & \makecell { 0.009 \\ (0.009) } & \makecell { 0.033 \\ (0.021) } & \makecell { 0.008 \\ (0.014) } & \makecell { 0.010 \\ (0.016) } & \makecell { \textbf{0.004} \\ (0.011) } \\
Readmission & \makecell { 0.022 \\ (0.011) } & \makecell { 0.005 \\ (0.008) } & \makecell { 0.008 \\ (0.008) } & \makecell { 0.007 \\ (0.009) } & \makecell { 0.006 \\ (0.009) } & \makecell { \textbf{0.002} \\ (0.009) } & \makecell { 0.006 \\ (0.008) } \\

        \bottomrule
    \end{tabular}
\end{table}

\begin{table}[ht]
    \footnotesize
    \centering
    \setlength{\tabcolsep}{3pt}
    \setlength\extrarowheight{3pt}
    \caption{Phenotype racial gap in discriminative performance.\textit{Best performances are marked in \textbf{bold}.}}
    \label{tab:phenotype-race}
    \begin{tabular}{cc cccccc}
        \toprule
        \textbf{Phenotype} & \textbf{Best Baseline} & \makecell{\textbf{\textsc{CEHR}}\\\textbf{\textsc{BERT}}} & \makecell{\textbf{\textsc{CEHR}}\\\textbf{\textsc{GPT}}} & \makecell{\textbf{\textsc{Core}}\\\textbf{\textsc{Behrt}}} & 
        \textbf{\llama} & \textbf{\mamba}& \textbf{\motor} \\ \cmidrule{2-8}
AMI & \makecell { 0.134 \\ (0.041) } & \makecell { 0.083 \\ (0.034) } & \makecell { \textbf{0.025} \\ (0.026) } & \makecell { 0.031 \\ (0.031) } & \makecell { 0.061 \\ (0.024) } & \makecell { 0.028 \\ (0.026) } & \makecell { 0.059 \\ (0.028) } \\
CLL & \makecell { 0.100 \\ (0.088) } & \makecell { \textbf{0.035} \\ (0.101) } & \makecell { 0.243 \\ (0.140) } & \makecell { 0.097 \\ (0.143) } & \makecell { 0.160 \\ (0.165) } & \makecell { 0.037 \\ (0.128) } & \makecell { 0.041 \\ (0.151) } \\
Celiac & \makecell { 0.419 \\ (0.065) } & \makecell { 0.286 \\ (0.044) } & \makecell { 0.393 \\ (0.079) } & \makecell { \textbf{0.078} \\ (0.064) } & \makecell { 0.268 \\ (0.096) } & \makecell { 0.194 \\ (0.101) } & \makecell { 0.218 \\ (0.082) } \\
HTN & \makecell { 0.037 \\ (0.014) } & \makecell { 0.016 \\ (0.011) } & \makecell { \textbf{0.002} \\ (0.012) } & \makecell { 0.045 \\ (0.011) } & \makecell { 0.006 \\ (0.010) } & \makecell { 0.006 \\ (0.011) } & \makecell { 0.024 \\ (0.009) } \\
Ischemic Stroke & \makecell { 0.097 \\ (0.022) } & \makecell { 0.022 \\ (0.026) } & \makecell { \textbf{0.009} \\ (0.025) } & \makecell { 0.039 \\ (0.025) } & \makecell { 0.043 \\ (0.023) } & \makecell { 0.041 \\ (0.023) } & \makecell { 0.053 \\ (0.018) } \\
MASLD & \makecell { 0.046 \\ (0.032) } & \makecell { 0.047 \\ (0.021) } & \makecell { 0.046 \\ (0.025) } & \makecell { 0.027 \\ (0.021) } & \makecell { 0.036 \\ (0.020) } & \makecell { 0.050 \\ (0.025) } & \makecell { \textbf{0.016} \\ (0.022) } \\
Osteoporosis & \makecell { 0.108 \\ (0.019) } & \makecell { 0.085 \\ (0.016) } & \makecell { 0.061 \\ (0.016) } & \makecell { 0.045 \\ (0.023) } & \makecell { 0.046 \\ (0.021) } & \makecell { \textbf{0.011} \\ (0.019) } & \makecell { 0.043 \\ (0.016) } \\
Pancreatic Cancer & \makecell { 0.393 \\ (0.057) } & \makecell { 0.245 \\ (0.038) } & \makecell { 0.274 \\ (0.055) } & \makecell { 0.030 \\ (0.046) } & \makecell { \textbf{0.010} \\ (0.054) } & \makecell { 0.011 \\ (0.041) } & \makecell { 0.227 \\ (0.040) } \\
SLE & \makecell { 0.170 \\ (0.095) } & \makecell { 0.088 \\ (0.090) } & \makecell { 0.104 \\ (0.063) } & \makecell { 0.272 \\ (0.084) } & \makecell { \textbf{0.008} \\ (0.085) } & \makecell { 0.088 \\ (0.083) } & \makecell { 0.052 \\ (0.085) } \\
Schizophrenia & \makecell { 0.072 \\ (0.012) } & \makecell { 0.055 \\ (0.016) } & \makecell { 0.055 \\ (0.013) } & \makecell { 0.158 \\ (0.012) } & \makecell { \textbf{0.015} \\ (0.016) } & \makecell { 0.043 \\ (0.012) } & \makecell { 0.070 \\ (0.014) } \\
T2DM & \makecell { 0.140 \\ (0.025) } & \makecell { 0.079 \\ (0.025) } & \makecell { 0.069 \\ (0.023) } & \makecell { 0.074 \\ (0.024) } & \makecell { 0.113 \\ (0.020) } & \makecell { 0.128 \\ (0.023) } & \makecell { \textbf{0.043} \\ (0.018) } \\
        \bottomrule
    \end{tabular}
    \vspace{2pt}
    \newline
    AMI: Acute myocardial infarction. CLL: Chronic lymphocytic leukemia. HTN: Hypertension. MASLD: Metabolic dysfunction-associated steatotic liver disease. SLE: Systemic lupus erythematosus. T2DM: Type 2 diabetes mellitus.
\end{table}

\clearpage
\begin{table}[ht]
    \footnotesize
    \centering
    \setlength{\tabcolsep}{3pt}
    \setlength\extrarowheight{3pt}
    \caption{Outcome healthcare utilization gap in discriminative performance. \textit{Best performances are marked in \textbf{bold}.}}
    \label{tab:outcome-utilization}
    \begin{tabular}{ccccccccc}
        \toprule
         & \textbf{Best Baseline} & \makecell{\textbf{\textsc{CEHR}}\\\textbf{\textsc{BERT}}} & \makecell{\textbf{\textsc{CEHR}}\\\textbf{\textsc{GPT}}} & \makecell{\textbf{\textsc{Core}}\\\textbf{\textsc{Behrt}}} & 
        \textbf{\llama} & \textbf{\mamba}& \textbf{\motor} \\ \cmidrule{2-8}
    LOS & \makecell { 0.071 \\ (0.005) } & \makecell { 0.060 \\ (0.004) } & \makecell { \textbf{0.028} \\ (0.003) } & \makecell { 0.070 \\ (0.004) } & \makecell { 0.047 \\ (0.004) } & \makecell { 0.048 \\ (0.004) } & \makecell { 0.031 \\ (0.003) } \\
Mortality & \makecell { 0.077 \\ (0.010) } & \makecell { 0.068 \\ (0.008) } & \makecell { \textbf{0.040} \\ (0.005) } & \makecell { 0.087 \\ (0.012) } & \makecell { 0.060 \\ (0.007) } & \makecell { 0.054 \\ (0.009) } & \makecell { 0.045 \\ (0.005) } \\
Readmission & \makecell { 0.078 \\ (0.027) } & \makecell { 0.068 \\ (0.026) } & \makecell { 0.042 \\ (0.029) } & \makecell { 0.041 \\ (0.006) } & \makecell { \textbf{0.039} \\ (0.006) } & \makecell { 0.043 \\ (0.006) } & \makecell { 0.053 \\ (0.027) } \\
        \bottomrule
    \end{tabular}
\end{table}

\begin{table}[ht]
    \footnotesize
    \centering
    \setlength{\tabcolsep}{3pt}
    \setlength\extrarowheight{3pt}
    \caption{Phenotype healthcare utilization gap in discriminative performance.\textit{Best performances are marked in \textbf{bold}.}}
    \label{tab:phenotype-utilization}
    \begin{tabular}{cc cccccc}
        \toprule
        \textbf{Phenotype} & \textbf{Best Baseline} & \makecell{\textbf{\textsc{CEHR}}\\\textbf{\textsc{BERT}}} & \makecell{\textbf{\textsc{CEHR}}\\\textbf{\textsc{GPT}}} & \makecell{\textbf{\textsc{Core}}\\\textbf{\textsc{Behrt}}} & 
        \textbf{\llama} & \textbf{\mamba} & \textbf{\motor} \\ \cmidrule{2-8}
    AMI & \makecell { 0.261 \\ (0.021) } & \makecell { 0.113 \\ (0.017) } & \makecell { 0.179 \\ (0.011) } & \makecell { 0.201 \\ (0.017) } & \makecell { 0.122 \\ (0.011) } & \makecell { 0.100 \\ (0.052) } & \makecell { \textbf{0.081} \\ (0.020) } \\
CLL & \makecell { 0.260 \\ (0.057) } & \makecell { 0.416 \\ (0.063) } & \makecell { \textbf{0.017} \\ (0.063) } & \makecell { 0.103 \\ (0.051) } & \makecell { 0.070 \\ (0.068) } & \makecell { 0.058 \\ (0.069) } & \makecell { 0.248 \\ (0.064) } \\
Celiac & \makecell { 0.173 \\ (0.111) } & \makecell { 0.116 \\ (0.099) } & \makecell { 0.158 \\ (0.109) } & \makecell { 0.174 \\ (0.146) } & \makecell { 0.113 \\ (0.146) } & \makecell { 0.171 \\ (0.113) } & \makecell { \textbf{0.068} \\ (0.099) } \\
HTN & \makecell { 0.080 \\ (0.025) } & \makecell { 0.038 \\ (0.024) } & \makecell { \textbf{0.009} \\ (0.008) } & \makecell { 0.024 \\ (0.008) } & \makecell { 0.028 \\ (0.029) } & \makecell { 0.025 \\ (0.008) } & \makecell { 0.037 \\ (0.024) } \\
Ischemic Stroke & \makecell { 0.130 \\ (0.029) } & \makecell { 0.008 \\ (0.022) } & \makecell { 0.013 \\ (0.028) } & \makecell { 0.027 \\ (0.023) } & \makecell { 0.013 \\ (0.024) } & \makecell { \textbf{0.008} \\ (0.025) } & \makecell { 0.053 \\ (0.016) } \\
MASLD & \makecell { 0.221 \\ (0.106) } & \makecell { 0.213 \\ (0.066) } & \makecell { 0.076 \\ (0.130) } & \makecell { \textbf{0.043} \\ (0.092) } & \makecell { 0.237 \\ (0.017) } & \makecell { 0.052 \\ (0.155) } & \makecell { 0.181 \\ (0.018) } \\
Osteoporosis & \makecell { 0.247 \\ (0.080) } & \makecell { 0.101 \\ (0.103) } & \makecell { \textbf{0.039} \\ (0.080) } & \makecell { 0.231 \\ (0.050) } & \makecell { 0.092 \\ (0.081) } & \makecell { 0.170 \\ (0.045) } & \makecell { 0.046 \\ (0.061) } \\
Pancreatic Cancer & \makecell { 0.315 \\ (0.060) } & \makecell { 0.116 \\ (0.054) } & \makecell { 0.157 \\ (0.054) } & \makecell { 0.163 \\ (0.048) } & \makecell { 0.135 \\ (0.039) } & \makecell { \textbf{0.091} \\ (0.050) } & \makecell { 0.098 \\ (0.046) } \\
SLE & \makecell { 0.157 \\ (0.118) } & \makecell { 0.254 \\ (0.072) } & \makecell { 0.083 \\ (0.103) } & \makecell { 0.230 \\ (0.070) } & \makecell { 0.017 \\ (0.098) } & \makecell { \textbf{0.010} \\ (0.091) } & \makecell { 0.011 \\ (0.105) } \\
Schizophrenia & \makecell { 0.226 \\ (0.026) } & \makecell { 0.082 \\ (0.036) } & \makecell { 0.057 \\ (0.043) } & \makecell { 0.051 \\ (0.012) } & \makecell { 0.075 \\ (0.032) } & \makecell { \textbf{0.021} \\ (0.011) } & \makecell { 0.054 \\ (0.033) } \\
T2DM & \makecell { 0.234 \\ (0.104) } & \makecell { \textbf{0.086} \\ (0.019) } & \makecell { 0.153 \\ (0.071) } & \makecell { 0.104 \\ (0.110) } & \makecell { 0.207 \\ (0.024) } & \makecell { 0.118 \\ (0.065) } & \makecell { 0.158 \\ (0.019) } \\
        \bottomrule
    \end{tabular}
    \vspace{2pt}
    \newline
    AMI: Acute myocardial infarction. CLL: Chronic lymphocytic leukemia. HTN: Hypertension. MASLD: Metabolic dysfunction-associated steatotic liver disease. SLE: Systemic lupus erythematosus. T2DM: Type 2 diabetes mellitus.
\end{table}

\clearpage
\newpage
\subsection{Detailed baseline results}
For completeness, Tables~\ref{tab:outcome-overall-baseline} and~\ref{tab:phenotype-overall-baseline} detail the discriminative performances for each considered baseline model. All previous tables and figures reflect the best performance across these different models.

\begin{table}[ht]
    \footnotesize
    \centering
    \setlength{\tabcolsep}{3.5pt}
    \caption{Outcome prediction results for considered baselines.}
    \label{tab:outcome-overall-baseline}
    \begin{tabular}{ccccc}
        \toprule
        \textbf{Task} &
         \textbf{\femr - GBM} & \textbf{\femr - LR} & \textbf{\medstab - XGB} & \textbf{\medstab - LR} \\
        \cmidrule{2-5}
LOS & \makecell { 0.808 \\ (0.002) } & \makecell { 0.728 \\ (0.002) } & \makecell { \textbf{0.829} \\ (0.002) } & \makecell { 0.799 \\ (0.002) } \\
Mortality & \makecell { 0.896 \\ (0.004) } & \makecell { 0.833 \\ (0.006) } & \makecell { \textbf{0.914} \\ (0.003) } & \makecell { 0.878 \\ (0.005) } \\
Readmission & \makecell { 0.746 \\ (0.003) } & \makecell { 0.683 \\ (0.004) } & \makecell { \textbf{0.765} \\ (0.003) } & \makecell { 0.707 \\ (0.003) } \\

        \bottomrule
    \end{tabular}
    \vspace{2pt}
    \newline
    {LR = Logistic Regression, GBM = LightGBM, XGB = XGBoost}
\end{table}

\begin{table}[ht]
    \footnotesize
    \centering
    \setlength{\tabcolsep}{3.5pt}
    \caption{Phenotype prediction results for considered baselines.}
    \label{tab:phenotype-overall-baseline}
    \begin{tabular}{ccccc}
        \toprule
                \textbf{Phenotype}
         & \textbf{\femr - GBM} & \textbf{\femr - LR} & \textbf{\medstab - XGB} & \textbf{\medstab - LR} \\
        \cmidrule{2-5}
    AMI & \makecell { 0.772 \\ (0.010) } & \makecell { 0.665 \\ (0.014) } & \makecell { \textbf{0.828} \\ (0.009) } & \makecell { 0.687 \\ (0.013) } \\
CLL & \makecell { 0.491 \\ (0.032) } & \makecell { \textbf{0.720} \\ (0.026) } & \makecell { 0.657 \\ (0.031) } & \makecell { 0.686 \\ (0.033) } \\
Celiac & \makecell { 0.566 \\ (0.044) } & \makecell { 0.531 \\ (0.045) } & \makecell { 0.539 \\ (0.038) } & \makecell { \textbf{0.581} \\ (0.042) } \\
HTN & \makecell { 0.701 \\ (0.003) } & \makecell { 0.578 \\ (0.004) } & \makecell { \textbf{0.731} \\ (0.004) } & \makecell { 0.582 \\ (0.005) } \\
Ischemic Stroke & \makecell { 0.850 \\ (0.008) } & \makecell { 0.718 \\ (0.012) } & \makecell { \textbf{0.882} \\ (0.008) } & \makecell { 0.751 \\ (0.014) } \\
MASLD & \makecell { 0.660 \\ (0.010) } & \makecell { 0.554 \\ (0.011) } & \makecell { \textbf{0.674} \\ (0.008) } & \makecell { 0.514 \\ (0.010) } \\
Osteoporosis & \makecell { 0.674 \\ (0.006) } & \makecell { 0.533 \\ (0.008) } & \makecell { \textbf{0.707} \\ (0.006) } & \makecell { 0.536 \\ (0.009) } \\
Pancreatic Cancer & \makecell { 0.673 \\ (0.024) } & \makecell { 0.470 \\ (0.030) } & \makecell { \textbf{0.729} \\ (0.016) } & \makecell { 0.418 \\ (0.028) } \\
SLE & \makecell { 0.698 \\ (0.041) } & \makecell { 0.395 \\ (0.027) } & \makecell { \textbf{0.716} \\ (0.027) } & \makecell { 0.452 \\ (0.032) } \\
Schizophrenia & \makecell { \textbf{0.807} \\ (0.004) } & \makecell { 0.605 \\ (0.006) } & \makecell { 0.752 \\ (0.005) } & \makecell { 0.615 \\ (0.005) } \\
T2DM & \makecell { \textbf{0.714} \\ (0.010) } & \makecell { 0.577 \\ (0.010) } & \makecell { 0.706 \\ (0.008) } & \makecell { 0.591 \\ (0.010) } \\
        \bottomrule
    \end{tabular}
    \vspace{2pt}
    \newline
    {LR = Logistic Regression, GBM = LightGBM, XGB = XGBoost}
\end{table}

\newpage
\section{\mimic detailed results}~\label{appendix:additional_results:mimic}

\subsection{Average performance comparison}
Table~\ref{tab:leaderboard_perf_mimic} presents the average performance across tasks across both small and large numbers of labeled samples observed in \mimic. These results echo the same conclusions as \cumc.
\begin{table}[!ht]
    \footnotesize
    \centering
    \setlength{\tabcolsep}{2pt}
    \caption{Average performance on the proposed dataset stratified per metrics, sorted by average ranking across all metrics.}
    \label{tab:leaderboard_perf_mimic}
    \begin{tabular}{cccc ccc}
        \toprule
        Foundation & \multicolumn{3}{c}{Large data regime} & \multicolumn{3}{c}{Small data regime} \\
        \cmidrule(lr){2-4} \cmidrule(lr){5-7}
         Model & Calibration & Discrimination & Fairness &Calibration & Discrimination & Fairness \\
        \cmidrule{2-7}
\motor & 0.01 (0.01) & 0.79 (0.08) & 0.04 (0.04) & 0.11 (0.07) & 0.70 (0.07) & 0.05 (0.05) \\
\cehrgpt & 0.00 (0.00) & 0.78 (0.08) & 0.05 (0.07) & 0.07 (0.03) & 0.65 (0.10) & 0.06 (0.07) \\
\mamba & 0.00 (0.00) & 0.75 (0.08) & 0.04 (0.05) & 0.09 (0.08) & 0.62 (0.13) & 0.04 (0.04) \\
\llama & 0.00 (0.00) & 0.76 (0.08) & 0.05 (0.05) & 0.06 (0.03) & 0.61 (0.11) & 0.05 (0.07) \\
\cehrbert & 0.00 (0.00) & 0.75 (0.08) & 0.06 (0.06) & 0.10 (0.07) & 0.64 (0.09) & 0.06 (0.06) \\
Best Baseline & 0.01 (0.01) & 0.76 (0.08) & 0.13 (0.11) & 0.06 (0.02) & 0.65 (0.07) & 0.11 (0.09) \\
\corebehrt & 0.01 (0.01) & 0.70 (0.07) & 0.06 (0.09) & 0.08 (0.03) & 0.61 (0.08) & 0.07 (0.06) \\
         \bottomrule
    \end{tabular}
\end{table}

\newpage
\subsection{Task-specific performances}
\begin{figure}[ht]
    \makebox[\textwidth][l]{
    \includegraphics[height=100px]{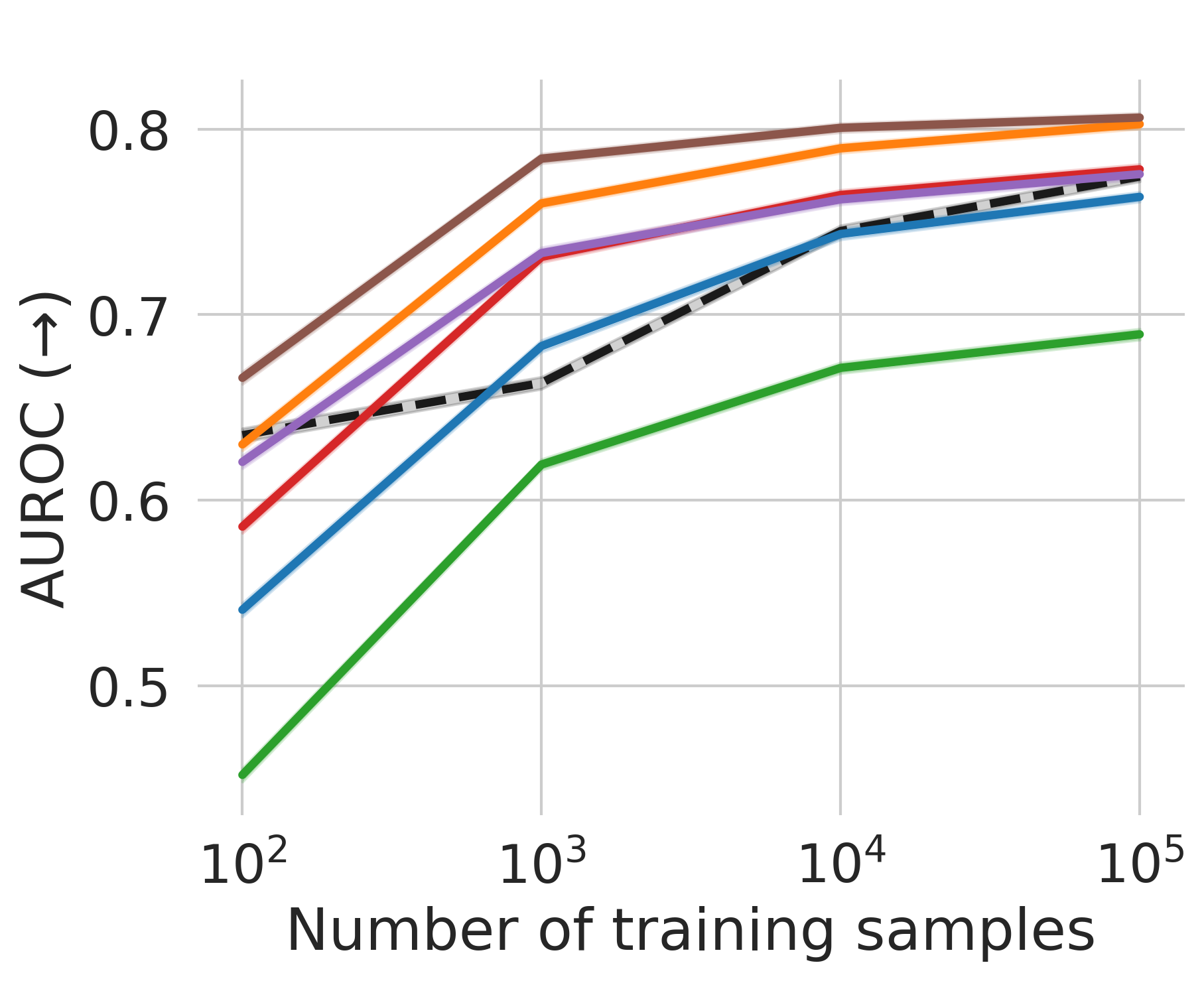}    
    \includegraphics[height=100px]{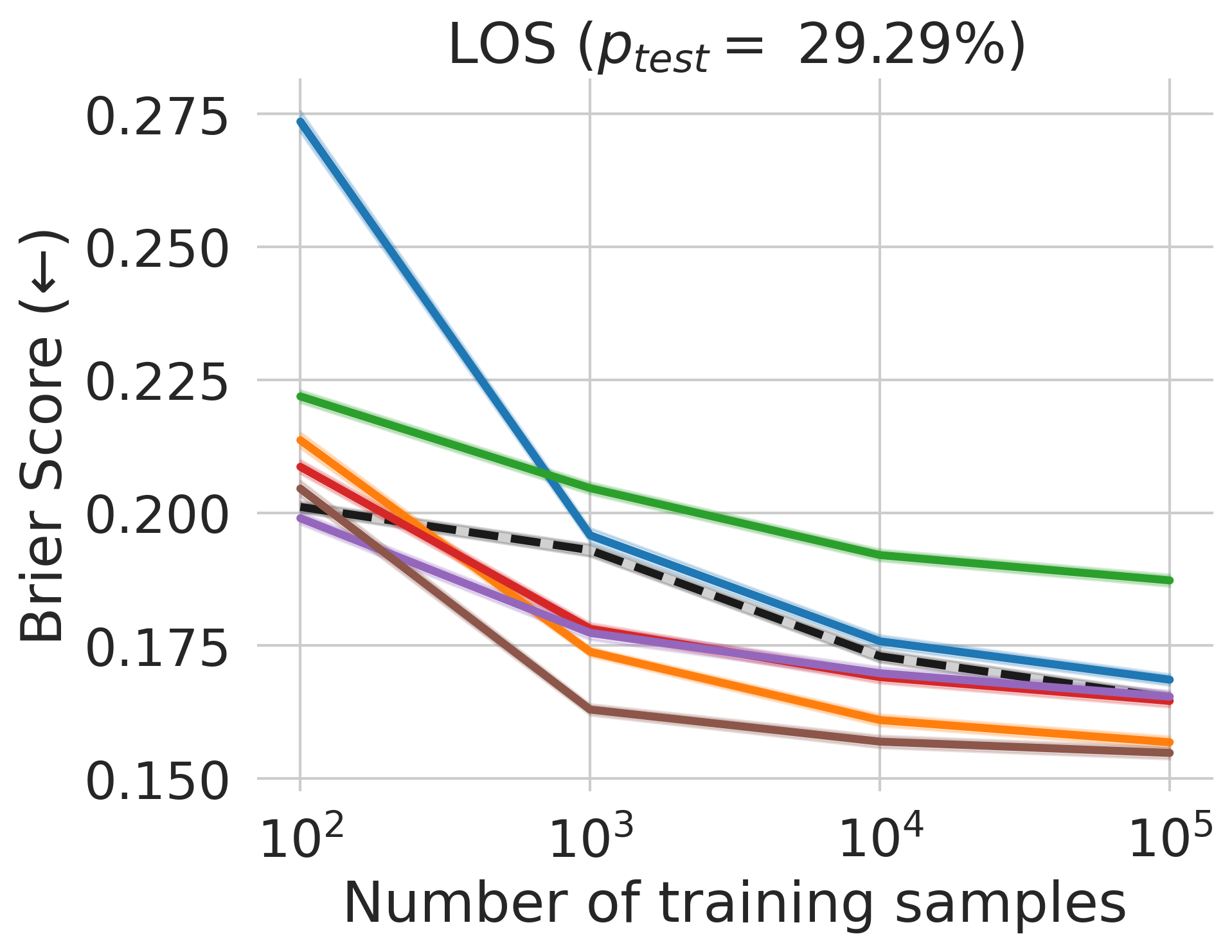}
    \includegraphics[height=100px]{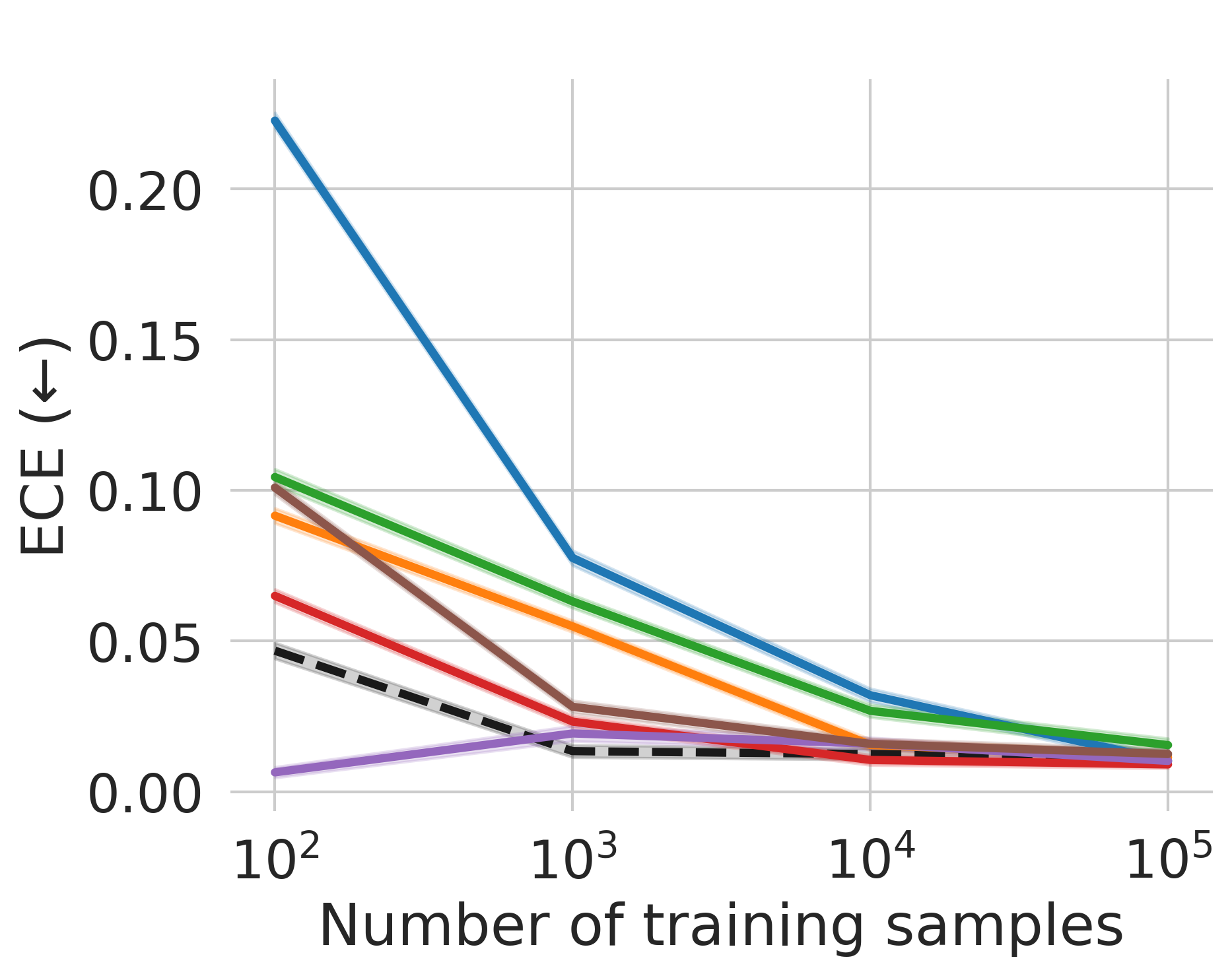}}

    \makebox[\textwidth][l]{
    \includegraphics[height=100px]{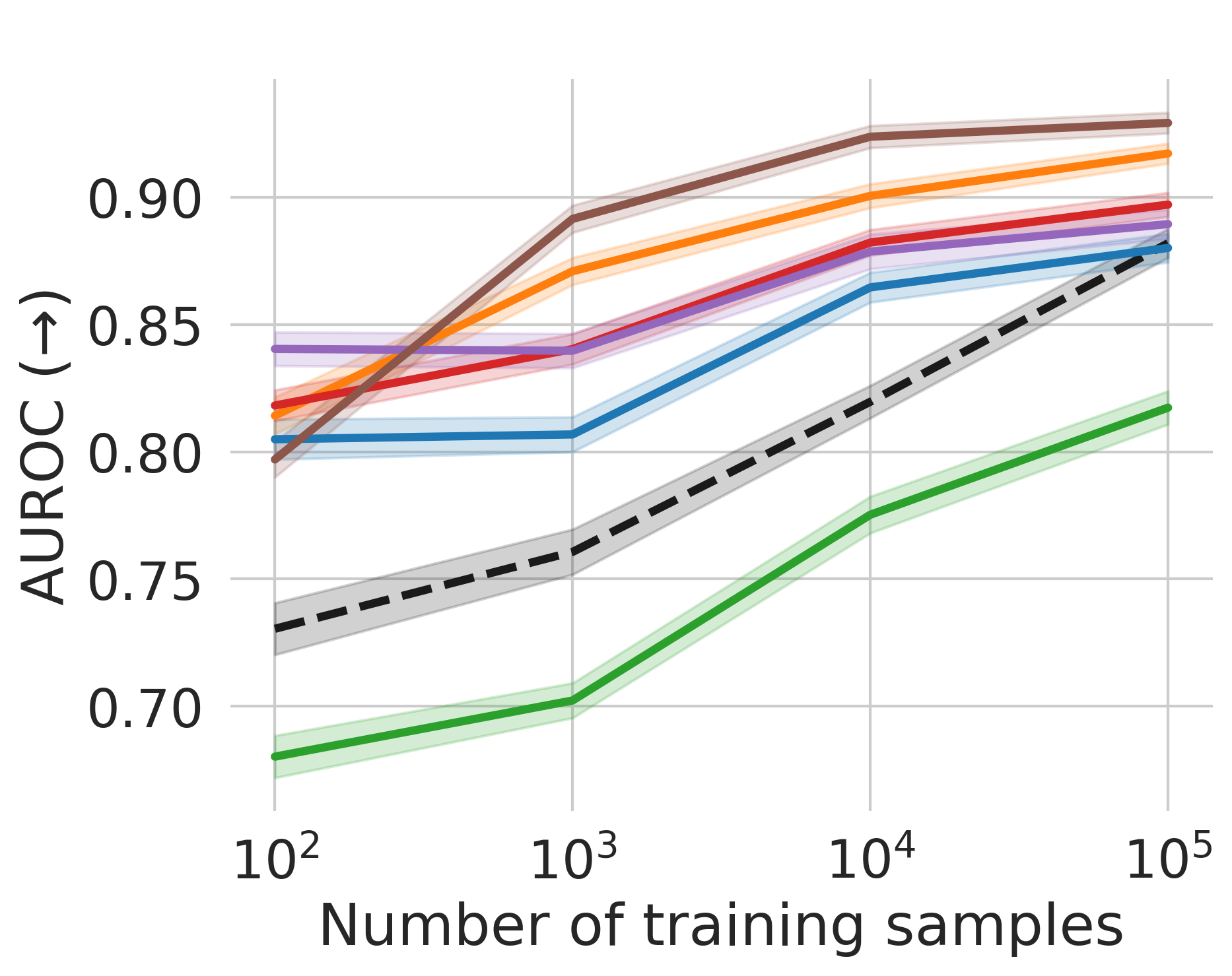}
    \includegraphics[height=100px]{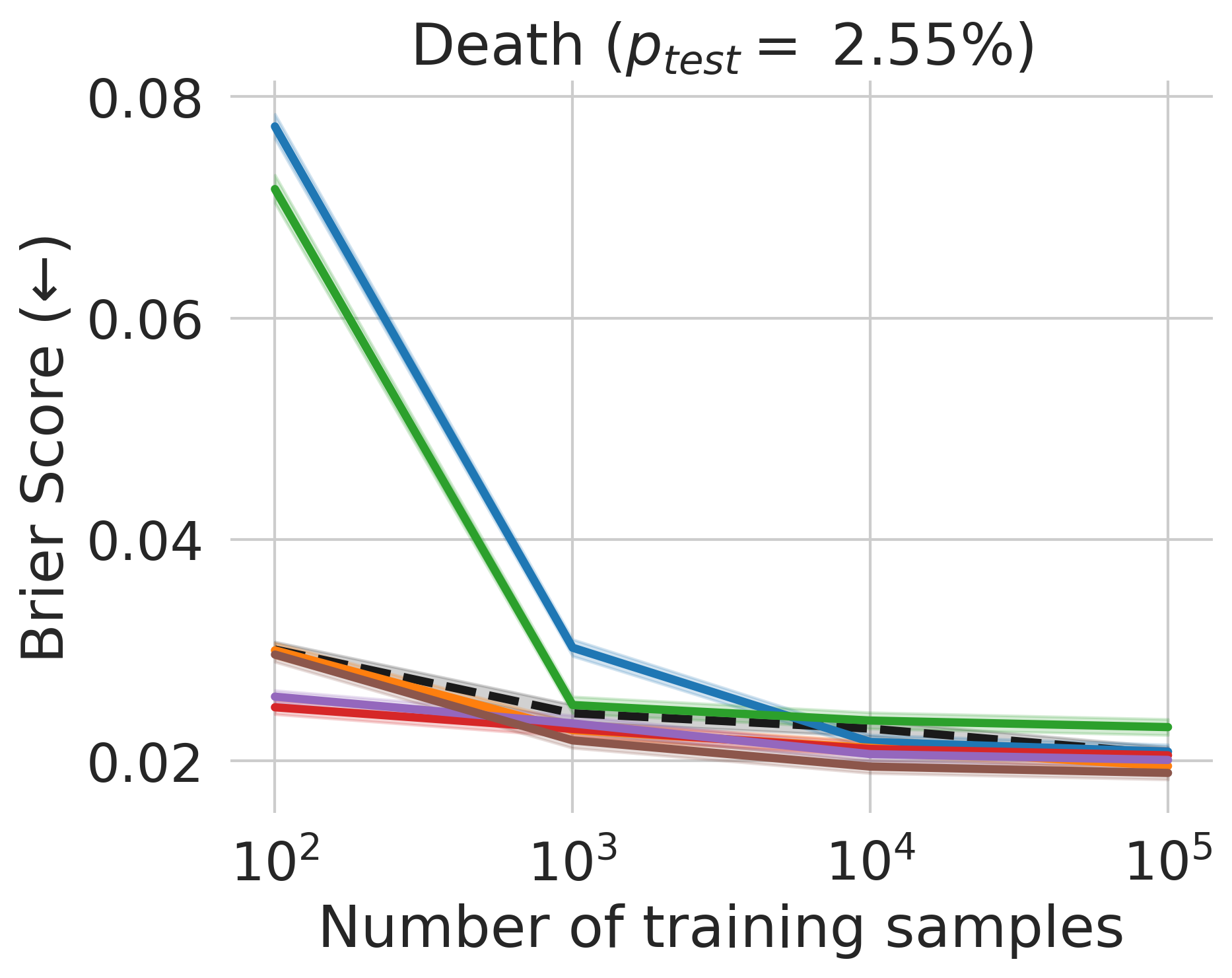}
    \includegraphics[height=100px]{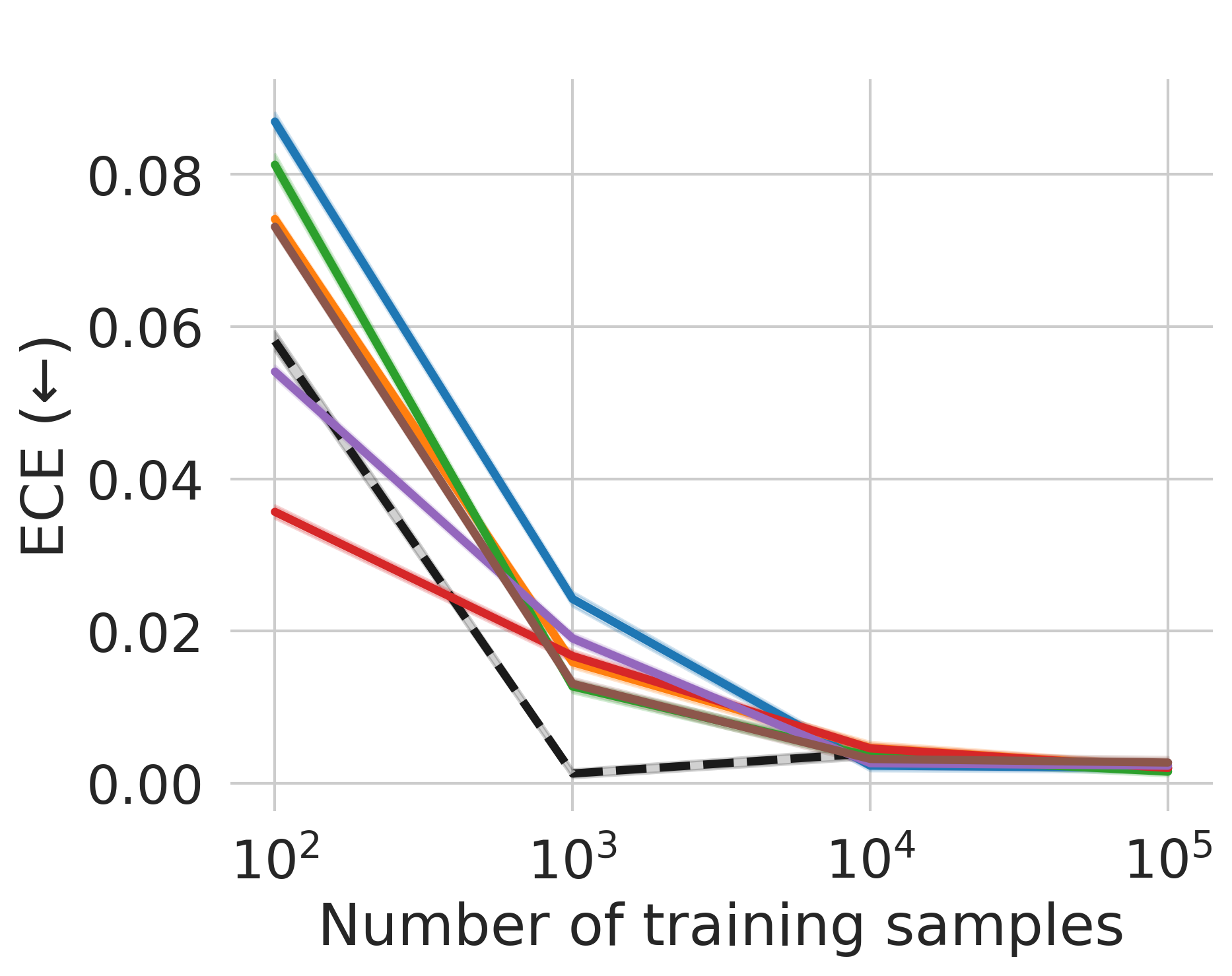}
    \includegraphics[height=100px]{figures/legend.png}}

    \makebox[\textwidth][l]{
    \includegraphics[height=100px]{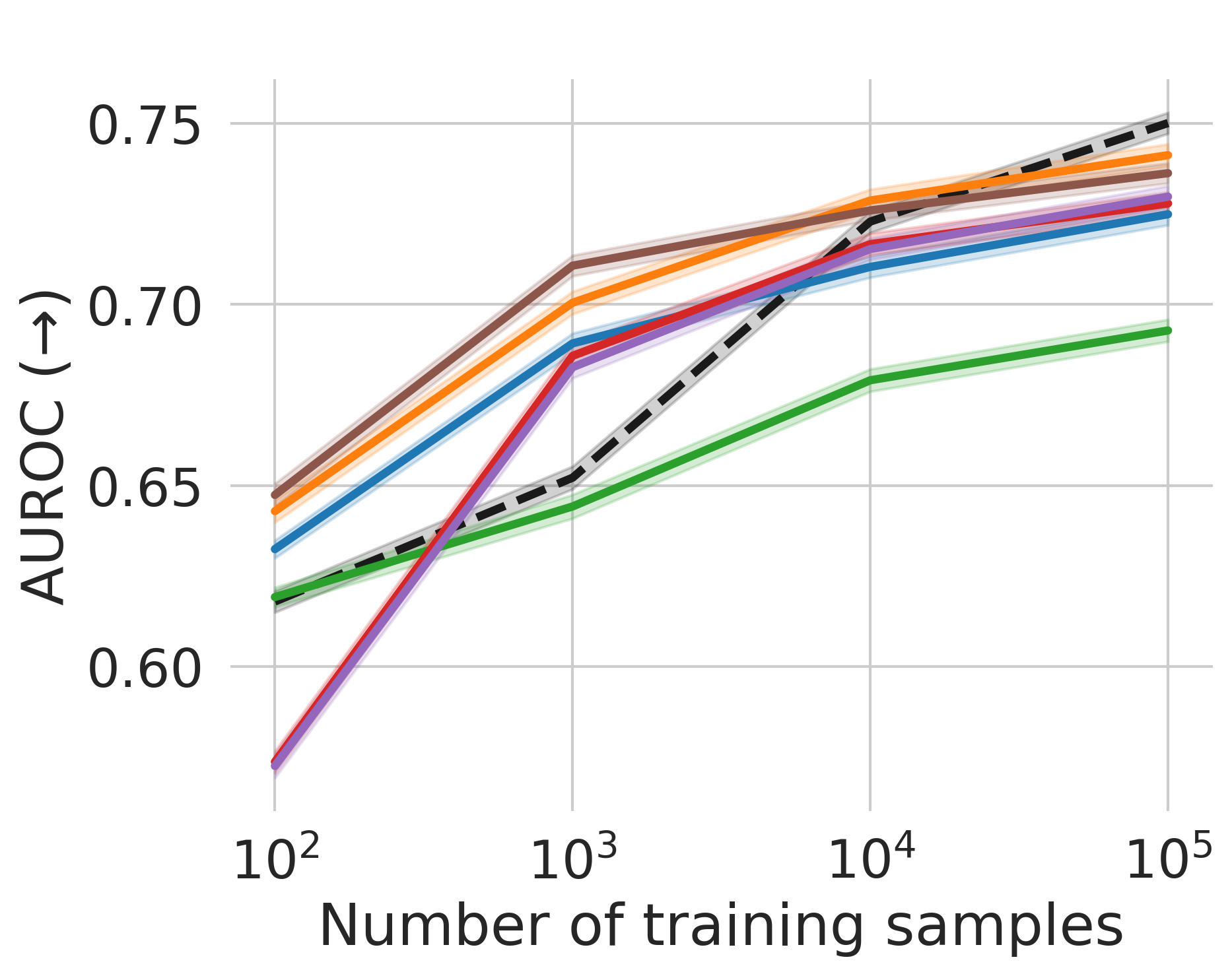}
    \includegraphics[height=100px]{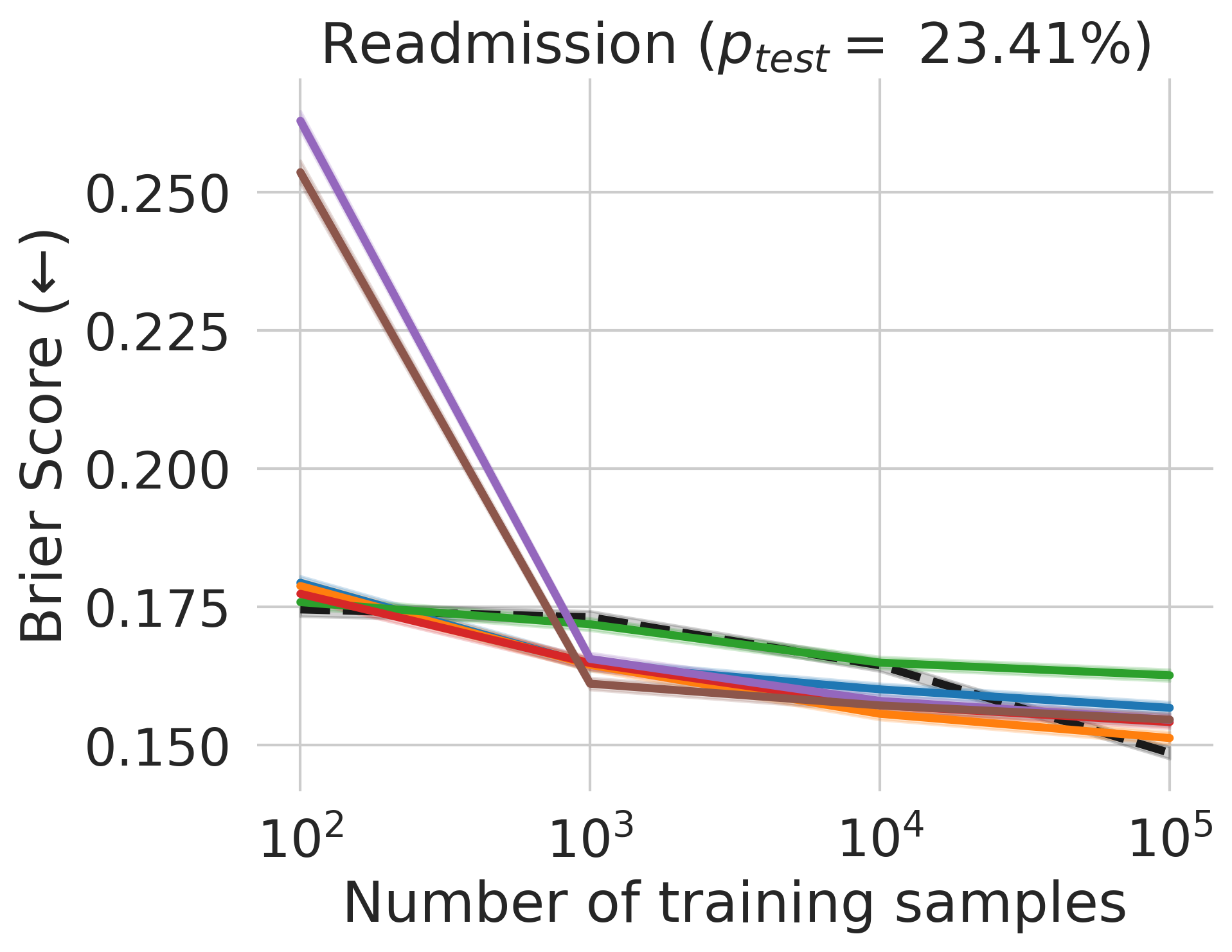}
    \includegraphics[height=100px]{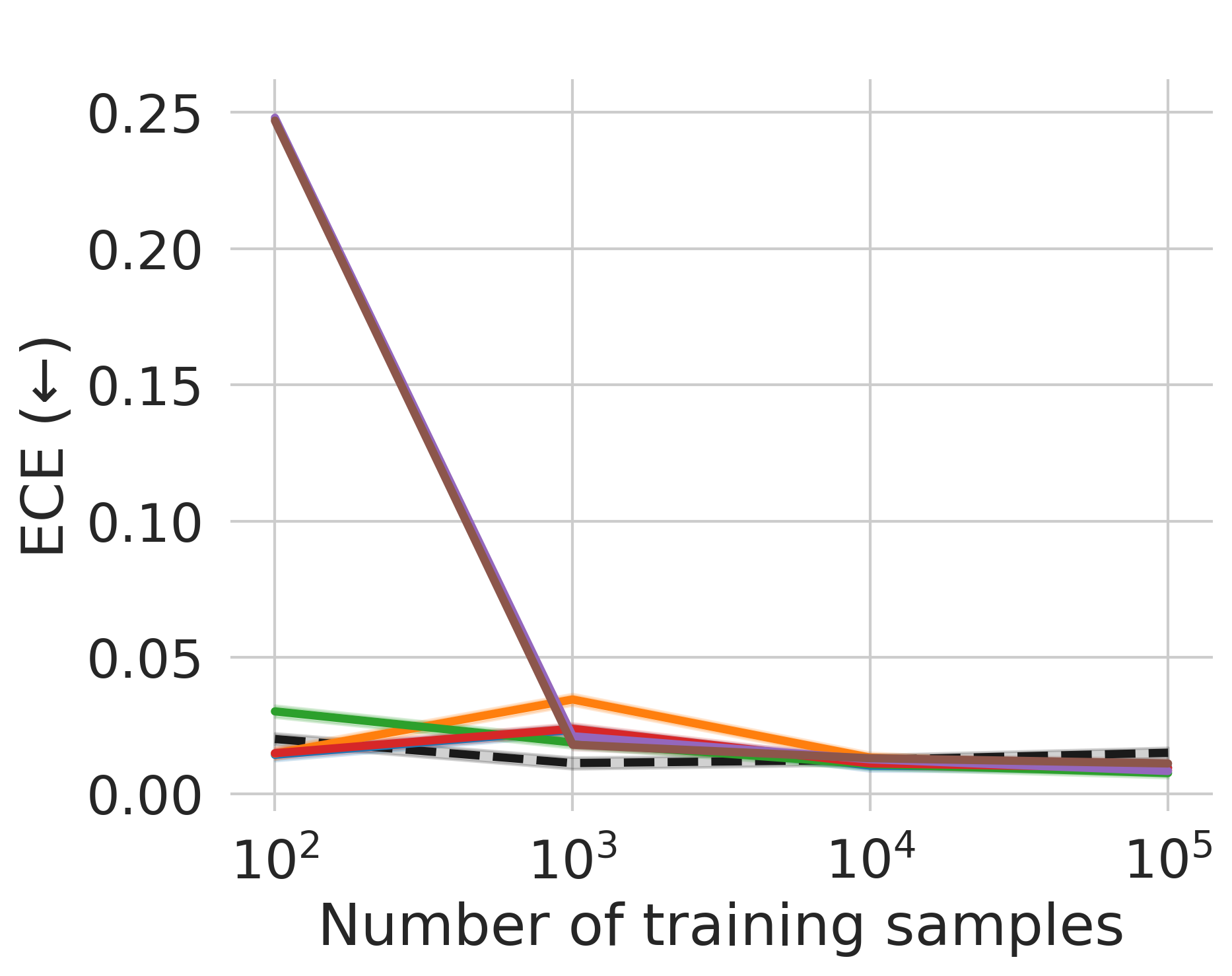}}

    \centering
    \caption{Discriminative performance and calibration results for all outcomes as the number of training points increases (Shaded area represents the bootstrapped standard deviation).}
    \label{fig:all_outcomes:mimic}
\end{figure}

\begin{figure}[ht]
    \makebox[\textwidth][l]{
    \includegraphics[height=100px]{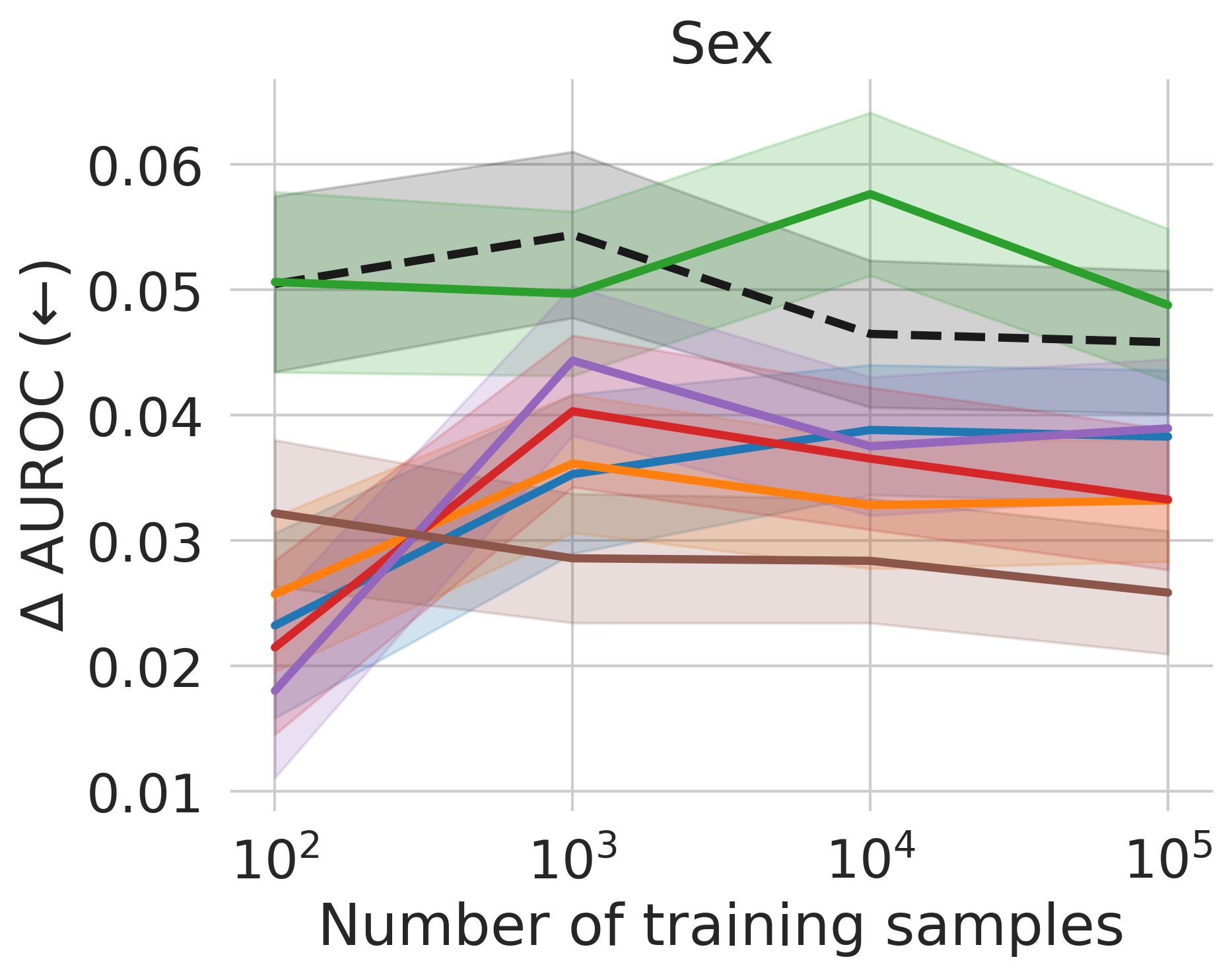}
    \includegraphics[height=100px]{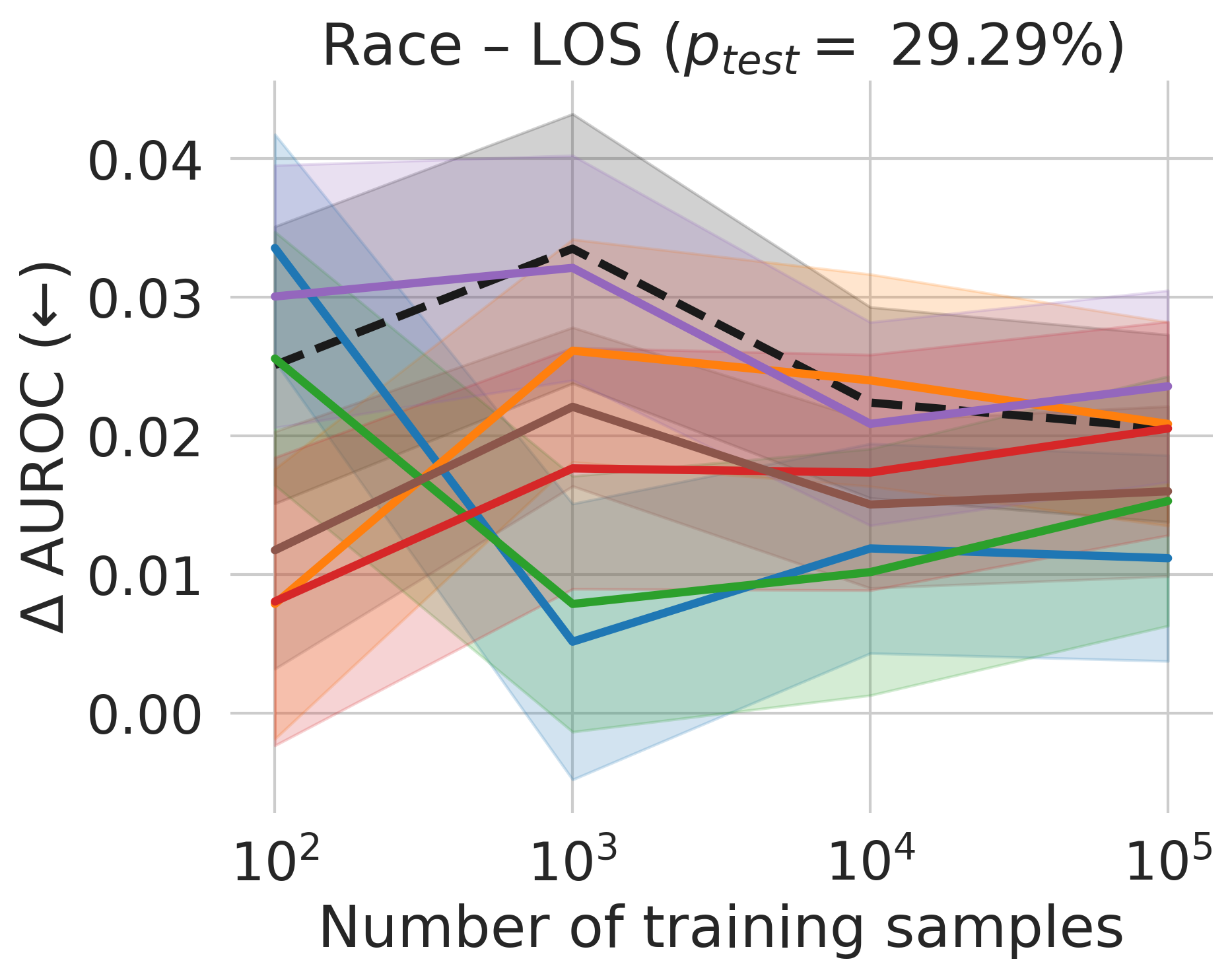}}

    \makebox[\textwidth][l]{
    \includegraphics[height=100px]{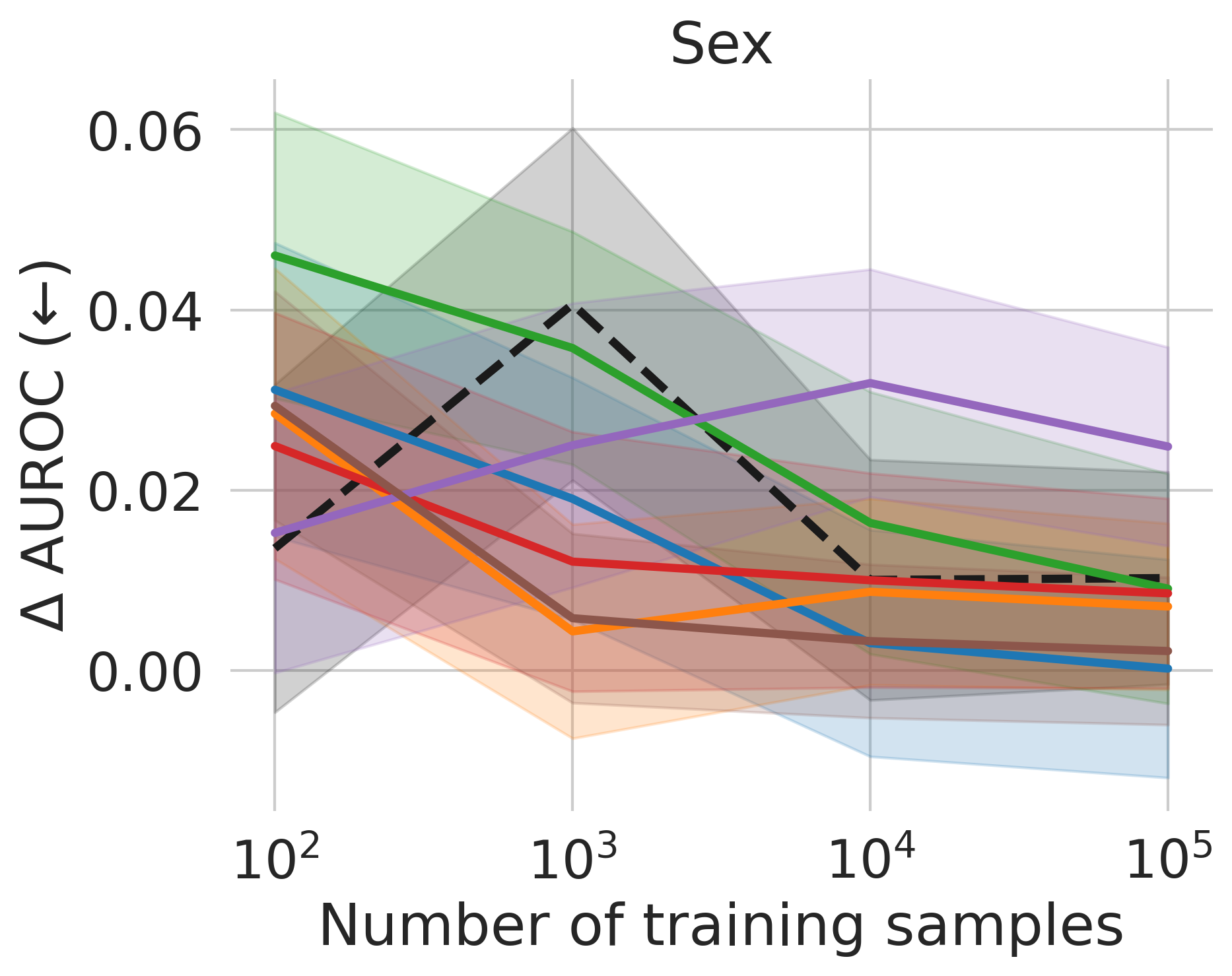}
    \includegraphics[height=100px]{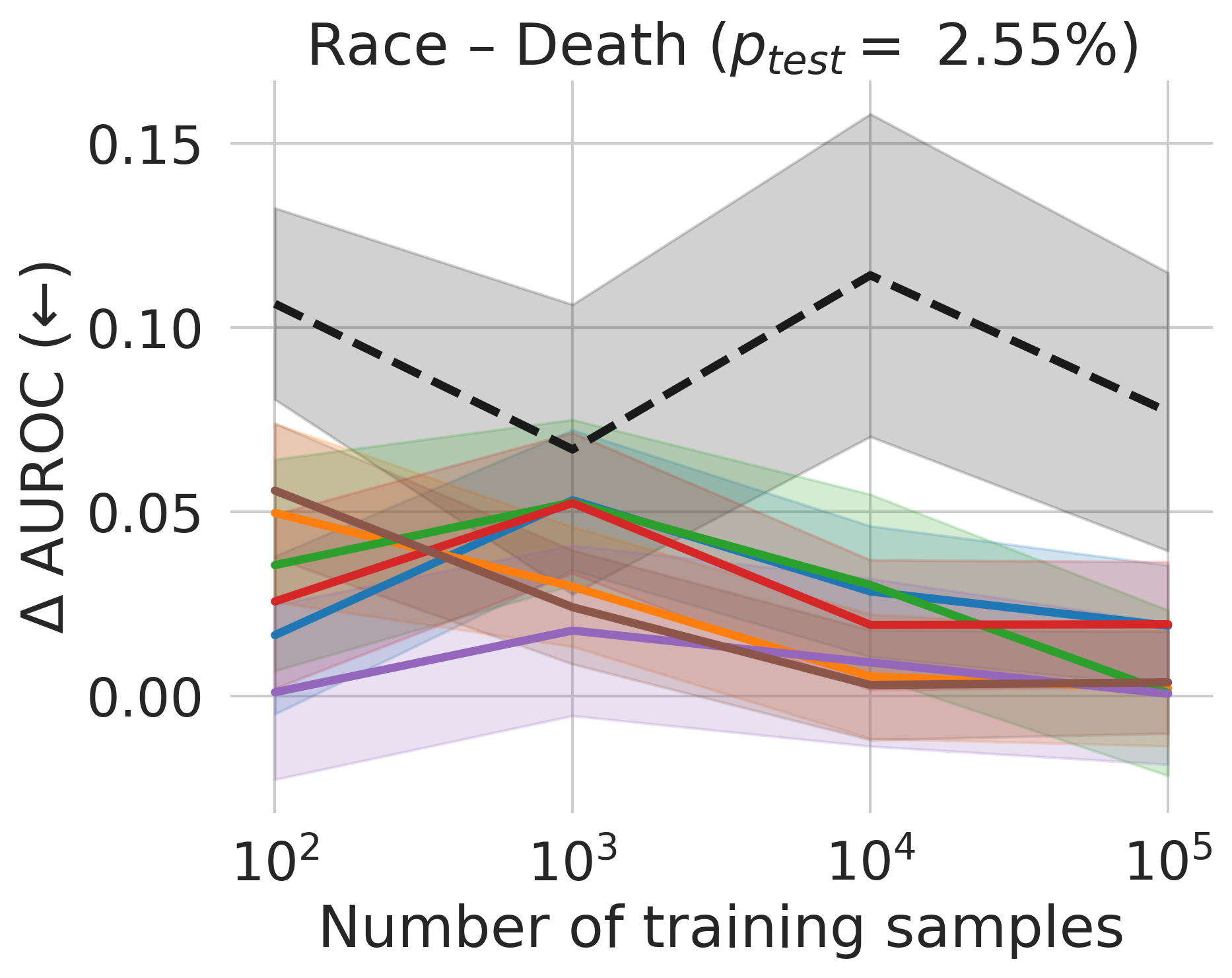}
    \includegraphics[height=100px]{figures/legend.png}}

    \makebox[\textwidth][l]{
    \includegraphics[height=100px]{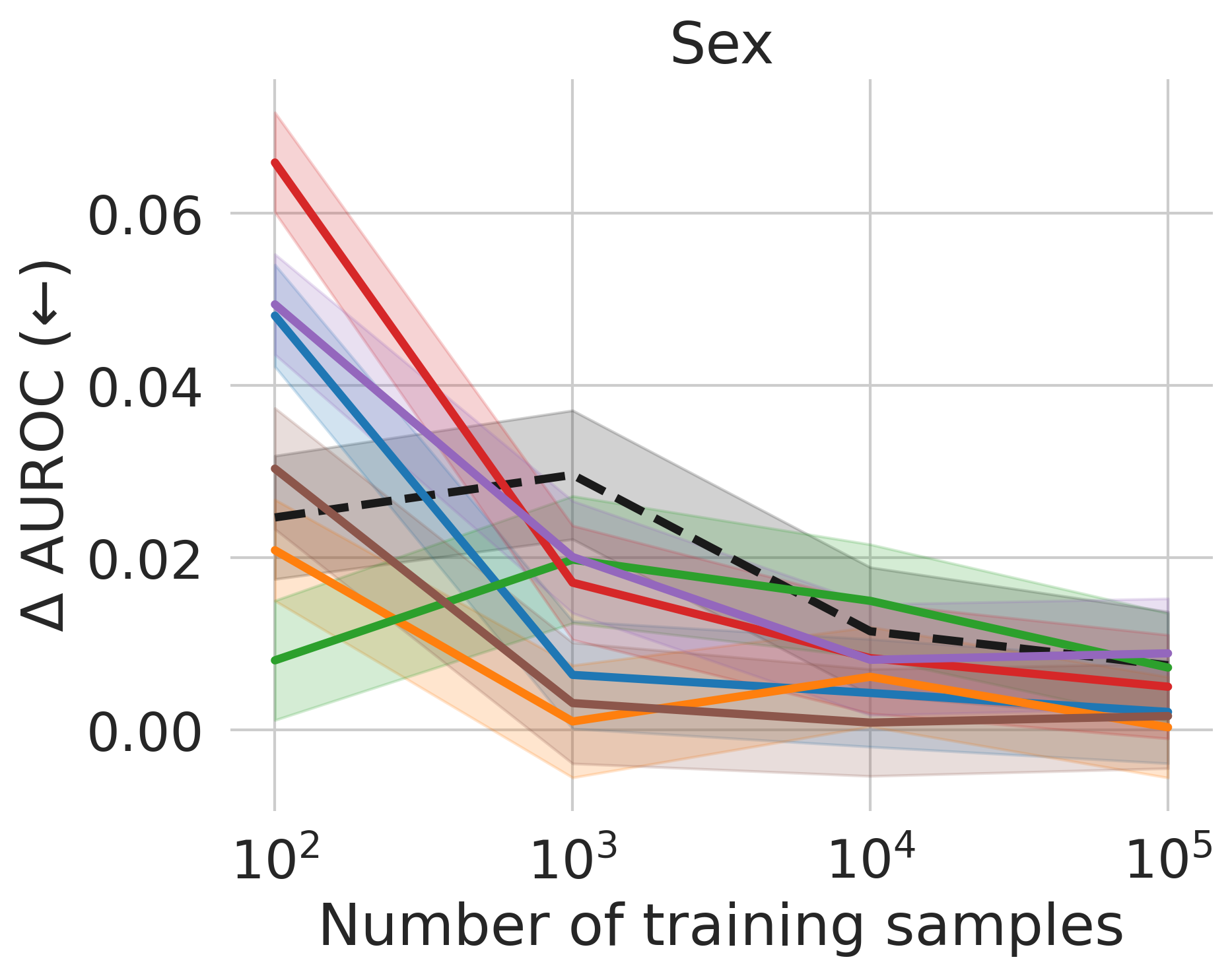}
    \includegraphics[height=100px]{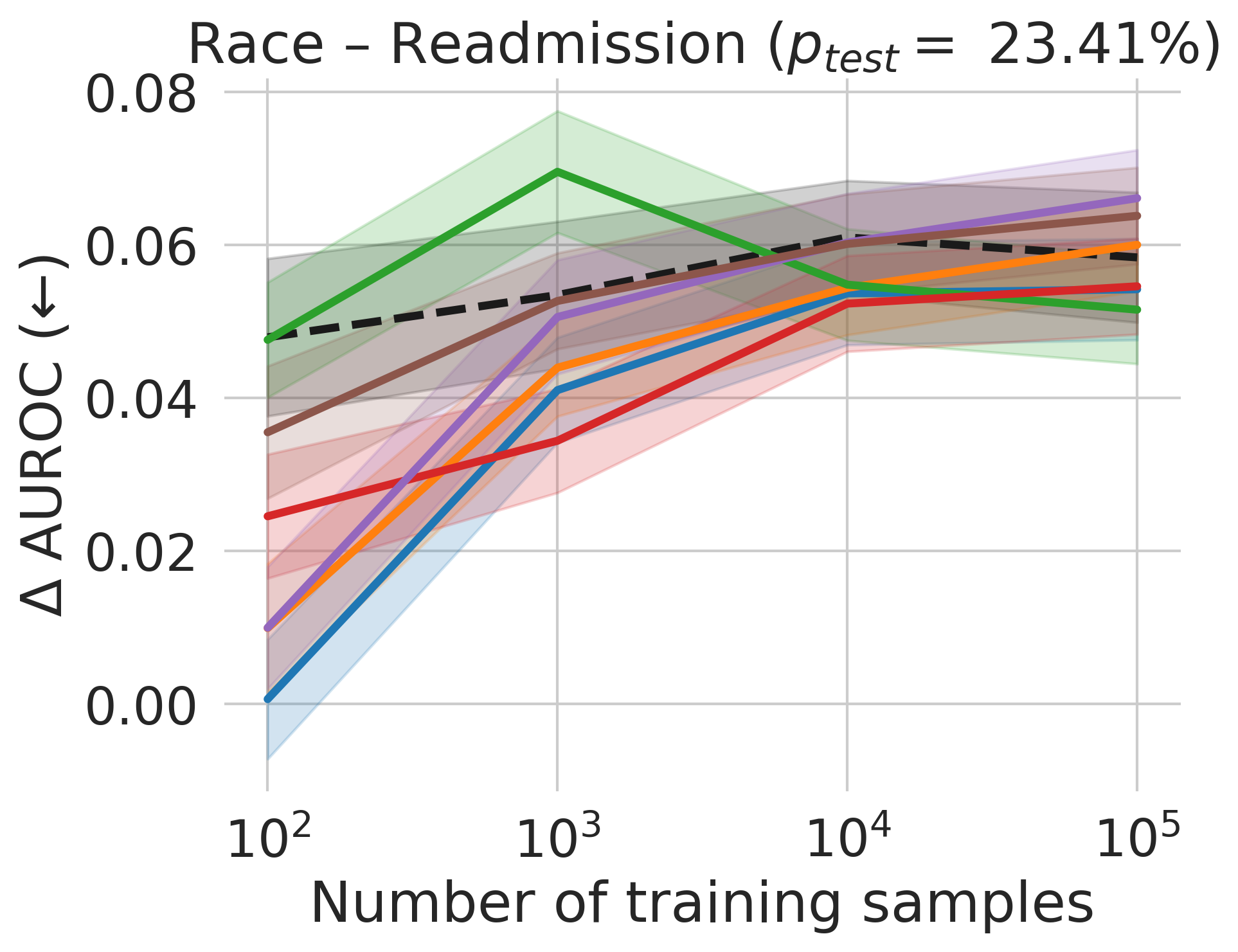}
    }

    \centering
    \caption{Discriminative performances and calibration results for all outcomes for an increasing number of training points (Shaded area represents bootstrapped standard deviation).}
    \label{fig:all_outcomes:mimic:fairness}
\end{figure}

\begin{figure}[ht]
    \makebox[\textwidth][l]{
    \includegraphics[height=100px]{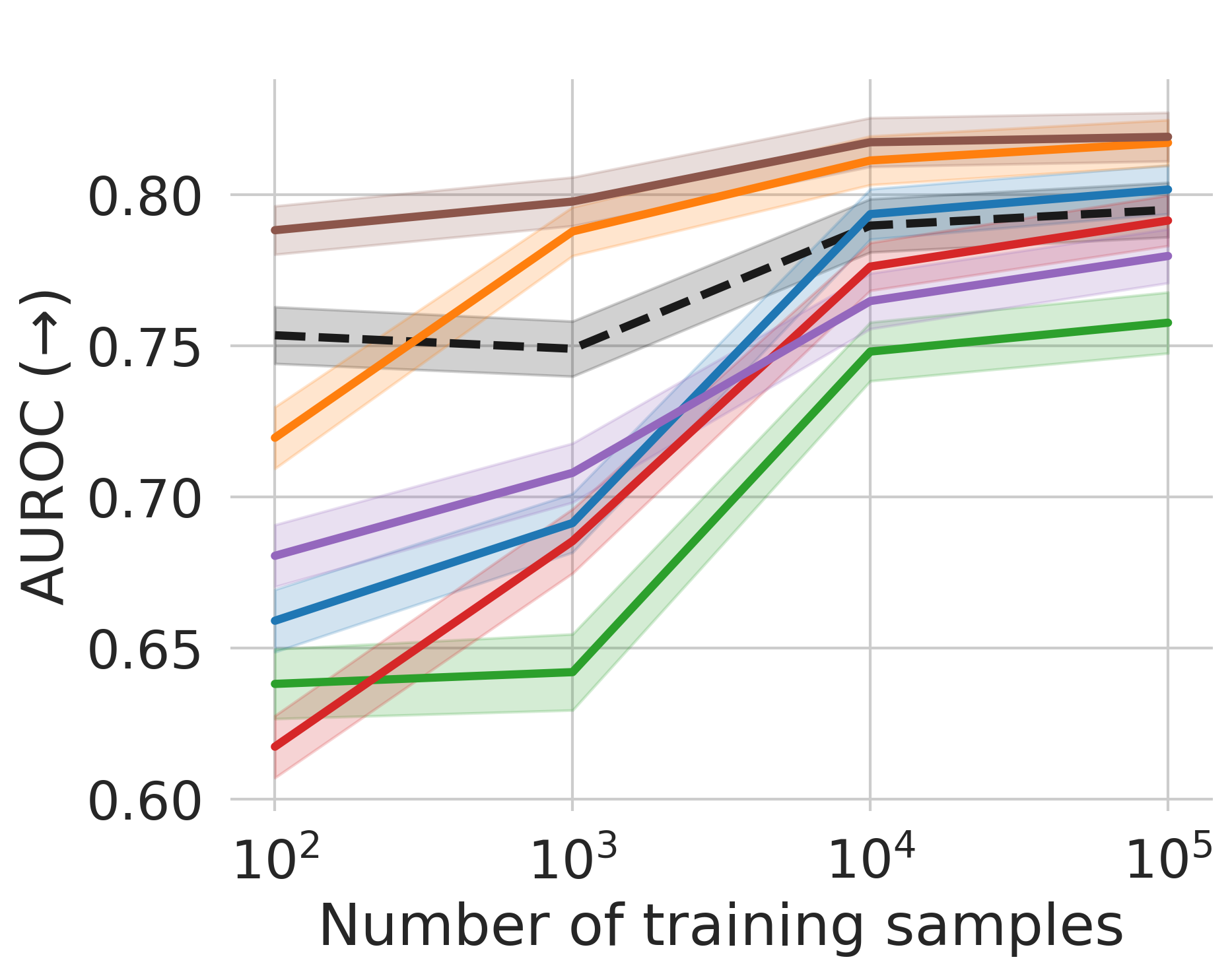}   
    \includegraphics[height=100px]{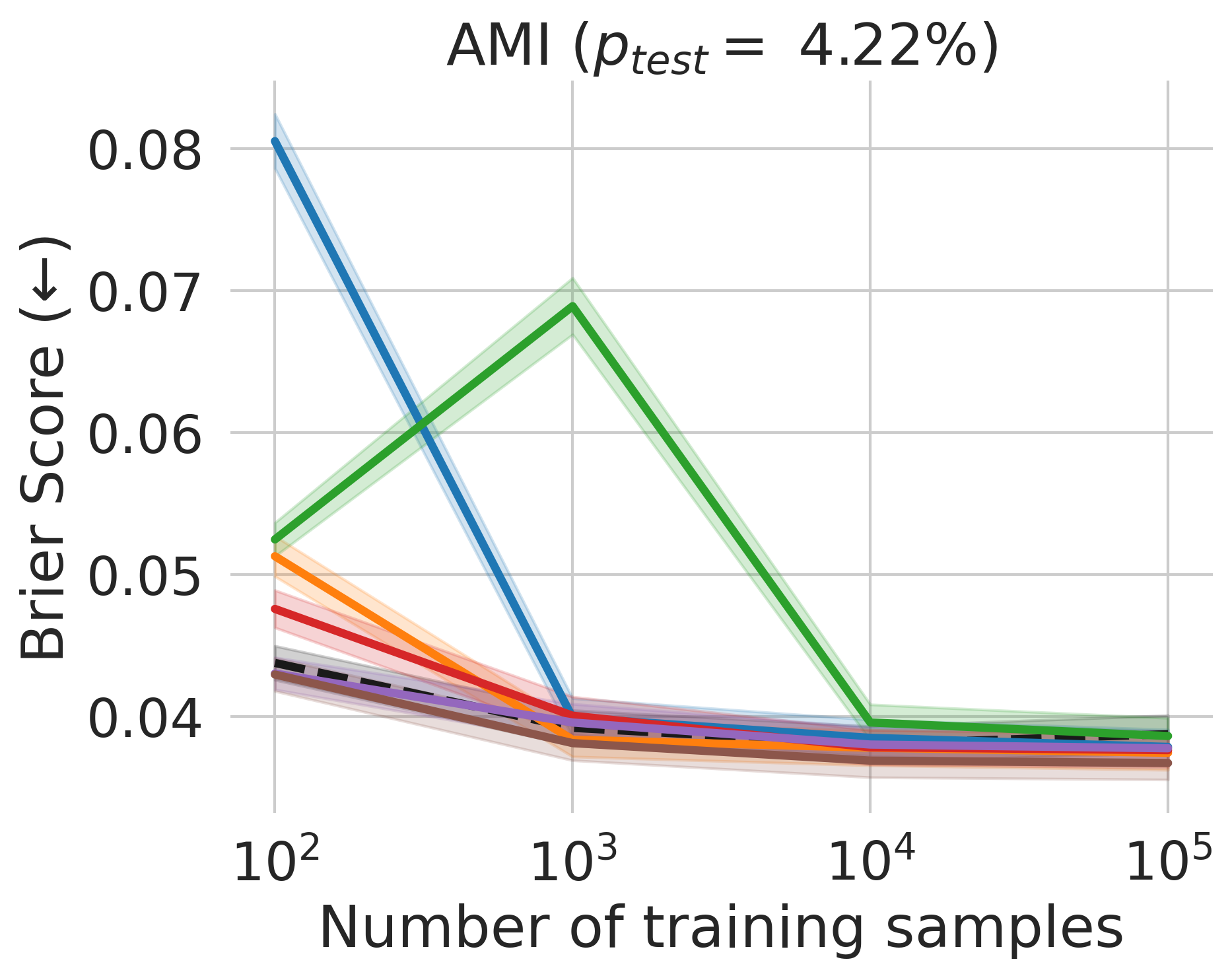}
    \includegraphics[height=100px]{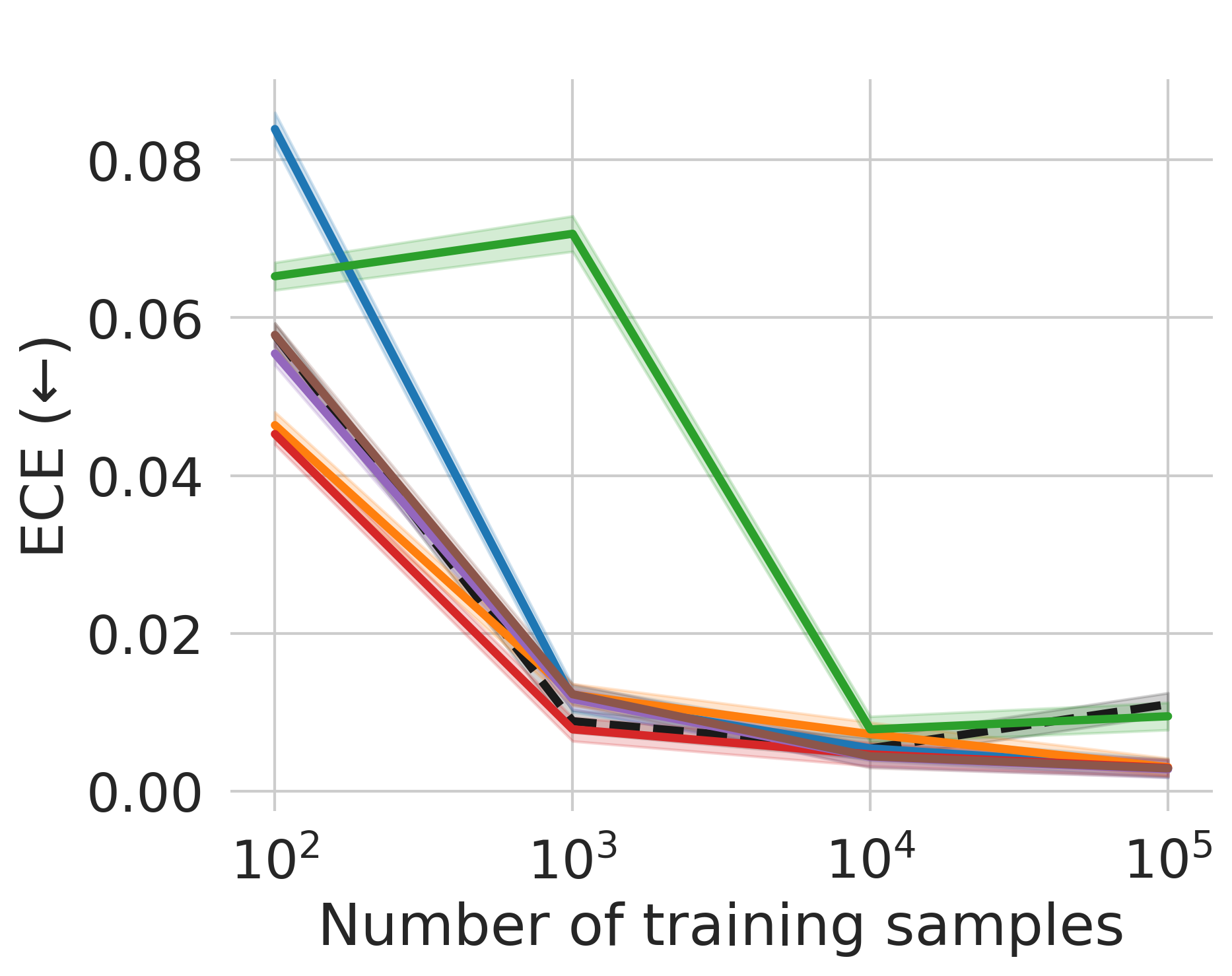}}
    
    \makebox[\textwidth][l]{
    \includegraphics[height=100px]{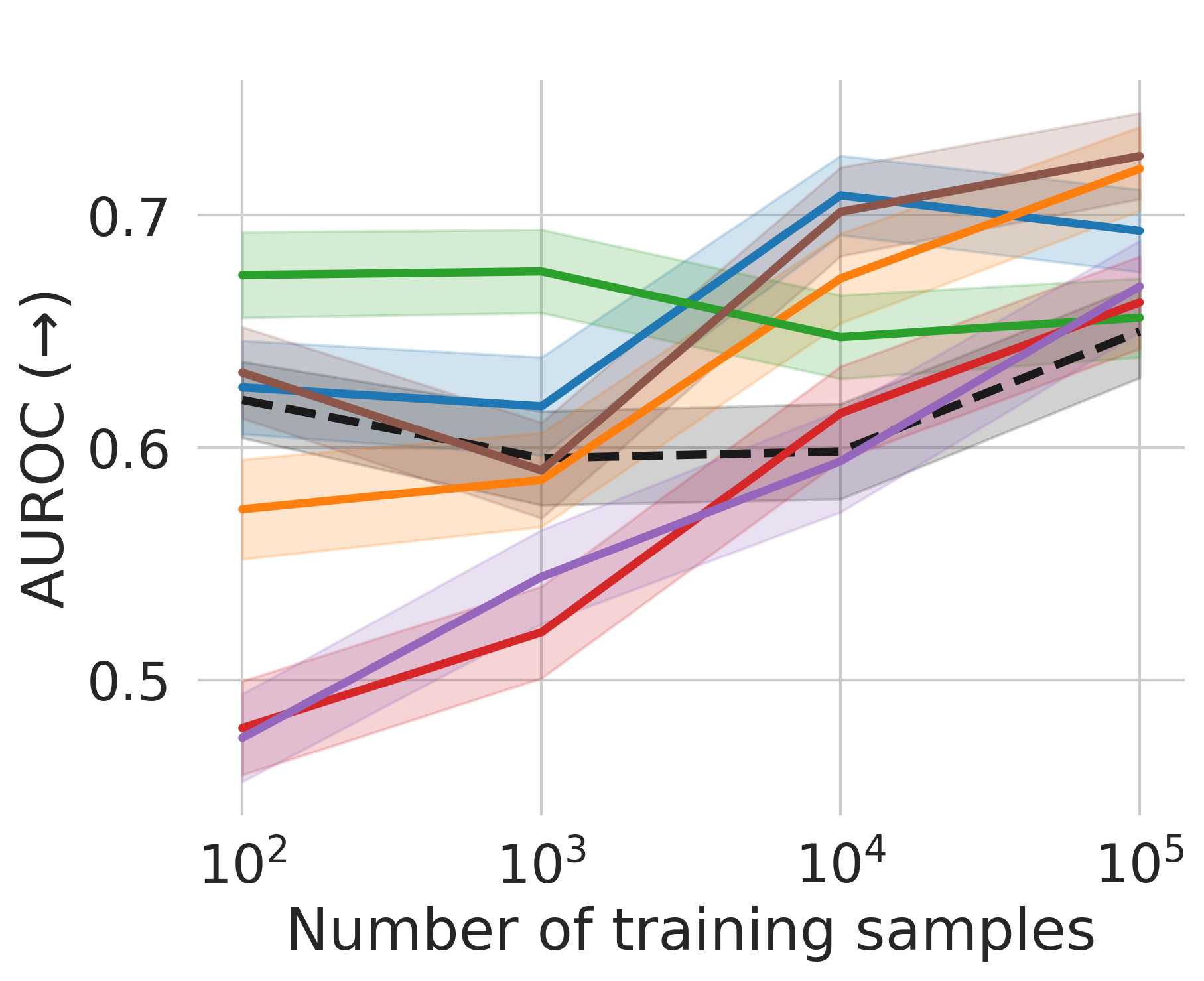}
    \includegraphics[height=100px]{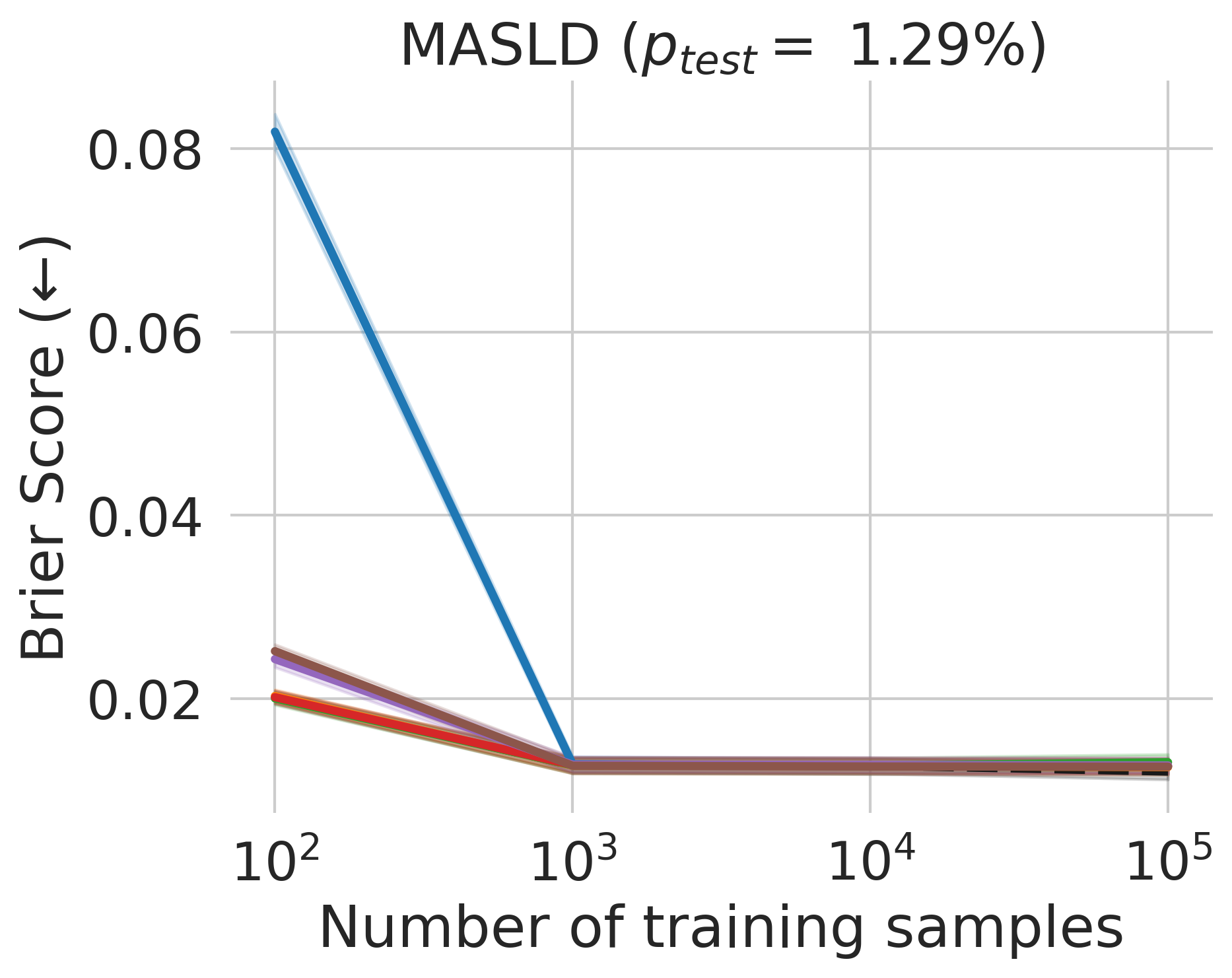}
    \includegraphics[height=100px]{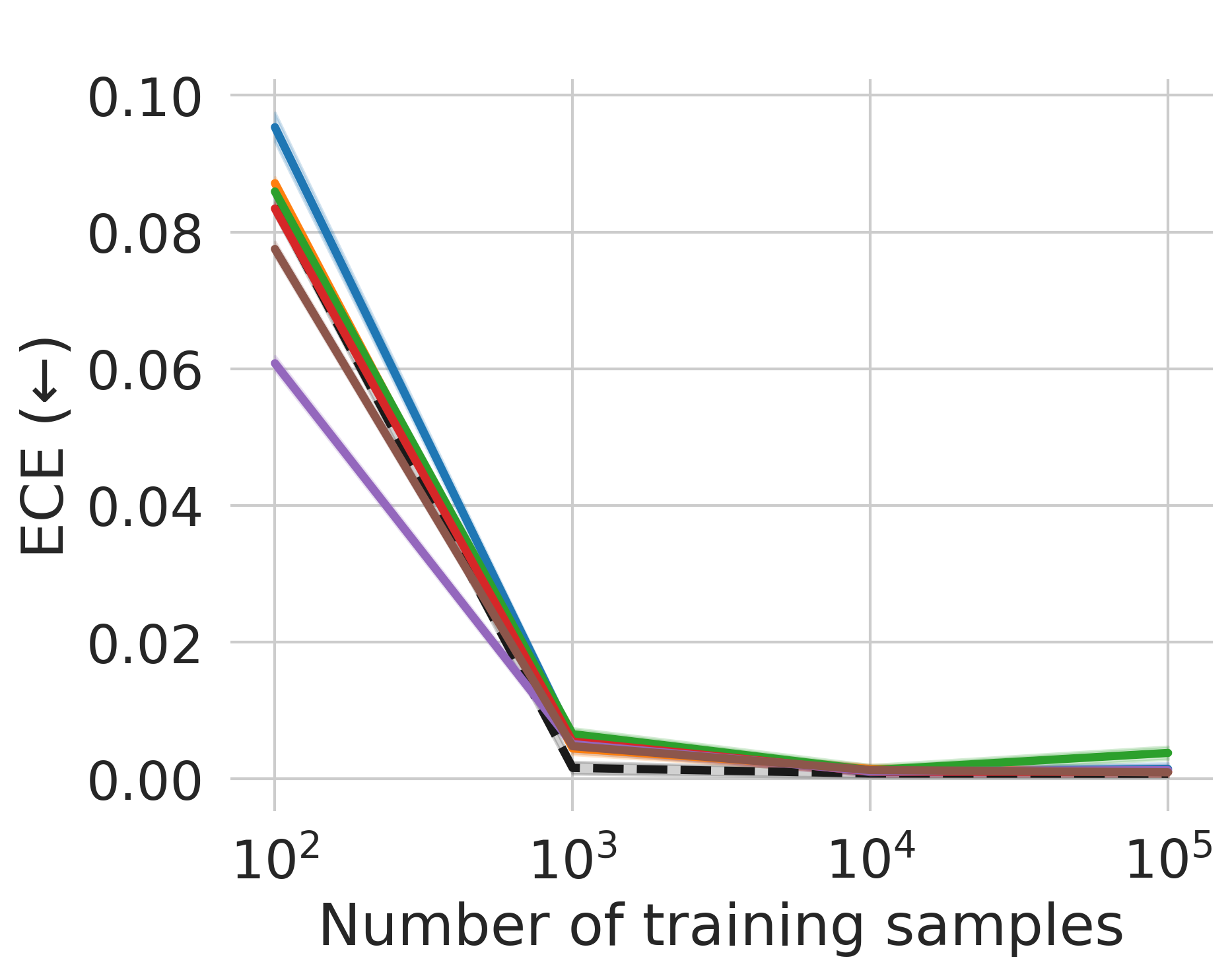}
    \includegraphics[height=100px]{figures/legend.png}}

    \makebox[\textwidth][l]{
    \includegraphics[height=100px]{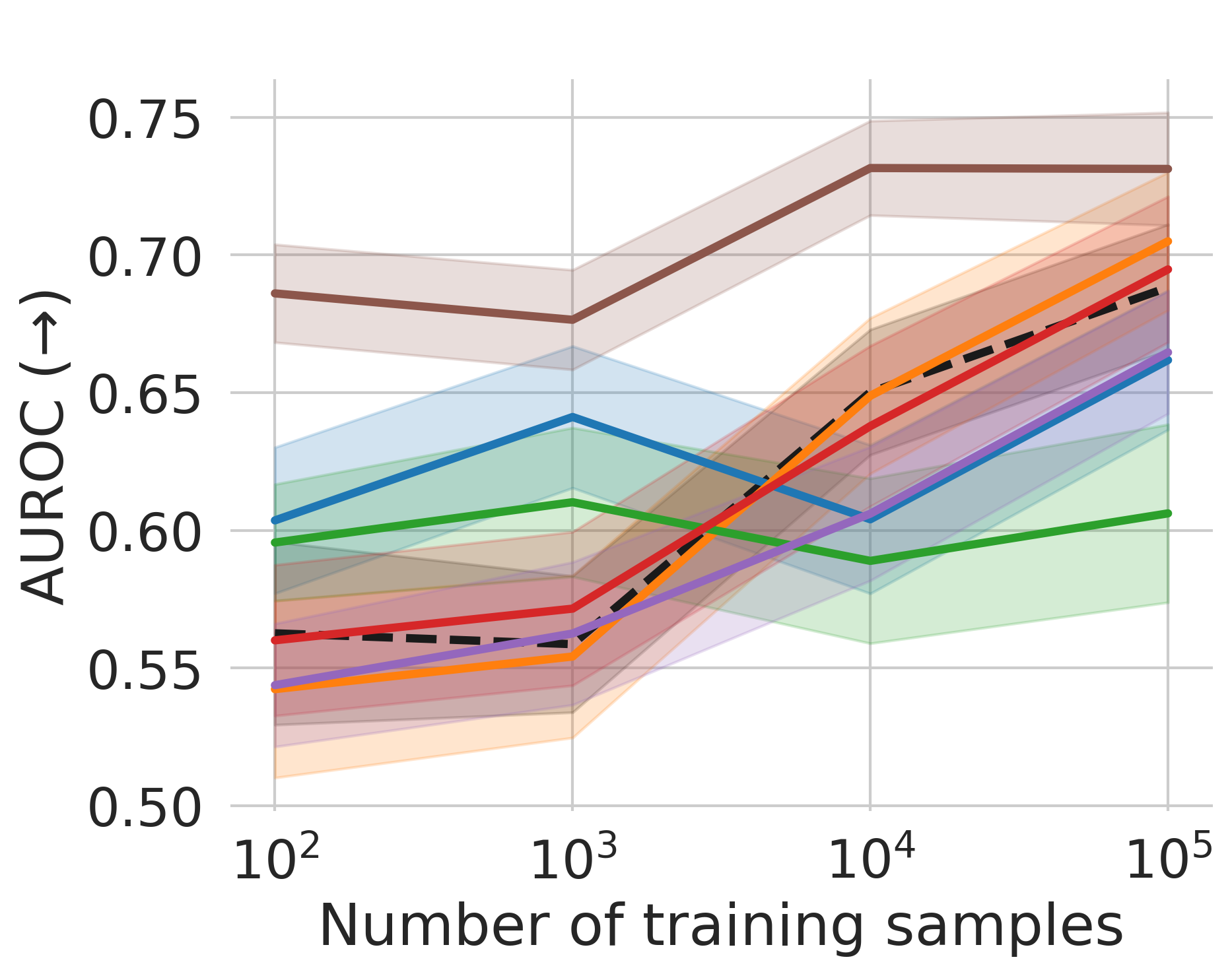}
    \includegraphics[height=100px]{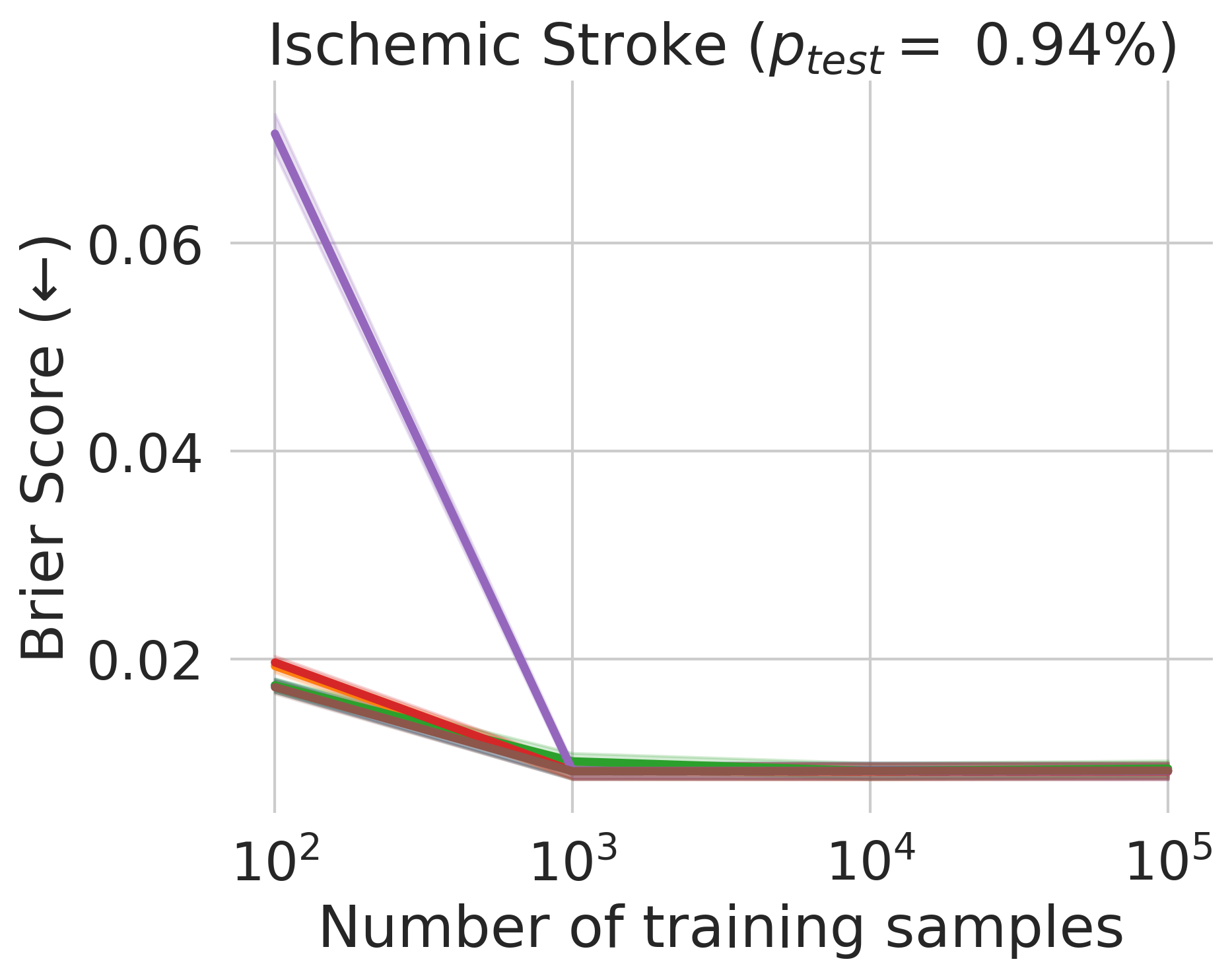}
    \includegraphics[height=100px]{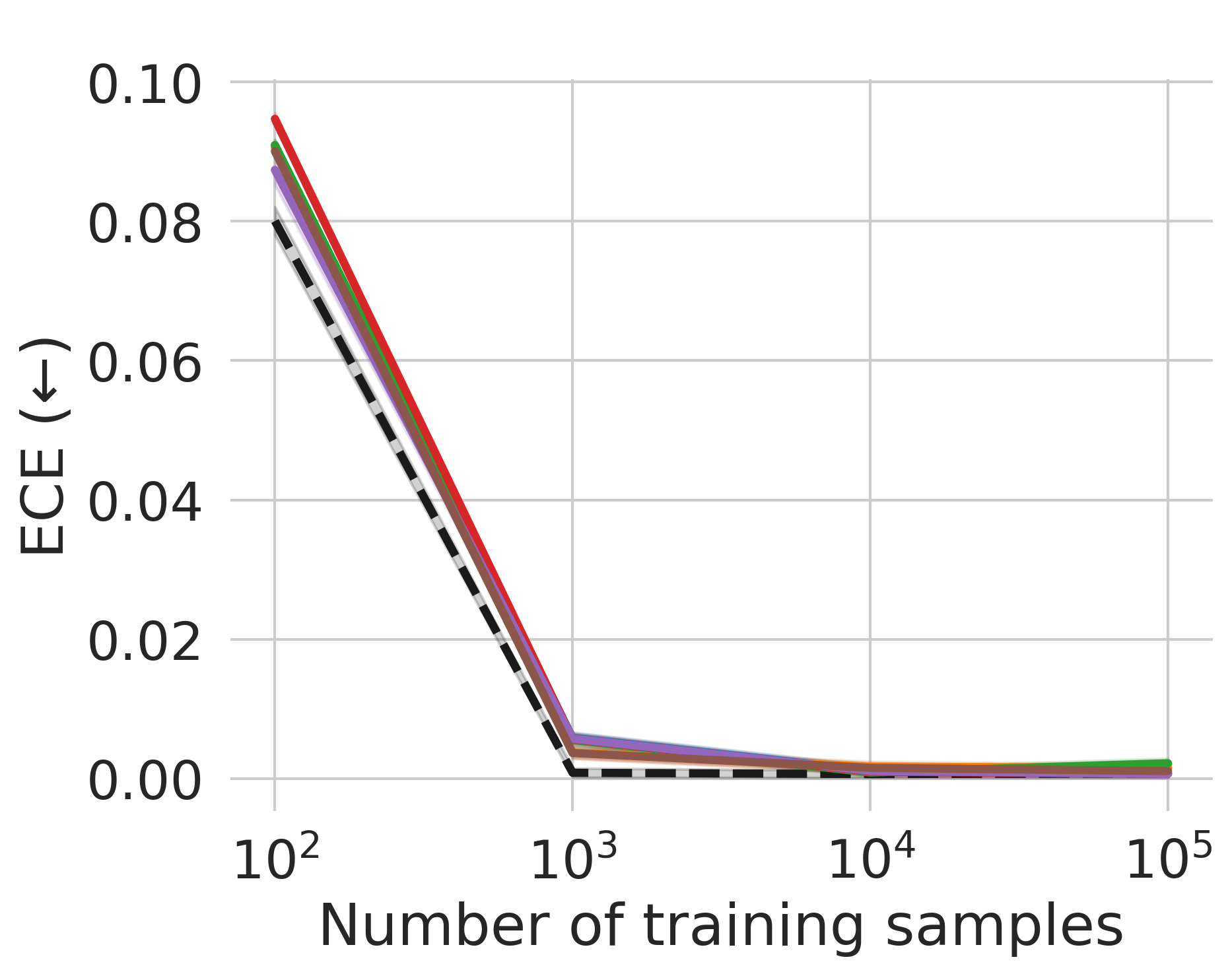}}

    \centering
    \caption{Discriminative performances and calibration results for all phenotypes for an increasing number of training points (Shaded area represents bootstrapped standard deviation).}
    \label{fig:all_phenotypes:mimic}
\end{figure}

\begin{figure}[ht]
    \makebox[\textwidth][l]{
    \includegraphics[height=100px]{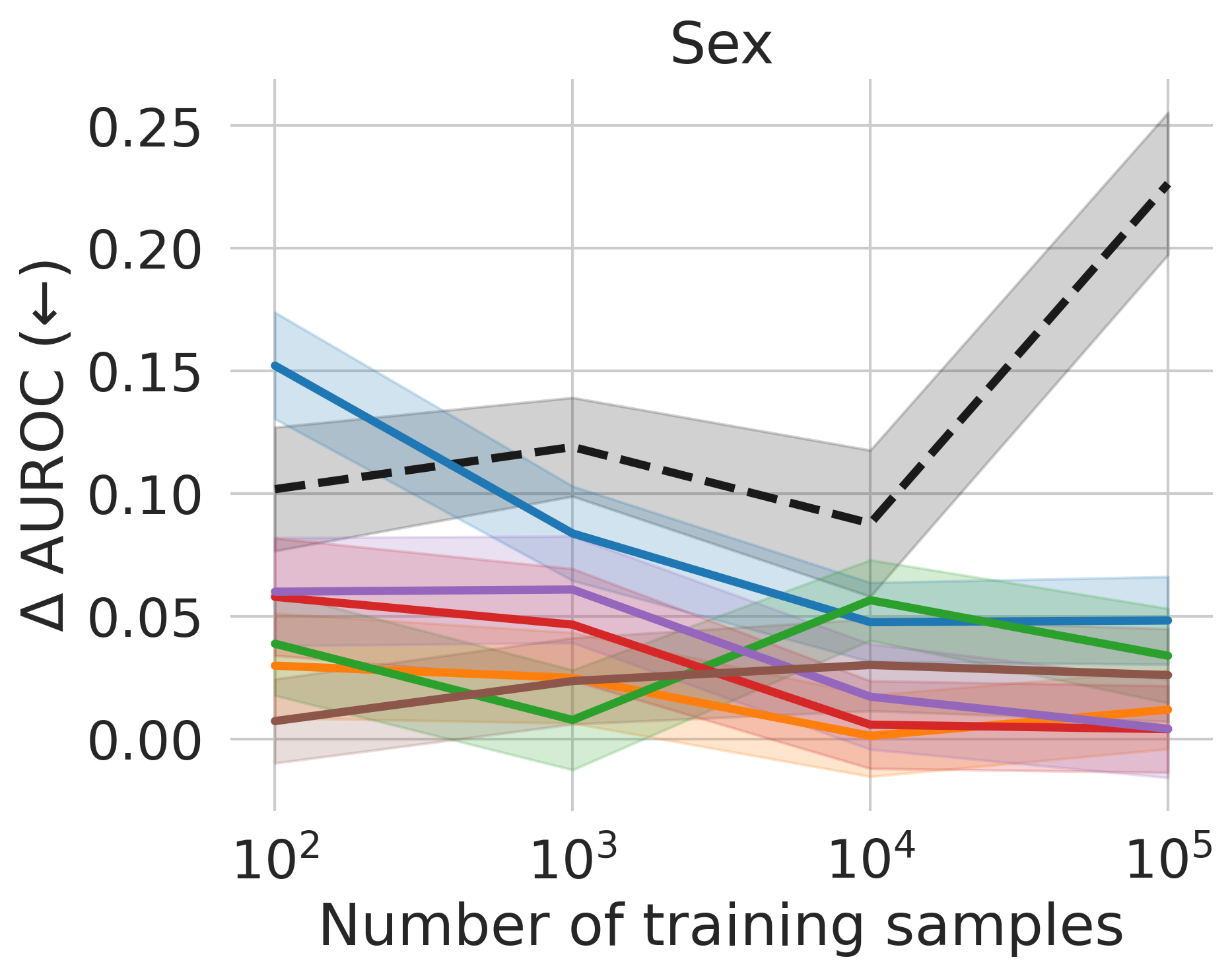}
    \includegraphics[height=100px]{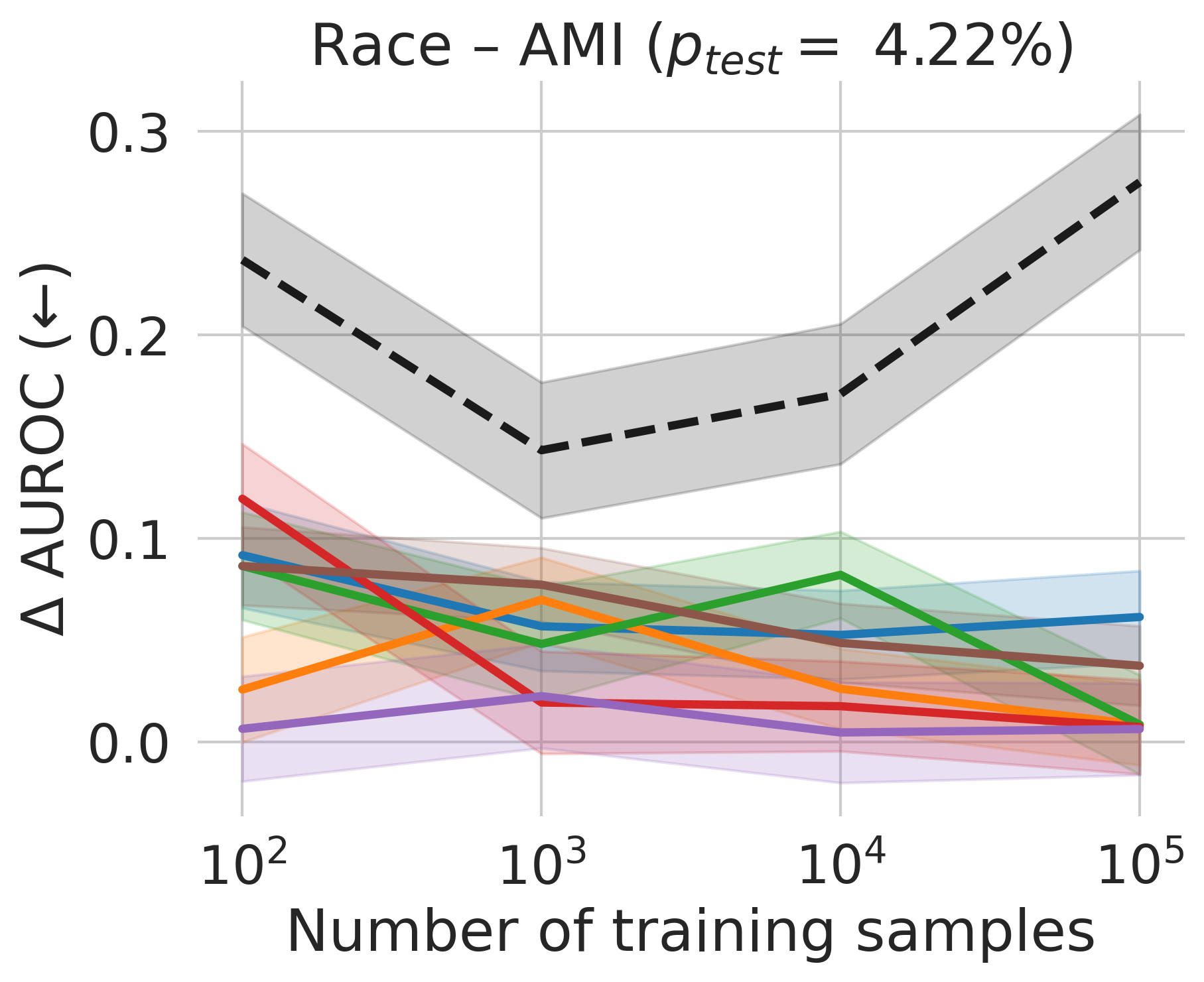}}

    \makebox[\textwidth][l]{
    \includegraphics[height=100px]{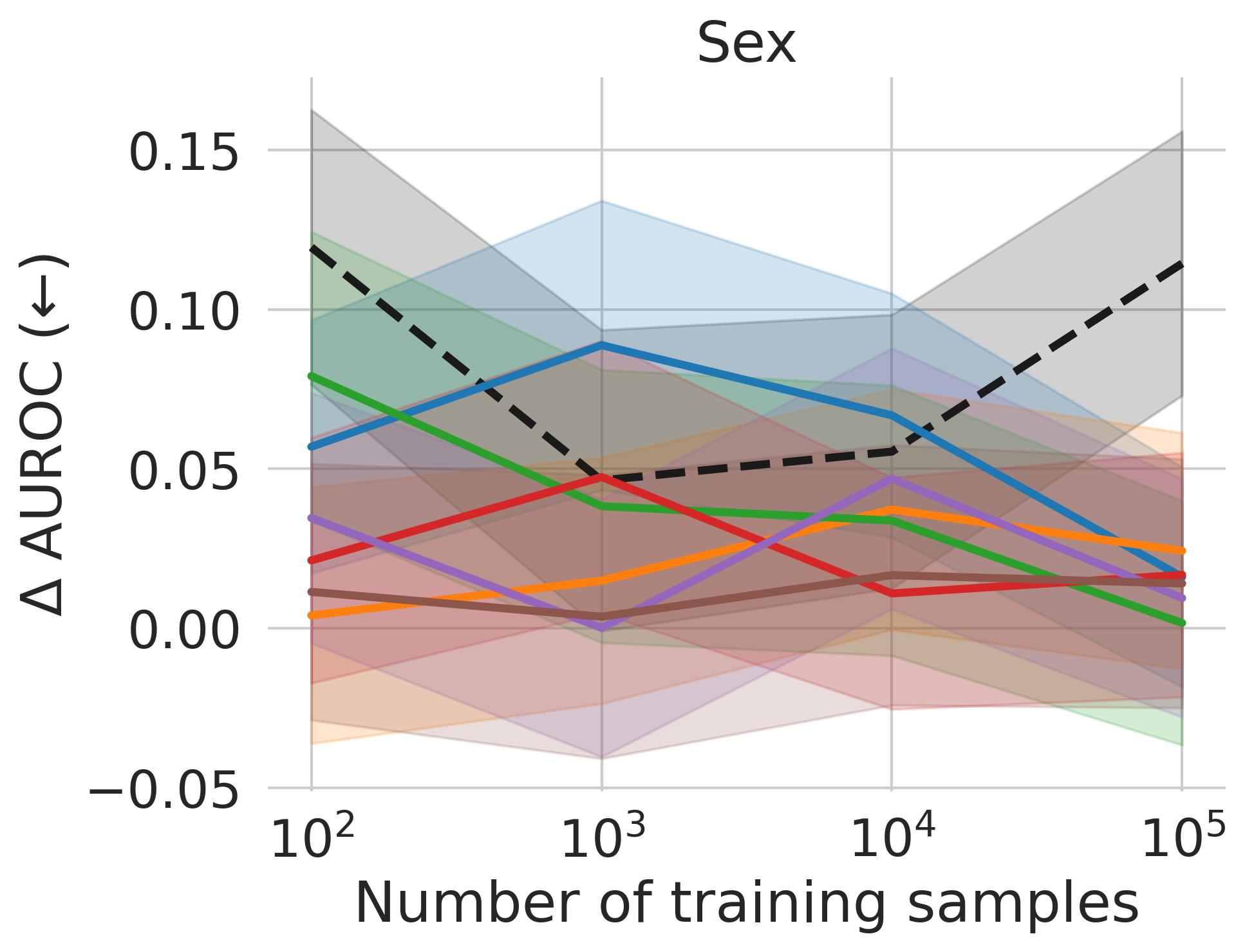}
    \includegraphics[height=100px]{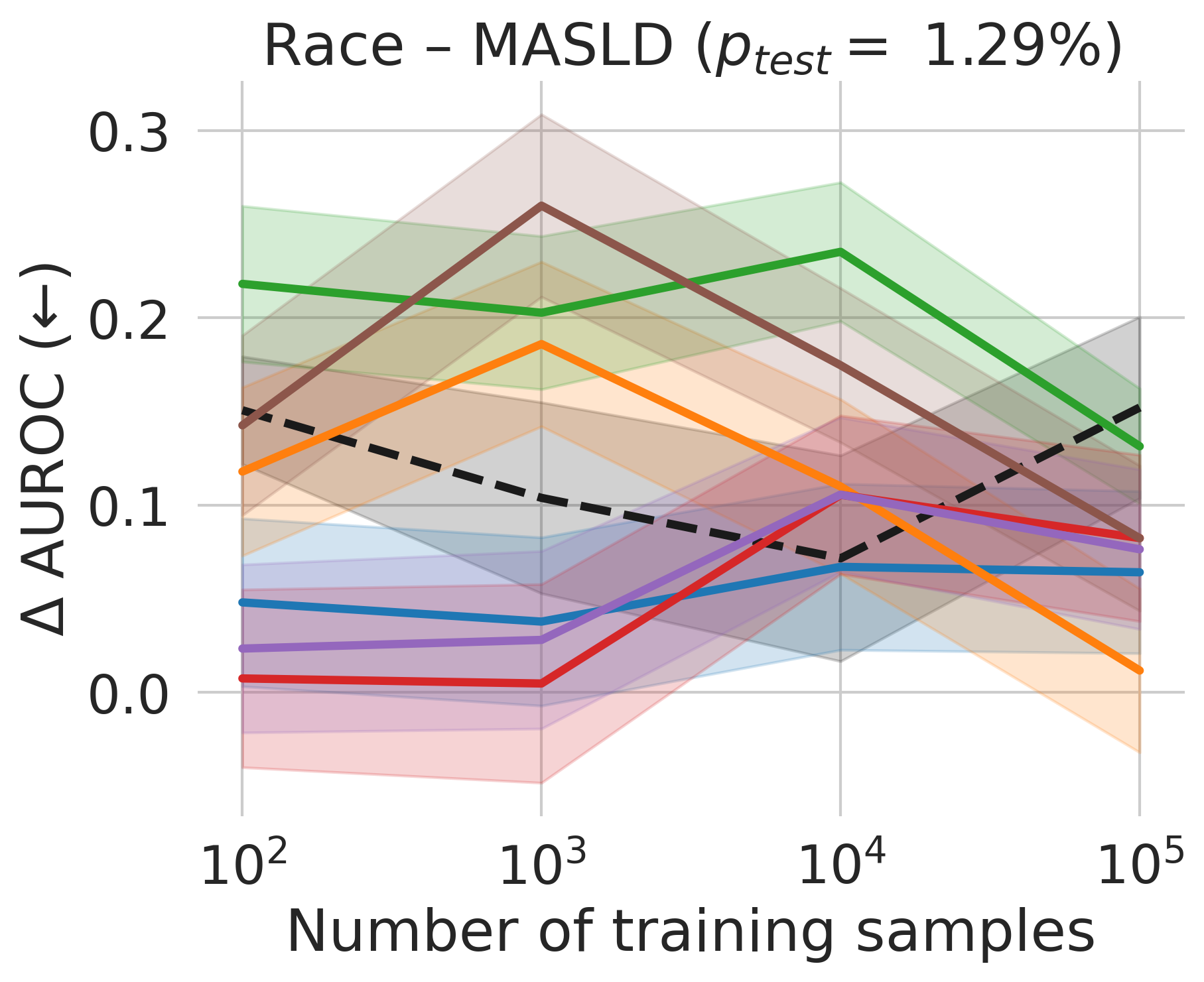}
    \includegraphics[height=100px]{figures/legend.png}}

    \makebox[\textwidth][l]{
    \includegraphics[height=100px]{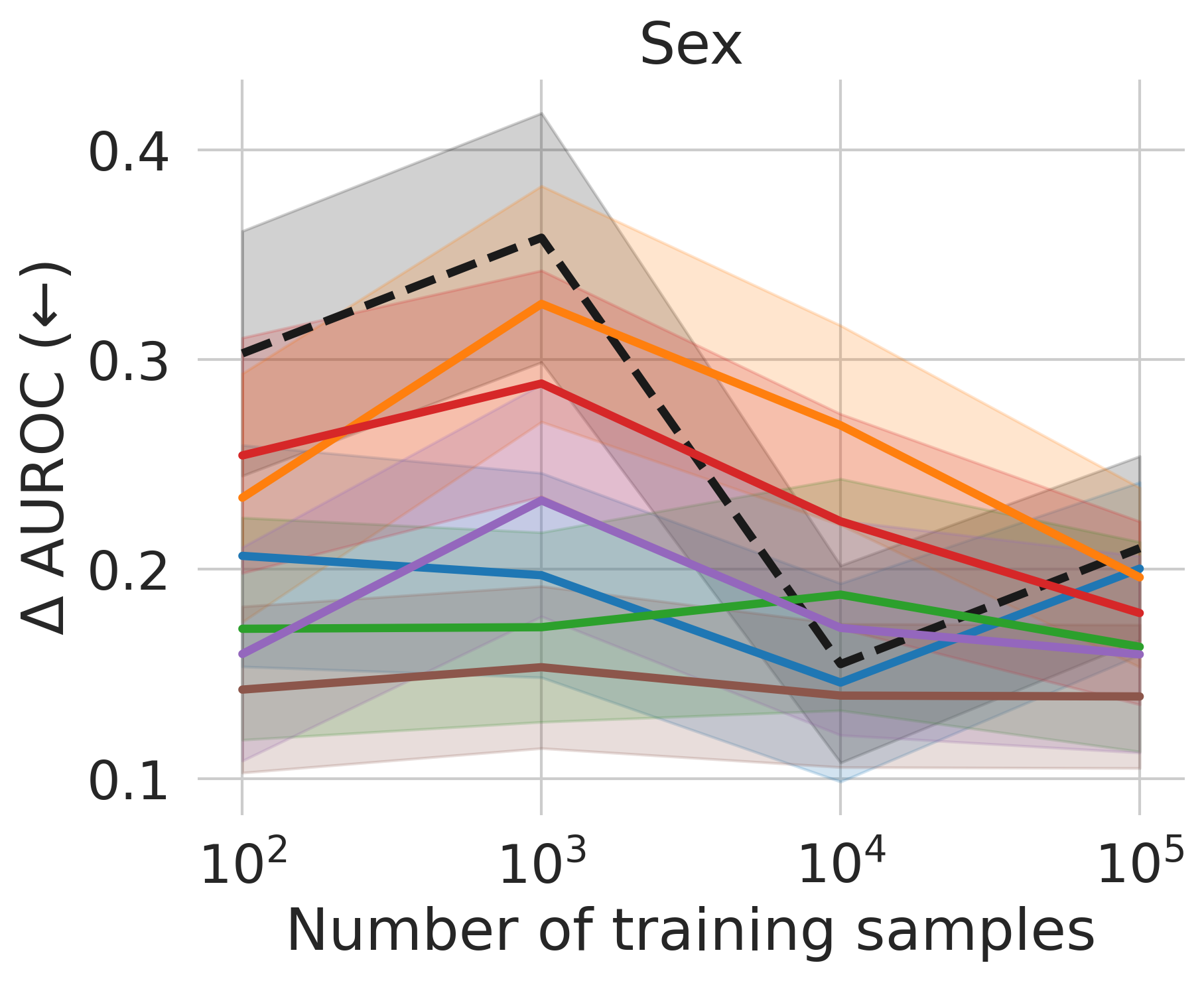}
    \includegraphics[height=100px]{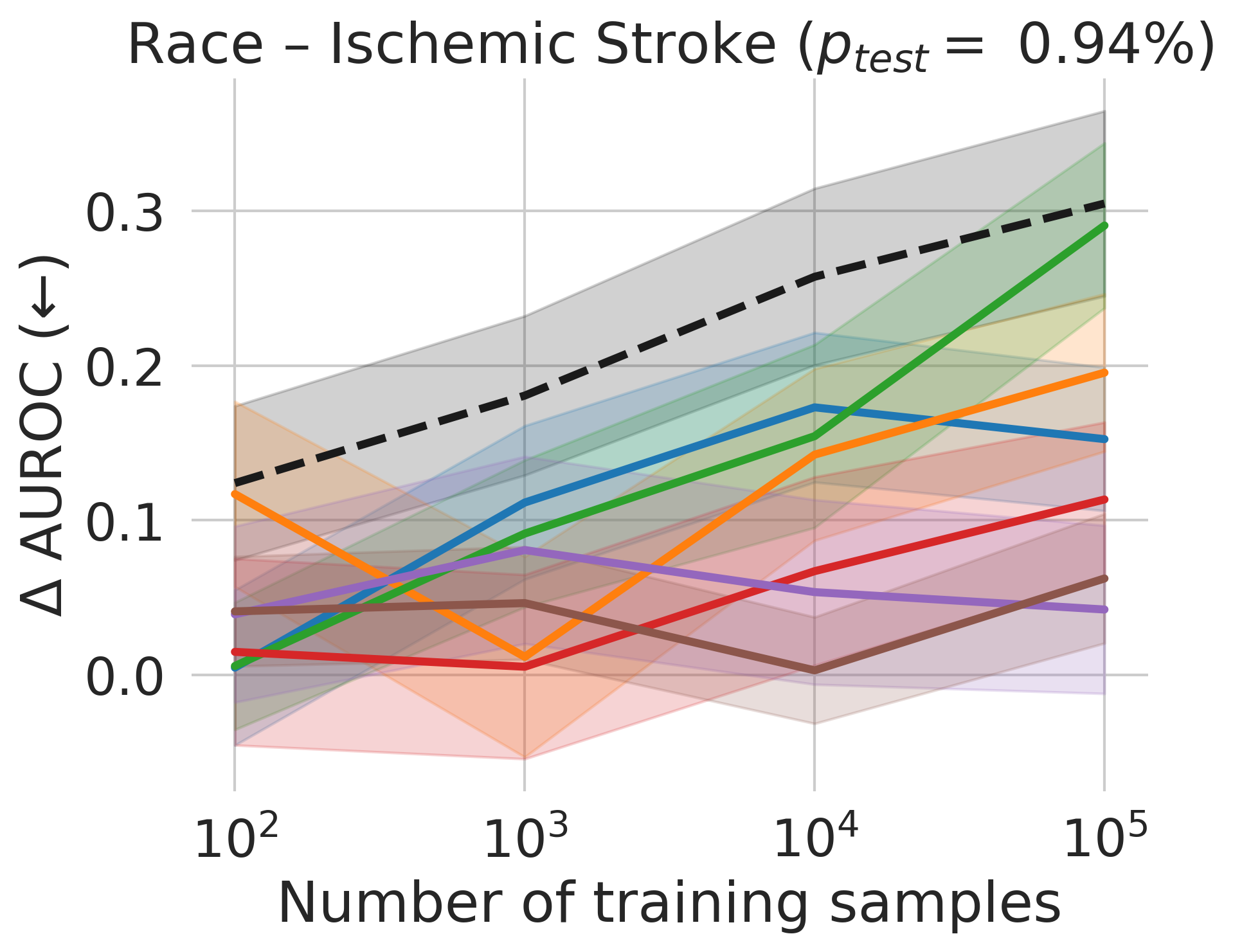}}

    \centering
    \caption{Discriminative performances and calibration results for all phenotypes for an increasing number of training points (Shaded area represents bootstrapped standard deviation).}
    \label{fig:all_phenotypes:mimic:fairness}
\end{figure}

Figure~\ref{fig:all_outcomes:mimic} presents the overall performance, including calibration for all patient outcomes in MIMIC-IV. Figure~\ref{fig:all_phenotypes:mimic} presents the overall performance, including calibration for all phenotypes in MIMIC-IV.
Fairness is assessed by measuring the discriminative gap across sex and race (see Figures~\ref{fig:all_outcomes:mimic:fairness} and~\ref{fig:all_phenotypes:mimic:fairness})\footnote{Due to the granularity of the available demographic data, we consider two groups of interest: Black and White patients.}. Note that healthcare utilization is not available in \mimic.

\clearpage
\subsection{Performances using the largest training set}
Table~\ref{tab:outcome-overall-performance:mimic} presents the discriminative tasks over all considered tasks when using all extracted training points. 
Similarly, Table~\ref{tab:outcome-overall-calibration:mimic} presents the calibration as measured by the Brier score, and Table~\ref{tab:outcome-overall-ece:mimic} presents calibration results as measured using ECE. 
Tables~\ref{tab:outcome-sex:mimic} and~\ref{tab:outcome-race:mimic} present the maximal discriminative performances for the sex and ethnicity splits.

\begin{table}[ht]
    \footnotesize
    \centering
    \setlength{\tabcolsep}{3pt}
    \setlength\extrarowheight{3pt}
    
    \caption{Outcome discriminative performance results. \textit{Best performances are marked in \textbf{bold}.}}
    \label{tab:outcome-overall-performance:mimic}
    \begin{tabular}{cccccccc}
        \toprule
         & \textbf{Best Baseline} & \makecell{\textbf{\textsc{CEHR}}\\\textbf{\textsc{BERT}}} & \makecell{\textbf{\textsc{CEHR}}\\\textbf{\textsc{GPT}}} & \makecell{\textbf{\textsc{Core}}\\\textbf{\textsc{Behrt}}} & 
        \textbf{\llama} & \textbf{\mamba} & \textbf{\motor} \\ \cmidrule{2-8}
    LOS & \makecell { 0.774 \\ (0.003) } & \makecell { 0.763 \\ (0.003) } & \makecell { 0.803 \\ (0.003) } & \makecell { 0.689 \\ (0.003) } & \makecell { 0.778 \\ (0.003) } & \makecell { 0.776 \\ (0.003) } & \makecell { \textbf{0.806} \\ (0.002) } \\
Mortality & \makecell { 0.882 \\ (0.006) } & \makecell { 0.880 \\ (0.006) } & \makecell { 0.917 \\ (0.004) } & \makecell { 0.817 \\ (0.007) } & \makecell { 0.897 \\ (0.005) } & \makecell { 0.889 \\ (0.006) } & \makecell { \textbf{0.929} \\ (0.004) } \\
Readmission & \makecell { \textbf{0.750} \\ (0.003) } & \makecell { 0.725 \\ (0.003) } & \makecell { 0.741 \\ (0.003) } & \makecell { 0.693 \\ (0.003) } & \makecell { 0.728 \\ (0.003) } & \makecell { 0.730 \\ (0.003) } & \makecell { 0.736 \\ (0.003) } \\

            \cmidrule{2-8}
    AMI & \makecell { 0.795 \\ (0.009) } & \makecell { 0.802 \\ (0.008) } & \makecell { 0.817 \\ (0.008) } & \makecell { 0.758 \\ (0.010) } & \makecell { 0.791 \\ (0.008) } & \makecell { 0.780 \\ (0.009) } & \makecell { \textbf{0.819} \\ (0.008) } \\
MASLD & \makecell { 0.650 \\ (0.020) } & \makecell { 0.693 \\ (0.018) } & \makecell { 0.720 \\ (0.018) } & \makecell { 0.656 \\ (0.017) } & \makecell { 0.662 \\ (0.020) } & \makecell { 0.669 \\ (0.020) } & \makecell { \textbf{0.725} \\ (0.018) } \\
Ischemic Stroke & \makecell { 0.688 \\ (0.023) } & \makecell { 0.662 \\ (0.025) } & \makecell { 0.705 \\ (0.025) } & \makecell { 0.606 \\ (0.032) } & \makecell { 0.695 \\ (0.026) } & \makecell { 0.665 \\ (0.022) } & \makecell { \textbf{0.731} \\ (0.020) } \\
    
        \bottomrule
    \end{tabular}
        \vspace{2pt}
    \newline
    AMI: Acute myocardial infarction. MASLD: Metabolic dysfunction-associated steatotic liver disease. 
\end{table}

\begin{table}[ht]
    \footnotesize
    \centering
    \setlength{\tabcolsep}{3pt}
    \setlength\extrarowheight{3pt}
    \caption{Outcome Brier score results. \textit{Best performances are marked in \textbf{bold}.}}
    \label{tab:outcome-overall-calibration:mimic}
    \begin{tabular}{cccccccc}
        \toprule
         & \textbf{Best Baseline} & \makecell{\textbf{\textsc{CEHR}}\\\textbf{\textsc{BERT}}} & \makecell{\textbf{\textsc{CEHR}}\\\textbf{\textsc{GPT}}} & \makecell{\textbf{\textsc{Core}}\\\textbf{\textsc{Behrt}}} & 
        \textbf{\llama} & \textbf{\mamba} & \textbf{\motor} \\ \cmidrule{2-8}
     LOS & \makecell { 0.165 \\ (0.001) } & \makecell { 0.169 \\ (0.001) } & \makecell { 0.157 \\ (0.001) } & \makecell { 0.187 \\ (0.001) } & \makecell { 0.165 \\ (0.001) } & \makecell { 0.165 \\ (0.001) } & \makecell { \textbf{0.155} \\ (0.001) } \\
Mortality & \makecell { 0.021 \\ (0.001) } & \makecell { 0.021 \\ (0.001) } & \makecell { 0.020 \\ (0.001) } & \makecell { 0.023 \\ (0.001) } & \makecell { 0.021 \\ (0.001) } & \makecell { 0.020 \\ (0.001) } & \makecell { \textbf{0.019} \\ (0.001) } \\
Readmission & \makecell { \textbf{0.149} \\ (0.001) } & \makecell { 0.157 \\ (0.001) } & \makecell { 0.151 \\ (0.001) } & \makecell { 0.163 \\ (0.001) } & \makecell { 0.154 \\ (0.001) } & \makecell { 0.155 \\ (0.001) } & \makecell { 0.155 \\ (0.001) } \\
            \cmidrule{2-8}
   AMI & \makecell { 0.039 \\ (0.001) } & \makecell { 0.038 \\ (0.001) } & \makecell { 0.037 \\ (0.001) } & \makecell { 0.039 \\ (0.001) } & \makecell { 0.038 \\ (0.001) } & \makecell { 0.038 \\ (0.001) } & \makecell { \textbf{0.037} \\ (0.001) } \\
MASLD & \makecell { \textbf{0.012} \\ (0.001) } & \makecell { 0.013 \\ (0.001) } & \makecell { 0.013 \\ (0.001) } & \makecell { 0.013 \\ (0.001) } & \makecell { 0.013 \\ (0.001) } & \makecell { 0.013 \\ (0.001) } & \makecell { 0.013 \\ (0.001) } \\
Ischemic Stroke & \makecell { 0.009 \\ (0.001) } & \makecell { 0.009 \\ (0.001) } & \makecell { 0.009 \\ (0.001) } & \makecell { 0.009 \\ (0.001) } & \makecell { \textbf{0.009} \\ (0.001) } & \makecell { 0.009 \\ (0.001) } & \makecell { 0.009 \\ (0.001) } \\

            \bottomrule
    \end{tabular}
            \vspace{2pt}
    \newline
    AMI: Acute myocardial infarction. MASLD: Metabolic dysfunction-associated steatotic liver disease. 
\end{table}

\begin{table}[ht]
    \footnotesize
    \centering
    \setlength{\tabcolsep}{3pt}
    \setlength\extrarowheight{3pt}
    \caption{Outcome ECE results. \textit{Best performances are marked in \textbf{bold}.}}
    \label{tab:outcome-overall-ece:mimic}
    \begin{tabular}{cccccccc}
        \toprule
         & \textbf{Best Baseline} & \makecell{\textbf{\textsc{CEHR}}\\\textbf{\textsc{BERT}}} & \makecell{\textbf{\textsc{CEHR}}\\\textbf{\textsc{GPT}}} & \makecell{\textbf{\textsc{Core}}\\\textbf{\textsc{Behrt}}} & 
        \textbf{\llama} & \textbf{\mamba} & \textbf{\motor} \\ \cmidrule{2-8}
LOS & \makecell { 0.012 \\ (0.002) } & \makecell { 0.010 \\ (0.002) } & \makecell { 0.011 \\ (0.002) } & \makecell { 0.015 \\ (0.002) } & \makecell { \textbf{0.009} \\ (0.001) } & \makecell { 0.010 \\ (0.002) } & \makecell { 0.012 \\ (0.002) } \\
Mortality & \makecell { 0.002 \\ (0.001) } & \makecell { 0.002 \\ (0.001) } & \makecell { 0.002 \\ (0.001) } & \makecell { \textbf{0.002} \\ (0.001) } & \makecell { 0.002 \\ (0.001) } & \makecell { 0.002 \\ (0.001) } & \makecell { 0.003 \\ (0.001) } \\
Readmission & \makecell { 0.015 \\ (0.002) } & \makecell { 0.010 \\ (0.002) } & \makecell { 0.009 \\ (0.002) } & \makecell { \textbf{0.008} \\ (0.002) } & \makecell { 0.009 \\ (0.002) } & \makecell { 0.008 \\ (0.001) } & \makecell { 0.011 \\ (0.002) } \\

            \cmidrule{2-8}
    AMI & \makecell { 0.011 \\ (0.001) } & \makecell { 0.003 \\ (0.001) } & \makecell { 0.003 \\ (0.001) } & \makecell { 0.009 \\ (0.002) } & \makecell { 0.003 \\ (0.001) } & \makecell { \textbf{0.003} \\ (0.001) } & \makecell { 0.003 \\ (0.001) } \\
MASLD & \makecell { \textbf{0.001} \\ (0.001) } & \makecell { 0.001 \\ (0.001) } & \makecell { 0.001 \\ (0.001) } & \makecell { 0.004 \\ (0.001) } & \makecell { 0.001 \\ (0.001) } & \makecell { 0.001 \\ (0.001) } & \makecell { 0.001 \\ (0.001) } \\
Ischemic Stroke & \makecell { 0.001 \\ (0.001) } & \makecell { 0.001 \\ (0.001) } & \makecell { 0.002 \\ (0.001) } & \makecell { 0.002 \\ (0.001) } & \makecell { 0.001 \\ (0.000) } & \makecell { \textbf{0.001} \\ (0.001) } & \makecell { 0.001 \\ (0.000) } \\

        \bottomrule
    \end{tabular}
            \vspace{2pt}
    \newline
    AMI: Acute myocardial infarction. MASLD: Metabolic dysfunction-associated steatotic liver disease. 
\end{table}

\clearpage
\begin{table}[ht]
    \footnotesize
    \centering
    \setlength{\tabcolsep}{3pt}
    \setlength\extrarowheight{3pt}
    \caption{Outcome sex gap in discriminative performance. \textit{Best performances are marked in \textbf{bold}.}}
    \label{tab:outcome-sex:mimic}
    \begin{tabular}{cccccccc}
        \toprule
         & \textbf{Best Baseline} & \makecell{\textbf{\textsc{CEHR}}\\\textbf{\textsc{BERT}}} & \makecell{\textbf{\textsc{CEHR}}\\\textbf{\textsc{GPT}}} & \makecell{\textbf{\textsc{Core}}\\\textbf{\textsc{Behrt}}} & 
        \textbf{\llama} & \textbf{\mamba} & \textbf{\motor} \\ \cmidrule{2-8}
    LOS & \makecell { 0.046 \\ (0.006) } & \makecell { 0.038 \\ (0.005) } & \makecell { 0.033 \\ (0.005) } & \makecell { 0.049 \\ (0.006) } & \makecell { 0.033 \\ (0.006) } & \makecell { 0.039 \\ (0.006) } & \makecell { \textbf{0.026} \\ (0.005) } \\
Mortality & \makecell { 0.010 \\ (0.012) } & \makecell { \textbf{0.000} \\ (0.012) } & \makecell { 0.007 \\ (0.009) } & \makecell { 0.009 \\ (0.013) } & \makecell { 0.009 \\ (0.011) } & \makecell { 0.025 \\ (0.011) } & \makecell { 0.002 \\ (0.008) } \\
Readmission & \makecell { 0.008 \\ (0.006) } & \makecell { 0.002 \\ (0.006) } & \makecell { \textbf{0.000} \\ (0.006) } & \makecell { 0.007 \\ (0.006) } & \makecell { 0.005 \\ (0.006) } & \makecell { 0.009 \\ (0.006) } & \makecell { 0.002 \\ (0.006) } \\

            \cmidrule{2-8}
    AMI & \makecell { 0.226 \\ (0.029) } & \makecell { 0.048 \\ (0.018) } & \makecell { 0.012 \\ (0.016) } & \makecell { 0.034 \\ (0.019) } & \makecell { \textbf{0.004} \\ (0.018) } & \makecell { 0.004 \\ (0.020) } & \makecell { 0.026 \\ (0.019) } \\
MASLD & \makecell { 0.114 \\ (0.041) } & \makecell { 0.016 \\ (0.034) } & \makecell { 0.024 \\ (0.037) } & \makecell { \textbf{0.002} \\ (0.038) } & \makecell { 0.017 \\ (0.038) } & \makecell { 0.009 \\ (0.037) } & \makecell { 0.014 \\ (0.039) } \\
Ischemic Stroke & \makecell { 0.210 \\ (0.044) } & \makecell { 0.200 \\ (0.041) } & \makecell { 0.196 \\ (0.043) } & \makecell { 0.163 \\ (0.050) } & \makecell { 0.179 \\ (0.044) } & \makecell { 0.159 \\ (0.047) } & \makecell { \textbf{0.139} \\ (0.034) } \\

        \bottomrule
    \end{tabular}
            \vspace{2pt}
    \newline
    AMI: Acute myocardial infarction. MASLD: Metabolic dysfunction-associated steatotic liver disease. 
\end{table}

\begin{table}[ht]
    \footnotesize
    \centering
    \setlength{\tabcolsep}{3pt}
    \setlength\extrarowheight{3pt}
    \caption{Outcome racial gap in discriminative performance. \textit{Best performances are marked in \textbf{bold}.}}
    \label{tab:outcome-race:mimic}
    \begin{tabular}{cccccccc}
        \toprule
         & \textbf{Best Baseline} & \makecell{\textbf{\textsc{CEHR}}\\\textbf{\textsc{BERT}}} & \makecell{\textbf{\textsc{CEHR}}\\\textbf{\textsc{GPT}}} & \makecell{\textbf{\textsc{Core}}\\\textbf{\textsc{Behrt}}} & 
        \textbf{\llama} & \textbf{\mamba}  & \textbf{\motor} \\ \cmidrule{2-8}
LOS & \makecell { 0.021 \\ (0.007) } & \makecell { \textbf{0.011} \\ (0.007) } & \makecell { 0.021 \\ (0.007) } & \makecell { 0.015 \\ (0.009) } & \makecell { 0.021 \\ (0.008) } & \makecell { 0.024 \\ (0.007) } & \makecell { 0.016 \\ (0.006) } \\
Mortality & \makecell { 0.077 \\ (0.038) } & \makecell { 0.019 \\ (0.016) } & \makecell { 0.002 \\ (0.016) } & \makecell { 0.001 \\ (0.022) } & \makecell { 0.019 \\ (0.017) } & \makecell { \textbf{0.001} \\ (0.019) } & \makecell { 0.004 \\ (0.014) } \\
Readmission & \makecell { 0.058 \\ (0.009) } & \makecell { 0.054 \\ (0.007) } & \makecell { 0.060 \\ (0.006) } & \makecell { \textbf{0.052} \\ (0.007) } & \makecell { 0.055 \\ (0.006) } & \makecell { 0.066 \\ (0.006) } & \makecell { 0.064 \\ (0.006) } \\
            \cmidrule{2-8}
AMI & \makecell { 0.275 \\ (0.033) } & \makecell { 0.061 \\ (0.023) } & \makecell { 0.008 \\ (0.020) } & \makecell { 0.008 \\ (0.024) } & \makecell { 0.007 \\ (0.023) } & \makecell { \textbf{0.006} \\ (0.023) } & \makecell { 0.037 \\ (0.020) } \\
MASLD & \makecell { 0.152 \\ (0.048) } & \makecell { 0.064 \\ (0.043) } & \makecell { \textbf{0.012} \\ (0.044) } & \makecell { 0.131 \\ (0.031) } & \makecell { 0.082 \\ (0.044) } & \makecell { 0.076 \\ (0.043) } & \makecell { 0.082 \\ (0.039) } \\
Ischemic Stroke & \makecell { 0.305 \\ (0.060) } & \makecell { 0.152 \\ (0.046) } & \makecell { 0.195 \\ (0.051) } & \makecell { 0.291 \\ (0.053) } & \makecell { 0.113 \\ (0.050) } & \makecell { \textbf{0.042} \\ (0.054) } & \makecell { 0.062 \\ (0.042) } \\
        \bottomrule
    \end{tabular}
            \vspace{2pt}
    \newline
    AMI: Acute myocardial infarction. MASLD: Metabolic dysfunction-associated steatotic liver disease. 
\end{table}

\clearpage
\subsection{Detailed baseline results}
For completeness, Table~\ref{tab:outcome-overall-baseline:mimic} details the discriminative performances for each considered baseline model in \mimic. All previous tables and figures reflect the best performance across these different models.

\begin{table}[ht]
    \footnotesize
    \centering
    \setlength{\tabcolsep}{3.5pt}
    \caption{Outcome prediction results for considered baselines.}
    \label{tab:outcome-overall-baseline:mimic}
    \begin{tabular}{ccccc}
        \toprule
        \textbf{Task} &
         \textbf{\femr - GBM} & \textbf{\femr - LR} & \textbf{\medstab - XGB} & \textbf{\medstab - LR} \\
        \cmidrule{2-5}
   LOS & \makecell { 0.763 \\ (0.002) } & \makecell { 0.742 \\ (0.003) } & \makecell { \textbf{0.774} \\ (0.003) } & \makecell { 0.735 \\ (0.003) } \\
Mortality & \makecell { 0.875 \\ (0.005) } & \makecell { 0.834 \\ (0.008) } & \makecell { \textbf{0.882} \\ (0.006) } & \makecell { 0.861 \\ (0.007) } \\
Readmission & \makecell { 0.725 \\ (0.003) } & \makecell { 0.680 \\ (0.004) } & \makecell { \textbf{0.750} \\ (0.003) } & \makecell { 0.685 \\ (0.003) } \\

        \cmidrule{2-5}
  AMI & \makecell { 0.756 \\ (0.010) } & \makecell { 0.563 \\ (0.013) } & \makecell { \textbf{0.795} \\ (0.009) } & \makecell { 0.572 \\ (0.014) } \\
MASLD & \makecell { 0.592 \\ (0.019) } & \makecell { \textbf{0.650} \\ (0.020) } & \makecell { 0.599 \\ (0.022) } & \makecell { 0.575 \\ (0.021) } \\
Ischemic Stroke & \makecell { 0.657 \\ (0.027) } & \makecell { 0.481 \\ (0.035) } & \makecell { \textbf{0.688} \\ (0.023) } & \makecell { 0.521 \\ (0.035) } \\

        \bottomrule
    \end{tabular}
    \vspace{2pt}
    \newline     
    AMI: Acute myocardial infarction. MASLD: Metabolic dysfunction-associated steatotic liver disease. 
    {LR = Logistic Regression, GBM = LightGBM, XGB = XGBoost}
\end{table}

\section*{Code availability}
The code for the proposed benchmark and task definitions is publicly available at \url{https://github.com/reAIM-Lab/ehr_foundation_model_benchmark/}.

%% file: references.bib
@misc{NIDDK_Type2DiabetesRiskFactors,
  author       = {{National Institute of Diabetes and Digestive and Kidney Diseases (NIDDK)}},
  title        = {Risk Factors for Type 2 Diabetes},
  year         = {2025},
  howpublished = {\url{https://www.niddk.nih.gov/health-information/diabetes/overview/risk-factors-type-2-diabetes}},
  note         = {Accessed: 2025-05-05}
}

@article{doi:10.1056/AIra2501253,
author = {Matthew B. A. McDermott  and Ethan Steinberg  and Jason A. Fries  and Robin P. van de Water  and Chao Pang  and Patrick Rockenschaub  and Pawel Renc  and Jungwoo Oh  and Kamilė Stankevičiūtė  and Justin Xu  and Tom J. Pollard  and Nassim Oufattole  and Michael Wornow  and Teya S. Bergamaschi  and Hyewon Jeong  and Simon A. Lee  and Vincent Jeanselme  and Kiril V. Klein  and Mikkel Odgaard  and Maria E. Montgomery  and Arkadiusz Sitek  and Mads Nielsen  and Jeffrey N. Chiang  and Noa Dagan  and Isaac Kohane  and Shalmali Joshi  and Edward Choi  and Nigam H. Shah },
title = {MEDS — An Emerging Data Standard and Ecosystem for Health AI Research},
journal = {NEJM AI},
volume = {3},
number = {6},
pages = {AIra2501253},
year = {2026},
doi = {10.1056/AIra2501253},
}

@ARTICLE{Unknown2025-ul,
  title     = "Benchmarking matters",
  author  = {{The Editors of Nature Biomedical Engineering}},
  journal   = "Nat. Biomed. Eng.",
  publisher = "Springer Science and Business Media LLC",
  volume    =  9,
  number    =  11,
  pages     = "1789--1790",
  month     =  nov,
  year      =  2025,
  language  = "en"
}

@misc{wang2025crisis,
  title={The crisis of biomedical foundation models},
  author={Wang, Fei},
  journal={Journal of Biomedical Informatics},
  pages={104917},
  year={2025},
  publisher={Elsevier}
}

@article{pang2025cehrxgptscalablemultitaskfoundation,
  title={CEHR-XGPT: A Scalable Multi-Task Foundation Model for Electronic Health Records},
  author={Pang, Chao and Park, Jiheum and Jiang, Xinzhuo and Pavinkurve, Nishanth Parameshwar and Kalluri, Krishna S and Joshi, Shalmali and Elhadad, No{\~A} and Natarajan, Karthik and others},
  journal={arXiv preprint arXiv:2509.03643},
  year={2025}
}

@article{renc2025foundation,
  title={Foundation model of electronic medical records for adaptive risk estimation},
  author={Renc, Pawel and Grzeszczyk, Michal K and Oufattole, Nassim and Goode, Deirdre and Jia, Yugang and Bieganski, Szymon and McDermott, Matthew BA and Was, Jaroslaw and Samir, Anthony E and Cunningham, Jonathan W and others},
  journal={GigaScience},
  volume={14},
  year={2025},
  publisher={Oxford University Press}
}

@misc{ACS_CLLRiskFactors,
  author       = {{American Cancer Society}},
  title        = {Risk Factors for Chronic Lymphocytic Leukemia},
  year         = {2024},
  howpublished = {\url{https://www.cancer.org/cancer/types/chronic-lymphocytic-leukemia/causes-risks-prevention/risk-factors.html}},
  note         = {Accessed: 2025-05-05}
}

@misc{UCSD_OsteoporosisRisk,
  author       = {{UC San Diego Health}},
  title        = {Osteoporosis Risk Factors},
  year         = {2024},
  howpublished = {\url{https://health.ucsd.edu/care/endocrinology-diabetes/osteoporosis/risk-factors/}},
  note         = {Accessed: 2025-05-05}
}

@article{Suchard2019,
  author       = {Marc A. Suchard and Martijn J. Schuemie and Harlan M. Krumholz and Seng Chan You and RuiJun Chen and Nicole Pratt and Christian G. Reich and Jon Duke and David Madigan and George Hripcsak and Patrick B. Ryan},
  title        = {Comprehensive comparative effectiveness and safety of first-line antihypertensive drug classes: a systematic, multinational, large-scale analysis},
  journal      = {The Lancet},
  year         = {2019},
  volume       = {394},
  number       = {10211},
  pages        = {1816--1826},
  doi          = {10.1016/S0140-6736(19)32317-7}
}

@misc{CDCStrokeRisk2024,
  author       = {{Centers for Disease Control and Prevention}},
  title        = {Risk Factors for Stroke},
  year         = {2024},
  howpublished = {\url{https://www.cdc.gov/stroke/risk-factors/index.html}},
  note         = {Accessed: 2025-05-05}
}

@article{Huh2022,
  author    = {Youn Huh and Yoon Jeong Cho and Ga Eun Nam},
  title     = {Recent Epidemiology and Risk Factors of Nonalcoholic Fatty Liver Disease},
  journal   = {Journal of Obesity \& Metabolic Syndrome},
  year      = {2022},
  volume    = {31},
  number    = {1},
  pages     = {17--27},
  doi       = {10.7570/jomes22021}
}

@article{Chung2025,
  author    = {Goh Eun Chung and Su Jong Yu and Jeong-Ju Yoo and Yuri Cho and Kyu-na Lee and Dong Wook Shin and Yoon Jun Kim and Jung-Hwan Yoon and Kyungdo Han and Eun Ju Cho},
  title     = {Metabolic dysfunction-associated steatotic liver disease increases cardiovascular disease risk in young adults},
  journal   = {Scientific Reports},
  year      = {2025},
  volume    = {15},
  pages     = {5777},
  doi       = {10.1038/s41598-025-89293-6}
}

@article{Suchard2021LEGENDT2DM,
  author = {Suchard, Marc A and Schuemie, Martijn J and Krumholz, Harlan M and You, Seoyoung C and Chen, Ren and Pratt, Nicole and Reich, Christian and Ryan, Patrick and Hripcsak, George},
  title = {Large-scale evidence generation and evaluation across a network of databases for type 2 diabetes mellitus (LEGEND-T2DM): a protocol for a series of multinational, real-world comparative cardiovascular effectiveness and safety studies},
  journal = {BMJ Open},
  volume = {11},
  number = {1},
  pages = {e043247},
  year = {2021},
  doi = {10.1136/bmjopen-2020-043247},
  publisher = {BMJ Publishing Group Ltd}
}

@article{guo2025foundation,
  title={Foundation models in bioinformatics},
  author={Guo, Fei and Guan, Renchu and Li, Yaohang and Liu, Qi and Wang, Xiaowo and Yang, Can and Wang, Jianxin},
  journal={National Science Review},
  year={2025},
  publisher={Oxford University Press}
}

@article{team2023gemini,
  title={Gemini: a family of highly capable multimodal models},
  author={Team, Gemini and Anil, Rohan and Borgeaud, Sebastian and Alayrac, Jean-Baptiste and Yu, Jiahui and Soricut, Radu and Schalkwyk, Johan and Dai, Andrew M and Hauth, Anja and Millican, Katie and others},
  journal={arXiv preprint arXiv:2312.11805},
  year={2023}
}

@article{gu2023mamba,
  title={Mamba: Linear-time sequence modeling with selective state spaces},
  author={Gu, Albert and Dao, Tri},
  journal={arXiv preprint arXiv:2312.00752},
  year={2023}
}

@article{grattafiori2024llama,
  title={The llama 3 herd of models},
  author={Grattafiori, Aaron and Dubey, Abhimanyu and Jauhri, Abhinav and Pandey, Abhinav and Kadian, Abhishek and Al-Dahle, Ahmad and Letman, Aiesha and Mathur, Akhil and Schelten, Alan and Vaughan, Alex and others},
  journal={arXiv preprint arXiv:2407.21783},
  year={2024}
}

@incollection{hripcsak2015observational,
  title={Observational Health Data Sciences and Informatics (OHDSI): opportunities for observational researchers},
  author={Hripcsak, George and Duke, Jon D and Shah, Nigam H and Reich, Christian G and Huser, Vojtech and Schuemie, Martijn J and Suchard, Marc A and Park, Rae Woong and Wong, Ian Chi Kei and Rijnbeek, Peter R and others},
  booktitle={MEDINFO 2015: eHealth-enabled Health},
  pages={574--578},
  year={2015},
  publisher={IOS Press}
}

@article{steinberg2020languagemodelseffectivepatient,
  title={Language models are an effective representation learning technique for electronic health record data},
  author={Steinberg, Ethan and Jung, Ken and Fries, Jason A and Corbin, Conor K and Pfohl, Stephen R and Shah, Nigam H},
  journal={Journal of biomedical informatics},
  volume={113},
  pages={103637},
  year={2021},
  publisher={Elsevier}
}

@inproceedings{10.5555/3294996.3295074,
author = {Ke, Guolin and Meng, Qi and Finley, Thomas and Wang, Taifeng and Chen, Wei and Ma, Weidong and Ye, Qiwei and Liu, Tie-Yan},
title = {LightGBM: a highly efficient gradient boosting decision tree},
year = {2017},
isbn = {9781510860964},
publisher = {Curran Associates Inc.},
address = {Red Hook, NY, USA},
booktitle = {Proceedings of the 31st International Conference on Neural Information Processing Systems},
pages = {3149–3157},
numpages = {9},
location = {Long Beach, California, USA},
series = {NIPS'17}
}

@inproceedings{Chen_2016, series={KDD ’16},
   title={XGBoost: A Scalable Tree Boosting System},
   DOI={10.1145/2939672.2939785},
   booktitle={Proceedings of the 22nd ACM SIGKDD International Conference on Knowledge Discovery and Data Mining},
   publisher={ACM},
   author={Chen, Tianqi and Guestrin, Carlos},
   year={2016},
   month=aug, pages={785–794},
   collection={KDD ’16} }

@inproceedings{
arnrich2024medical,
title={Medical Event Data Standard ({MEDS}): Facilitating Machine Learning for Health},
author={Bert Arnrich and Edward Choi and Jason Alan Fries and Matthew B.A. McDermott and Jungwoo Oh and Tom Pollard and Nigam Shah and Ethan Steinberg and Michael Wornow and Robin van de Water},
booktitle={ICLR 2024 Workshop on Learning from Time Series For Health},
year={2024}
}

@misc{oufattole2024medstabautomatedtabularizationbaseline,
      title={MEDS-Tab: Automated tabularization and baseline methods for MEDS datasets}, 
      author={Nassim Oufattole and Teya Bergamaschi and Aleksia Kolo and Hyewon Jeong and Hanna Gaggin and Collin M. Stultz and Matthew B. A. McDermott},
      year={2024},
      eprint={2411.00200},
      archivePrefix={arXiv},
      primaryClass={cs.LG},
}

@article{radford2018improving,
  title={Improving language understanding by generative pre-training},
  author={Radford, Alec and Narasimhan, Karthik and Salimans, Tim and Sutskever, Ilya and others},
  year={2018},
  publisher={San Francisco, CA, USA}
}

@article{henderson2023foundation,
  title={Foundation models and fair use},
  author={Henderson, Peter and Li, Xuechen and Jurafsky, Dan and Hashimoto, Tatsunori and Lemley, Mark A and Liang, Percy},
  journal={Journal of Machine Learning Research},
  volume={24},
  number={400},
  pages={1--79},
  year={2023}
}

@inproceedings{
steinberg2024motor,
title={{MOTOR}: A Time-To-Event Foundation Model For Structured Medical Records},
author={Ethan Steinberg and Yizhe Xu and Jason Alan Fries and Nigam Shah},
booktitle={The Twelfth International Conference on Learning Representations},
year={2024}
}

@inproceedings{wornowcontext,
  title={Context Clues: Evaluating Long Context Models for Clinical Prediction Tasks on EHR Data},
  author={Wornow, Michael and Bedi, Suhana and Hernandez, Miguel Angel Fuentes and Steinberg, Ethan and Fries, Jason Alan and Re, Christopher and Koyejo, Sanmi and Shah, Nigam},
  year={2024},
  booktitle={The Thirteenth International Conference on Learning Representations}
}

@article{linden2022prevalence,
  title={Prevalence and pattern of acute and chronic multimorbidity across all body systems and age groups in primary health care},
  author={Linden, Michael and Linden, Ulrike and Goretzko, David and Gensichen, Jochen},
  journal={Scientific Reports},
  volume={12},
  number={1},
  pages={272},
  year={2022},
  publisher={Nature Publishing Group UK London}
}

@article{blecker2021hospitalizations,
  title={Hospitalizations for chronic disease and acute conditions in the time of COVID-19},
  author={Blecker, Saul and Jones, Simon A and Petrilli, Christopher M and Admon, Andrew J and Weerahandi, Himali and Francois, Fritz and Horwitz, Leora I},
  journal={JAMA internal medicine},
  volume={181},
  number={2},
  pages={269--271},
  year={2021},
  publisher={American Medical Association}
}

@article{mcdermott2021reproducibility,
  title={Reproducibility in machine learning for health research: Still a ways to go},
  author={McDermott, Matthew BA and Wang, Shirly and Marinsek, Nikki and Ranganath, Rajesh and Foschini, Luca and Ghassemi, Marzyeh},
  journal={Science Translational Medicine},
  volume={13},
  number={586},
  pages={eabb1655},
  year={2021},
  publisher={American Association for the Advancement of Science}
}

@article{li2020behrt,
  title={BEHRT: transformer for electronic health records},
  author={Li, Yikuan and Rao, Shishir and Solares, Jos{\'e} Roberto Ayala and Hassaine, Abdelaali and Ramakrishnan, Rema and Canoy, Dexter and Zhu, Yajie and Rahimi, Kazem and Salimi-Khorshidi, Gholamreza},
  journal={Scientific reports},
  volume={10},
  number={1},
  pages={7155},
  year={2020},
  publisher={Nature Publishing Group UK London}
}

@article{steinberg2021language,
  title={Language models are an effective representation learning technique for electronic health record data},
  author={Steinberg, Ethan and Jung, Ken and Fries, Jason A and Corbin, Conor K and Pfohl, Stephen R and Shah, Nigam H},
  journal={Journal of biomedical informatics},
  volume={113},
  pages={103637},
  year={2021},
  publisher={Elsevier}
}

@article{wornow2024ehrshot,
  title={Ehrshot: An ehr benchmark for few-shot evaluation of foundation models},
  author={Wornow, Michael and Thapa, Rahul and Steinberg, Ethan and Fries, Jason and Shah, Nigam},
  journal={Advances in Neural Information Processing Systems},
  volume={36},
  year={2024}
}

@article{bommasani2021opportunities,
  title={On the opportunities and risks of foundation models},
  author={Bommasani, Rishi and Hudson, Drew A and Adeli, Ehsan and Altman, Russ and Arora, Simran and von Arx, Sydney and Bernstein, Michael S and Bohg, Jeannette and Bosselut, Antoine and Brunskill, Emma and others},
  journal={arXiv preprint arXiv:2108.07258},
  year={2021}
}

@article{wornow2023shaky,
  title={The shaky foundations of large language models and foundation models for electronic health records},
  author={Wornow, Michael and Xu, Yizhe and Thapa, Rahul and Patel, Birju and Steinberg, Ethan and Fleming, Scott and Pfeffer, Michael A and Fries, Jason and Shah, Nigam H},
  journal={npj Digital Medicine},
  volume={6},
  number={1},
  pages={135},
  year={2023},
  publisher={Nature Publishing Group UK London}
}

@article{mcdermott2024event,
  title={Event Stream GPT: a data pre-processing and modeling library for generative, pre-trained transformers over continuous-time sequences of complex events},
  author={McDermott, Matthew and Nestor, Bret and Argaw, Peniel and Kohane, Isaac S},
  journal={Advances in Neural Information Processing Systems},
  volume={36},
  year={2024}
}

@article{chen2020ethical,
author = {Irene Y. Chen and Emma Pierson and Sherri Rose and Shalmali Joshi and Kadija Ferryman and Marzyeh Ghassemi},
title = {Ethical Machine Learning in Healthcare},
journal = {Annual Review of Biomedical Data Science},
year = {2021}
}

@article{chen2023multimodal,
  title={Multimodal clinical benchmark for emergency care (mc-bec): A comprehensive benchmark for evaluating foundation models in emergency medicine},
  author={Chen, Emma and Kansal, Aman and Chen, Julie and Jin, Boyang Tom and Reisler, Julia and Kim, David E and Rajpurkar, Pranav},
  journal={Advances in Neural Information Processing Systems},
  volume={36},
  pages={45794--45811},
  year={2023}
}

@article{renc2024zero,
  title={Zero shot health trajectory prediction using transformer},
  author={Renc, Pawel and Jia, Yugang and Samir, Anthony E and Was, Jaroslaw and Li, Quanzheng and Bates, David W and Sitek, Arkadiusz},
  journal={npj Digital Medicine},
  volume={7},
  number={1},
  pages={256},
  year={2024},
  publisher={Nature Publishing Group UK London}
}

@misc{kazemi_time2vec_2019,
	title = {{Time2Vec}: {Learning} a {Vector} {Representation} of {Time}},
	shorttitle = {{Time2Vec}},
	doi = {10.48550/arXiv.1907.05321},
	urldate = {2025-03-26},
	publisher = {arXiv},
	author = {Kazemi, Seyed Mehran and Goel, Rishab and Eghbali, Sepehr and Ramanan, Janahan and Sahota, Jaspreet and Thakur, Sanjay and Wu, Stella and Smyth, Cathal and Poupart, Pascal and Brubaker, Marcus},
	month = jul,
	year = {2019},
	note = {arXiv:1907.05321 [cs]},
}

@article{he2024foundation,
  title={Foundation model for advancing healthcare: challenges, opportunities and future directions},
  author={He, Yuting and Huang, Fuxiang and Jiang, Xinrui and Nie, Yuxiang and Wang, Minghao and Wang, Jiguang and Chen, Hao},
  journal={IEEE Reviews in Biomedical Engineering},
  year={2024},
  publisher={IEEE}
}

@inproceedings{pang2021cehr,
  title={CEHR-BERT: Incorporating temporal information from structured EHR data to improve prediction tasks},
  author={Pang, Chao and Jiang, Xinzhuo and Kalluri, Krishna S and Spotnitz, Matthew and Chen, RuiJun and Perotte, Adler and Natarajan, Karthik},
  booktitle={Machine Learning for Health},
  pages={239--260},
  year={2021},
  organization={PMLR}
}

@article{rasmy2021med,
  title={Med-BERT: pretrained contextualized embeddings on large-scale structured electronic health records for disease prediction},
  author={Rasmy, Laila and Xiang, Yang and Xie, Ziqian and Tao, Cui and Zhi, Degui},
  journal={NPJ digital medicine},
  volume={4},
  number={1},
  pages={86},
  year={2021},
  publisher={Nature Publishing Group UK London}
}

@article{odgaard2024core,
  title={Core-behrt: A carefully optimized and rigorously evaluated behrt},
  author={Odgaard, Mikkel and Klein, Kiril Vadimovic and Thysen, Sanne M{\o}ller and Jimenez-Solem, Espen and Sillesen, Martin and Nielsen, Mads},
  journal={arXiv preprint arXiv:2404.15201},
  year={2024}
}

@article{kraljevic2024foresight,
  title={Foresight—a generative pretrained transformer for modelling of patient timelines using electronic health records: a retrospective modelling study},
  author={Kraljevic, Zeljko and Bean, Dan and Shek, Anthony and Bendayan, Rebecca and Hemingway, Harry and Yeung, Joshua Au and Deng, Alexander and Baston, Alfred and Ross, Jack and Idowu, Esther and others},
  journal={The Lancet Digital Health},
  volume={6},
  number={4},
  pages={e281--e290},
  year={2024},
  publisher={Elsevier}
}

@article{fallahpour2024ehrmamba,
  title={Ehrmamba: Towards generalizable and scalable foundation models for electronic health records},
  author={Fallahpour, Adibvafa and Alinoori, Mahshid and Ye, Wenqian and Cao, Xu and Afkanpour, Arash and Krishnan, Amrit},
  journal={arXiv preprint arXiv:2405.14567},
  year={2024}
}

@article{hur2023genhpf,
  title={Genhpf: General healthcare predictive framework for multi-task multi-source learning},
  author={Hur, Kyunghoon and Oh, Jungwoo and Kim, Junu and Kim, Jiyoun and Lee, Min Jae and Cho, Eunbyeol and Moon, Seong-Eun and Kim, Young-Hak and Atallah, Louis and Choi, Edward},
  journal={IEEE Journal of Biomedical and Health Informatics},
  volume={28},
  number={1},
  pages={502--513},
  year={2023},
  publisher={IEEE}
}

@article{stang2010advancing,
  title={Advancing the science for active surveillance: rationale and design for the Observational Medical Outcomes Partnership},
  author={Stang, Paul E and Ryan, Patrick B and Racoosin, Judith A and Overhage, J Marc and Hartzema, Abraham G and Reich, Christian and Welebob, Emily and Scarnecchia, Thomas and Woodcock, Janet},
  journal={Annals of internal medicine},
  volume={153},
  number={9},
  pages={600--606},
  year={2010},
  publisher={American College of Physicians}
}

@article{yang2023transformehr,
  title={TransformEHR: transformer-based encoder-decoder generative model to enhance prediction of disease outcomes using electronic health records},
  author={Yang, Zhichao and Mitra, Avijit and Liu, Weisong and Berlowitz, Dan and Yu, Hong},
  journal={Nature communications},
  volume={14},
  number={1},
  pages={7857},
  year={2023},
  publisher={Nature Publishing Group UK London}
}

@book{barocas2023fairness,
  title={Fairness and machine learning: Limitations and opportunities},
  author={Barocas, Solon and Hardt, Moritz and Narayanan, Arvind},
  year={2023},
  publisher={MIT press}
}

@article{marks2025closer,
  title={A closer look at benchmarking self-supervised pre-training with image classification},
  author={Marks, Markus and Knott, Manuel and Kondapaneni, Neehar and Cole, Elijah and Defraeye, Thijs and Perez-Cruz, Fernando and Perona, Pietro},
  journal={International Journal of Computer Vision},
  pages={1--13},
  year={2025},
  publisher={Springer}
}

@article{shah2023creation,
  title={Creation and adoption of large language models in medicine},
  author={Shah, Nigam H and Entwistle, David and Pfeffer, Michael A},
  journal={Jama},
  volume={330},
  number={9},
  pages={866--869},
  year={2023},
  publisher={American Medical Association}
}

@article{moor2023foundation,
  title={Foundation models for generalist medical artificial intelligence},
  author={Moor, Michael and Banerjee, Oishi and Abad, Zahra Shakeri Hossein and Krumholz, Harlan M and Leskovec, Jure and Topol, Eric J and Rajpurkar, Pranav},
  journal={Nature},
  volume={616},
  number={7956},
  pages={259--265},
  year={2023},
  publisher={Nature Publishing Group UK London}
}

@article{kraljevic2021medgpt,
  title={MedGPT: Medical concept prediction from clinical narratives},
  author={Kraljevic, Zeljko and Shek, Anthony and Bean, Daniel and Bendayan, Rebecca and Teo, James and Dobson, Richard},
  journal={arXiv preprint arXiv:2107.03134},
  year={2021}
}

@article{jeong2023event,
  title={Event-based contrastive learning for medical time series},
  author={Jeong, Hyewon and Oufattole, Nassim and Mcdermott, Matthew and Balagopalan, Aparna and Jangeesingh, Bryan and Ghassemi, Marzyeh and Stultz, Collin},
  journal={arXiv preprint arXiv:2312.10308},
  year={2023}
}

@inproceedings{devlin2019bert,
  title={Bert: Pre-training of deep bidirectional transformers for language understanding},
  author={Devlin, Jacob and Chang, Ming-Wei and Lee, Kenton and Toutanova, Kristina},
  booktitle={Proceedings of the 2019 conference of the association for computational linguistics: human language technologies},
  volume = {1},
  pages={4171--4186},
  year={2019}
}

@article{joshi2025ai,
  title={AI as an intervention: improving clinical outcomes relies on a causal approach to AI development and validation},
  author={Joshi, Shalmali and Urteaga, I{\~n}igo and van Amsterdam, Wouter AC and Hripcsak, George and Elias, Pierre and Recht, Benjamin and Elhadad, No{\'e}mie and Fackler, James and Sendak, Mark P and Wiens, Jenna and others},
  journal={Journal of the American Medical Informatics Association},
  pages={ocae301},
  year={2025},
  publisher={Oxford University Press}
}

@article{finnerty2024prevalence,
  title={Prevalence and incidence measures for schizophrenia among commercial health insurance and medicaid enrollees},
  author={Finnerty, Molly T and Khan, Atif and You, Kai and Wang, Rui and Gu, Gyojeong and Layman, Deborah and Chen, Qingxian and Elhadad, No{\'e}mie and Joshi, Shalmali and Appelbaum, Paul S and others},
  journal={Schizophrenia},
  volume={10},
  number={1},
  pages={68},
  year={2024},
  publisher={Nature Publishing Group UK London}
}

@article{fu2025foundation,
  title={A foundation model of transcription across human cell types},
  author={Fu, Xi and Mo, Shentong and Buendia, Alejandro and Laurent, Anouchka P and Shao, Anqi and Alvarez-Torres, Maria del Mar and Yu, Tianji and Tan, Jimin and Su, Jiayu and Sagatelian, Romella and others},
  journal={Nature},
  pages={1--9},
  year={2025},
  publisher={Nature Publishing Group UK London}
}

@article{touvron2023llama,
  title={Llama: Open and efficient foundation language models},
  author={Touvron, Hugo and Lavril, Thibaut and Izacard, Gautier and Martinet, Xavier and Lachaux, Marie-Anne and Lacroix, Timoth{\'e}e and Rozi{\`e}re, Baptiste and Goyal, Naman and Hambro, Eric and Azhar, Faisal and others},
  journal={arXiv preprint arXiv:2302.13971},
  year={2023}
}

@inproceedings{das2024decoder,
  title={A decoder-only foundation model for time-series forecasting},
  author={Das, Abhimanyu and Kong, Weihao and Sen, Rajat and Zhou, Yichen},
  booktitle={Forty-first International Conference on Machine Learning},
  year={2024}
}

@inproceedings{
dai2023instructblip,
title={Instruct{BLIP}: Towards general-purpose vision-language models with instruction tuning},
author={Wenliang Dai and Junnan Li and Dongxu Li and Anthony Tiong and Junqi Zhao and Weisheng Wang and Boyang Li and Pascale Fung and Steven Hoi},
booktitle={Thirty-seventh Conference on Neural Information Processing Systems},
year={2023},
}

@inproceedings{
xu2025aces,
title={{ACES}: Automatic Cohort Extraction System for Event-Stream Datasets},
author={Justin Xu and Jack Gallifant and Alistair Johnson and Matthew B.A. McDermott},
booktitle={The Thirteenth International Conference on Learning Representations},
year={2025},
}

@article{pang2024cehr,
  title={CEHR-GPT: Generating electronic health records with chronological patient timelines},
  author={Pang, Chao and Jiang, Xinzhuo and Pavinkurve, Nishanth Parameshwar and Kalluri, Krishna S and Minto, Elise L and Patterson, Jason and Zhang, Linying and Hripcsak, George and G{\"u}rsoy, Gamze and Elhadad, No{\'e}mie and others},
  journal={arXiv preprint arXiv:2402.04400},
  year={2024}
}

@article{yang2022large,
  title={A large language model for electronic health records},
  author={Yang, Xi and Chen, Aokun and PourNejatian, Nima and Shin, Hoo Chang and Smith, Kaleb E and Parisien, Christopher and Compas, Colin and Martin, Cheryl and Costa, Anthony B and Flores, Mona G and others},
  journal={NPJ digital medicine},
  volume={5},
  number={1},
  pages={194},
  year={2022},
  publisher={Nature Publishing Group UK London}
}

@article{swerdel2019phevaluator,
  title={PheValuator: Development and evaluation of a phenotype algorithm evaluator},
  author={Swerdel, Joel N and Hripcsak, George and Ryan, Patrick B},
  journal={Journal of Biomedical Informatics},
  volume={97},
  pages={103258},
  year={2019},
  publisher={Elsevier}
}

@article{lee2021glovephenotyping,
  title={Comparative effectiveness of medical concept embedding for feature engineering in phenotyping},
  author={Lee, Junghwan and Liu, Cong and Kim, Jae Hyun and Butler, Alex and Shang, Ning and Pang, Chao and Natarajan, Karthik and Ryan, Patrick B and Ta, Casey and Weng, Chunhua},
  journal={JAMIA Open},
  volume={4},
  pages={2},
  year={2021}}

@inproceedings{pennington2014glove,
    title = "{G}lo{V}e: Global Vectors for Word Representation",
    author = "Pennington, Jeffrey  and
      Socher, Richard  and
      Manning, Christopher",
    editor = "Moschitti, Alessandro  and
      Pang, Bo  and
      Daelemans, Walter",
    booktitle = "Proceedings of the 2014 Conference on Empirical Methods in Natural Language Processing ({EMNLP})",
    month = oct,
    year = "2014",
    address = "Doha, Qatar",
    publisher = "Association for Computational Linguistics",
    doi = "10.3115/v1/D14-1162",
    pages = "1532--1543"
}

@misc{UptoDateCeliac,
    author = {{UpToDate}},
  title = {Diagnosis of celiac disease in adults},
  howpublished = {\url{https://www.uptodate.com/contents/diagnosis-of-celiac-disease-in-adults}},
  note = {Accessed: 2025-05-06}
}

@misc{UptoDateAMI,
    author = {{UpToDate}},
  title = {Diagnosis of acute myocardial infarction},
  howpublished = {\url{https://www.uptodate.com/contents/diagnosis-of-acute-myocardial-infarction}},
  note = {Accessed: 2025-05-06}
}

@article{swerdel_ramcharran_hardin_2023, title={Using a data-driven approach for the development and evaluation of phenotype algorithms for systemic lupus erythematosus}, volume={18}, ISSN={1932-6203}, DOI={10.1371/journal.pone.0281929}, number={2}, journal={PLOS ONE}, publisher={PLOS ONE}, author={Swerdel, Joel N. and Ramcharran, Darmendra and Hardin, Jill}, year={2023}, pages={e0281929} }

@misc{UptoDateSLE,
    author = {{UpToDate}},
  title = {Systemic lupus erythematosus in adults: Clinical manifestations and diagnosis},
  howpublished = {\url{https://www.uptodate.com/contents/systemic-lupus-erythematosus-in-adults-clinical-manifestations-and-diagnosis}},
  note = {Accessed: 2025-05-06}
}

@misc{UptoDatePC,
    author = {{UpToDate}},
  title = {Clinical manifestations, diagnosis, and staging of exocrine pancreatic cancer},
  howpublished = {\url{https://www.uptodate.com/contents/clinical-manifestations-diagnosis-and-staging-of-exocrine-pancreatic-cancer}},
  note = {Accessed: 2025-05-06}
}

@misc{PancreaticCancerPresentation,
    author = {Hripcsak, George and Chen, RuiJun and Falconer, Thomas},
  title = {Feasibility of Large-Scale
Observational Cancer
Research using the OHDSI
Network—Aim 2 Findings},
  howpublished = {\url{https://www.ohdsi.org/wp-content/uploads/2015/04/NCI-FinalPresentation-OHDSI.pdf}},
  note = {Accessed: 2025-05-06}
}

@misc{UptoDateHTNEpi,
    author = {{UpToDate}},
  title = {Hypertension in adults: Epidemiology},
  howpublished = {\url{https://www.uptodate.com/contents/the-prevalence-and-control-of-hypertension-in-adults}},
  note = {Accessed: 2025-05-06}
}

@misc{UptoDateHTNAdult,
    author = {{UpToDate}},
  title = {Overview of hypertension in adults},
  howpublished = {\url{https://www.uptodate.com/contents/overview-of-hypertension-in-adults}},
  note = {Accessed: 2025-05-06}
}

@article{kaplan2020scaling,
  title={Scaling laws for neural language models},
  author={Kaplan, Jared and McCandlish, Sam and Henighan, Tom and Brown, Tom B and Chess, Benjamin and Child, Rewon and Gray, Scott and Radford, Alec and Wu, Jeffrey and Amodei, Dario},
  journal={arXiv preprint arXiv:2001.08361},
  year={2020}
}

@article{rodriguez2019stand,
  title={Stand-alone artificial intelligence for breast cancer detection in mammography: comparison with 101 radiologists},
  author={Rodriguez-Ruiz, Alejandro and L{\aa}ng, Kristina and Gubern-Merida, Albert and Broeders, Mireille and Gennaro, Gisella and Clauser, Paola and Helbich, Thomas H and Chevalier, Margarita and Tan, Tao and Mertelmeier, Thomas and others},
  journal={JNCI: Journal of the National Cancer Institute},
  volume={111},
  number={9},
  pages={916--922},
  year={2019},
  publisher={Oxford University Press}
}

@article{eisemann2025nationwide,
  title={Nationwide real-world implementation of AI for cancer detection in population-based mammography screening},
  author={Eisemann, Nora and Bunk, Stefan and Mukama, Trasias and Baltus, Hannah and Elsner, Susanne A and Gomille, Timo and Hecht, Gerold and Heywang-K{\"o}brunner, Sylvia and Rathmann, Regine and Siegmann-Luz, Katja and others},
  journal={Nature medicine},
  volume={31},
  number={3},
  pages={917--924},
  year={2025},
  publisher={Nature Publishing Group US New York}
}

@inproceedings{beede2020human,
  title={A human-centered evaluation of a deep learning system deployed in clinics for the detection of diabetic retinopathy},
  author={Beede, Emma and Baylor, Elizabeth and Hersch, Fred and Iurchenko, Anna and Wilcox, Lauren and Ruamviboonsuk, Paisan and Vardoulakis, Laura M},
  booktitle={Proceedings of the 2020 CHI conference on human factors in computing systems},
  pages={1--12},
  year={2020}
}

@article{johnson2023mimic,
  title={MIMIC-IV, a freely accessible electronic health record dataset},
  author={Johnson, Alistair EW and Bulgarelli, Lucas and Shen, Lu and Gayles, Alvin and Shammout, Ayad and Horng, Steven and Pollard, Tom J and Hao, Sicheng and Moody, Benjamin and Gow, Brian and others},
  journal={Scientific data},
  volume={10},
  number={1},
  pages={1},
  year={2023},
  publisher={Nature Publishing Group UK London}
}

@article{pickhardt2023opportunistic,
  title={Opportunistic screening: radiology scientific expert panel},
  author={Pickhardt, Perry J and Summers, Ronald M and Garrett, John W and Krishnaraj, Arun and Agarwal, Sheela and Dreyer, Keith J and Nicola, Gregory N},
  journal={Radiology},
  volume={307},
  number={5},
  pages={e222044},
  year={2023},
  publisher={Radiological Society of North America}
}

@article{al2023machine,
  title={Machine learning for ECG diagnosis and risk stratification of occlusion myocardial infarction},
  author={Al-Zaiti, Salah S and Martin-Gill, Christian and Z{\`e}gre-Hemsey, Jessica K and Bouzid, Zeineb and Faramand, Ziad and Alrawashdeh, Mohammad O and Gregg, Richard E and Helman, Stephanie and Riek, Nathan T and Kraevsky-Phillips, Karina and others},
  journal={Nature Medicine},
  volume={29},
  number={7},
  pages={1804--1813},
  year={2023},
  publisher={Nature Publishing Group US New York}
}

@ARTICLE{Ahsan2022-mlearlydx,
  title     = "Machine-learning-based disease diagnosis: A comprehensive review",
  author    = "Ahsan, Md Manjurul and Luna, Shahana Akter and Siddique, Zahed",
  journal   = "Healthcare (Basel)",
  publisher = "MDPI AG",
  volume    =  10,
  number    =  3,
  pages     = "541",
  month     =  mar,
  year      =  2022,
  copyright = "https://creativecommons.org/licenses/by/4.0/",
  language  = "en"
}

@ARTICLE{Rajagopal2024-mlopscost,
  title     = "Machine learning operations in health care: A scoping review",
  author    = "Rajagopal, Anjali and Ayanian, Shant and Ryu, Alexander J and
               Qian, Ray and Legler, Sean R and Peeler, Eric A and Issa,
               Meltiady and Coons, Trevor J and Kawamoto, Kensaku",
  journal   = "Mayo Clin. Proc. Digit. Health",
  publisher = "Elsevier BV",
  volume    =  2,
  number    =  3,
  pages     = "421--437",
  month     =  sep,
  year      =  2024,
  copyright = "http://creativecommons.org/licenses/by-nc-nd/4.0/",
  language  = "en"
}

@ARTICLE{Feng2025-genomicrobustness,
  title     = "Benchmarking {DNA} foundation models for genomic and genetic
               tasks",
  author    = "Feng, Haonan and Wu, Lang and Zhao, Bingxin and Huff, Chad and
               Zhang, Jianjun and Wu, Jia and Lin, Lifeng and Wei, Peng and Wu,
               Chong",
  journal   = "Nat. Commun.",
  publisher = "Springer Science and Business Media LLC",
  volume    =  16,
  number    =  1,
  pages     = "10780",
  month     =  nov,
  year      =  2025,
  copyright = "https://creativecommons.org/licenses/by/4.0",
  language  = "en"
}

@ARTICLE{Bian2025-imagingrobustness,
  title     = "Artificial intelligence in medical imaging: From task-specific
               models to large-scale foundation models",
  author    = "Bian, Yueyan and Li, Jin and Ye, Chuyang and Jia, Xiuqin and
               Yang, Qi",
  journal   = "Chin. Med. J. (Engl.)",
  publisher = "Ovid Technologies (Wolters Kluwer Health)",
  volume    =  138,
  number    =  6,
  pages     = "651--663",
  month     =  mar,
  year      =  2025,
  language  = "en"
}

@ARTICLE{Timilsina2025-harmonizingfms,
  title     = "Harmonizing foundation models in healthcare: A comprehensive
               survey of their roles, relationships, and impact in artificial
               intelligence's advancing terrain",
  author    = "Timilsina, Mohan and Buosi, Samuele and Razzaq, Muhammad Asif
               and Haque, Rafiqul and Judge, Conor and Curry, Edward",
  journal   = "Comput. Biol. Med.",
  publisher = "Elsevier BV",
  volume    =  189,
  number    =  109925,
  pages     = "109925",
  month     =  may,
  year      =  2025,
  copyright = "http://creativecommons.org/licenses/by/4.0/",
  language  = "en"
}

@article{wynants2020prediction,
  title={Prediction models for diagnosis and prognosis of covid-19: systematic review and critical appraisal},
  author={Wynants, Laure and Van Calster, Ben and Collins, Gary S and Riley, Richard D and Heinze, Georg and Schuit, Ewoud and Albu, Elena and Arshi, Banafsheh and Bellou, Vanesa and Bonten, Marc MJ and others},
  journal={bmj},
  volume={369},
  year={2020},
  publisher={British Medical Journal Publishing Group}
}

@article{zeiberg2019machine,
  title={Machine learning for patient risk stratification for acute respiratory distress syndrome},
  author={Zeiberg, Daniel and Prahlad, Tejas and Nallamothu, Brahmajee K and Iwashyna, Theodore J and Wiens, Jenna and Sjoding, Michael W},
  journal={PloS one},
  volume={14},
  number={3},
  pages={e0214465},
  year={2019},
  publisher={Public Library of Science San Francisco, CA USA}
}

@article{radford2019language,
  title={Language models are unsupervised multitask learners},
  author={Radford, Alec and Wu, Jeffrey and Child, Rewon and Luan, David and Amodei, Dario and Sutskever, Ilya and others},
  journal={OpenAI blog},
  volume={1},
  number={8},
  pages={9},
  year={2019}
}
